\documentclass[10pt]{article} 
\usepackage[accepted]{tmlr}

\usepackage{amsmath,amsfonts,bm}

\def\eqref#1{equation~\ref{#1}}

\def\1{\bm{1}}

\DeclareMathAlphabet{\mathsfit}{\encodingdefault}{\sfdefault}{m}{sl}
\SetMathAlphabet{\mathsfit}{bold}{\encodingdefault}{\sfdefault}{bx}{n}

\usepackage{algorithm}
\usepackage{algorithmic}
\usepackage{hyperref}
\usepackage{url}
\usepackage{wrapfig}
\usepackage{booktabs}
\usepackage{multirow}
\usepackage{graphicx}
\usepackage{enumitem}
\usepackage{adjustbox}
\usepackage{subcaption}
\usepackage{lipsum}  
\usepackage{multicol}
\usepackage{longtable}
\usepackage{xcolor}
\usepackage{amssymb}
\usepackage{tikz}
\usepackage{oplotsymbl}

\title{Navigating Noise: A Study of How Noise Influences\\ Generalisation and Calibration of Neural Networks}

\author{\name Martin Ferianc\thanks{Joint first authors.} \email martin.ferianc.19@ucl.ac.uk \\
      \addr Department of Electronic and Electrical Engineering\\
      University College London
      \and
      \name Ondrej Bohdal\footnotemark[1]  \email ondrej.bohdal@ed.ac.uk \\
      \addr School of Informatics\\
      University of Edinburgh
      \and
      \name Timothy Hospedales \email t.hospedales@ed.ac.uk \\
      \addr School of Informatics\\
      University of Edinburgh\\
      Samsung AI Center Cambridge
      \and
      \name Miguel Rodrigues \email m.rodrigues@ucl.ac.uk \\
      \addr Department of Electronic and Electrical Engineering\\
      University College London
      \and}

\def\month{04}  
\def\year{2024} 
\def\openreview{\url{https://openreview.net/forum?id=zn3fB4VVF0}} 

\begin{document}

\maketitle

\begin{abstract}
    Enhancing the generalisation abilities of neural networks (NNs) through integrating noise such as MixUp or Dropout during training has emerged as a powerful and adaptable technique. 
    Despite the proven efficacy of noise in NN training, there is no consensus regarding which noise sources, types and placements yield maximal benefits in generalisation and confidence calibration. 
    This study thoroughly explores diverse noise modalities to evaluate their impacts on NN's generalisation and calibration under in-distribution or out-of-distribution settings, paired with experiments investigating the metric landscapes of the learnt representations across a spectrum of NN architectures, tasks, and datasets.
    Our study shows that AugMix and weak augmentation exhibit cross-task effectiveness in computer vision, emphasising the need to tailor noise to specific domains.
    Our findings emphasise the efficacy of combining noises and successful hyperparameter transfer within a single domain but the difficulties in transferring the benefits to other domains. 
    Furthermore, the study underscores the complexity of simultaneously optimising for both generalisation and calibration, emphasising the need for practitioners to carefully consider noise combinations and hyperparameter tuning for optimal performance in specific tasks and datasets.
\end{abstract}

\section{Introduction}\label{sec:introduction}

Neural networks (NNs) have demonstrated remarkable capabilities across various tasks, yet they often grapple with overfitting to training data, resulting in suboptimal generalisation performance on unseen samples~\citep{srivastava2014dropout,bishop1995training, sietsma1991creating}. 
Addressing this issue, conventional techniques such as weight decay~\citep{krogh1991simple} and early stopping~\citep{prechelt2002early} have been employed to regularise NN training. 
Alongside these methods, the introduction of noise during the NN's training has emerged as a potent strategy to enhance generalisation~\citep{sietsma1991creating,neelakantan2017adding, camuto2021understanding, kukavcka2017regularization}.
The concept of noise injections refers to the deliberate introduction of artificial perturbations into different aspects of NN training. 
Note that this is distinct from the concept of noise in the data itself which originates from the data collection process~\citep{song2022learning}.
Diverging from weight decay and early stopping that modulate the model's search within the hypothesis space, noise injections embrace randomness during training, fostering exploration of a broader array of representations~\citep{he2019parametric}. 
The appeal of noise injections extends further due to their versatile applicability across diverse tasks, datasets, and NN architectures. 
These attributes establish noise injections as a convenient approach for enhancing NN's generalisation.

In addition to generalisation, confidence calibration is a desirable model property, especially in safety-critical applications where confidence scores must be aligned with the model's accuracy to make informed decisions~\citep{guo2017calibration}.
Empirically, noise injections have been shown to improve confidence calibration by improving the generalisation of the NNs in previously unseen circumstances and inherently reducing overconfidence in predictions~\citep{guo2017calibration, muller2019does, hendrycks2019augmix, zhang2017mixup, gal2016dropout}.
However, the relationship between generalisation and calibration is not straightforward, and the two properties are often at odds with each other~\citep{guo2017calibration}.

Various noise injection methodologies have been proposed, encompassing \textbf{activation} techniques such as Dropout~\citep{srivastava2014dropout, gal2016dropout} and Gaussian Dropout~\citep{kingma2015variational}, \textbf{weight} noises such as DropConnect~\citep{wan2013regularization} or additive Gaussian noise~\citep{blundell2015weight}, \textbf{target} methods such as label smoothing~\citep{szegedy2016inception}, \textbf{input-target} strategies exemplified by MixUp~\citep{zhang2017mixup}, \textbf{input} modifications such as AugMix~\citep{hendrycks2019augmix} or the standard horizontal flipping and center cropping~\citep{krizhevsky2009learning},
\textbf{model} approaches including weight perturbation~\citep{ash2020warm}, and \textbf{gradient} perturbations involving Gaussian noise~\citep{neelakantan2017adding}.
Despite the diversity of these techniques, comprehensive and fair comparisons are scarce, leaving a gap in understanding which approach is helpful for specific datasets, tasks and models in conjunction with generalisation and calibration. 

This study aims to systematically and comprehensively investigate the effects of widely used noise injection methods on NN generalisation and calibration across multiple datasets, tasks, and architectures.
This exploration is predicated on the premise that while generalisation focuses on reducing overfitting and improving the model's predictive accuracy, calibration deals with aligning the model's confidence with its actual performance.
Rather than focusing on improving state-of-the-art performance, we aim to provide a holistic view of the effects of noise injections on NNs' generalisation and calibration for the benefit of practitioners.
To this end, we present the following contributions:

\begin{enumerate}[leftmargin=*]
\item The first systematic empirical investigation into the impact of noise injections on NN generalisation and calibration across diverse datasets, tasks and NN architectures. 
Our exploration extends to evaluation under in-distribution (ID) and out-of-distribution (OOD) scenarios and their transferability across architectures and datasets.
\item A methodological framework for simultaneously combining various noise injection approaches.
\item Visualisation of the learnt representation landscape across noises, jointly comparing calibration and generalisation performance. 
\end{enumerate}

Our investigation reveals that certain types of noise aid in generalisation by introducing robustness against overfitting and variability in data and potentially improve calibration by mitigating overconfidence in predictions. 
The findings show that AugMix, weak augmentation and Dropout prove effective across diverse tasks, emphasising their versatility. 
Task-specific nuances in noise effectiveness, such as AugMix's superiority in computer vision (CV), Dropout in natural language processing (NLP) and Gaussian noise in tabular data regression, highlight the need for tailored approaches.
Combining noises, careful hyperparameter tuning, and task-specific considerations are crucial for optimising NN's performance.
Our code is publicly available at \url{https://github.com/martinferianc/noise}.

\section{Related Work}\label{sec:related_work}

In this study, we consider artificial addition of noise into various facets of NN training -- including \textbf{input}, \textbf{target}, \textbf{input-target}, \textbf{activations}, \textbf{weights}, \textbf{gradients}, and \textbf{model} parameters.
The noise application is denoted by $\alpha_{\textrm{<place>}}(\cdot, \delta)$, where $\alpha_{\textrm{<place>}}$ is the noise application methodology which can be executed at different places, e.g. $\alpha_{\textrm{input}}$ for input noise, $\alpha_{\textrm{target}}$ for target noise along with $\cdot$ arbitrary arguments, depending on the noise injection methodology.
For example, under this definition, additive input Gaussian noise samples a Gaussian and adds it to the input $x$ as $\alpha_{\textrm{input}}(x, \delta)=x+\epsilon;\epsilon \sim \mathcal{N}(0, \sigma^2)$, where $\sigma^2$ is the hyperparameter in $\delta$.
Note that this study focuses on \textit{artificial} noise injections, which are purposely introduced during training, and not on \textit{natural} noise, inherent in the data, e.g. label noise in the targets where the classifying label is incorrect.
The natural noise needs to be addressed separately and we refer the reader to~\citet{song2022learning} for a review of strategies for learning with noisy labels.
Under different noise placements, we review several noise injection strategies.
The review focused on the most fundamental noise injection methodologies, which constitute the building blocks of more complex approaches and represent the noise injection category.

\textbf{Input Noise}:
Pioneering work by~\citet{sietsma1991creating} demonstrated the benefits of training with added input Gaussian noise, while~\citet{bishop1995training} established its linkage to regularisation in the least squares problems.
In CV, weak augmentation, such as random cropping and horizontal flipping, has improved generalisation~\citep{krizhevsky2009learning}.
AugMix, domain-specific to CV, applies a sequence of image processing operations to the input, bolstering robustness in OOD settings.
From the adversarial robustness domain, ODS augments inputs conditioned on the prediction, aiming to diversify the inputs~\citep{tashiro2020diversity}. 
\textbf{Target Noise}:
Label smoothing~\citep{pereyra2017regularizing} softens the one-hot classification targets by replacing the targets with a categorical distribution with the most mass on the correct class and the rest spread across the other classes through the addition of constant uniform noise, effectively improving NN's robustness~\citep{muller2019does}.
This differs from label noise, where the entire probability mass is on an incorrect class, and the NN must learn to ignore the errors~\citep{song2022learning}.
\textbf{Input-Target Noise}:
Variants of MixUp have exhibited efficacy in augmenting both generalisation and calibration~\citep{zhang2017mixup, muller2019does, guo2019augmenting, yao2022c, guo2017calibration}.
MixUp adds noise to both the input and the target via linear interpolation between two samples and their targets, while CMixUp expands this approach to regression problems.
\textbf{Activation Noise:} 
Widespread activation noise includes Dropout or Gaussian noise injections. 
Dropout~\citep{srivastava2014dropout, noh2017regularizing} randomly deactivates activations through 0-1 noise, while Gaussian noise injections add noise to activations~\citep{kingma2015variational,devries2017dataset}.
Bayesian NNs~\citep{gal2016dropout} incorporate these injections during training and evaluation, in contrast to our work's focus solely on their application in training. 
\textbf{Weight Noise:}
Unlike Dropout, DropConnect~\citep{wan2013regularization} randomly deactivates weights or connections between neurons, while Gaussian noise injections add noise to weights~\citep{blundell2015weight}.
Note that we do not model the variance of the Gaussian noise through learnable parameters, as in~\citep{blundell2015weight}, but rather fix it through a searchable hyperparameter.
We do this to ensure a fair comparison with other noise injection approaches, such as Dropout, which do not have learnable parameters and would require changing the model architecture to accommodate them.
\textbf{Gradient Noise:} 
Annealed Gaussian noise added to gradients during training has demonstrated its efficacy in improving NN generalisation~\cite{neelakantan2017adding, welling2011bayesian, zhou2019toward, chaudhari2015energy, wu2020noisy}.  
\textbf{Model Noise:}
A recent contribution, Gaussian noise injection through periodic weight shrinking and perturbation~\cite{ash2020warm}, improves retraining generalisation.

In previous work, the impact of noise per injection type was studied. 
\citet{poole2014analyzing} show that injecting noise at different layers of autoencoders implements various regularisation techniques and can improve feature learning and classification performance.
~\citet{cohen2019certified} show that smoothing classifiers with Gaussian noise naturally induces robustness in the L2 norm.
~\citet{wei2020implicit} disentangle and analytically characterise the explicit regularisation effect from modifying the expected training objective and the implicit regularisation effect from the stochasticity of Dropout noise in NNs.
~\citet{camuto2021understanding, camuto2020explicit} show that training NNs with Gaussian noise injections on inputs and activations regularises them to learn lower frequency functions, improves generalisation and calibration on unseen data but also confers robustness to perturbation.
On one hand,~\citet{jang2021noise} show that training NNs on noisy images can improve their robustness and match human behavioural and neural responses.
On the other hand,~\citet{geirhos2018generalisation} demonstrate that adding specific noise to the input can surpass humans in generalisation on that specific noise, but not to other types of noise, while human vision is robust to a wide range of noise types.
The results of~\citet{geirhos2018generalisation} are confirmed by the results of~\citet{kang2019transfer}, who show that robustness against one type of noise does not necessarily transfer to robustness against other types of noise.
Furthermore,~\citet{kang2019transfer} consider adversarial training, where a model is trained to be robust against noise-based adversarial attacks~\citep{goodfellow2014explaining}.
An adversarial attack is a specific type of noise injection during evaluation, where the noise is designed to fool the model.
In comparison, our work focuses on the generalisation and confidence calibration performance of NNs with respect to domain shift in the data distribution, rather than adversarial attacks.
We consider enhancing robustness to adversarial attacks through artificial noise injections as future work.
Moreover,~\citep{kukavcka2017regularization} provided a taxonomy of regularisation in NNs, covering multiple noise-based approaches. 

The closest work to ours is~\citep{chun2020empirical}, which considered regularisation commonly used during training and its impact on generalisation, confidence calibration and out-of-distribution detection in computer vision.
While their focus was not noise-specific, as in our work, they overlap with our work by considering input noise: weak augmentation and Gaussian noise, target noise: label smoothing~\citep{muller2019does}, input-target noise: MixUp~\citep{zhang2017mixup}.
They show that common regularisation techniques improve generalisation, confidence calibration and out-of-distribution detection.
In comparison to~\citep{chun2020empirical}, our work focuses on a broader set of noise injections, network architectures, datasets and tasks, evaluation of the weight landscapes, and in-depth noise combinations paired with comprehensive hyperparameter search.

Past work has studied noise injection techniques in isolation, mainly focused on generalisation alone, lacked comprehensive hyperparameter optimisation, and rarely evaluated the robustness of distribution shift. 
For example, only MixUp, AugMix and label smoothing have been studied in calibration~\citep{guo2017calibration, muller2019does, guo2019augmenting, yao2022c, chun2020empirical}.
An exception to this is~\citet{chun2020empirical}, who studied generalisation, calibration and out-of-distribution detection for some noise injections.
While promising, these methods require further unified analysis to determine their relationships across datasets, tasks, architectures and across a broader set of noise injections.
Our work addresses these gaps by \textit{1.)} studying the impact across datasets, tasks and architectures; \textit{2.)} benchmarking the impact of noise injections' hyperparameters on transferability between datasets and architectures; \textit{3.)} studying confidence-calibration in addition to generalisation; \textit{4.)} performing a comprehensive hyperparameter search with fair comparisons; \textit{5.)} evaluating robustness to distribution shift; \textit{6.)} providing a methodological framework for combining and tuning various noise injection approaches across categories; and lastly \textit{7.)} visualising the learnt representation or learning landscape across noise injections in 1D or 2D~\citep{goodfellow2014qualitatively, li2018visualizing} across both generalisation and calibration.

\section{Methodology}\label{sec:methodology}

We establish a structured methodology to investigate noise injections' effects on NNs.
The noise types are divided into \textbf{input}, \textbf{input-target}, \textbf{target}, \textbf{activation}, \textbf{weight}, \textbf{gradient} and \textbf{model}, and we enable their conditional deployment through probabilities $\{p_{noise}^{i}\}_{i=1}^S$ in the range $0 \leq p_{noise}^{i} \leq 1$, where $S$ denotes the number of noises. 

The training allows simultaneous consideration of $S$ noise types, each associated with specific hyperparameters $\{\delta^i\}_{i=1}^S$ and an application function $\{\alpha^i_{\textrm{<place>}}(\cdot, \delta)\}_{i=1}^S$, where $\alpha^i_{\textrm{<place>}}$ is the noise application methodology which can be executed at different places, e.g. $\alpha_{\textrm{input}}$ for input noise, $\alpha_{\textrm{target}}$ for target noise along with $\cdot$ arbitrary arguments, depending on the noise injection methodology.
The different noise types implement only the relevant $\alpha^i_{\textrm{<place>}}(\cdot, \delta)$ function, while others are ignored.
We encourage the reader to refer to the code for the implementation details for each noise type.
The probabilities $\{p_{noise}^{i}\}_{i=1}^S$ allow us to tune the frequency of applying each noise type, while the hyperparameters $\{\delta^i\}_{i=1}^S$ enable us to adjust the magnitude of each noise type.
This enables us to tune both the magnitude and frequency of noise injections, unlike, for example, Dropout~\citep{srivastava2014dropout}, which only allows the tuning of the magnitude, and it is applied every batch.
The tuning of the frequency allows us to avoid conflicts between noises, as it can be set to 0 if the noise is conflicting with other noises.

Algorithm~\ref{alg:training} provides a comprehensive overview of the training process, executed throughout $E$ epochs with $L$ batches processed per epoch. 
For every batch, input and target data $(x_{b}, y_{b})$ are randomly drawn from the training dataset $\mathcal{D} = \{(x_{b}, y_{b})\}_{b=1}^{L}$. 
For each noise in $S$, we sample a uniform random variable $\epsilon \sim U(0,1)$, and if $\epsilon < p_{noise}^{i}$, we enable noise $i$ with hyperparameters $\delta^i$ for the current batch $b$.

\begin{wrapfigure}{r}{0.62\textwidth}
\begin{minipage}{0.62\textwidth}
\vspace{-2em}
\begin{algorithm}[H]
\caption{Training of a Neural Network with Noise}
\label{alg:training}
\begin{algorithmic}[1]
\REQUIRE Training dataset $\mathcal{D} = \{(x_{b}, y_{b})\}_{b=1}^{L}$, $L$ batches, number of epochs $E$, network depth $D$, weights $W = \{W^{d}\}_{d=1}^D$, hidden states $z_b=\{z^d_b\}_{d=1}^D$, activations $\phi=\{\phi^d(\cdot)\}_{d=1}^D$, weighted operations $f=\{f^d(\cdot, W^{d})\}_{d=1}^D$, $S$ noise types, probabilities of applying noise to a batch $p_{noise}=\{p_{noise}^{i}\}_{i=1}^S$, Noise hyperparameters $\delta=\{\delta^i\}_{i=1}^S$, Noise application functions $\alpha_{\textrm{<place>}}=\{\alpha^i_{\textrm{<place>}}(\cdot, \delta)\}_{i=1}^S$.

\STATE Initialise $W$ randomly
\FOR{$e=1$ to $E$}
    \FOR{$b=1$ to $L$}
        \STATE Randomly select a batch $(x_{b}, y_{b})$ from $\mathcal{D}$
        \STATE Sample $\epsilon=\{\epsilon^{i} \sim U(0,1)\}_{i=1}^S$ 
        \STATE Set toggles $t=\{t^{i} = \epsilon^{i} < p_{noise}^{i}\}_{i=1}^S$
        \STATE \textbf{Input noise}: $\textrm{apply}\_\alpha_{\textrm{input}}(x_{b}, t, \delta)$
        \STATE \textbf{Target noise}: $\textrm{apply}\_\alpha_{\textrm{target}}(y_{b}, t, \delta)$
        \STATE \textbf{Input-target noise}: $\textrm{apply}\_\alpha_{\textrm{input-target}}(x_b, y_{b}, t, \delta)$
        \STATE $z^0_b= x_{b}$
        \FOR{$d=1$ to $D$}
            \STATE \textbf{Weight noise}: $\textrm{apply}\_\alpha_{\textrm{weight}}(W^{d}, t, \delta)$
            \STATE Compute hidden state $z^d_b = f^d(z_b^{d-1}, W^{d})$
            \STATE \textbf{Activation noise}: $\textrm{apply}\_\alpha_{\textrm{activation}}(z^d_b, t, \delta)$ if $d<D$
            \STATE $z^d_b = \phi^d(z^d_b)$
        \ENDFOR
        \STATE Assign predictions $\hat{y}_{b} = z^d_b$
        \STATE Compute loss $\mathcal{L}(\hat{y}^{i}, y^{i})$ and gradients $\nabla_W \mathcal{L}$
        \STATE \textbf{Gradient noise}: $\textrm{apply}\_\alpha_{\textrm{gradient}}(\nabla_W \mathcal{L}, t, \delta)$
        \STATE Update weights $W$
    \ENDFOR
    \STATE \textbf{Model noise}: $\textrm{apply}\_\alpha_{\textrm{model}}(W, e, t, \delta)$
\ENDFOR
\STATE \textbf{Procedure:} $\textrm{apply}\_\alpha_{\textrm{<place>}}(\cdot, t, \delta)$
\FOR{$i=1$ to $S$}
    \IF{$t^{i}$ and $\alpha^i_{\textrm{<place>}}$ exists}
        \STATE $\alpha^i_{\textrm{<place>}}(\cdot, \delta^i)$
    \ENDIF
\ENDFOR
\end{algorithmic}
\end{algorithm}
\end{minipage}
\vspace{-0.5cm}
\end{wrapfigure}

For each noise in $S$, we sample a uniform random variable $\epsilon \sim U(0,1)$, and if $\epsilon < p_{noise}^{i}$, we enable noise $i$ through setting the toggle $t^{i}$ to 1 for the current batch $b$.
The enabled noises are applied in the order: \textit{1.)} input, target, input-target, \textit{2.)} weights, \textit{3.)} activations, \textit{4.)} gradients and \textit{5.)} model through the $\textrm{apply}\_\alpha_{\textrm{<place>}}(\cdot, t, \delta)$ procedure.
The procedure sequentially iterates over the noise types in $S$ and applies the noise if the noise is enabled and the application function $\alpha^i_{\textrm{<place>}}$ exists.
The user specifies the order of the noises in $S$.

Our approach accounts for networks of depth $D$, denoted by $\{f^d(\cdot, W^{d})\}_{d=1}^D$, involving weights together with biases $W = \{W^{d}\}_{d=1}^D$ and activations $\{\phi^d(\cdot)\}_{d=1}^D$ to produce hidden states $\{z^d_b\}_{d=1}^D$. $z^b_0$ corresponds to the input $x_{b}$, while $z^D_b$ represents the output prediction $\hat{y}_{b}$.

For \textbf{input} noise, we explore AugMix, ODS, weak augmentation: random cropping and horizontal flipping, and additive Gaussian noise injections~\citep{hendrycks2019augmix,tashiro2020diversity,sietsma1991creating}. 
For \textbf{input-target} we explore MixUp and CMixUp~\citep{zhang2017mixup,yao2022c}. 
For \textbf{target} noise, we consider label smoothing, and the target noise also inherently involves MixUp and CMixUp~\citep{zhang2017mixup,yao2022c,muller2019does}. 
The \textbf{activation} noise examines Dropout and additive Gaussian noise~\citep{srivastava2014dropout,kingma2015variational} prior to activations for all linear or convolutional layers, except the last layer.
For \textbf{weight} noise, we consider Gaussian noise added to the weights~\citep{blundell2015weight} or DropConnect~\citep{wan2013regularization} for all linear or convolutional layers, except the last layer.
We consider \textbf{gradient} Gaussian noise added to all gradients of the loss function~\citep{neelakantan2017adding}. 
After the update of the weights, the \textbf{model} noise is applied to the weights, for which we consider shrinking the weights and adding Gaussian noise~\citep{ash2020warm}, but not in the last 25\% of the training epochs.
Out of these noises, label smoothing, MixUp and ODS are exclusive to classification, and CMixUp is applicable only in regression. 
AugMix and weak augmentation are exclusive to the CV data.
The other noises are broadly applicable across tasks.
\section{Experiments}\label{sec:experiments}

Next, in Section~\ref{sec:experiments:settings} we present the concrete datasets, tasks and architectures used in our experiments, followed by experiments on ID data in Section~\ref{sec:experiments:in-domain}, OOD data in Section~\ref{sec:experiments:out-of-domain}, combined noises in Section~\ref{sec:experiments:combination}, transferability in Section~\ref{sec:experiments:transferability} and lastly the metric landscape visualisations in Section~\ref{sec:experiments:landscapes}.

\subsection{Experimental Settings}\label{sec:experiments:settings}

\textbf{Tasks, Architectures and Datasets:} 
We consider various setups, including computer vision (CV) classification and regression, tabular data classification and regression, and natural language processing (NLP) classification. 
For CV classification we include datasets such as CIFAR-10, CIFAR-100~\citep{krizhevsky2009learning}, SVHN~\citep{netzer2011reading}, and TinyImageNet~\citep{le2015tiny}, along with neural architectures such as a fully-connected (FC) net and ResNet~\citep{he2016deep}. 
For CV regression, we introduce a rotated version of CIFAR-100 to predict the rotation angle, and we also use the WikiFace dataset~\citep{rothe2015dex}, where the aim is to predict the age based on the image of the face. 
We use the ResNet model in both cases. 
We deem the rotation prediction task compelling to evaluate since it is a common task in the literature for self-supervised pre-training~\citep{gidaris2018unsupervised}. 
In the realm of tabular data classification and regression, we use an FC network and evaluate noises on diverse datasets, including Wine, Toxicity, Abalone, Students, Adult for classification and Concrete, Energy, Boston, Wine, Yacht for regression~\citep{asuncion2007uci}.
We explore NLP classification using the NewsGroup and SST-2 datasets~\citep{Lang1995Newsweeder,socher2013recursive} paired with global pooling convolutional NN~\citep{kim2014convolutional} and a transformer~\citep{vaswani2017attention}.
The Appendix details the datasets and architectures and gives the complete numerical results.
 
\textbf{Metrics:}
To assess the effectiveness of the noise injection methods in classification, we measure their performance using three metrics: Error ($\downarrow, \%$), Expected Calibration Error (ECE)~\citep{guo2017calibration} ($\downarrow, \%$) with 10 bins and the categorical Negative Log-Likelihood (NLL) ($\downarrow$).
For regression, we use the Mean Squared Error (MSE) ($\downarrow$) and the Gaussian NLL ($\downarrow$).
We test the generalisation of the models by evaluating their performance on the ID test set. 
For CV classification and regression, we test the robustness of the models by assessing their performance on an OOD test set by applying corruptions~\citep{hendrycks2019benchmarking} to the ID test set. 
These corruptions include, for example, adding snow or fog to the image, changing the brightness or saturation of the image or blurring the image across 5 intensities.
We created the OOD test set for tabular data by adding or multiplying the inputs with Gaussian or Uniform noise or by zeroing some of the input features with Bernoulli noise, similarly across 5 intensities.
While vision data has ImageNet-C~\citep{hendrycks2019benchmarking}, to the best of our knowledge, there is no similar benchmark for tabular data. 
Our methodology for introducing perturbations and zeroing out features is designed to simulate a wide range of potential distribution shifts in real-world scenarios, such as instrumentation errors, missing data, and adversarial attacks. 
We crafted the OOD evaluation to be similar to~\citep{hendrycks2019benchmarking}, in terms of the magnitude and severity of the noise, allowing us to systematically evaluate the robustness of the models.
To summarise the results, we collect the results for each approach for each dataset and metric and rank them relative to the no noise baseline. 
For example, -1 means that the approach is one rank better than the no noise baseline, and 1 means that the approach is one rank worse than the no noise baseline.
We then average the ranks across the datasets for each task and metric.

\textbf{Hyperparameter Optimisation:}
We first tune the learning rate and L2 regularisation of a no noise network, which are reused when tuning the HPs of each noise injection method.
By tuning the learning rate and L2 regularisation, we wanted to simulate a realistic scenario where the practitioner seeks to add noise to their existing model and does not want to jointly tune the model's hyperparameters and the noise injection method. 
The tuning was performed with 1 seed, and the winning hyperparameters were retrained 3 times with different seeds.
10\% of the training data was used as the validation set to select the best model, with validation NLL used as the selection objective to combine both generalisation and calibration.
The tuning is performed using model-based Tree-structured Parzen Estimator method~\citep{bergstra2011tpe} with successive halving pruning strategy~\citep{jamieson2016non}. We evaluate 50 trials for each setting, which allows us to manage the trade-off between compute costs and a reasonable number of trials.

\subsection{In-Domain Evaluation}\label{sec:experiments:in-domain}

\begin{figure}
  \centering
  \begin{subfigure}{0.48\textwidth}
    \includegraphics[width=\textwidth]{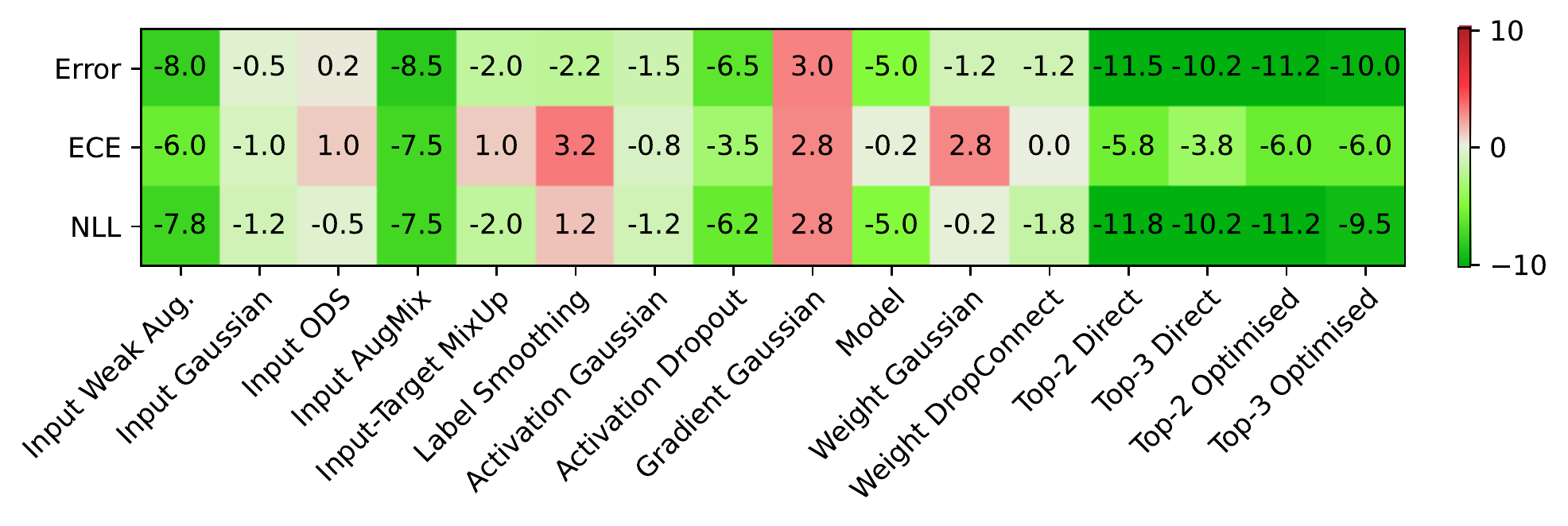}
    \vspace{-0.70cm}
    \caption{CV classification.}
    \label{fig:in_domain:cv_classification}
  \end{subfigure}
  \hfill
  \begin{subfigure}{0.48\textwidth}
    \includegraphics[width=\textwidth]{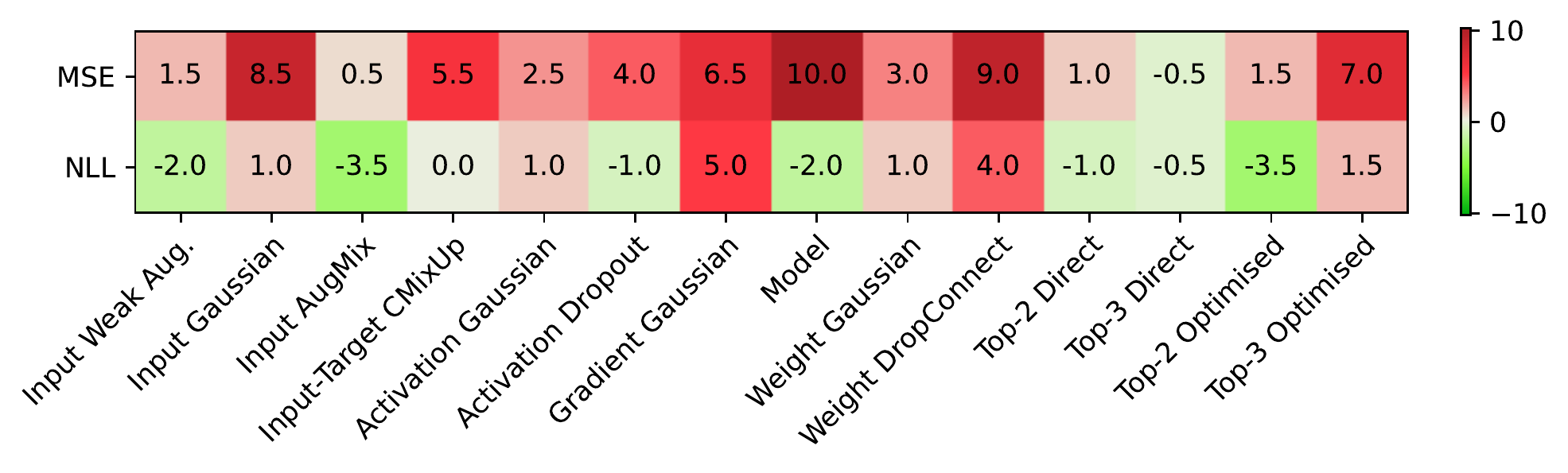}
    \vspace{-0.70cm}
    \caption{CV regression.}
    \label{fig:in_domain:cv_regression}
  \end{subfigure}
  \begin{subfigure}{0.48\textwidth}
    \includegraphics[width=\textwidth]{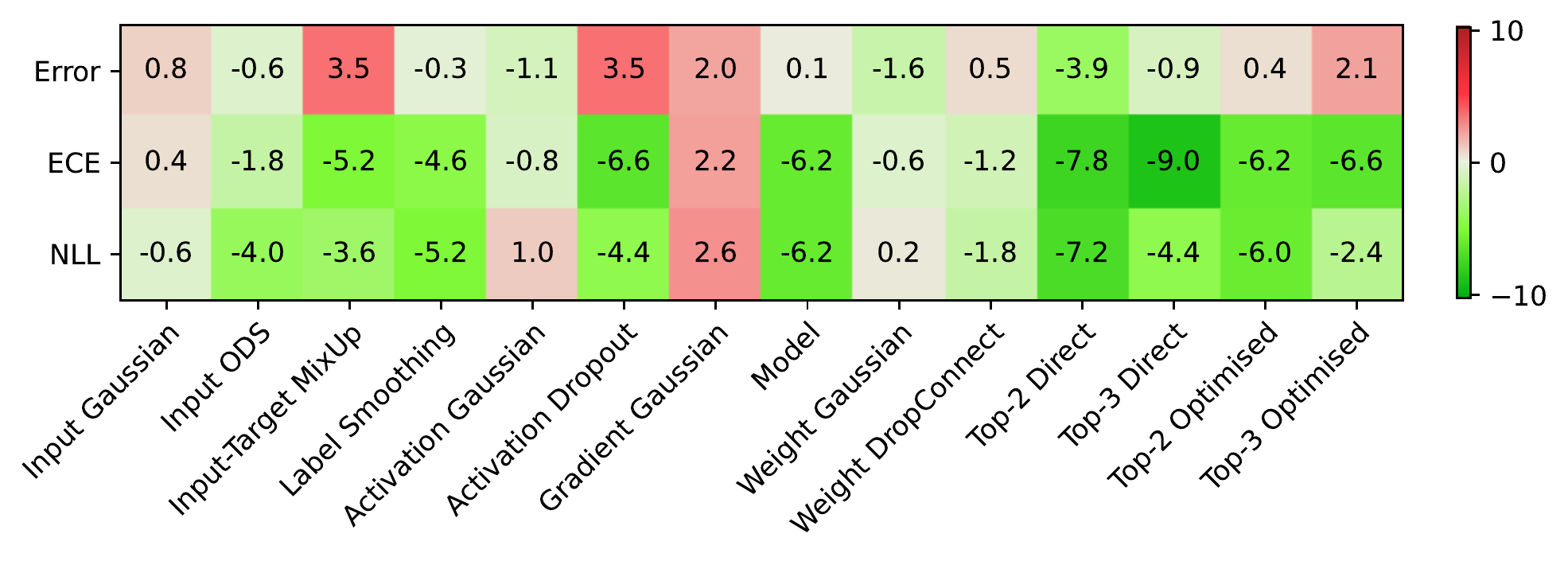}
    \vspace{-0.70cm}
    \caption{Tabular classification.}
    \label{fig:in_domain:tab_classification}
  \end{subfigure}
  \hfill
  \begin{subfigure}{0.48\textwidth}
    \includegraphics[width=\textwidth]{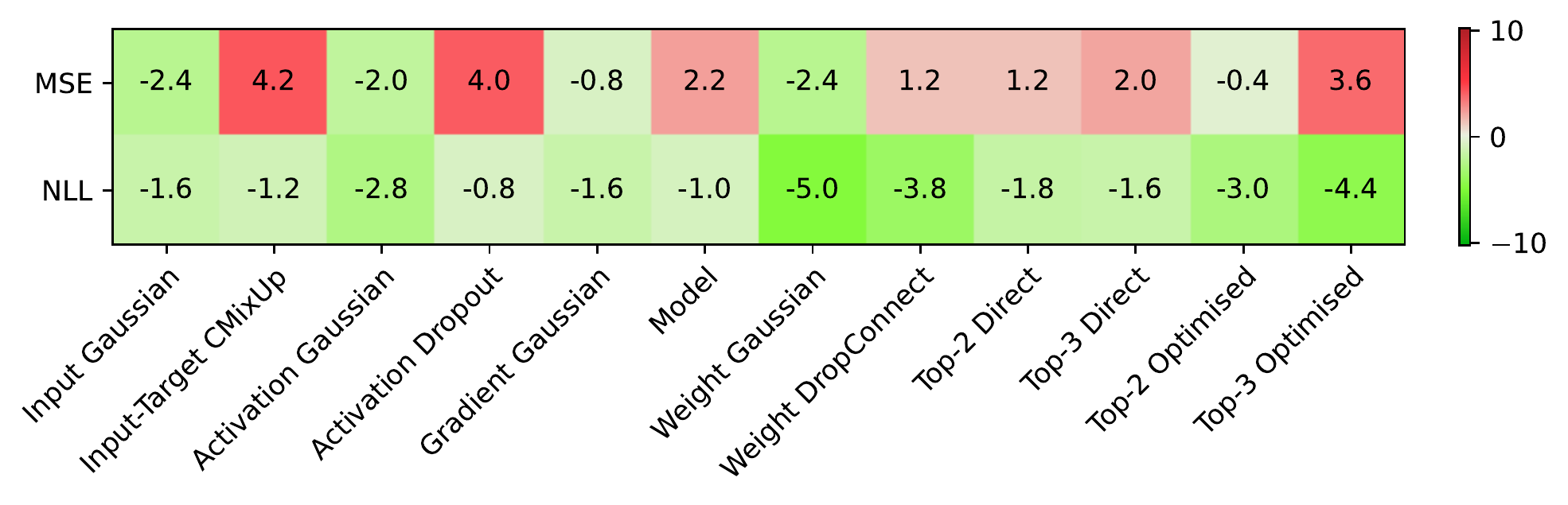}
    \vspace{-0.70cm}
    \caption{Tabular regression.}
    \label{fig:in_domain:tabular_regression}
  \end{subfigure}
  \hfill
  \begin{subfigure}{0.48\textwidth}
    \includegraphics[width=\textwidth]{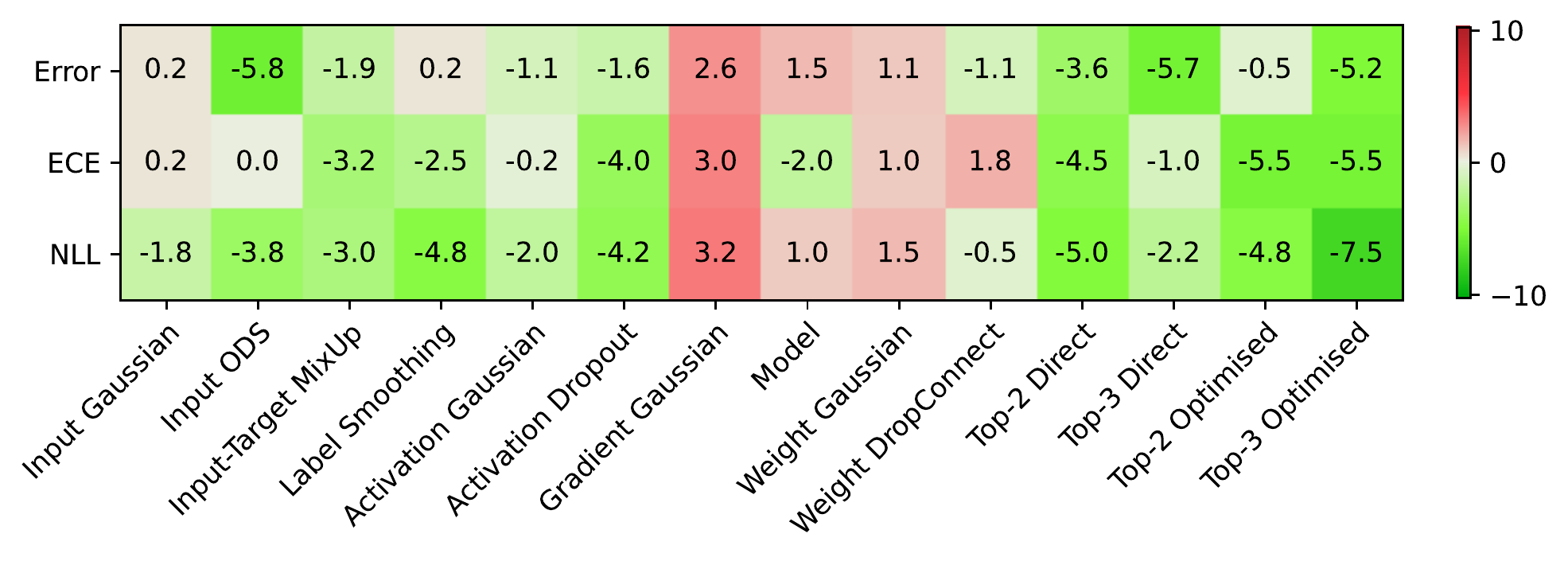}
    \vspace{-0.70cm}
    \caption{NLP classification.}
    \label{fig:in_domain:nlp_classification}
  \end{subfigure}
  \vspace{-0.25cm}
  \caption{In-domain evaluation of the differences in rankings compared to not using any noise.}
  \label{fig:in_domain}
\end{figure}

In Figure~\ref{fig:in_domain}, we show the in-domain (ID) performance of NNs trained with various noise injection methods across CV classification and regression, tabular data classification and regression, and NLP classification.
Overall, we observe that the noise injection methods significantly improve the generalisation and calibration in many cases, but different noise types are needed for various tasks.
In CV classification, almost all noises improve the error rate, with many simultaneously improving calibration. 
The most beneficial noises are AugMix, weak augmentation and Dropout. 
MixUp and label smoothing are a surprise to a certain extent as they improved generalisation but not calibration. 
In CV regression, improving generalisation was challenging, with no improvement. 
However, several noises have improved NLL, with AugMix, weak augmentation, and Dropout achieving the best balance. 
These results suggest that image augmentation broadly benefits CV, confirming expectations.

Several noises have improved the error rate to a lesser extent or kept it at a similar level in tabular data classification. 
In contrast, almost all noises have improved ECE and NLL. 
The improvements were particularly impactful in several cases, with model noise, label smoothing, and Dropout being the best. 
While ODS is designed to improve adversarial robustness, it improved ECE and NLL while slightly improving error rates. 
All noises improve NLL for tabular regression, and some significantly improve MSE.
Gaussian noises applied to the weights, activations, or inputs are the most useful types of noise for improving the two metrics.
In NLP classification, about half of the noises improve error, with some also improving calibration simultaneously. 
The best noises are Dropout, label smoothing and ODS, which differs from what was the best for CV. 
These noises significantly lowered error and NLL, while MixUp and model noise were particularly useful for reducing ECE. 
ODS was beneficial for improving error and calibration via NLL, which can be a surprise as this technique was not previously considered for improving generalisation or calibration.

\begin{figure}[t]
  \centering
  \begin{subfigure}{0.48\textwidth}
    \includegraphics[width=\textwidth]{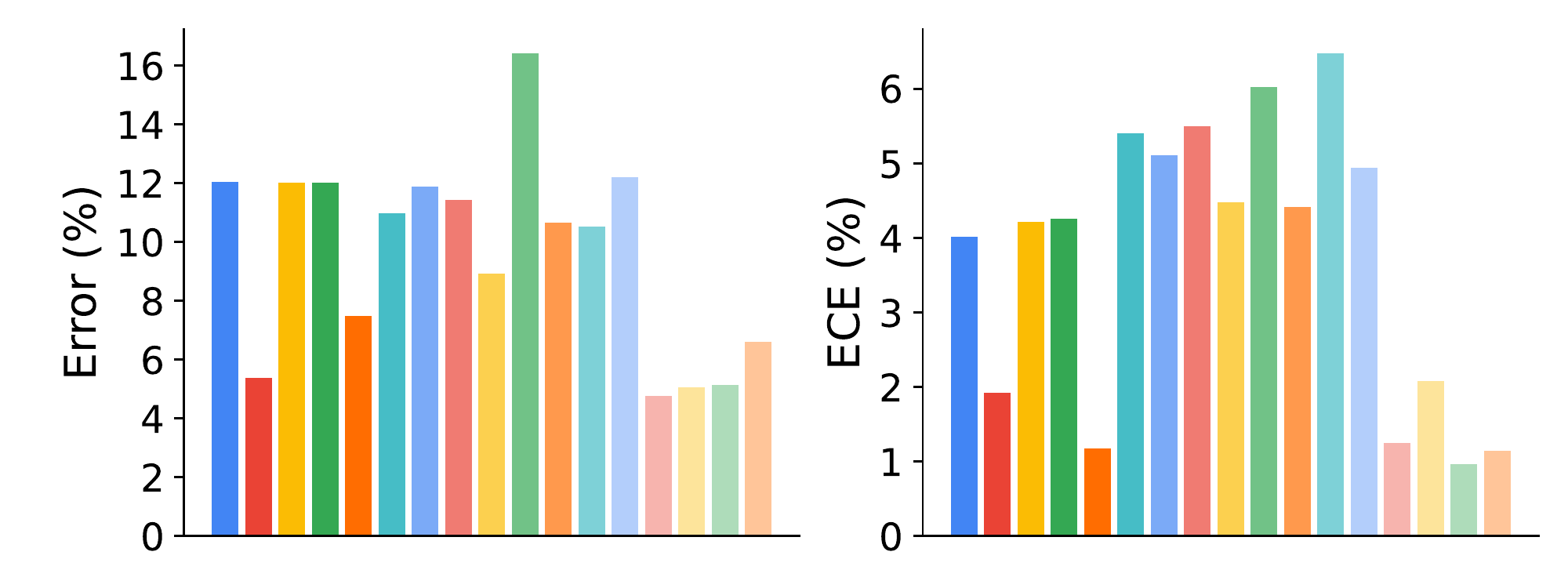}
    \vspace{-0.70cm}
    \caption{CIFAR-10 CV classification.}
    \label{fig:in_domain:cv_classification:cifar10}
  \end{subfigure}
  \hfill
  \begin{subfigure}{0.48\textwidth}
    \includegraphics[width=\textwidth]{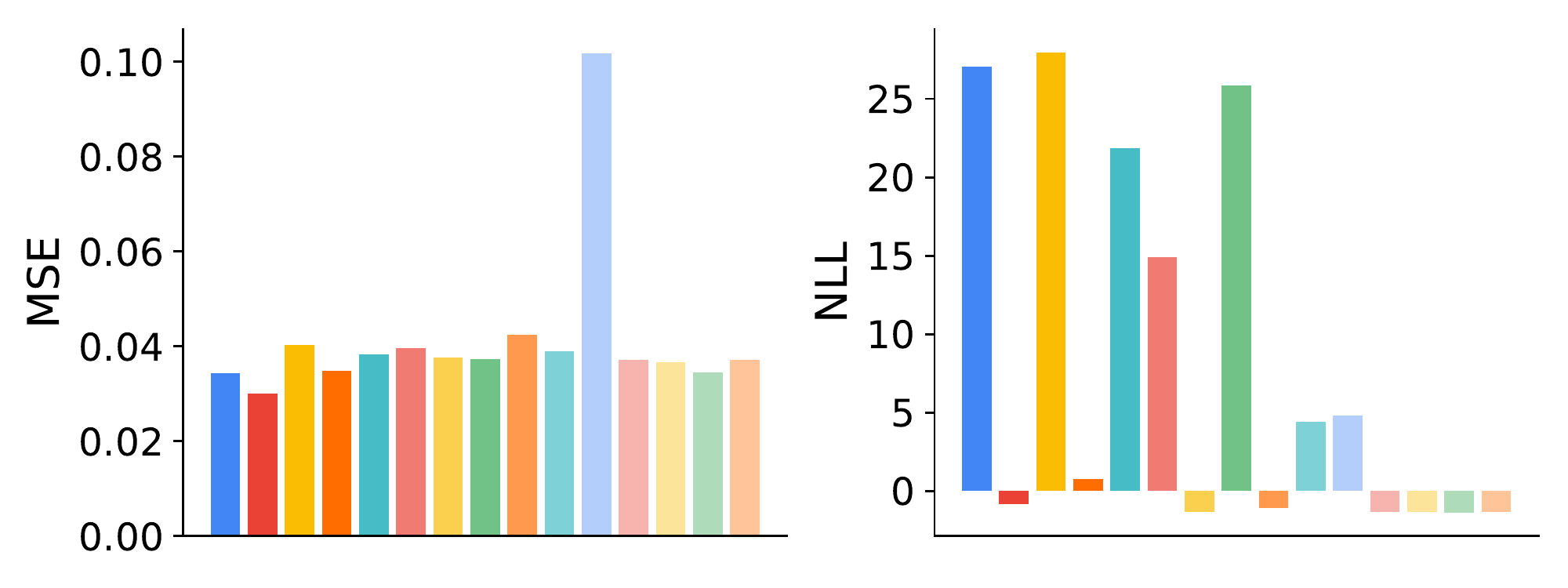}
    \vspace{-0.70cm}
    \caption{WikiFace CV regression.}
    \label{fig:in_domain:cv_regression:wiki_face}
  \end{subfigure}
  \begin{subfigure}{0.48\textwidth}
    \includegraphics[width=\textwidth]{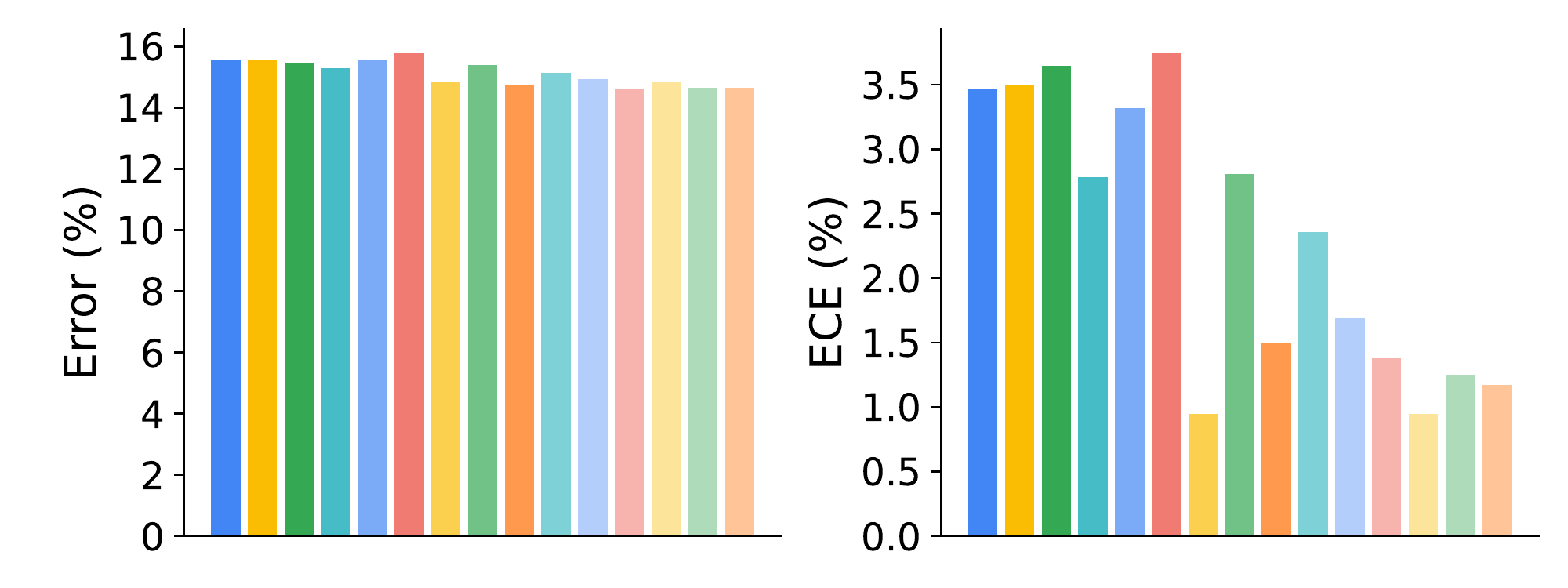}
    \vspace{-0.70cm}
    \caption{Adult tabular data classification.}
    \label{fig:in_domain:tab_classification:adult}
  \end{subfigure}
  \hfill
  \begin{subfigure}{0.48\textwidth}
    \includegraphics[width=\textwidth]{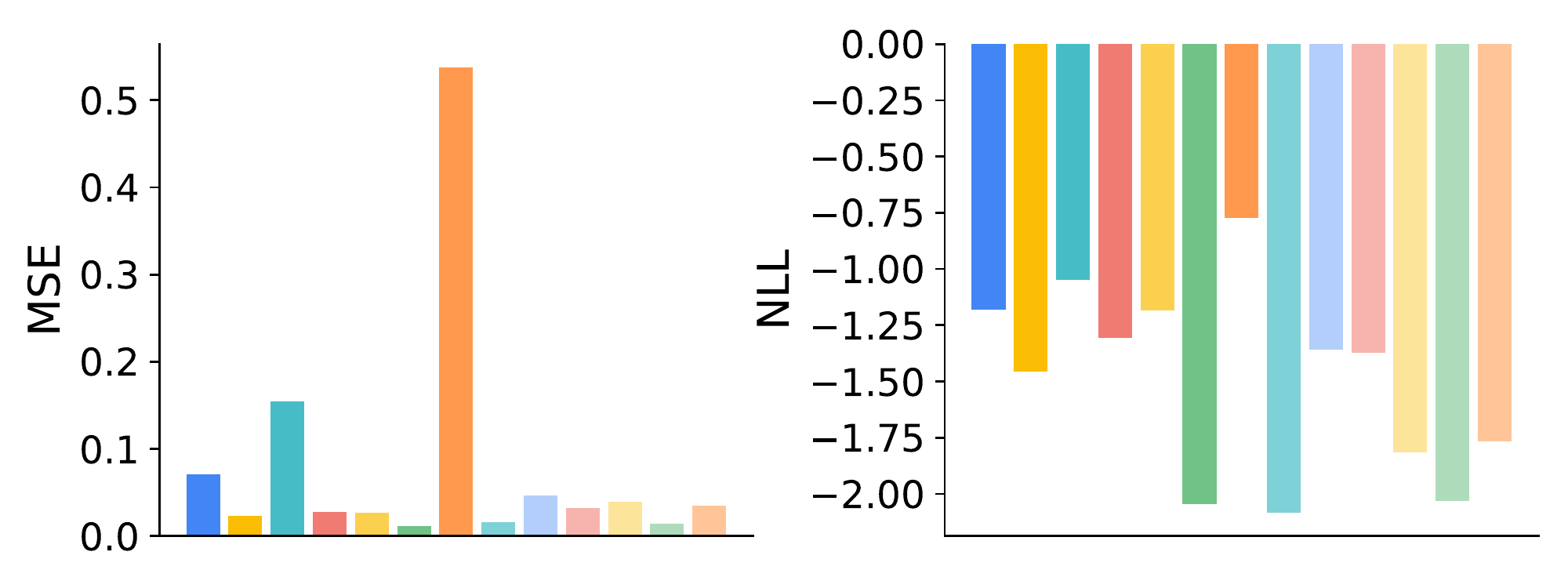}
    \vspace{-0.70cm}
    \caption{Yacht tabular data regression.}
    \label{fig:in_domain:tabular_regression:yacht}
  \end{subfigure}
  \hfill
  \begin{subfigure}{0.48\textwidth}
    \includegraphics[width=\textwidth]{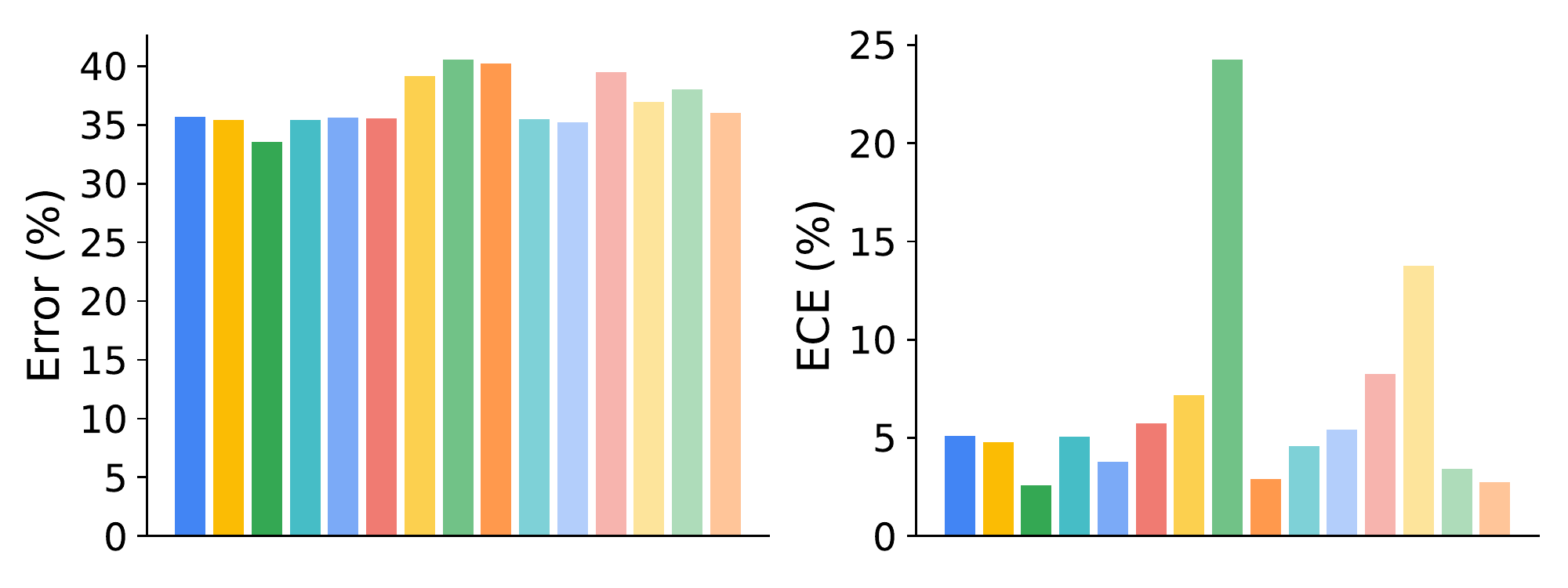}
    \vspace{-0.70cm}
    \caption{NewsGroup NLP classification.}
    \label{fig:in_domain:nlp_classification:newsgroup}
  \end{subfigure}
  \hfill
  \begin{subfigure}{\textwidth}
    \includegraphics[width=\textwidth]{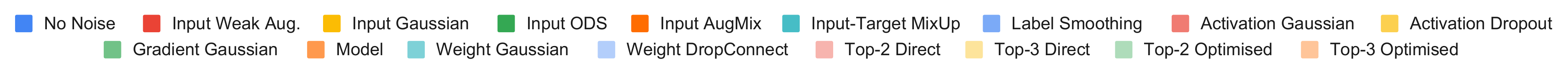}
  \end{subfigure}
  \vspace{-0.65cm}
  \caption{Detailed in-domain performance of NNs trained with various noises across the five tasks.}
  \label{fig:in_domain:cifar_10:wiki_face:adult:yacht:newsgroup}
\end{figure}

In Figure~\ref{fig:in_domain:cifar_10:wiki_face:adult:yacht:newsgroup}, we show detailed results for selecting representative datasets across the 5 tasks.
We see the improvements in error can be large for CIFAR-10, for example, halving it in some of the best cases -- weak augmentation and AugMix, with Dropout also leading to a few percentage point improvements. 
The situation is similar for ECE, where weak augmentation and AugMix make the ECE one-half or one-third. 
Many errors are slightly better, but certain noises, such as MixUp, label smoothing, or Gaussian noise added to the activations, worsen the calibration. 
For WikiFace, there are more minor improvements in error from weak augmentation and AugMix with overall similar MSE across different noises. 
Still, the differences in calibration as measured using NLL can be considerable, with most noises improving the NLL significantly.

Moving the focus to tabular data, most noises applied to the Adult classification dataset improve the error marginally. 
In contrast, many improve ECE significantly, with the best ones being Dropout, model noise and DropConnect. 
Most noises have significantly improved MSE for the Yacht regression dataset, but CMixUp and model noise led to significant increases. 
The best ones have been gradient Gaussian and Gaussian noise added to the weights. 
NLL has been improved in several cases, including gradient Gaussian and weight Gaussian, demonstrating solid improvements in MSE and NLL. 
The errors stay similar for NLP classification on NewsGroup using the global pooling CNN model. 
ODS leads to the best improvement, while several noises, specifically Dropout, gradient Gaussian, and model noise, lead to worse generalisation. 
ODS and label smoothing have also noticeably improved ECE.

\textbf{Main Observations:} The noises are effective across various tasks and datasets. 
The shortlist of the most effective methods is AugMix and weak augmentation in CV, model noise, and Gaussian noise added to weights for tabular data and dropout in NLP.
Different task types benefit from different types of noise.

\subsection{Out-of-Domain Evaluation}\label{sec:experiments:out-of-domain}

\begin{figure}[t]
  \centering
  \begin{subfigure}{0.48\textwidth}
   \includegraphics[width=1.0\textwidth]{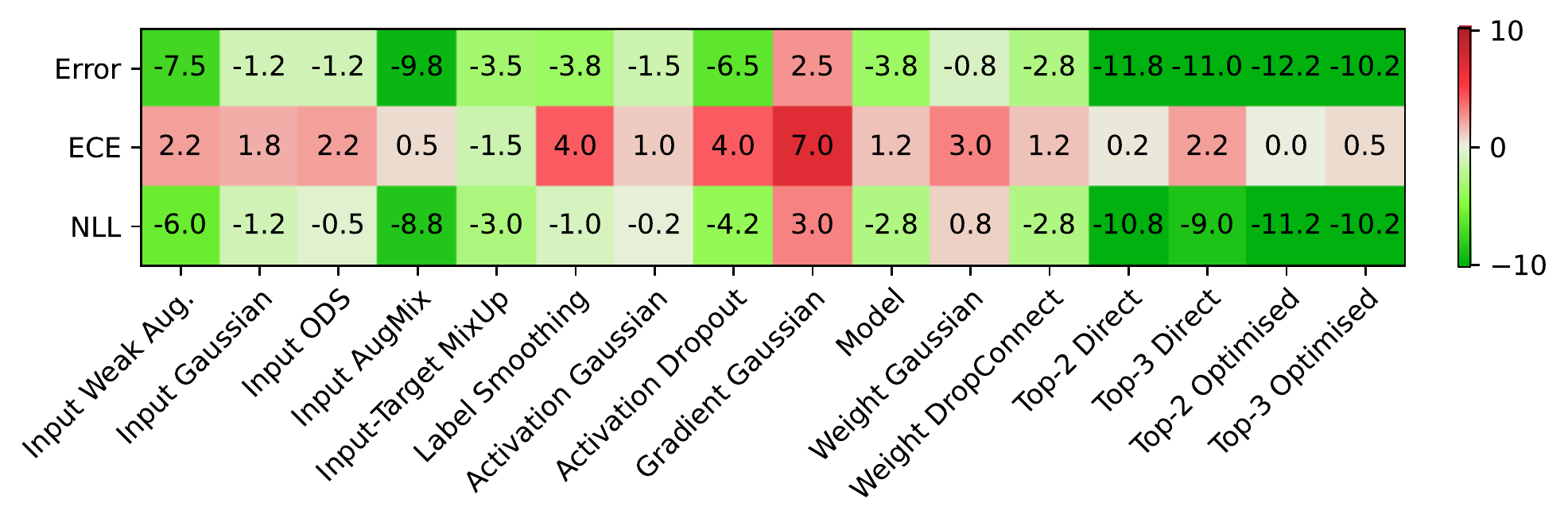}
   \vspace{-0.70cm}
   \caption{CV classification.}
   \label{fig:out_domain:cv_classification}
   \end{subfigure}
   \hfill
   \begin{subfigure}{0.48\textwidth}
   \includegraphics[width=1.0\textwidth]{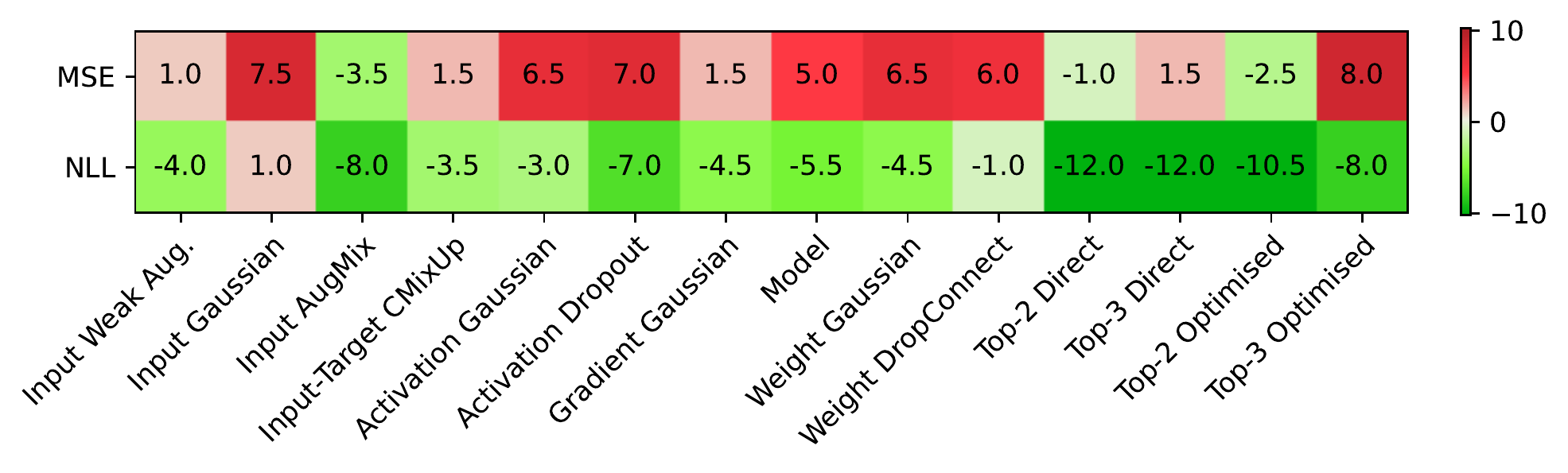}
   \vspace{-0.70cm}
   \caption{CV regression.}
   \label{fig:out_domain:cv_regression}
   \end{subfigure}
   \begin{subfigure}{0.48\textwidth}
   \includegraphics[width=1.0\textwidth]{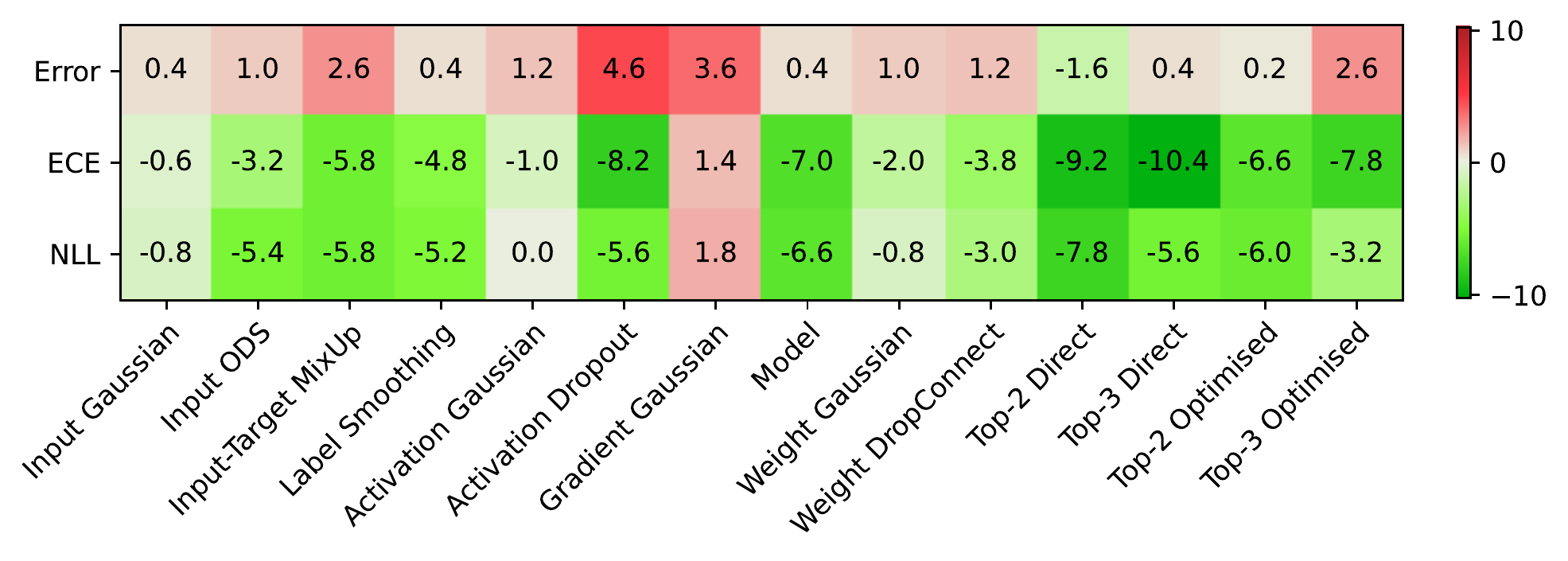}
   \vspace{-0.70cm}
   \caption{Tabular classification.}
   \label{fig:out_domain:tab_classification}
   \end{subfigure}
   \hfill
   \begin{subfigure}{0.48\textwidth}
   \includegraphics[width=1.0\textwidth]{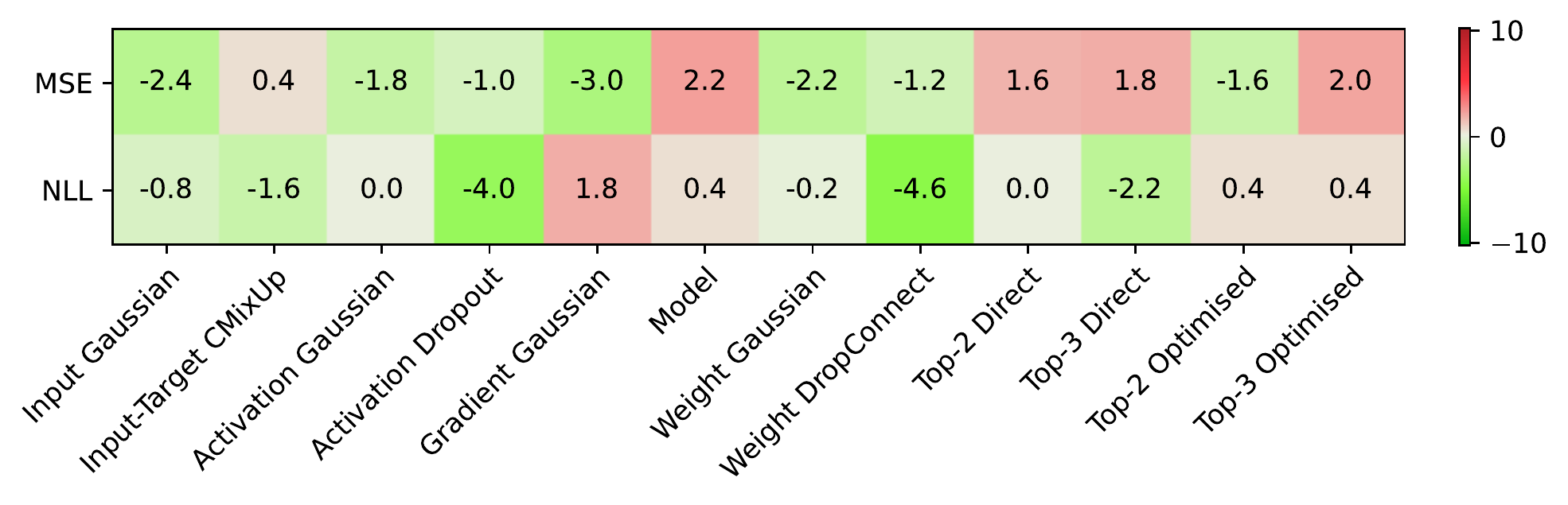}
   \vspace{-0.70cm}
   \caption{Tabular regression.}
   \label{fig:out_domain:tab_regression}
   \end{subfigure}
   \caption{OOD evaluation of the differences in rankings compared to not using any noise.}
   \label{fig:out_domain}
\end{figure}

\begin{table}
\begin{subtable}{0.48\textwidth}
\centering
\begin{sc}
\begin{adjustbox}{max width=\linewidth}
\begin{tabular}{lccccc}
\toprule
\textbf{Metric} & \textbf{SVHN} & \textbf{CIFAR-10} & \textbf{CIFAR-100} & \textbf{TinyImageNet} & \textbf{Average}  \\
\midrule
Error & 0.912\tiny{$\pm$ 0.000} & 0.706\tiny{$\pm$ 0.000} & 0.897\tiny{$\pm$ 0.000} & 0.912\tiny{$\pm$ 0.000} & 0.857 \\
ECE & 1.000\tiny{$\pm$ 0.000} & 0.618\tiny{$\pm$ 0.000} & 0.250\tiny{$\pm$ 0.177} & 0.118\tiny{$\pm$ 0.542} & 0.496 \\
NLL & 0.971\tiny{$\pm$ 0.000} & 0.515\tiny{$\pm$ 0.003} & 0.868\tiny{$\pm$ 0.000} & 0.882\tiny{$\pm$ 0.000} & 0.809 \\
\bottomrule
\end{tabular}
\end{adjustbox}
\end{sc}
\caption{CV classification.}
\label{tab:cv_classification_iid_ood_kendall_tau}
\end{subtable}
\begin{subtable}{0.48\textwidth}
\centering
\begin{sc}
\begin{adjustbox}{max width=0.8\linewidth}
\begin{tabular}{lccc}
\toprule
\textbf{Metric} & \textbf{Rotated CIFAR-100} & \textbf{WikiFace} & \textbf{Average} \\
\midrule
MSE & 0.257\tiny{$\pm$ 0.202} & 0.581\tiny{$\pm$ 0.002} & 0.419 \\
NLL & -0.238\tiny{$\pm$ 0.239} & 0.810\tiny{$\pm$ 0.000} & 0.286 \\
\bottomrule
\end{tabular}
\end{adjustbox}
\end{sc}
\caption{CV regression.}
\label{tab:cv_regression_iid_ood_kendall_tau}
\vspace{0.4em}
\end{subtable}
\hspace{\fill}
\begin{subtable}{0.48\textwidth}
\centering

\begin{sc}
\begin{adjustbox}{max width=\linewidth}
\begin{tabular}{lcccccc}
\toprule
\textbf{Metric} & \textbf{Wine} & \textbf{Toxicity} & \textbf{Abalone} & \textbf{Students}  & \textbf{Adult} & \textbf{Average}  \\
\midrule
Error & 0.861\tiny{$\pm$ 0.000} & 0.743\tiny{$\pm$ 0.000} & 0.529\tiny{$\pm$ 0.006} & 0.532\tiny{$\pm$ 0.007} & 0.924\tiny{$\pm$ 0.000} & 0.718 \\
ECE & 0.905\tiny{$\pm$ 0.000} & 0.905\tiny{$\pm$ 0.000} & 0.867\tiny{$\pm$ 0.000} & 0.867\tiny{$\pm$ 0.000} & 0.962\tiny{$\pm$ 0.000} & 0.901 \\
NLL & 0.924\tiny{$\pm$ 0.000} & 0.981\tiny{$\pm$ 0.000} & 0.790\tiny{$\pm$ 0.000} & 0.810\tiny{$\pm$ 0.000} & 0.962\tiny{$\pm$ 0.000} & 0.893 \\
\bottomrule
\end{tabular}
\end{adjustbox}
\end{sc}
\caption{Tabular data classification.}
\label{tab:tab_classification_iid_ood_kendall_tau}
\end{subtable}
\begin{subtable}{0.48\textwidth}
\centering
\begin{sc}
\begin{adjustbox}{max width=\linewidth}
\begin{tabular}{lcccccc}
\toprule
\textbf{Metric} & \textbf{Energy} & \textbf{Boston} & \textbf{Wine} & \textbf{Yacht}  & \textbf{Concrete} & \textbf{Average}  \\
\midrule
MSE & 0.667\tiny{$\pm$ 0.001} & 0.923\tiny{$\pm$ 0.000} & -0.564\tiny{$\pm$ 0.007} & 0.641\tiny{$\pm$ 0.002} & 0.872\tiny{$\pm$ 0.000} & 0.508 \\
NLL & -0.615\tiny{$\pm$ 0.003} & 0.974\tiny{$\pm$ 0.000} & 0.154\tiny{$\pm$ 0.510} & 0.590\tiny{$\pm$ 0.004} & 0.436\tiny{$\pm$ 0.042} & 0.308 \\
\bottomrule
\end{tabular}
\end{adjustbox}
\end{sc}
\caption{Tabular data regression.}
\label{tab:tab_regression_iid_ood_kendall_tau}
\end{subtable}
\caption{Kendall Tau correlation between ID and OOD rankings of different noise types for various tasks.}
\label{tab:iid_ood_kendall_tau}
\end{table}

We evaluate the performance on the ID test set and an augmented OOD set, including an average over visual corruptions across 19 categories and 5 severities~\citep{hendrycks2019benchmarking}.
Likewise, we average the performance across 5 categories and 5 severities for tabular data.
The summary of the results is in Figure~\ref{fig:out_domain}, with analysis of correlations between ID and OOD rankings via Kendall Tau score in Table~\ref{tab:iid_ood_kendall_tau}.
For CV classification, we observe that the generalisation improvements also remain for OOD, but improving calibration in terms of ECE turns out to be much more challenging. 
The overall ranking of the best noises remains similar, with AugMix and weak augmentation remaining the best. 
MixUp rose to prominence thanks to the best OOD calibration and improved errors and NLL. 
Analysis of Kendall Tau correlation in Table~\ref{tab:cv_classification_iid_ood_kendall_tau} shows that ID and OOD rankings are strongly correlated for error and NLL, while only moderately for ECE.
CV regression is similar to classification ranking the best noises, with only AugMix leading to improvements in OOD generalisation. 
However, calibration is improved by most noises, with AugMix excelling. 
Only a minor correlation exists between ID and OOD rankings for MSE and NLL metrics.
For tabular classification, the noises generally improve ECE and NLL but not the error rate under OOD settings, with model noise, label smoothing, and Dropout being the best. 
This suggests all of these are among the best noises for both ID and OOD. 
ID and OOD rankings show a strong correlation overall.
Several noises improve OOD generalisation and calibration for tabular regression, with DropConnect, Dropout and Gaussian noise added to the inputs, leading to the best overall improvements. The ID and OOD ranking Kendall Tau correlation is low in this case. MixUp and CMixUp have improved OOD calibration for both tabular classification and regression.

\begin{figure}[t]
  \centering
  \begin{subfigure}{0.48\textwidth}
    \includegraphics[width=\textwidth]{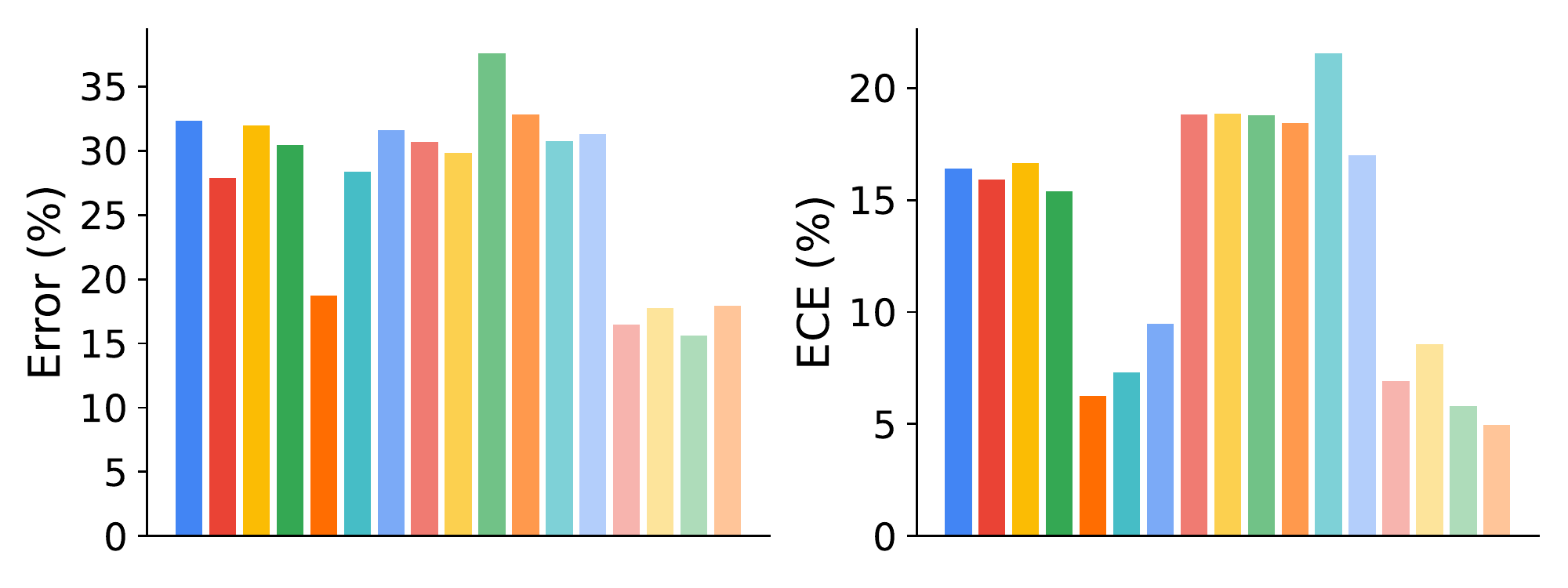}
    \vspace{-0.70cm}
    \caption{CIFAR-10 CV classification.}
    \label{fig:out_domain:cv_classification:cifar_10}
  \end{subfigure}
  \begin{subfigure}{0.48\textwidth}
    \includegraphics[width=\textwidth]{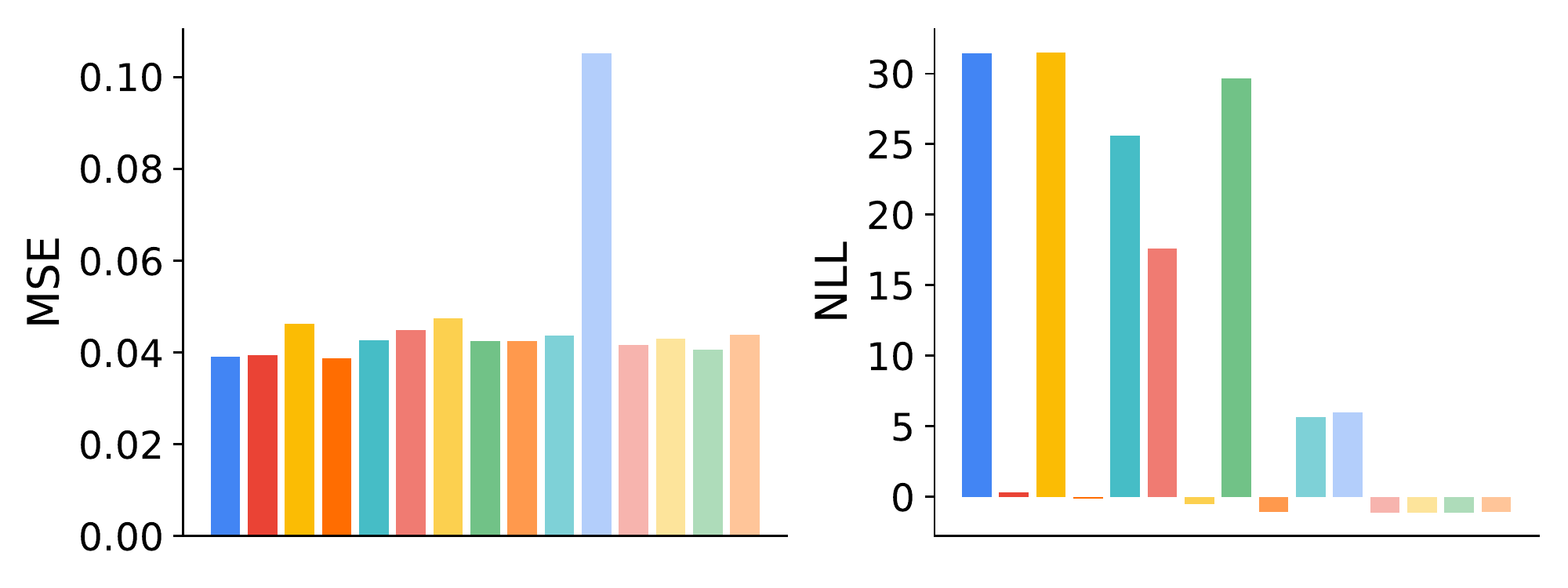}
    \vspace{-0.70cm}
    \caption{WikiFace CV regression.}
    \label{fig:out_domain:cv_regression:wiki_face}
  \end{subfigure}
  \hfill
  \begin{subfigure}{0.48\textwidth}
    \includegraphics[width=\textwidth]{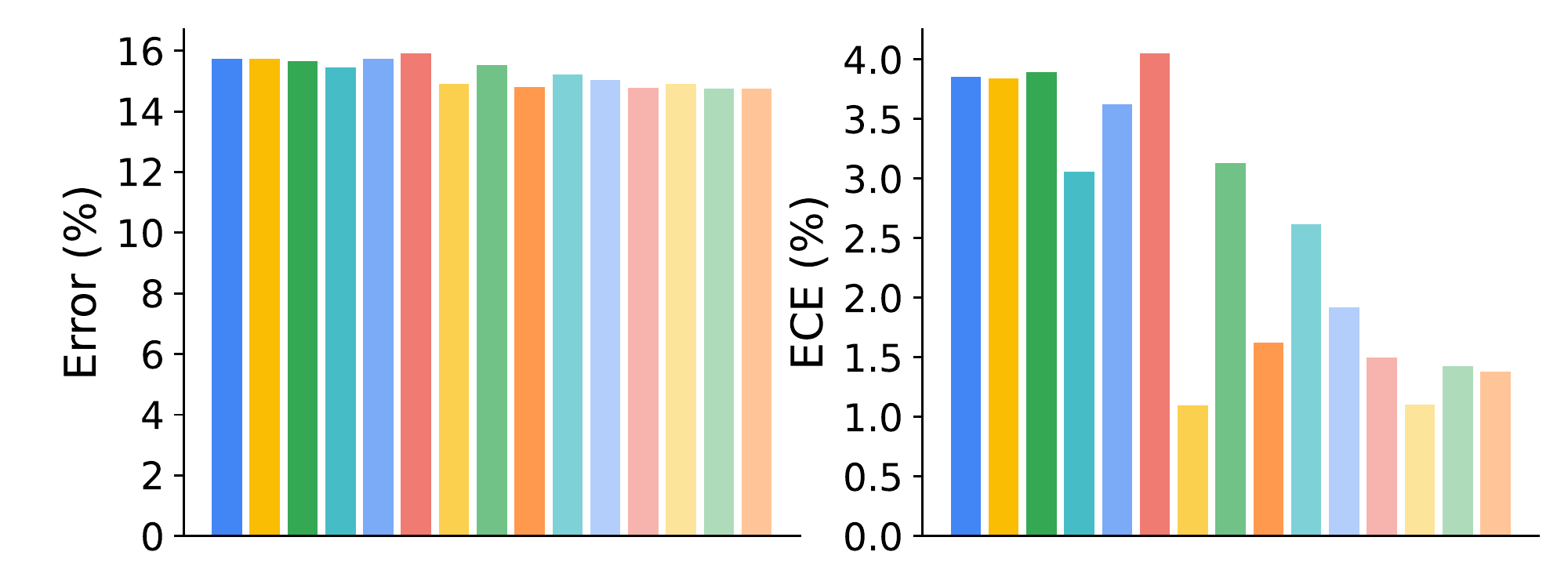}
    \vspace{-0.70cm}
    \caption{Adult tabular data classification.}
    \label{fig:out_domain:tab_classification:adult}
  \end{subfigure}
  \begin{subfigure}{0.48\textwidth}
    \includegraphics[width=\textwidth]{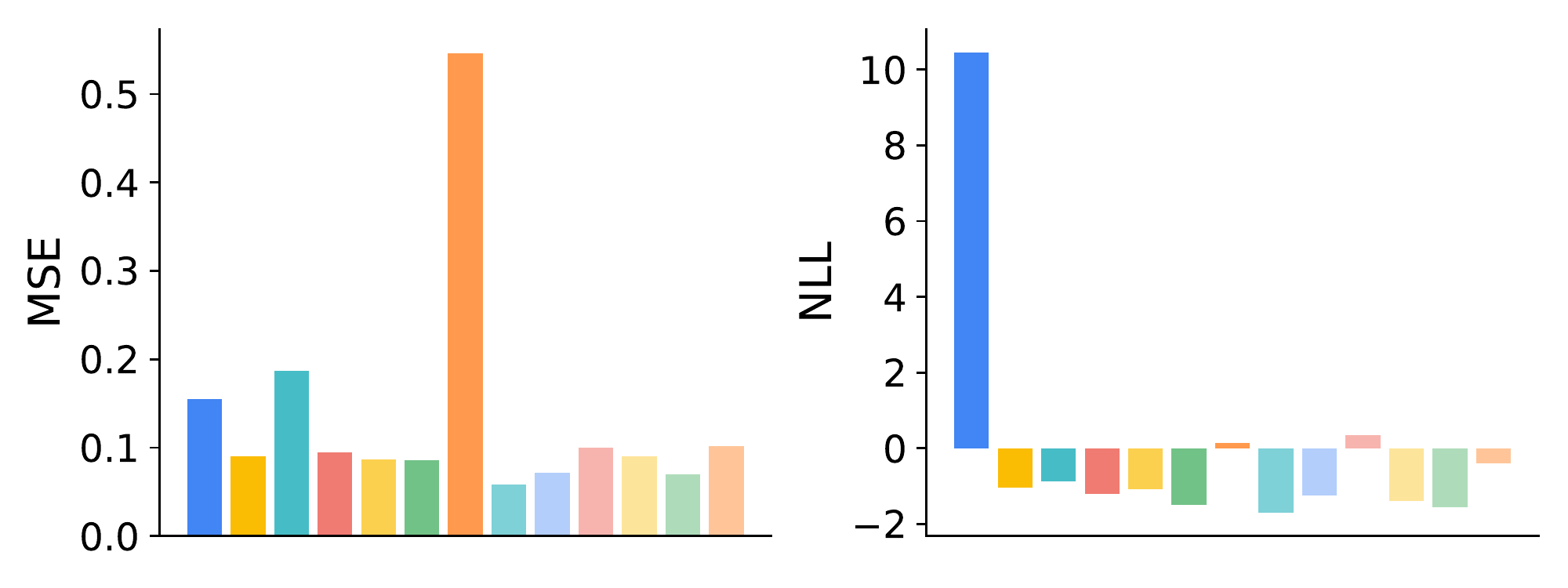}
    \vspace{-0.70cm}
    \caption{Yacht tabular data regression.}
    \label{fig:out_domain:tab_regression:yacht}
  \end{subfigure}
  \begin{subfigure}{\textwidth}
    \includegraphics[width=\textwidth]{tmlr_analysis_images/legend.pdf}
  \end{subfigure}
  \vspace{-0.65cm}
  \caption{Detailed OOD performance of NNs trained with various noises across the four tasks.}
  \label{fig:out_domain:cifar_10:wiki_face:adult:yacht}
\end{figure}

We study selected representative datasets regarding OOD performance in Figure~\ref{fig:out_domain:cifar_10:wiki_face:adult:yacht}.
OOD results on CIFAR-10 show that AugMix significantly improves both error and ECE, making ECE one-third of the no noise equivalent. 
MixUp leads to similarly considerable improvements in ECE and more minor yet significant improvements in error. 
Several noises, e.g., Dropout and Gaussian noise, added to activations or weights lead to a few percentages worse ECE.
On WikiFace, most OOD MSE values are similar, but OOD calibration in NLL is improved significantly for several noises, including AugMix, weak augmentation or Dropout. 
Improvements in generalisation for the Adult tabular classification dataset are minor, but the improvements in calibration can be significant, for example, Dropout and model noise halving the OOD ECE value. For the Yacht tabular regression dataset, the improvements in generalisation have been more critical, with the same being true for calibration measured in terms of OOD NLL.

We include an additional OOD investigation on TinyImageNet, where we study the performance on black and white sketches. We use the same classes as in TinyImageNet~\citep{wang2019learning} and report the full details and results in the Appendix. 
The shift from natural images to sketches represents a more significant domain shift than our standard OOD shifts~\citep{hendrycks2019benchmarking}, and hence also tests the generalisation of the noises to a larger degree. 
We do this to disambiguate the performance of AugMix, which uses augmentations that may be similar, but are still distinct~\citep{hendrycks2019augmix}, to the ones used for our OOD evaluation. 
The results clearly show that AugMix obtains similar rankings for all three metrics on both the synthetic OOD and sketch domain evaluations. 
Overall we see strong Kendall Tau correlation in terms of the error across all noises, but smaller in terms of ECE and NLL.

\textbf{Main Observations:}
We see consistent improvements in OOD generalisation and calibration for tabular data. 
Errors and NLL are improved for CV classification, but calibration is generally not improved when measured via ECE. 
CV regression usually sees improvements in OOD NLL only. 
The best ID noise types have often remained the best OOD, but overall, the correlations between ID and OOD rankings were not high in all cases. MixUp, or CMixUp for regression, showed surprising behaviour as it was much more helpful for improving OOD calibration than ID calibration.

\subsection{Combination of Noises}\label{sec:experiments:combination}
Next, we evaluate the combination of noises.
We construct them from empirical combinations of the Top 2 or 3 noises from the ID evaluation for each task, based on average rank across respective datasets and metrics.
We consider two cases: \textit{1.)} the found hyperparameters of the noises are directly applied, and \textit{2.)} the hyperparameters of the noises are jointly tuned.
We utilise the same 50-trial budget to tune the selected noises jointly.
We restricted our experiments to combinations of two or three noises, as we empirically found that tuning all the noises jointly in our study is not feasible even with a larger computational budget.
We consider the individual noise hyperparameter tuning and then the direct application or joint tuning of their combination to be a realistic scenario where a practitioner has a limited compute budget and wants to improve their model's performance~\citep{tuningplaybookgithub} iteratively.

The results are already in Figures~\ref{fig:in_domain},~\ref{fig:in_domain:cifar_10:wiki_face:adult:yacht:newsgroup},~\ref{fig:out_domain} and~\ref{fig:out_domain:cifar_10:wiki_face:adult:yacht} and denoted as Top-2 Direct, Top-3 Direct for \textit{1.)}, Top-2 Optimised and Top-3 Optimised for \textit{2.)}.
The combinations for Top-2 and Top-3 are in Table~\ref{tab:combinations}. 
To simplify the analysis of how effective the different combinations of noises are, we compute their average rank improvement compared to no noise and report it in Table~\ref{tab:combinations_rank}.
Notice that when we choose a combination of noises to involve noises from the same category, for example, ODS and input Gaussian are both input noises, these are applied sequentially.
\begin{table}[ht]
\centering
\scalebox{0.9}{
\begin{tabular}{cll}
\toprule
\textbf{Task} & \textbf{Top-2} & \textbf{Third Method} \\
\midrule
CV classification & \underline{Input AugMix}, \underline{Input Weak Augmentation}  & Activation Dropout \\
NLP classification & Activation Dropout, Target Label Smoothing & Input ODS \\
Tabular classification & Model, Target Label Smoothing & Activation Dropout \\
CV regression & \underline{Input AugMix}, \underline{Input Weak Augmentation}  & Activation Dropout \\
Tabular regression & Weight Gaussian, Activation Gaussian & Input Gaussian \\
\bottomrule
\end{tabular}}
\caption{Top task and noise combinations. \underline{Underlined} methods are from the same type.}
\label{tab:combinations}
\end{table}
\begin{table}[ht]
\centering
\begin{sc}
\begin{adjustbox}{max width=0.7\linewidth}
\begin{tabular}{lcccc}
\toprule
\textbf{Scenario} & \textbf{Top-2 Direct} & \textbf{Top-3 Direct} & \textbf{Top-2 Optimised} & \textbf{Top-3 Optimised}  \\
\midrule
ID & -4.75 & -3.69 & -4.34 & -3.30 \\
OOD & -5.24 & -4.43 & -5.00 & -2.59 \\
\bottomrule
\end{tabular}
\end{adjustbox}
\end{sc}
\caption{Average rank improvement over no noise for the different combination strategies.}
\label{tab:combinations_rank}
\end{table}

We can draw several observations from Table~\ref{tab:combinations_rank}. 
\textit{1.)} Directly combining hyperparameters for the top two or three noises is a good strategy when considering the same budget for hyperparameter tuning as for one noise. A significantly larger budget is likely needed for jointly optimising hyperparameters of multiple noises.
\textit{2.)} A combination of two noises performs better than a combination of three noises, suggesting there may be negative interactions when too many noise sources are used without extensive hyperparameter tuning.
\textit{3.)} The behaviour of different combination strategies is consistent across ID and OOD settings.


Commenting on the overall performance of the combinations of noises, the combinations are typically better for classification tasks than the individual noises. 
Still, the opposite may be true for regression.
As observed in Figures~\ref{fig:in_domain:cv_classification},~\ref{fig:in_domain:tab_classification} and \ref{fig:in_domain:nlp_classification}, the combinations are consistently ranked lowest in comparison to using no noise for classification, showing the effectiveness of the combinations. 
However, Figures~\ref{fig:in_domain:cv_regression} and \ref{fig:in_domain:tabular_regression} show that regression can benefit from only using one noise at a time. 
OOD analysis in Figure~\ref{fig:out_domain} confirms the benefits of combinations of noises for classification tasks, and it also shows that it can be beneficial for regression, contrary to the ID behaviour. 
The combinations are generally ranked lower and can improve calibration and generalisation, as seen in lower MSE, NLL, or error and ECE simultaneously.

\textbf{Main Observations:} 
The combination of noises can improve both calibration and generalisation simultaneously.
Directly combining two noises is better than three, as too many can lead to conflicts. Combining noises directly with their hyperparameters is generally reasonable and a significantly larger tuning budget for optimising hyperparameters would be needed for optimising their hyperparameters jointly.

\subsection{Transferability of Hyperparameters Across Datasets and Models}\label{sec:experiments:transferability}

Furthermore, we evaluate the transferability of the hyperparameters across datasets and models.
We consider two cases: the transfer of hyperparameters to a new dataset and the transfer of hyperparameters to a new architecture.
For the dataset transfer, we consider the following combinations: SVHN to CIFAR-10, CIFAR-10 to CIFAR-100, CIFAR-100 to TinyImageNet, and 3 tabular regression datasets combinations, Concrete to Energy, Boston to Wine, Yacht to Concrete.
We consider the following combinations for the architecture transfer: FC to ResNet-18 for SVHN and ResNet-18 to ResNet-34 for CIFAR-10, CIFAR-100 and TinyImageNet. 
We use a NN with an additional layer for tabular data, i.e., five layers instead of four.

\begin{figure}[t]
  \centering
    \begin{subfigure}{0.48\textwidth}
    \includegraphics[width=1.0\textwidth]{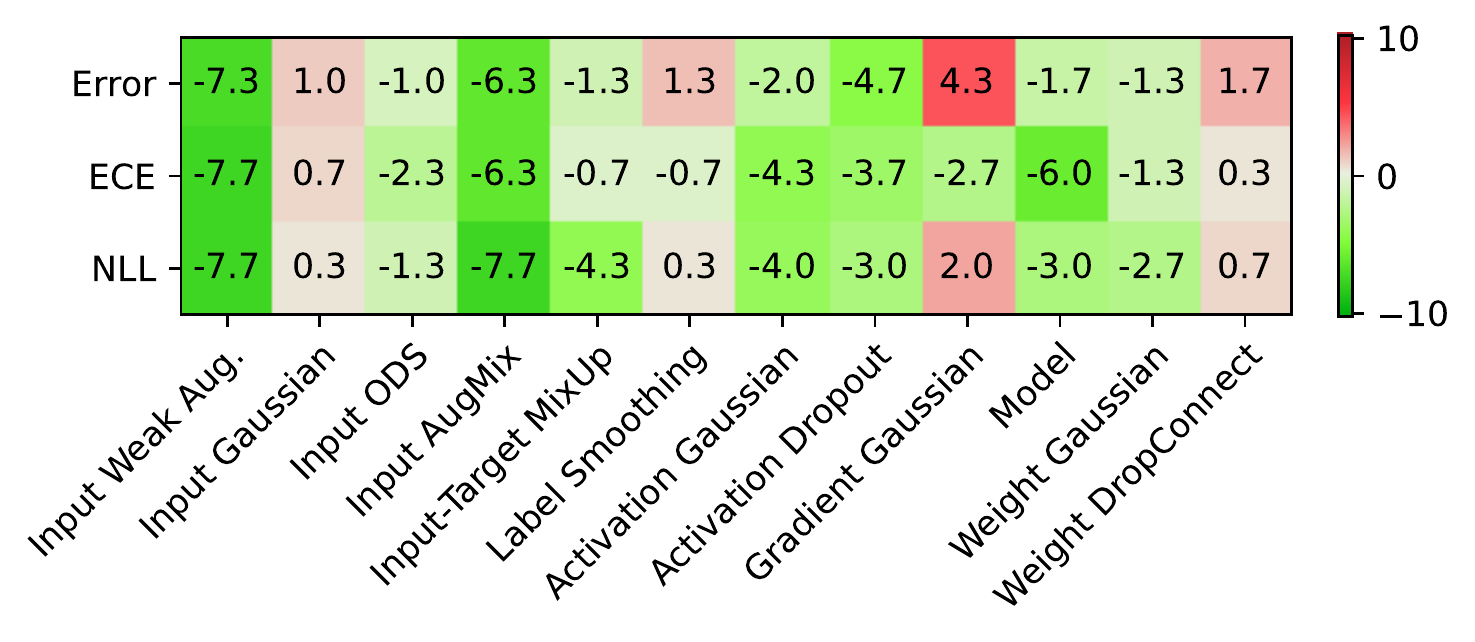}
    \vspace{-0.7cm}
    \caption{ID CV classification in dataset transfer.}
    \label{fig:transfer:dataset:id:cv_classification}
    \end{subfigure}
    \hfill
    \begin{subfigure}{0.48\textwidth}
    \includegraphics[width=1.0\textwidth]{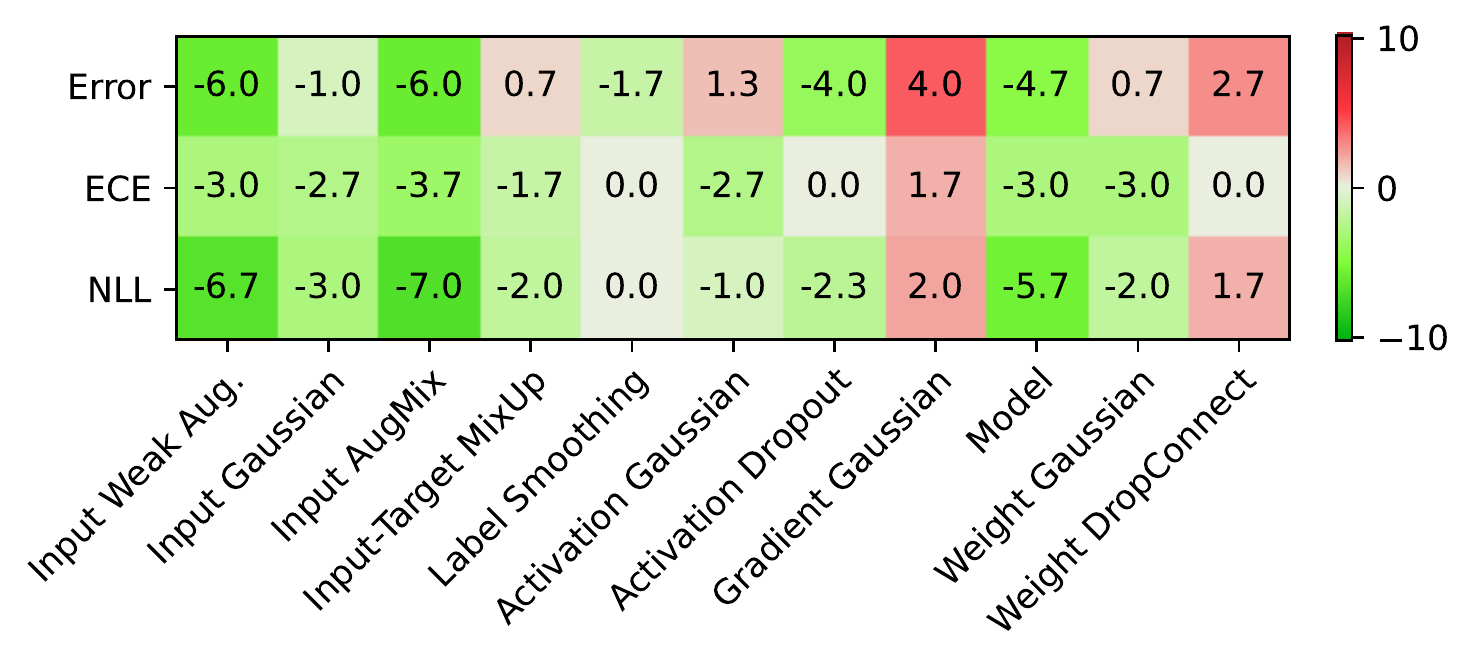}
    \vspace{-0.7cm}
    \caption{ID CV classification in architecture transfer.}
    \label{fig:transfer:architecture:id:cv_classification}
    \end{subfigure}
   \begin{subfigure}{0.48\textwidth}
   \includegraphics[width=1.0\textwidth]{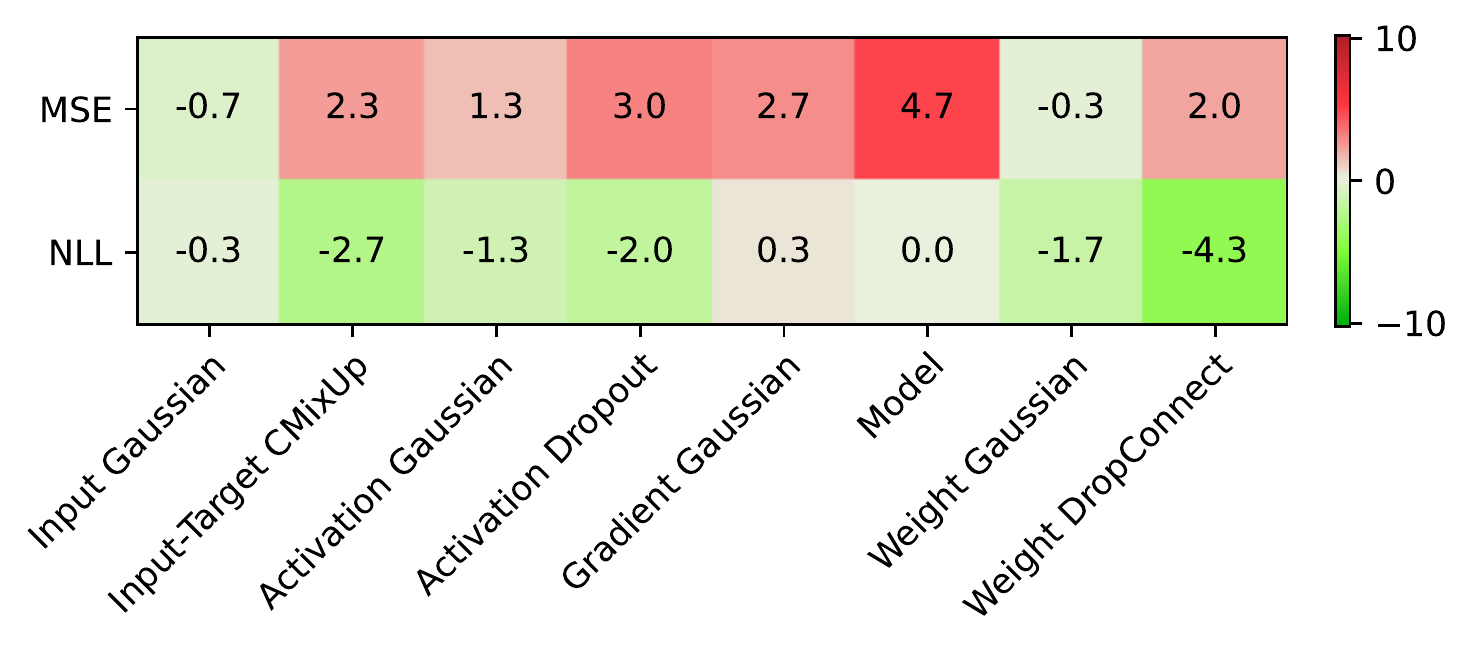}
    \vspace{-0.7cm}
    \caption{ID tabular regression in dataset transfer.}
    \label{fig:transfer:dataset:id:tabular_regression}
   \end{subfigure}
    \hfill
   \begin{subfigure}{0.48\textwidth}
   \includegraphics[width=1.0\textwidth]{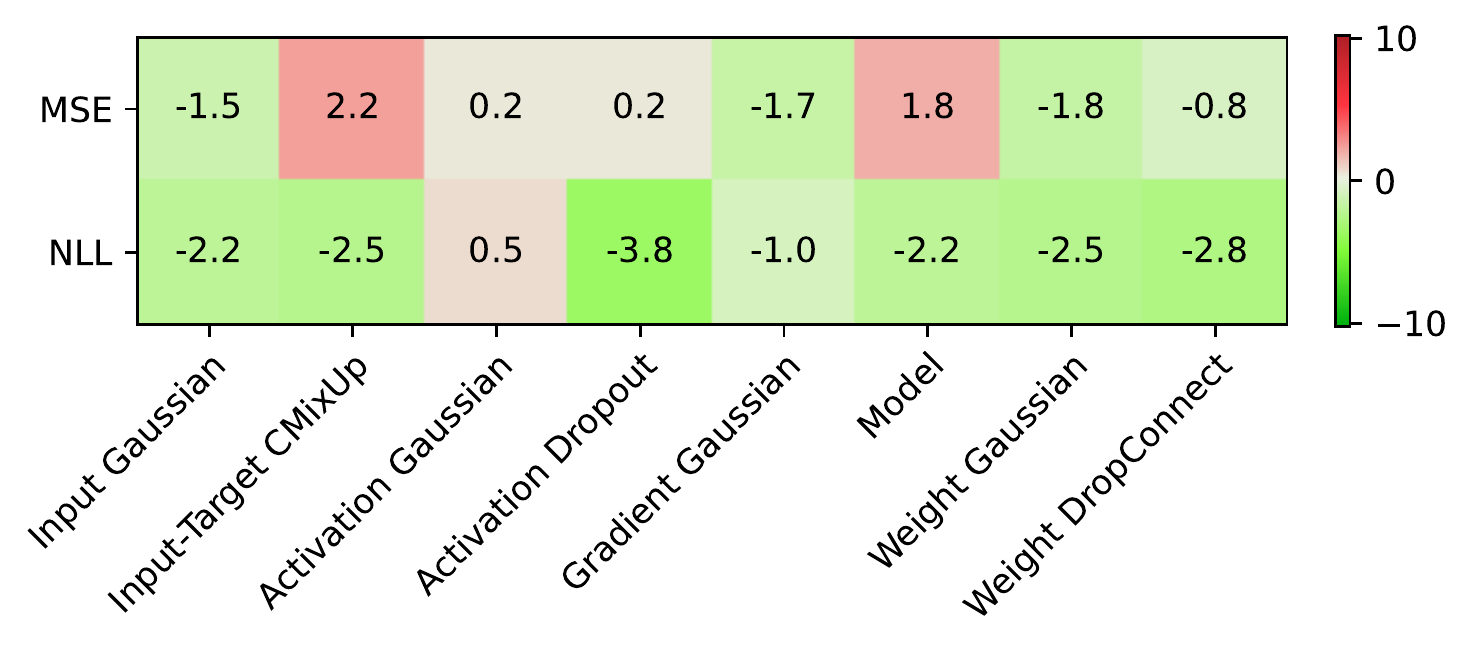}
    \vspace{-0.7cm}
    \caption{ID tabular regression in architecture transfer.}
    \label{fig:transfer:architecture:id:tabular_regression}
   \end{subfigure}
    \caption{Transfer of hyperparameters on in-domain (ID) data.}
   \label{fig:transfer}
\end{figure}

\begin{figure}[ht!]
  \centering
   \begin{subfigure}{0.48\textwidth}\includegraphics[width=\textwidth]{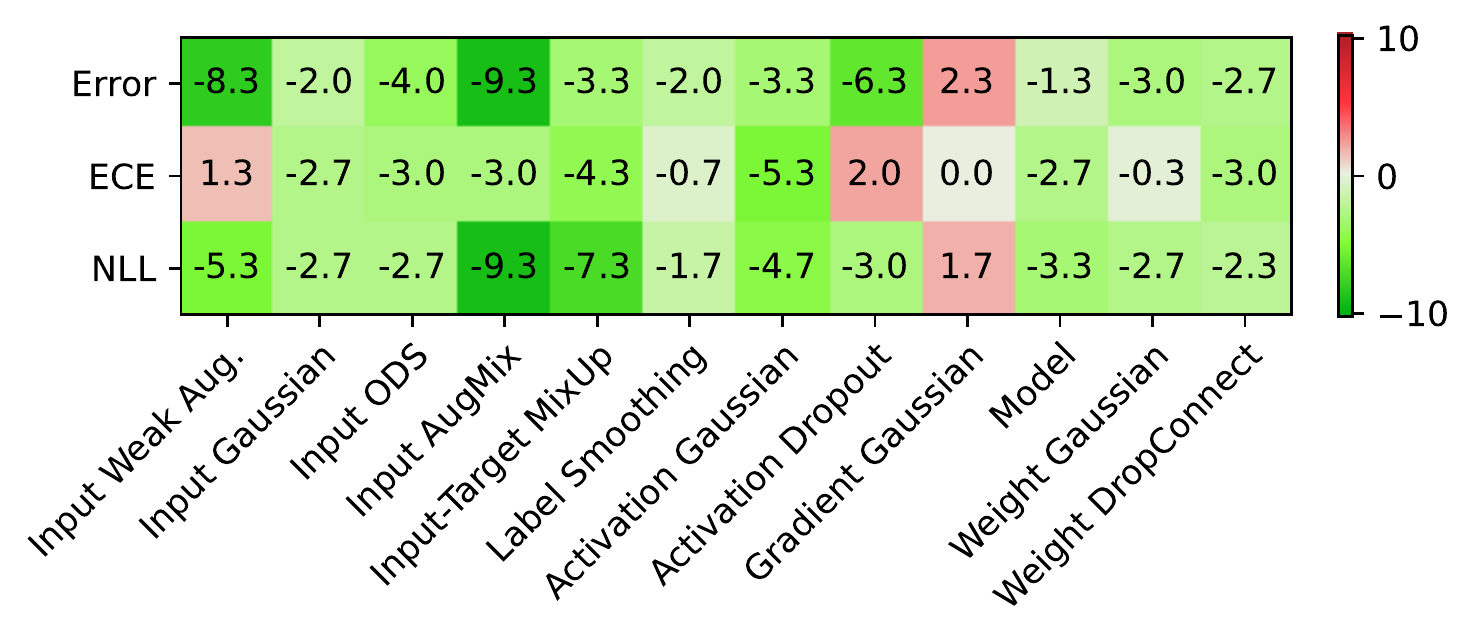}
    \vspace{-0.7cm}
    \caption{OOD CV classification in dataset transfer.}
    \label{fig:transfer:dataset:ood:cv_classification}
    \end{subfigure}
    \hfill
   \begin{subfigure}{0.48\textwidth}\includegraphics[width=\textwidth]{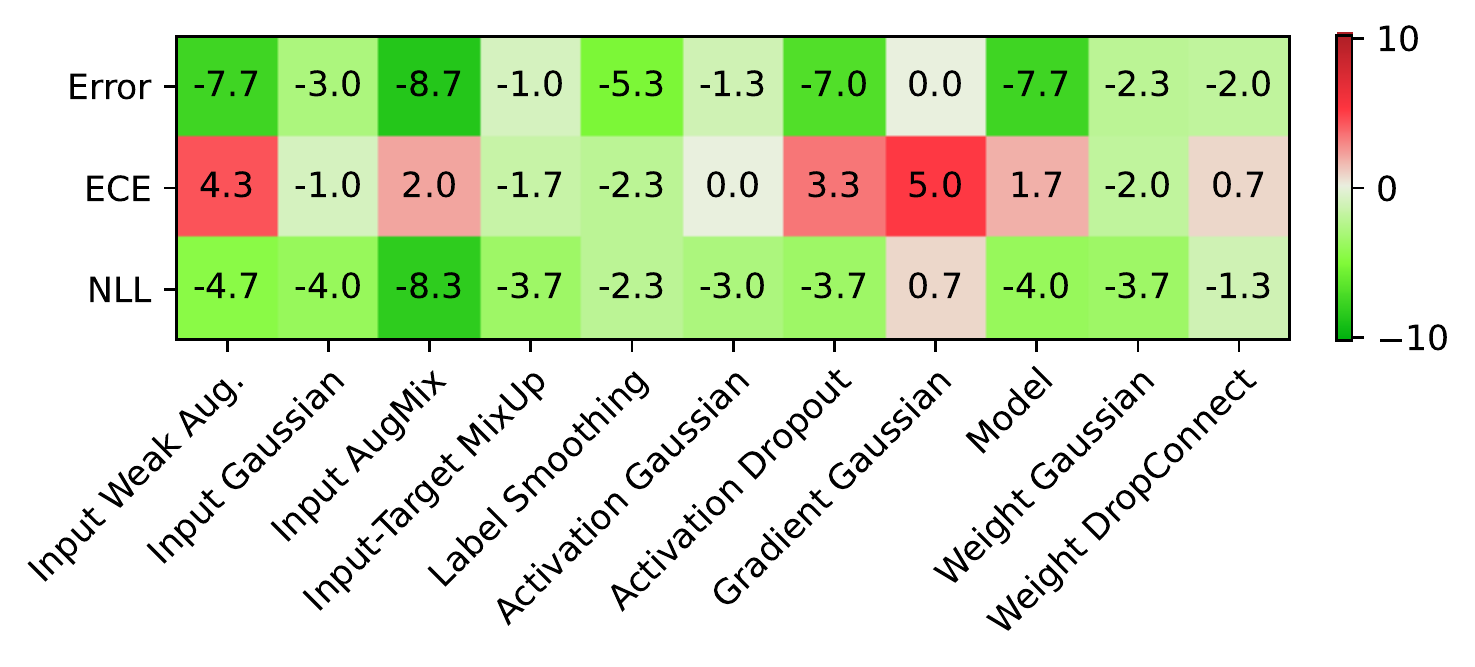}
    \vspace{-0.7cm}
    \caption{OOD CV classification in architecture transfer.}
    \label{fig:transfer:architecture:ood:cv_classification}
    \end{subfigure}
   \begin{subfigure}{0.48\textwidth}\includegraphics[width=\textwidth]{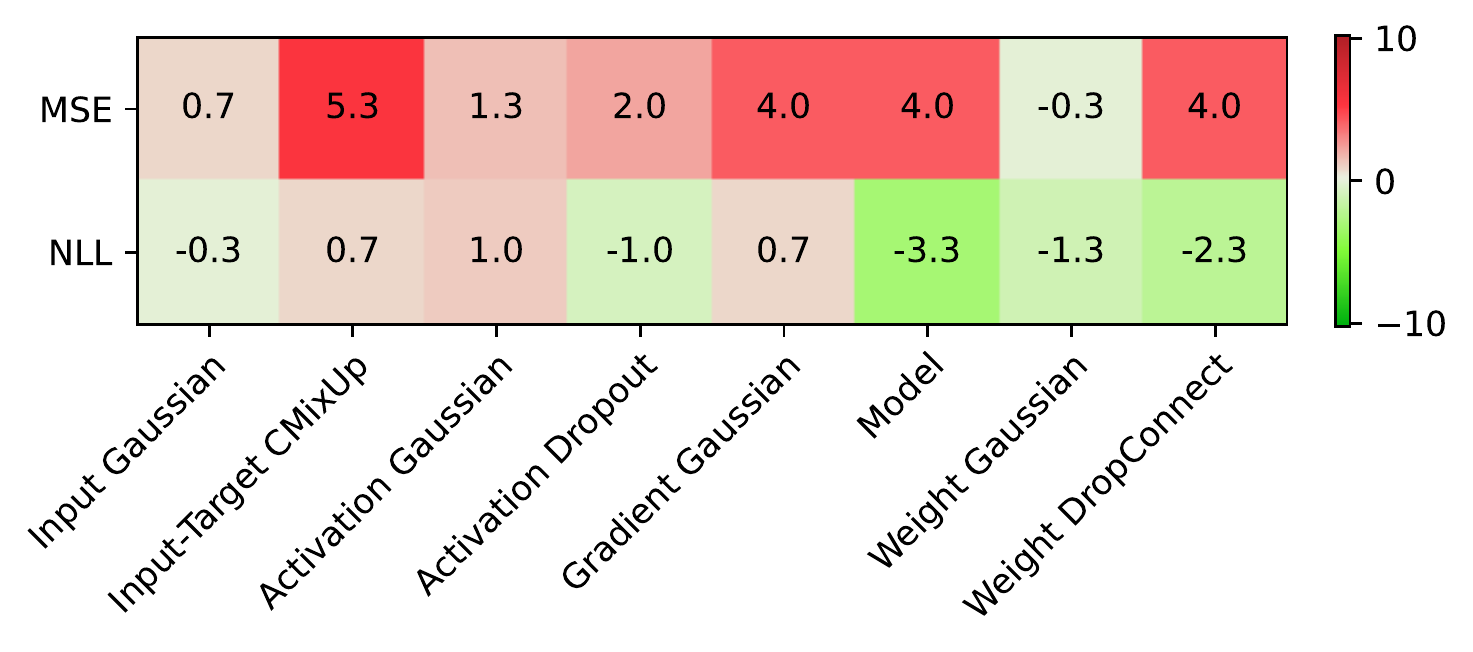}
    \vspace{-0.7cm}
    \caption{OOD tabular regression in dataset transfer.}
    \label{fig:transfer:dataset:ood:tabular_regression}
    \end{subfigure}
    \hfill
   \begin{subfigure}{0.48\textwidth}\includegraphics[width=\textwidth]{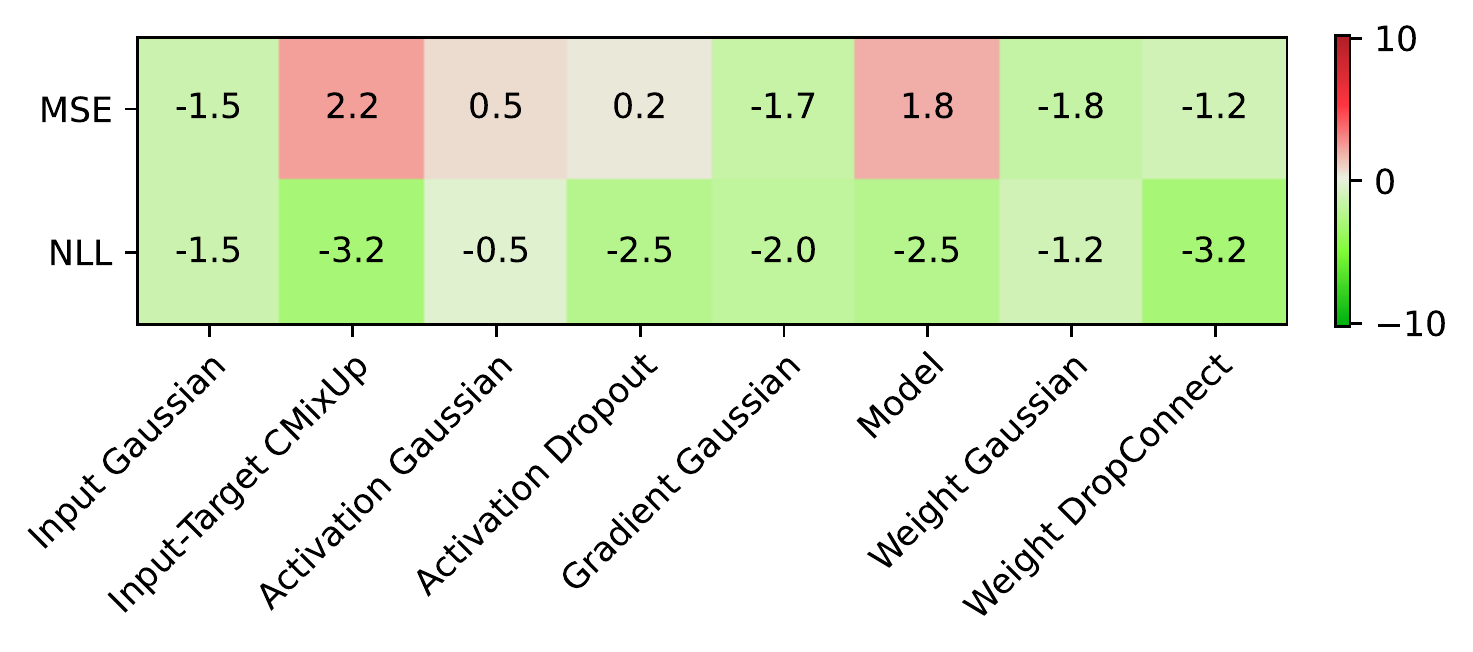}
    \vspace{-0.7cm}
    \caption{OOD tabular regression in architecture transfer.}
    \label{fig:transfer:architecture:ood:tabular_regression}
    \end{subfigure}
    \vspace{-0.25cm}
    \caption{Transfer of hyperparameters on out-of-domain (OOD) data.}
   \label{fig:ood_transfer}
\end{figure}

\subsubsection{Dataset Transfer}\label{sec:experiments:transfer:dataset}
Figures~\ref{fig:transfer:dataset:id:cv_classification} and~\ref{fig:transfer:dataset:id:tabular_regression} show the dataset transfer results for ID settings, with OOD settings shown in Figures~~\ref{fig:transfer:dataset:ood:cv_classification} and~\ref{fig:transfer:dataset:ood:tabular_regression}. 
We observe generally good transferability of hyperparameters across datasets for CV classification in ID and OOD settings. 
In particular, weak augmentation, AugMix and Dropout lead to solid improvements in the ID setting. 
AugMix also excels in OOD scenarios under dataset transfer, but weak augmentation and Dropout are not as strong in calibration measured using ECE. 
Certain noises are less transferable, including Gaussian noise added to the input and DropConnect. 
Hyperparameters for noise in tabular regression are less transferable because of worse generalisation measured using MSE.

\textbf{Main Observations:} The transfer of hyperparameters from dataset to dataset generally works well for CV classification. 
However, caution is advised as it is not the case for all noise types. 
For tabular data regression, tuning of hyperparameters is recommended.

\subsubsection{Architecture Transfer}\label{sec:experiments:transfer:architecture}
In Figures~\ref{fig:transfer:architecture:id:cv_classification} and~\ref{fig:transfer:architecture:id:tabular_regression}, we show the ID results for the architecture transfer, with Figures~\ref{fig:transfer:architecture:ood:cv_classification} and~\ref{fig:transfer:architecture:ood:tabular_regression} reporting the OOD results.
The transferability of noise hyperparameters is lower than across datasets for CV classification, but it is still successful, especially for weak augmentation and AugMix for ID settings. 
Transfer of hyperparameters for tabular data regression works for certain noise types in the ID setting, including adding Gaussian noise to the input or the weights, which are among the Top-3 noises for tabular data regression.

\textbf{Main Observations:} Transfer of hyperparameters across architectures appears more challenging than across datasets but can be successful in some instances. 
Caution is advised, and tuning is recommended.

\subsection{Learnt Representation Landscapes}\label{sec:experiments:landscapes}
\begin{figure}[t]
\centering
\begin{subfigure}{0.21\textwidth}
	\centering
	\includegraphics[width=\textwidth]{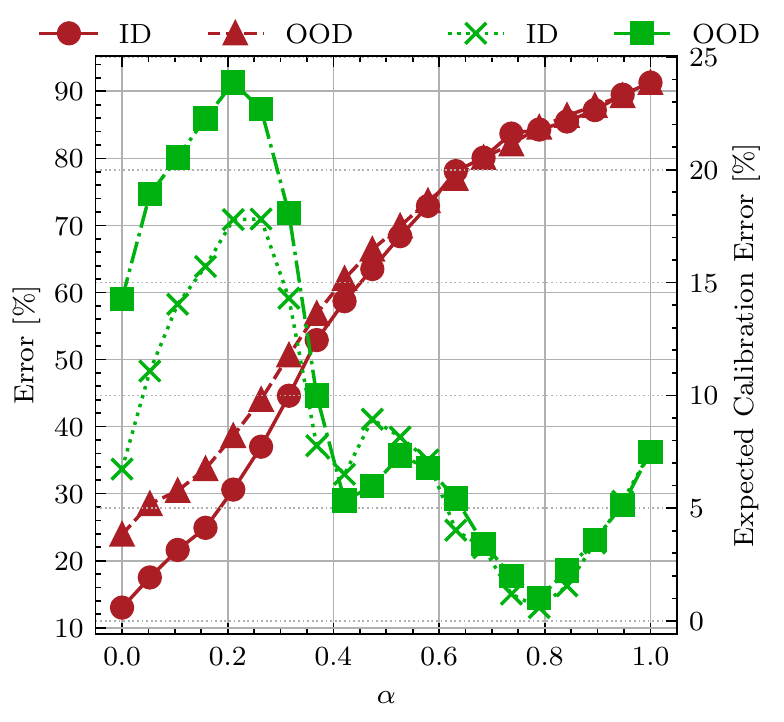}
        \vspace{-0.7cm}
	\caption{Error and ECE.}
	\label{fig:loss_landscape:cifar10-resnet-vanilla-lin_error_ece}
\end{subfigure}
\begin{subfigure}{0.21\textwidth}
	\centering
	\includegraphics[width=\textwidth]{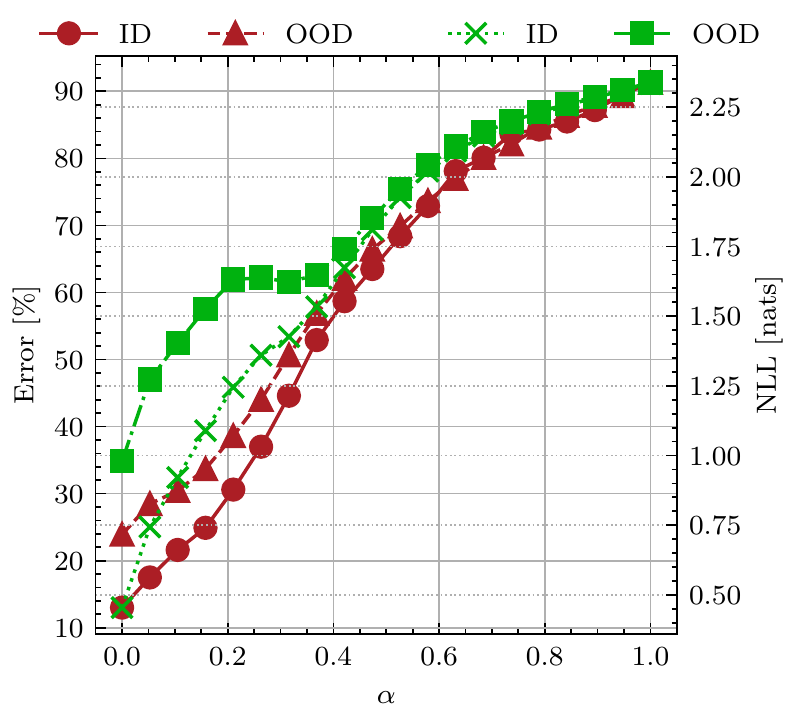}
        \vspace{-0.7cm}
	\caption{Error and NLL.}
	\label{fig:loss_landscape:cifar10-resnet-vanilla-lin_error_nll}
\end{subfigure}
\begin{subfigure}{0.27\textwidth}
	\centering
	\includegraphics[width=\textwidth]{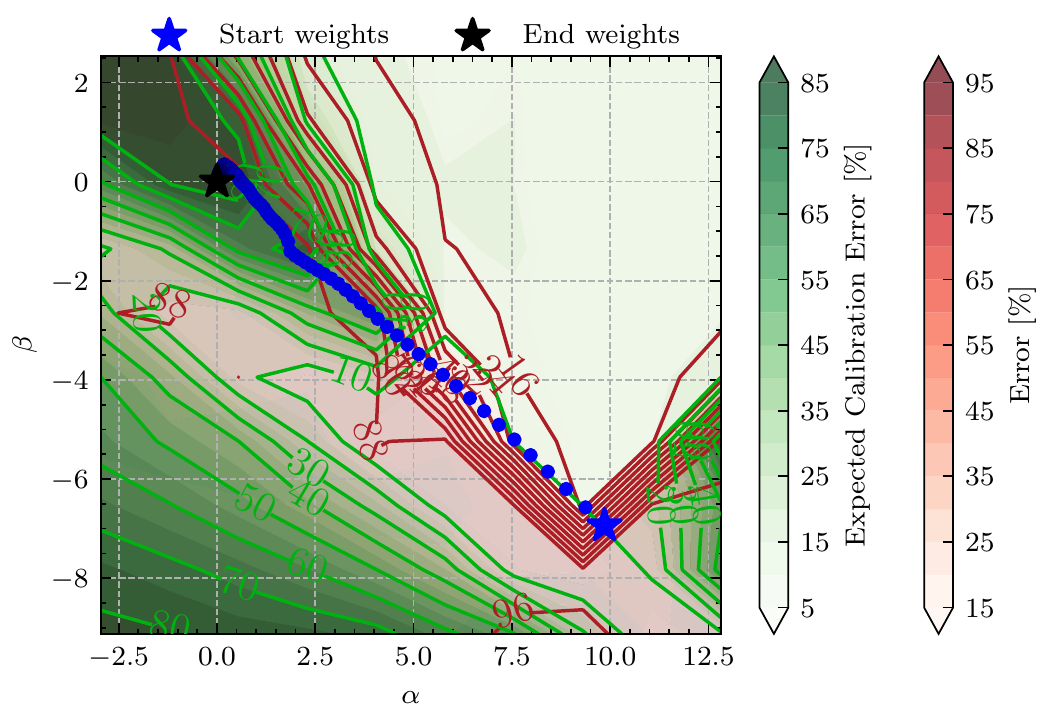}
        \vspace{-0.7cm}
	\caption{Error and ECE on ID.}
	\label{fig:loss_landscape:cifar10-resnet-vanilla-test_2d_error_ece}
\end{subfigure}
\begin{subfigure}{0.27\textwidth}
	\centering
	\includegraphics[width=\textwidth]{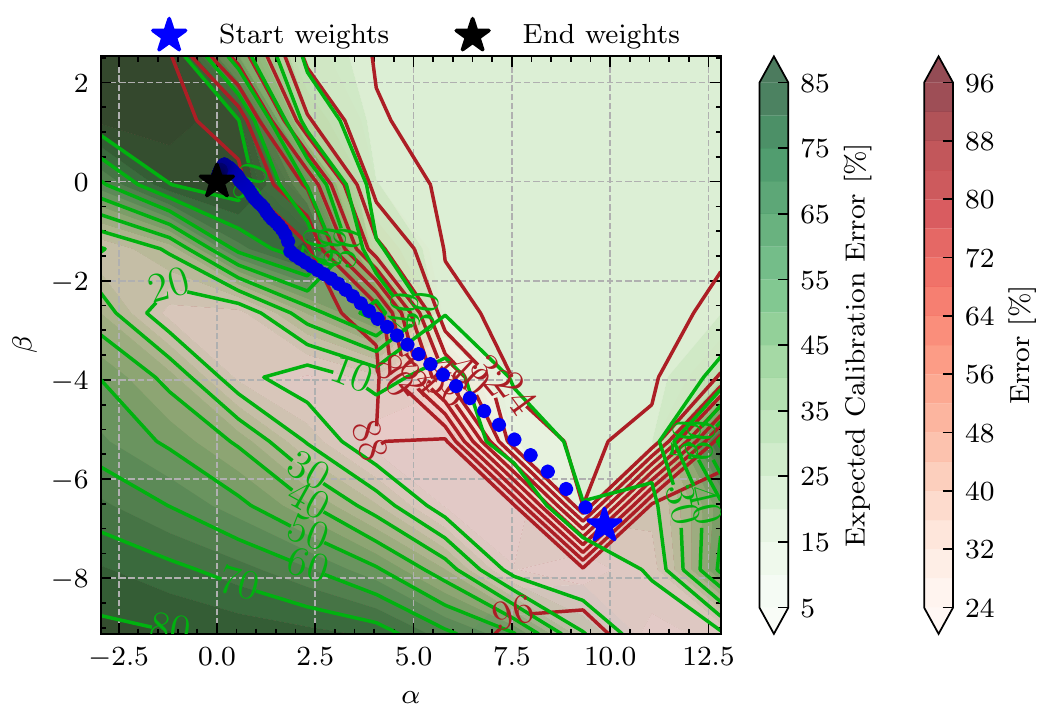}
         \vspace{-0.7cm}
	\caption{Error and ECE on OOD.}
	\label{fig:loss_landscape:cifar10-resnet-vanilla-test_2d_aug_error_ece}
\end{subfigure}
\hfill
\begin{subfigure}{0.21\textwidth}
	\centering
	\includegraphics[width=\textwidth]{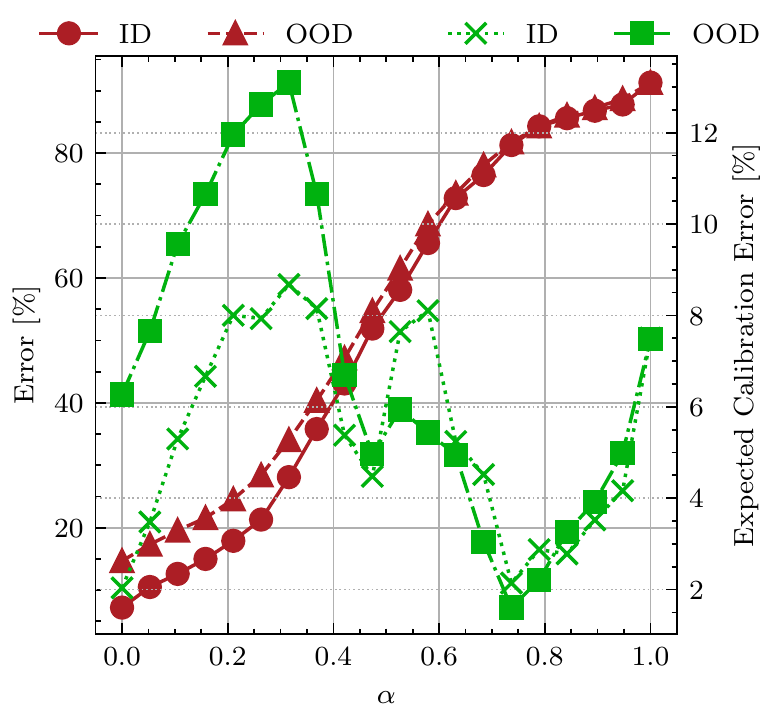}
         \vspace{-0.7cm}
	\caption{Error and ECE.}
	\label{fig:loss_landscape:cifar10-resnet-input_augmix-lin_error_ece}
\end{subfigure}
\begin{subfigure}{0.21\textwidth}
	\centering
	\includegraphics[width=\textwidth]{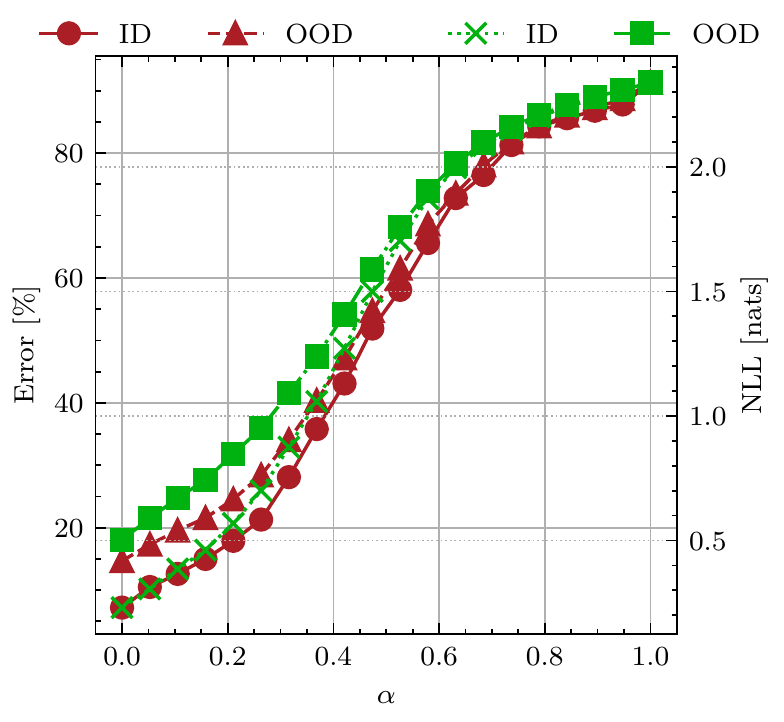}
         \vspace{-0.7cm}
	\caption{Error and NLL.}
	\label{fig:loss_landscape:cifar10-resnet-input_augmix-lin_error_nll}
\end{subfigure}
\begin{subfigure}{0.27\textwidth}
	\centering
	\includegraphics[width=\textwidth]{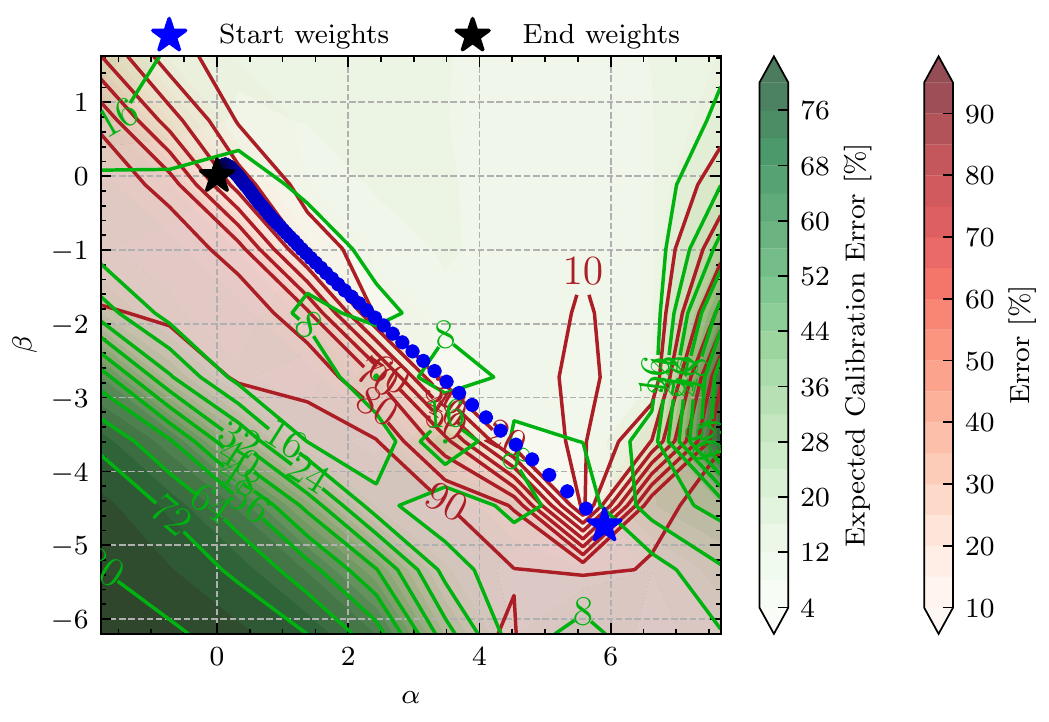}
         \vspace{-0.7cm}
	\caption{Error and ECE on ID.}
	\label{fig:loss_landscape:cifar10-resnet-input_augmix-test_2d_error_ece}
\end{subfigure}
\begin{subfigure}{0.27\textwidth}
	\centering
	\includegraphics[width=\textwidth]{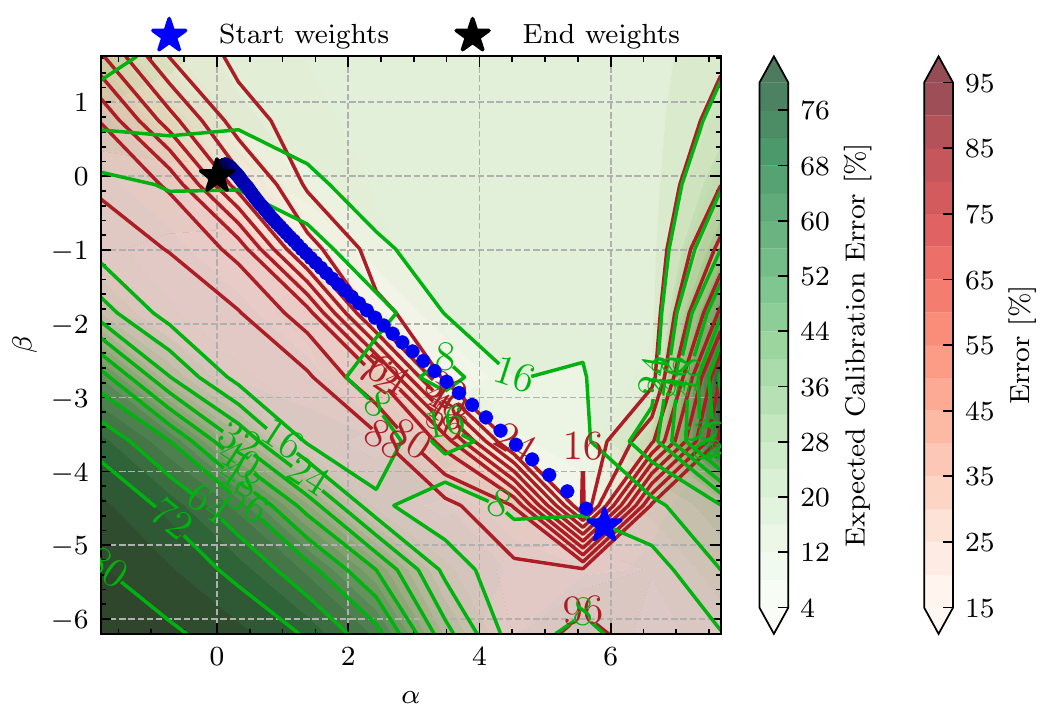}
         \vspace{-0.7cm}
	\caption{Error and ECE on OOD.}
	\label{fig:loss_landscape:cifar10-resnet-input_augmix-test_2d_aug_error_ece}
\end{subfigure}
\caption{No noise (top) and Input AugMix (bottom) on CIFAR-10.}
\label{fig:loss_landscape:cifar10-resnet-input_augmix}
\end{figure}

\begin{figure}[h!]
\centering
\begin{subfigure}{0.2\textwidth}
	\centering
	\includegraphics[width=\textwidth]{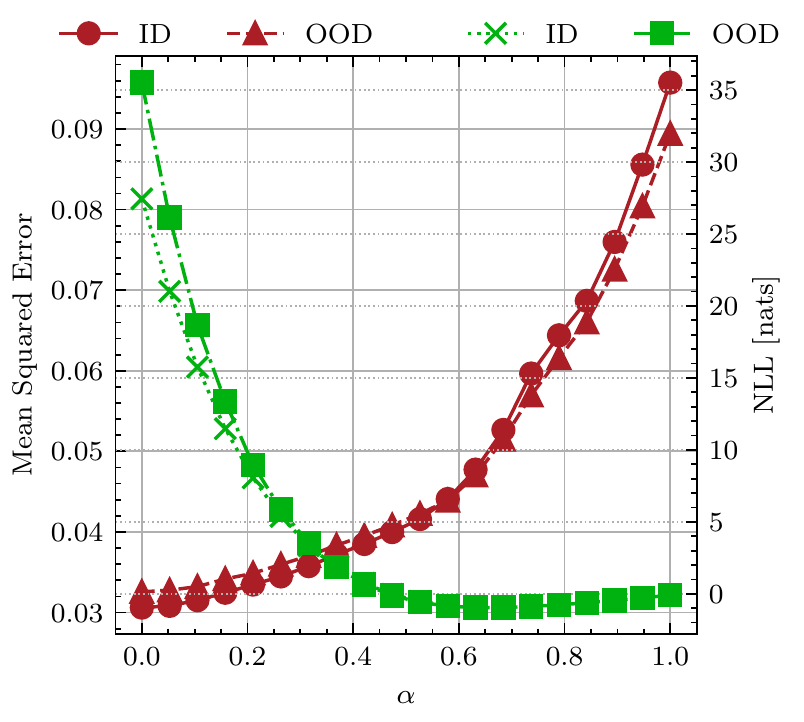}
         \vspace{-0.7cm}
	\caption{MSE and NLL.}
	\label{fig:loss_landscape:wiki_face-resnet-vanilla-lin_mse_nll}
\end{subfigure}
\begin{subfigure}{0.3\textwidth}
	\centering
	\includegraphics[width=\textwidth]{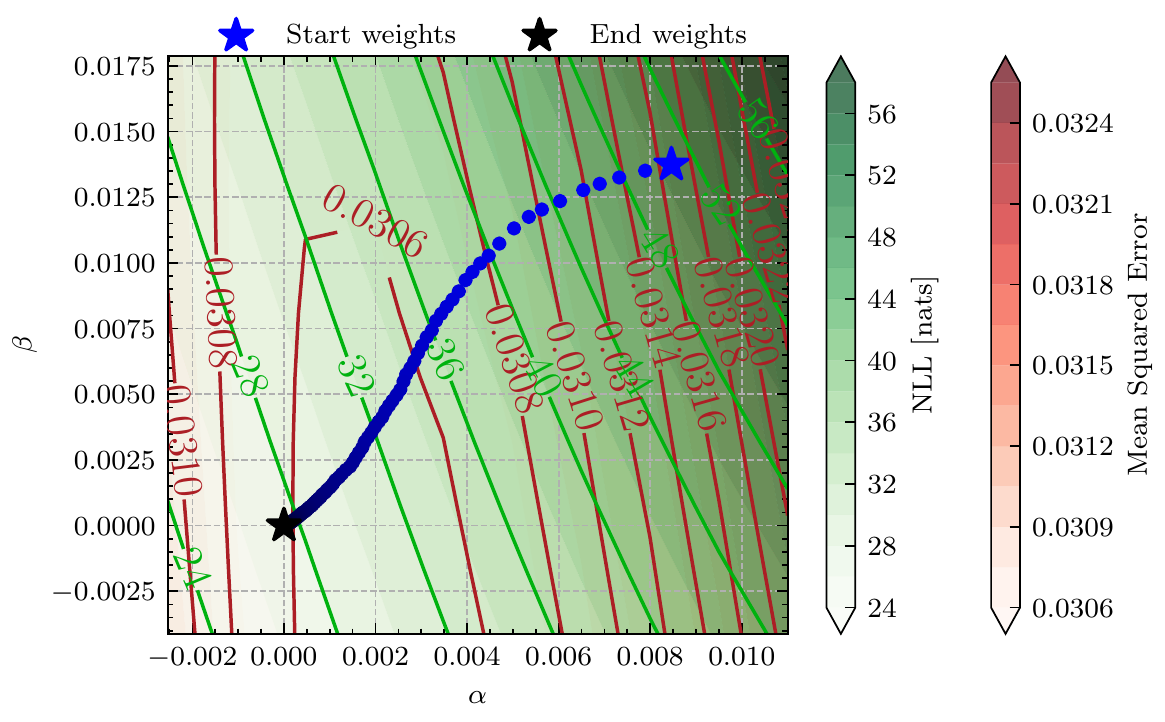}
         \vspace{-0.7cm}
	\caption{MSE and NLL on ID.}
	\label{fig:loss_landscape:wiki_face-resnet-vanilla-test_2d_mse_nll}
\end{subfigure}
\begin{subfigure}{0.3\textwidth}
	\centering
	\includegraphics[width=\textwidth]{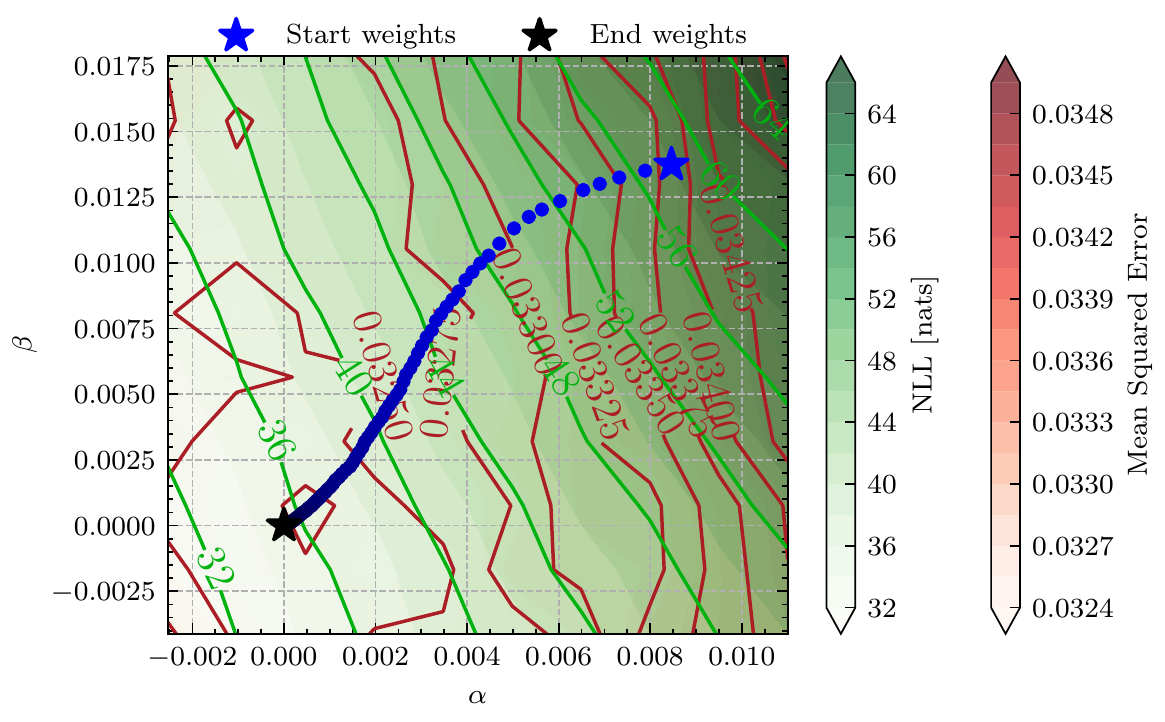}
         \vspace{-0.7cm}
	\caption{MSE and NLL on OOD.}
	\label{fig:loss_landscape:wiki_face-resnet-vanilla-test_2d_aug_mse_nll}
\end{subfigure}
\hfill
\begin{subfigure}{0.2\textwidth}
	\centering
	\includegraphics[width=\textwidth]{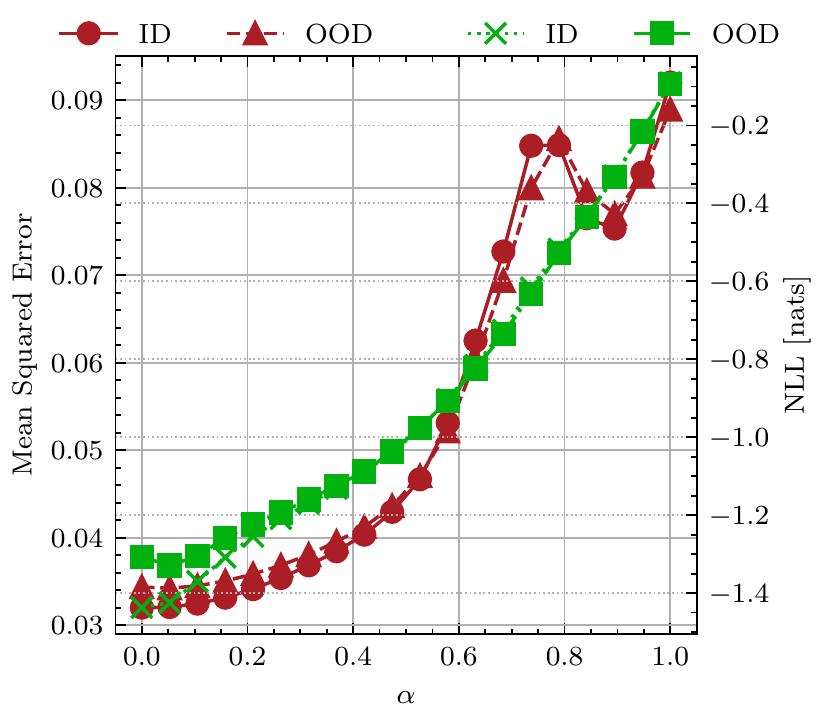}
         \vspace{-0.7cm}
	\caption{MSE and NLL.}
	\label{fig:loss_landscape:wiki_face-resnet-activation_dropout-lin_mse_nll}
\end{subfigure}
\begin{subfigure}{0.3\textwidth}
	\centering
	\includegraphics[width=\textwidth]{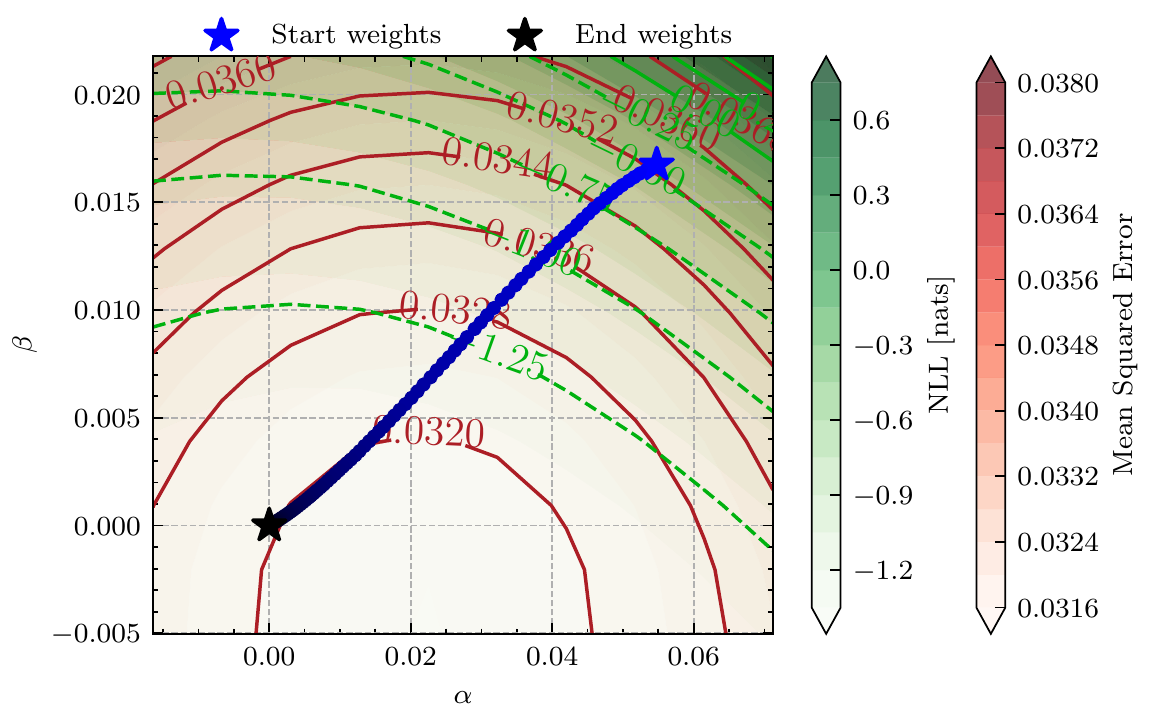}
         \vspace{-0.7cm}
	\caption{MSE and NLL on ID.}
	\label{fig:loss_landscape:wiki_face-resnet-activation_dropout-test_2d_mse_nll}
\end{subfigure}
\begin{subfigure}{0.3\textwidth}
	\centering
	\includegraphics[width=\textwidth]{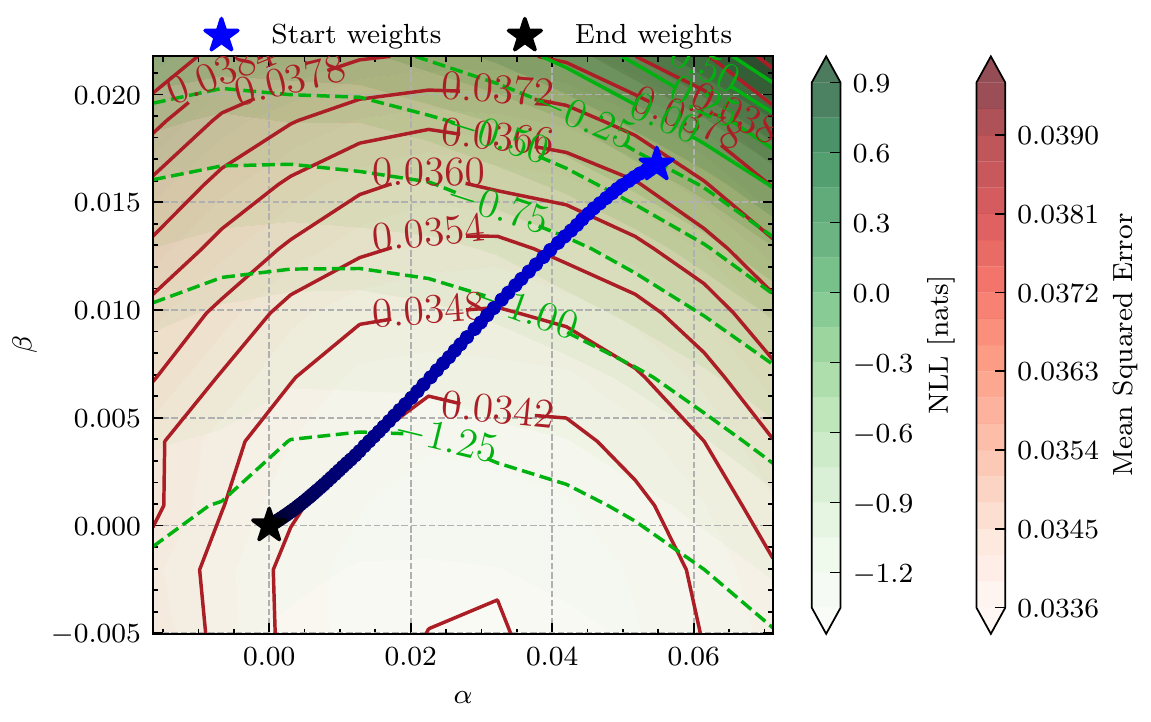}
         \vspace{-0.7cm}
	\caption{MSE and NLL on OOD.}
	\label{fig:loss_landscape:wiki_face-resnet-activation_dropout-test_2d_aug_mse_nll}
\end{subfigure}
\caption{No noise (top) and Activation Dropout (bottom) on WikiFace.}
\label{fig:loss_landscape:wiki_face-resnet-activation_dropout}
\end{figure}

We study the learnt representation landscapes of NNs trained with various noises through the lenses of ID and OOD performance in terms of error, ECE, NLL or MSE.
We consider the noises individually, with the ID-found hyperparameters starting from the same weight initialisation for fairness.
We visualise linear interpolation modulated through an $\alpha$ parameter between the final, $\alpha=0$ and initial model, $\alpha=1$~\citep{goodfellow2014qualitatively}.
The interpolation empirically investigates the smoothness of the training process.
We also visualise the landscape in 2D~\citep{li2018visualizing, Holbrook_2020} by saving the network after each epoch and concatenating the weights. 
Instead of using random coordinates, we use the first two principal components of the weights as the coordinates. 
We normalise them based on the magnitude of the original weights, and we project all the weights onto these two components in the vicinity of $\alpha$ and $\beta$.
The 2D visualisations show us the exploration and exploitation of the training process.
In Figures~\ref{fig:loss_landscape:cifar10-resnet-input_augmix} and~\ref{fig:loss_landscape:wiki_face-resnet-activation_dropout}, we compare the metric landscapes of no noise with AugMix and Dropout noises, respectively on CIFAR-10 and WikiFace datasets.
We used 20 points for linear interpolation and 100 points for the 2D plots across five selected OOD augmentations and 1000 test data samples for compute efficiency.
In red, we show the error or MSE; in green, we offer the NLL or ECE.
In the 1D plots, \tikz{\pgfuseplotmark{*}} and $\blacktriangle$ stand for ID and OOD error or MSE, and $\times$ and $\blacksquare$ stand for ID and OOD ECE or NLL.
In the 2D plots, the darker combined contours signify worse performance than the lighter parts, and the $\starletfill$ in blue or black denotes the start or end weights, respectively.
The Appendix contains the metric landscapes for all other noises, tabular classification -- Adult, and regression -- Yacht datasets.

Observing Figures~\ref{fig:loss_landscape:cifar10-resnet-input_augmix} and~\ref{fig:loss_landscape:wiki_face-resnet-activation_dropout}, we first notice the ID and OOD results are similar, with the OOD results being slightly worse across all metrics.
This includes both the 1D plots and the 2D plots.
Second, as seen in Figures~\ref{fig:loss_landscape:cifar10-resnet-vanilla-lin_error_ece} and~\ref{fig:loss_landscape:cifar10-resnet-vanilla-lin_error_nll}, the curves for error, representing generalisation, and ECE or NLL, representing confidence calibration, do not share the same shapes or curvatures.
MSE and NLL curves in Figure~\ref{fig:loss_landscape:wiki_face-resnet-vanilla-lin_mse_nll} are more similar than error and ECE curves.
Looking at the 1D plots, for example in Figures~\ref{fig:loss_landscape:cifar10-resnet-vanilla-lin_error_ece}~\ref{fig:loss_landscape:cifar10-resnet-vanilla-lin_error_nll} and~\ref{fig:loss_landscape:wiki_face-resnet-vanilla-lin_mse_nll}, the error or MSE can be more smoothly interpolated than ECE or NLL. Figure~\ref{fig:loss_landscape:wiki_face-resnet-vanilla-lin_mse_nll} shows models trained without noise can become overconfident, reflected in large NLL and small MSE.
Adding noise such as Dropout can fix this, leading to low NLL for the final model in Figure~\ref{fig:loss_landscape:wiki_face-resnet-activation_dropout-lin_mse_nll}.  
Looking at the 2D plots in Figures~\ref{fig:loss_landscape:cifar10-resnet-vanilla-test_2d_error_ece} and~\ref{fig:loss_landscape:wiki_face-resnet-vanilla-test_2d_mse_nll}, the error or MSE valley is wider than the ECE or NLL valley, and they are not aligned.
From a detailed comparison between no noise and AugMix or Dropout in Figures~\ref{fig:loss_landscape:cifar10-resnet-input_augmix} and~\ref{fig:loss_landscape:wiki_face-resnet-activation_dropout}, we observe that AugMix and Dropout can smoothen the optimisation in the 1D plots, but not for ECE, and decrease the gap between ID and OOD performance.
The 2D plots show that AugMix and Dropout can explore broader metric landscapes than no noise, shown in ranges of $\alpha$ and $\beta$ in the 2D plots, and marginally align the error or MSE with NLL.
Seen in the lightness of the 2D contour plots, the noises navigate lower NLL or ECE landscapes than no noise. 

Our general observations considering both CV and tabular datasets show that while noises such as AugMix, weak augmentation, MixUp or activation and weight noises based around Dropout can smoothen the optimisation regarding error or MSE, they rarely smoothen the optimisation regarding ECE.
The metric landscapes often look similar to no noise, but the optimisation ends in more profound valleys.
Across the datasets and tasks, label smoothing, input additive Gaussian and ODS have minimal effect on the 2D landscapes or 1D interpolation.
The model shrink and perturb make the optimisation more ``stairs-like'', and the metric landscape explored is broader.
Together with gradient Gaussian noise, the shrink and perturb noises explore broader metric landscapes than the others.
No method drastically changes the metric landscape or the interpolation from the default, but they can make the optimisation smoother or broader.


\textbf{Main Observations:} The metric landscapes for error or MSE and ECE or NLL are different, and the noises can smoothen the optimisation in terms of error or MSE but not necessarily in terms of ECE or NLL.
When a model trained without noise is overconfident, adding noise to the training can resolve it and lead to a significantly better-calibrated model at the end of training.

\section{Conclusion}\label{sec:discussion}

\textbf{Key Takeaways}:
Noise injection methods can improve NN performance across various tasks and datasets. 
This is despite the fact L2 regularisation was already tuned to prevent overfitting, indicating noise injection methods can provide additional benefits beyond standard regularisation.
The methods were not equally efficient across all tasks and datasets, with significant differences in performance between regression and classification.
Nevertheless, out of the considered noise types, the proposed methodology identified at least one noise or a combination of noises that demonstrated improvements in both calibration and generalisation over the no noise baseline for all tasks.
The most effective noise for CV was AugMix, model shrink and perturb and Gaussian noise added to weights for tabular data classification and regression, respectively. 
At the same time, Dropout and label smoothing worked the best for NLP. 
Even though ODS was not designed to improve calibration and generalisation, it has shown promising performance in several cases.
Combining noises outperformed individual noises in most classification cases, with regression often benefitting from using only one noise at a time.
While directly combining hyperparameters of noises is a reasonable strategy, tuning them can still be valuable if a large budget is used.
The noises improved ID and OOD performance, but the ID rankings were sometimes inconsistent with the OOD evaluation. 
AugMix remained highly ranked for robustness, demonstrating that domain-specific inductive biases are beneficial when crafting noise injection methods as they can improve generalisation and calibration.
Note that our aim was not to add new knowledge about AugMix's performance against image corruptions, given its well-documented performance~\citep{hendrycks2019augmix}, but to demonstrate that inductive biases should be considered when generating noise for better performance.
The visualisation showed noises can smoothen the optimisation in terms of error or MSE but not necessarily in terms of ECE or NLL. 
It also showed noise can help mitigate overconfidence.
Overall, the results indicate practitioners should consider combining noises, e.g. AugMix and Dropout, and tuning hyperparameters for their specific problem.

\textbf{Limitations}:
To conduct this study, we had to restrict the experiments' scope.
Our scope was limited to experimental datasets, tasks such as classification and regression, and standard NN architectures.
Testing on more complex data and downstream tasks such as object detection, segmentation, or reinforcement learning would reveal more profound impacts of noise injection.
Furthermore, diving deeper into one particular domain, such as NLP, could provide more insights into the effectiveness of noise injection methods.
Moreover, we also limited the optimisation to SGD with momentum and a cosine learning rate schedule, which were tuned beforehand to make the hyperparameter search tractable.
To draw practical conclusions, we evaluated the noise performance by minimising the NLL rather than exploring all possible settings.
The costs associated with adjusting and evaluating different noises limited the scope of the experiments. 
Consequently, certain noises might prove more effective with more thorough tuning and a larger budget. 
This is especially true for noise combinations, where the number of possible combinations grows exponentially with the number of noises.
Nonetheless, the existing findings offer valuable insights for practitioners by giving a preliminary indication of the most promising noise sources. 
This enables users to concentrate their efforts and compute the budget required for tuning. 
For example, AugMix, incorporating domain-specific inductive biases, was transferable and effective even with a limited budget.
Developing methods to choose hyperparameters without the need for extensive tuning would enhance the accessibility of these techniques.
Lastly, the out-of-distribution evaluation was focused on synthetic augmentations, which may not fully capture the real-world distribution shift, as noticed in~\citep{taori2020measuring} for computer vision tasks.

\textbf{Future Directions}:
The strong performance of AugMix highlights the potential for developing specialised, domain-specific noise techniques such as DeepAugment~\citep{hendrycks2021many}.
For example, tailored domain-specific noise methods could benefit tabular data-based problems and NLP. 
Future work should also explore specific data-architecture noise interactions, as the transferability of hyperparameters was limited.
Inspired by the annealed gradient noise, annealing noise levels overtraining may also prove helpful, as early noise could encourage robustness. 
In contrast, low late-stage noise could enable convergence on a high-accuracy solution.
The potential for combining noises from the same category should also be investigated further.
The noises affected the entire architecture, but targeting noise injection methods that only affect specific layers or sections of the network may be possible, requiring more or less regularisation.
Specific noise-based approaches for simultaneously exploring the generalisation and confidence calibration trade-off should be investigated further.
As demonstrated across image and tabular domains, the empirical differences between many methods seem minor.
This indicates that there are more fundamental determinants of performance.
Therefore, theoretical analysis of noise injection methods would be beneficial in understanding the underlying mechanisms and providing guidance for future research.
Lastly, testing on wider-scale out-of-distribution datasets and real-world distribution shifts would provide a more comprehensive evaluation of the noise injection methods' impact on robustness.
We hope our study and framework, embedded in our codebase, will assist further research in this area.

\section*{Acknowledgements}
Martin Ferianc was sponsored through a scholarship from the Institute of Communications and Connected Systems at UCL. Ondrej Bohdal was supported by the EPSRC Centre for Doctoral Training in Data Science, funded by the UK Engineering and Physical Sciences Research Council (grant EP/L016427/1) and the University of Edinburgh.

\newpage
\bibliography{bib.bib}

\begin{thebibliography}{62}
\providecommand{\natexlab}[1]{#1}
\providecommand{\url}[1]{\texttt{#1}}
\expandafter\ifx\csname urlstyle\endcsname\relax
  \providecommand{\doi}[1]{doi: #1}\else
  \providecommand{\doi}{doi: \begingroup \urlstyle{rm}\Url}\fi

\bibitem[Ash \& Adams(2020)Ash and Adams]{ash2020warm}
Jordan Ash and Ryan~P Adams.
\newblock On warm-starting neural network training.
\newblock In \emph{NeurIPS}, 2020.

\bibitem[Asuncion \& Newman(2007)Asuncion and Newman]{asuncion2007uci}
Arthur Asuncion and David Newman.
\newblock Uci machine learning repository, 2007.

\bibitem[Bergstra et~al.(2011)Bergstra, Bardenet, Bengio, and
  K\'{e}gl]{bergstra2011tpe}
James Bergstra, R\'{e}mi Bardenet, Yoshua Bengio, and Bal\'{a}zs K\'{e}gl.
\newblock Algorithms for hyper-parameter optimization.
\newblock In \emph{NeurIPS}, 2011.

\bibitem[Bishop(1995)]{bishop1995training}
Chris~M Bishop.
\newblock Training with noise is equivalent to tikhonov regularization.
\newblock \emph{Neural computation}, 7\penalty0 (1):\penalty0 108--116, 1995.

\bibitem[Blundell et~al.(2015)Blundell, Cornebise, Kavukcuoglu, and
  Wierstra]{blundell2015weight}
Charles Blundell, Julien Cornebise, Koray Kavukcuoglu, and Daan Wierstra.
\newblock Weight uncertainty in neural network.
\newblock In \emph{ICML}, 2015.

\bibitem[Camuto(2021)]{camuto2021understanding}
Alexander Camuto.
\newblock \emph{Understanding Gaussian noise injections in neural networks}.
\newblock PhD thesis, University of Oxford, 2021.

\bibitem[Camuto et~al.(2020)Camuto, Willetts, Simsekli, Roberts, and
  Holmes]{camuto2020explicit}
Alexander Camuto, Matthew Willetts, Umut Simsekli, Stephen~J Roberts, and
  Chris~C Holmes.
\newblock Explicit regularisation in gaussian noise injections.
\newblock In \emph{NeurIPS}, 2020.

\bibitem[Chaudhari \& Soatto(2015)Chaudhari and Soatto]{chaudhari2015energy}
Pratik Chaudhari and Stefano Soatto.
\newblock On the energy landscape of deep networks.
\newblock \emph{arXiv}, 2015.

\bibitem[Chun et~al.(2020)Chun, Oh, Yun, Han, Choe, and Yoo]{chun2020empirical}
Sanghyuk Chun, Seong~Joon Oh, Sangdoo Yun, Dongyoon Han, Junsuk Choe, and
  Youngjoon Yoo.
\newblock {An empirical evaluation on robustness and uncertainty of
  regularization methods}.
\newblock \emph{arXiv}, 2020.

\bibitem[Cohen et~al.(2019)Cohen, Rosenfeld, and Kolter]{cohen2019certified}
Jeremy Cohen, Elan Rosenfeld, and Zico Kolter.
\newblock Certified adversarial robustness via randomized smoothing.
\newblock In \emph{ICML}, 2019.

\bibitem[DeVries \& Taylor(2017)DeVries and Taylor]{devries2017dataset}
Terrance DeVries and Graham~W Taylor.
\newblock Dataset augmentation in feature space.
\newblock \emph{arXiv}, 2017.

\bibitem[Gal \& Ghahramani(2016)Gal and Ghahramani]{gal2016dropout}
Yarin Gal and Zoubin Ghahramani.
\newblock Dropout as a bayesian approximation: Representing model uncertainty
  in deep learning.
\newblock In \emph{ICML}, 2016.

\bibitem[Geirhos et~al.(2018)Geirhos, Temme, Rauber, Sch{\"u}tt, Bethge, and
  Wichmann]{geirhos2018generalisation}
Robert Geirhos, Carlos~RM Temme, Jonas Rauber, Heiko~H Sch{\"u}tt, Matthias
  Bethge, and Felix~A Wichmann.
\newblock {Generalisation in humans and deep neural networks}.
\newblock In \emph{NeurIPS}, 2018.

\bibitem[Gidaris et~al.(2018)Gidaris, Singh, and
  Komodakis]{gidaris2018unsupervised}
Spyros Gidaris, Praveer Singh, and Nikos Komodakis.
\newblock Unsupervised representation learning by predicting image rotations.
\newblock In \emph{ICLR}, 2018.

\bibitem[Godbole et~al.(2023)Godbole, Dahl, Gilmer, Shallue, and
  Nado]{tuningplaybookgithub}
Varun Godbole, George~E. Dahl, Justin Gilmer, Christopher~J. Shallue, and
  Zachary Nado.
\newblock Deep learning tuning playbook, 2023.
\newblock URL \url{http://github.com/google-research/tuning_playbook}.
\newblock Version 1.0.

\bibitem[Goodfellow et~al.(2014)Goodfellow, Vinyals, and
  Saxe]{goodfellow2014qualitatively}
Ian~J Goodfellow, Oriol Vinyals, and Andrew~M Saxe.
\newblock Qualitatively characterizing neural network optimization problems.
\newblock \emph{arXiv}, 2014.

\bibitem[Goodfellow et~al.(2015)Goodfellow, Shlens, and
  Szegedy]{goodfellow2014explaining}
Ian~J Goodfellow, Jonathon Shlens, and Christian Szegedy.
\newblock {Explaining and harnessing adversarial examples}.
\newblock In \emph{ICLR}, 2015.

\bibitem[Guo et~al.(2017)Guo, Pleiss, Sun, and Weinberger]{guo2017calibration}
Chuan Guo, Geoff Pleiss, Yu~Sun, and Kilian~Q Weinberger.
\newblock On calibration of modern neural networks.
\newblock In \emph{ICML}, 2017.

\bibitem[Guo et~al.(2019)Guo, Mao, and Zhang]{guo2019augmenting}
Hongyu Guo, Yongyi Mao, and Richong Zhang.
\newblock Augmenting data with mixup for sentence classification: An empirical
  study.
\newblock \emph{arXiv}, 2019.

\bibitem[He et~al.(2016)He, Zhang, Ren, and Sun]{he2016deep}
Kaiming He, Xiangyu Zhang, Shaoqing Ren, and Jian Sun.
\newblock Deep residual learning for image recognition.
\newblock In \emph{CVPR}, 2016.

\bibitem[He et~al.(2019)He, Rakin, and Fan]{he2019parametric}
Zhezhi He, Adnan~Siraj Rakin, and Deliang Fan.
\newblock Parametric noise injection: Trainable randomness to improve deep
  neural network robustness against adversarial attack.
\newblock In \emph{CVPR}, 2019.

\bibitem[Hendrycks \& Dietterich(2019)Hendrycks and
  Dietterich]{hendrycks2019benchmarking}
Dan Hendrycks and Thomas Dietterich.
\newblock Benchmarking neural network robustness to common corruptions and
  perturbations.
\newblock In \emph{ICLR}, 2019.

\bibitem[Hendrycks et~al.(2020)Hendrycks, Mu, Cubuk, Zoph, Gilmer, and
  Lakshminarayanan]{hendrycks2019augmix}
Dan Hendrycks, Norman Mu, Ekin~D Cubuk, Barret Zoph, Justin Gilmer, and Balaji
  Lakshminarayanan.
\newblock Augmix: A simple data processing method to improve robustness and
  uncertainty.
\newblock In \emph{ICLR}, 2020.

\bibitem[Hendrycks et~al.(2021)Hendrycks, Basart, Mu, Kadavath, Wang, Dorundo,
  Desai, Zhu, Parajuli, Guo, et~al.]{hendrycks2021many}
Dan Hendrycks, Steven Basart, Norman Mu, Saurav Kadavath, Frank Wang, Evan
  Dorundo, Rahul Desai, Tyler Zhu, Samyak Parajuli, Mike Guo, et~al.
\newblock The many faces of robustness: A critical analysis of
  out-of-distribution generalization.
\newblock In \emph{CVPR}, 2021.

\bibitem[Holbrook(2020)]{Holbrook_2020}
Ryan Holbrook, 2020.
\newblock URL
  \url{https://mathformachines.com/posts/visualizing-the-loss-landscape/}.

\bibitem[Jamieson \& Talwalkar(2016)Jamieson and Talwalkar]{jamieson2016non}
Kevin Jamieson and Ameet Talwalkar.
\newblock Non-stochastic best arm identification and hyperparameter
  optimization.
\newblock In \emph{AISTATS}, 2016.

\bibitem[Jang et~al.(2021)Jang, McCormack, and Tong]{jang2021noise}
Hojin Jang, Devin McCormack, and Frank Tong.
\newblock Noise-trained deep neural networks effectively predict human vision
  and its neural responses to challenging images.
\newblock \emph{PLoS biology}, 19\penalty0 (12):\penalty0 e3001418, 2021.

\bibitem[Kang et~al.(2019)Kang, Sun, Brown, Hendrycks, and
  Steinhardt]{kang2019transfer}
Daniel Kang, Yi~Sun, Tom Brown, Dan Hendrycks, and Jacob Steinhardt.
\newblock {Transfer of adversarial robustness between perturbation types}.
\newblock \emph{arXiv}, 2019.

\bibitem[Kim(2014)]{kim2014convolutional}
Yoon Kim.
\newblock Convolutional neural networks for sentence classification.
\newblock \emph{arXiv}, 2014.

\bibitem[Kingma et~al.(2015)Kingma, Salimans, and
  Welling]{kingma2015variational}
Durk~P Kingma, Tim Salimans, and Max Welling.
\newblock Variational dropout and the local reparameterization trick.
\newblock In \emph{NeurIPS}, 2015.

\bibitem[Krizhevsky et~al.(2009)Krizhevsky, Hinton,
  et~al.]{krizhevsky2009learning}
Alex Krizhevsky, Geoffrey Hinton, et~al.
\newblock Learning multiple layers of features from tiny images.
\newblock Technical report, University of Toronto, 2009.

\bibitem[Krogh \& Hertz(1991)Krogh and Hertz]{krogh1991simple}
Anders Krogh and John Hertz.
\newblock A simple weight decay can improve generalization.
\newblock In \emph{NeurIPS}, 1991.

\bibitem[Kuka{\v{c}}ka et~al.(2017)Kuka{\v{c}}ka, Golkov, and
  Cremers]{kukavcka2017regularization}
Jan Kuka{\v{c}}ka, Vladimir Golkov, and Daniel Cremers.
\newblock Regularization for deep learning: A taxonomy.
\newblock \emph{arXiv}, 2017.

\bibitem[Lang(1995)]{Lang1995Newsweeder}
Ken Lang.
\newblock {Newsweeder: Learning to filter netnews}.
\newblock In \emph{ICML}, 1995.

\bibitem[Le \& Yang(2015)Le and Yang]{le2015tiny}
Ya~Le and Xuan Yang.
\newblock Tiny imagenet visual recognition challenge.
\newblock \emph{CS 231N}, 7\penalty0 (7):\penalty0 3, 2015.

\bibitem[Li et~al.(2018)Li, Xu, Taylor, Studer, and
  Goldstein]{li2018visualizing}
Hao Li, Zheng Xu, Gavin Taylor, Christoph Studer, and Tom Goldstein.
\newblock Visualizing the loss landscape of neural nets.
\newblock In \emph{NeurIPS}, 2018.

\bibitem[Loshchilov \& Hutter(2017)Loshchilov and Hutter]{loshchilov2016sgdr}
Ilya Loshchilov and Frank Hutter.
\newblock Sgdr: Stochastic gradient descent with warm restarts.
\newblock In \emph{ICLR}, 2017.

\bibitem[M{\"u}ller et~al.(2019)M{\"u}ller, Kornblith, and
  Hinton]{muller2019does}
Rafael M{\"u}ller, Simon Kornblith, and Geoffrey~E Hinton.
\newblock When does label smoothing help?
\newblock In \emph{NeurIPS}, 2019.

\bibitem[Neelakantan et~al.(2017)Neelakantan, Vilnis, Le, Kaiser, Kurach,
  Sutskever, and Martens]{neelakantan2017adding}
Arvind Neelakantan, Luke Vilnis, Quoc~V. Le, Lukasz Kaiser, Karol Kurach, Ilya
  Sutskever, and James Martens.
\newblock Adding gradient noise improves learning for very deep networks.
\newblock In \emph{OpenReview}, 2017.

\bibitem[Netzer et~al.(2011)Netzer, Wang, Coates, Bissacco, Wu, and
  Ng]{netzer2011reading}
Yuval Netzer, Tao Wang, Adam Coates, Alessandro Bissacco, Bo~Wu, and Andrew~Y
  Ng.
\newblock Reading digits in natural images with unsupervised feature learning.
\newblock In \emph{NIPS Workshop on Deep Learning and Unsupervised Feature
  Learning}, 2011.

\bibitem[Noh et~al.(2017)Noh, You, Mun, and Han]{noh2017regularizing}
Hyeonwoo Noh, Tackgeun You, Jonghwan Mun, and Bohyung Han.
\newblock Regularizing deep neural networks by noise: Its interpretation and
  optimization.
\newblock In \emph{NeurIPS}, 2017.

\bibitem[Pennington et~al.(2014)Pennington, Socher, and
  Manning]{pennington2014glove}
Jeffrey Pennington, Richard Socher, and Christopher~D Manning.
\newblock Glove: Global vectors for word representation.
\newblock In \emph{EMNLP}, 2014.

\bibitem[Pereyra et~al.(2017)Pereyra, Tucker, Chorowski, Kaiser, and
  Hinton]{pereyra2017regularizing}
Gabriel Pereyra, George Tucker, Jan Chorowski, {\L}ukasz Kaiser, and Geoffrey
  Hinton.
\newblock Regularizing neural networks by penalizing confident output
  distributions.
\newblock In \emph{ICLR Workshop}, 2017.

\bibitem[Poole et~al.(2014)Poole, Sohl-Dickstein, and
  Ganguli]{poole2014analyzing}
Ben Poole, Jascha Sohl-Dickstein, and Surya Ganguli.
\newblock Analyzing noise in autoencoders and deep networks.
\newblock \emph{arXiv}, 2014.

\bibitem[Prechelt(2002)]{prechelt2002early}
Lutz Prechelt.
\newblock Early stopping-but when?
\newblock In \emph{Neural Networks: Tricks of the trade}. Springer, 2002.

\bibitem[Rothe et~al.(2015)Rothe, Timofte, and Van~Gool]{rothe2015dex}
Rasmus Rothe, Radu Timofte, and Luc Van~Gool.
\newblock {DEX: Deep EXpectation of apparent age from a single image}.
\newblock In \emph{ICCV Workshop}, 2015.

\bibitem[Sietsma \& Dow(1991)Sietsma and Dow]{sietsma1991creating}
Jocelyn Sietsma and Robert~JF Dow.
\newblock Creating artificial neural networks that generalize.
\newblock \emph{Neural networks}, 4\penalty0 (1):\penalty0 67--79, 1991.

\bibitem[Socher et~al.(2013)Socher, Perelygin, Wu, Chuang, Manning, Ng, and
  Potts]{socher2013recursive}
Richard Socher, Alex Perelygin, Jean Wu, Jason Chuang, Christopher~D Manning,
  Andrew~Y Ng, and Christopher Potts.
\newblock Recursive deep models for semantic compositionality over a sentiment
  treebank.
\newblock In \emph{EMNLP}, 2013.

\bibitem[Song et~al.(2022)Song, Kim, Park, Shin, and Lee]{song2022learning}
Hwanjun Song, Minseok Kim, Dongmin Park, Yooju Shin, and Jae-Gil Lee.
\newblock {Learning from noisy labels with deep neural networks: A survey}.
\newblock \emph{Transactions on Neural Networks and Learning Systems}, 2022.

\bibitem[Srivastava et~al.(2014)Srivastava, Hinton, Krizhevsky, Sutskever, and
  Salakhutdinov]{srivastava2014dropout}
Nitish Srivastava, Geoffrey Hinton, Alex Krizhevsky, Ilya Sutskever, and Ruslan
  Salakhutdinov.
\newblock Dropout: a simple way to prevent neural networks from overfitting.
\newblock \emph{JMLR}, 2014.

\bibitem[Szegedy et~al.(2016)Szegedy, Vanhoucke, Ioffe, Shlens, and
  Wojna]{szegedy2016inception}
Christian Szegedy, Vincent Vanhoucke, Sergey Ioffe, Jon Shlens, and Zbigniew
  Wojna.
\newblock Rethinking the inception architecture for computer vision.
\newblock In \emph{CVPR}, 2016.

\bibitem[Taori et~al.(2020)Taori, Dave, Shankar, Carlini, Recht, and
  Schmidt]{taori2020measuring}
Rohan Taori, Achal Dave, Vaishaal Shankar, Nicholas Carlini, Benjamin Recht,
  and Ludwig Schmidt.
\newblock {Measuring robustness to natural distribution shifts in image
  classification}.
\newblock In \emph{NeurIPS}, 2020.

\bibitem[Tashiro et~al.(2020)Tashiro, Song, and Ermon]{tashiro2020diversity}
Yusuke Tashiro, Yang Song, and Stefano Ermon.
\newblock Diversity can be transferred: Output diversification for white-and
  black-box attacks.
\newblock In \emph{NeurIPS}, 2020.

\bibitem[Vaswani et~al.(2017)Vaswani, Shazeer, Parmar, Uszkoreit, Jones, Gomez,
  Kaiser, and Polosukhin]{vaswani2017attention}
Ashish Vaswani, Noam Shazeer, Niki Parmar, Jakob Uszkoreit, Llion Jones,
  Aidan~N Gomez, {\L}ukasz Kaiser, and Illia Polosukhin.
\newblock Attention is all you need.
\newblock In \emph{NeurIPS}, 2017.

\bibitem[Wan et~al.(2013)Wan, Zeiler, Zhang, Le~Cun, and
  Fergus]{wan2013regularization}
Li~Wan, Matthew Zeiler, Sixin Zhang, Yann Le~Cun, and Rob Fergus.
\newblock Regularization of neural networks using dropconnect.
\newblock In \emph{ICML}, 2013.

\bibitem[Wang et~al.(2019)Wang, Ge, Lipton, and Xing]{wang2019learning}
Haohan Wang, Songwei Ge, Zachary Lipton, and Eric~P Xing.
\newblock Learning robust global representations by penalizing local predictive
  power.
\newblock In \emph{NeurIPS}, 2019.

\bibitem[Wei et~al.(2020)Wei, Kakade, and Ma]{wei2020implicit}
Colin Wei, Sham Kakade, and Tengyu Ma.
\newblock The implicit and explicit regularization effects of dropout.
\newblock In \emph{ICML}, 2020.

\bibitem[Welling \& Teh(2011)Welling and Teh]{welling2011bayesian}
Max Welling and Yee~W Teh.
\newblock Bayesian learning via stochastic gradient langevin dynamics.
\newblock In \emph{ICML}, 2011.

\bibitem[Wu et~al.(2020)Wu, Hu, Xiong, Huan, Braverman, and Zhu]{wu2020noisy}
Jingfeng Wu, Wenqing Hu, Haoyi Xiong, Jun Huan, Vladimir Braverman, and
  Zhanxing Zhu.
\newblock On the noisy gradient descent that generalizes as {SGD}.
\newblock In \emph{ICML}, 2020.

\bibitem[Yao et~al.(2022)Yao, Wang, Zhang, Zou, and Finn]{yao2022c}
Huaxiu Yao, Yiping Wang, Linjun Zhang, James~Y Zou, and Chelsea Finn.
\newblock C-mixup: Improving generalization in regression.
\newblock In \emph{NeurIPS}, 2022.

\bibitem[Zhang et~al.(2018)Zhang, Cisse, Dauphin, and
  Lopez-Paz]{zhang2017mixup}
Hongyi Zhang, Moustapha Cisse, Yann~N Dauphin, and David Lopez-Paz.
\newblock mixup: Beyond empirical risk minimization.
\newblock In \emph{ICLR}, 2018.

\bibitem[Zhou et~al.(2019)Zhou, Liu, Li, Lin, Zhou, and Zhao]{zhou2019toward}
Mo~Zhou, Tianyi Liu, Yan Li, Dachao Lin, Enlu Zhou, and Tuo Zhao.
\newblock Toward understanding the importance of noise in training neural
  networks.
\newblock In \emph{ICML}, 2019.

\end{thebibliography}
\bibliographystyle{tmlr}

\appendix

\newpage

\section*{Appendix}

In the Appendix, we first provide the experimental settings and the hyperparameter ranges for all the experiments in Section~\ref{sec:appendix:settings}.
We then provide the full numerical results and visualisations for all the experiments in Section~\ref{sec:appendix:full_results}.

\section{Settings}\label{sec:appendix:settings}

\subsection{General Settings}\label{sec:appendix:general_settings}

We used stochastic gradient descent with a momentum of 0.9 to train all the networks. 
The learning rate and L2 regularisation were tuned and reused for each noise injection method. 
We used a cosine annealing learning rate schedule without restarts~\citep{loshchilov2016sgdr} for all experiments.
In most cases, we used gradient norm clipping of 20.0 to stabilise the training, with gradient clipping of 10.0 for tabular regression and 5.0 for WikiFace.
The batch size was set to 256 for all experiments.  
The final results are the average of 3 runs with 3 different seeds.
We used cross-entropy loss for all classification experiments.
For regression, we used the Gaussian negative log-likelihood (NLL) loss, where we modelled the variance as an additional output passed through an exponential function to ensure positivity.
We added a small $\epsilon$ of $1e^{-8}$ to the softmax probabilities to avoid NaNs. 
We clipped the variance between $1e^{-4}$ and $1e^{4}$ to avoid NaNs.
The hyperparameter ranges, and the sampling scale for each dataset-architecture pair are in Table~\ref{tab:hyperparameters}. 
The hyperparameters and implementations of all the noises and experiments can be found in the code, which will be open-sourced.
We used the default PyTorch weight initialisation for all layers. 

For the tabular OOD experiments, we constructed custom augmentations where we applied Gaussian or Uniform noise scaled by the magnitude of the input features across 5 severities for addition: [0.02, 0.04, 0.06, 0.08, 0.1] or multiplication [0.04, 0.08, 0.12, 0.16, 0.2] where the severity scaled the range or the standard deviation of the noise applied to the input. 
Additionally, we zeroed out some input features with probability [0.04, 0.08, 0.12, 0.16, 0.2], denoting 5 severities. 
In total, there were 5 different input shifts across 5 severities each.
To avoid label-flipping in applying these augmentations, we have introduced an empirically determined scaling factor for the severity of all noises for a particular dataset.
They multiply the [0.02, 0.04, 0.06, 0.08, 0.1] or [0.04, 0.08, 0.12, 0.16, 0.2] by a scaling factor to determine the severity of the noise applied to the input based on the dataset.
We use a K-nearest neighbour (KNN) classifier, specifically a 1-neighbour KNN, trained on a dataset's original, unmodified data. 
This classifier then predicts labels for the augmented data. 
We adjust the scaling factor for each dataset so that the KNN classifier's accuracy on the augmented data exceeds 99\% or the mean squared error is less than 0.01. 
This approach ensures that the augmentations are subtle enough to maintain the integrity of the data, meaning the nearest neighbour—the closest match in the original dataset—remains the same.
However, this is not a perfect solution, with an empirical guarantee that the augmentations are not too severe, and the nearest neighbour's label remains the same.
The found scaling factors are shown in Table \ref{tab:tabular_scaling_factors}.

Regarding noise implementation details, Dropout, DropConnect, additive weight or activation Gaussian noise, are applied to all linear and convolutional weights throughout the network, excluding the last layer and normalisation layers.
Both model and Gaussian gradient noise are implemented on all weights within the network, encompassing affine parameters in normalisation layers.
Rotation was omitted from AugMix, given that one of our tasks involved predicting the rotation angle.

\subsection{Vision Experiments}\label{sec:appendix:vision_settings}

For SVHN, we used a fully connected network with 4 hidden layers of 150 units followed by ReLU activations.
When we used ResNet-18 we used it with [64, 128, 256, 512] channels in 4 stages with [2, 2, 2, 2] blocks with strides [1, 2, 2, 2].
When we used ResNet-34, we used it with [64, 128, 256, 512] channels in 4 stages with [3, 4, 6, 3] blocks with strides [1, 2, 2, 2].
In all cases, we trained the networks for 200 epochs.
We only used 0-1 truncation followed by normalisation for each dataset without further data augmentations for training, validation and test sets.
For rotation experiments, we enabled uniform rotations between (0, 90°) degrees and rescaled the targets accordingly to [-1, 1]. 
Gaussian noise, motion blur, snow, elastic transformation, and JPEG compression were selected as OOD augmentations for visualisation experiments across all 5 severities.
For CIFAR-10, CIFAR-100, and SVHN we used the dedicated test sets as the test set, while for TinyImageNet we used the official validation set as the test set. We used 10\% of the training data to construct the validation sets.
For WikiFace, we used 10\% of the data as the test set and 10\% of the remaining data as the validation set and the rest as the training set.

For evaluation on the sketch domain, we utilise ImageNet-Sketch dataset from~\citet{wang2019learning} and derive our own TinyImageNet-Sketch dataset from it. 
As the ImageNet-Sketch images are not square, we crop the centre and then resize to $64\times64$, the same size as TinyImageNet. 
We only keep the images of the same 200 classes as used in TinyImageNet, allowing us to directly evaluate pre-trained models on the new TinyImageNet-Sketch dataset.

\subsection{NLP Classification Experiments}\label{sec:appendix:nlp_settings}
For our NLP experiments we used NewsGroup and binary SST datasets. In the NewsGroup dataset we aim to classify news texts into one of the 20 available categories based on the topic. The task in the binary SST dataset is to predict the sentiment of a movie review into 2 categories: positive or negative. For SST we only consider the text itself, rather than also considering the available parse trees, making the task more challenging.

Each dataset was first pre-processed with respect to glove embeddings~\citep{pennington2014glove} into embeddings of dimension 100 and sequence length 100 and 50 for NewsGroup and SST respectively.
In both cases, we trained the networks for 100 epochs.
We used the global-pooling convolutional network architecture from~\citet{kim2014convolutional} with planes [128, 128, 128] and a transformer decoder~\citep{vaswani2017attention} with embedding dimensions 100, 6 layers, 8 heads, 1024 hidden dimensions, 64 dimensions per head and no dropout.
For the NewsGroup experiments, we used about 5\% of the data as the test set, 10\% of the remaining data as the validation set and the rest as the training set. For the SST experiments, we used the original development set as the test set, for validation we took 10\% of the original training set and used the remainder as the training set. 
No OOD test was set for the NLP task due to a lack of suitable perturbations to construct OOD data.

\subsection{Tabular Regression Experiments}\label{sec:appendix:tabular_settings}

For the tabular experiments, we used a fully connected network with [100, 100, 100, 100] hidden units and ReLU activations.
In all cases, we trained the networks for 100 epochs.
We normalised the input features and targets to zero mean and unit variance by using the training set statistics and applied the same normalisation to the validation and test sets.
For the tabular experiments, we used 20\% of the data as the test set and 10\% of the remaining data as the validation set and the rest as the training set.
The regression targets were normalised to zero mean and unit variance. 


\section{Full Results}\label{sec:appendix:full_results}

We provide full results of all experiments in the paper, where the main reported value is the mean across 3 repetitions, followed by the standard deviation.
The ranks presented in the main body of the paper can be obtained by ranking the results in each table by the metric of interest.
Following the tables, there are the visualisations of metric landscapes for CIFAR-10, Adult, WikiFace and Yacht datasets.
We encourage the reader to look at our code for other datasets to regenerate them from there. 

\begin{table}[H]
\centering
\scalebox{1.0}{
\begin{tabular}{lcc}
\hline
\textbf{Hyperparameter ($\delta$)} & \textbf{Range} & \textbf{Scale} \\ \hline
Learning rate (LR) & $[10^{-4}, 10^{-1}]$ & Log\\
L2 weight & $[10^{-7}, 10^{-1}]$ & Log \\
Input Gaussian noise std. & $[10^{-4}, 10^{-1}]$ & Log  \\
Input AugMix alpha & $[0, 1]$ & Linear \\
Input AugMix severity & $[1, 10]$ & Linear \\
Input AugMix width & $[1, 5]$ & Linear \\
Input AugMix chain-depth & $[-1, 3]$ & Linear \\
Input ODS epsilon & $[10^{-4}, 10^{-1}]$ & Log  \\
Input ODS temperature & $[0.5, 5.0]$ & Log \\
Input-Target MixUp alpha & $[0, 1]$ & Linear\\
Input-Target CMixUp alpha & $[0, 1]$ & Linear\\
Input-Target CMixUp sigma & $[10^{-4}, 10^{2}]$ & Log \\
Target Label Smoothing& $[0, 0.25]$ & Linear \\
Activation Gaussian noise std & $[10^{-4}, 10^{-1}]$ & Log\\
Activation Dropout rate & $[0, 1]$ & Linear\\
Gradient Gaussian noise $\eta$ & $[0, 1]$ & Linear\\
Gradient Gaussian noise $\gamma$ & $[0, 1]$ & Linear\\
Weight Gaussian noise std & $[10^{-4}, 10^{-1}]$ & Log\\
Weight DropConnect rate & $[0, 1]$ & Linear\\
Model noise shrink factor & $[0.0, 1.0]$ & Linear\\
Model noise std & $[10^{-7}, 10^{-3}]$ & Log\\
Model noise frequency & $[0, 20]$ & Linear\\
\hline
\end{tabular}}
\caption{Hyperparameters (HPs) optimised for individual noises and their range.}
\label{tab:hyperparameters}
\end{table}

\begin{table}[h]
    \centering
    \begin{tabular}{lcc}
        \hline
        \textbf{Dataset} & \textbf{Task} & \textbf{Scaling Factor} \\ \hline
        Adult & Classification & 0.1438 \\
        Abalone & Classification & 0.0886 \\
        Concrete & Regression & 0.0207 \\
        Energy & Regression & 0.0546 \\
        Wine & Classification & 0.0078 \\
        Wine & Regression & 0.0048 \\
        Yacht & Regression & 0.0886 \\
        Toxicity & Classification & 0.3793 \\
        Students & Classification & 0.2336 \\
        Boston & Regression & 0.0886 \\ \hline
    \end{tabular}
    \caption{Scaling factors for tabular data.}
    \label{tab:tabular_scaling_factors}
\end{table}


\newpage

\begin{table}
\begin{center}
    \begin{small}
    \begin{sc}
\begin{adjustbox}{max width=\linewidth}
\begin{tabular}{lcccccccc}
\toprule
\multirow{2}{*}{\textbf{Noise Type}}& \multicolumn{2}{c}{\textbf{SVHN}} & \multicolumn{2}{c}{\textbf{CIFAR-10}}  & \multicolumn{2}{c}{\textbf{CIFAR-100}} & \multicolumn{2}{c}{\textbf{TinyImageNet}}\\
\cmidrule(lr){2-3} \cmidrule(lr){4-5} \cmidrule(lr){6-7}\cmidrule(lr){8-9}
 & \textbf{ID}  & \textbf{OOD} & \textbf{ID}  & \textbf{OOD} & \textbf{ID}  & \textbf{OOD} & \textbf{ID}  & \textbf{OOD}\\
\midrule
 No Noise                    & 16.33\tiny{$\pm$0.07} & 20.18\tiny{$\pm$0.09} & 12.04\tiny{$\pm$0.21}           & 32.32\tiny{$\pm$0.24} & 44.69\tiny{$\pm$0.84} & 61.52\tiny{$\pm$0.52} & 54.15\tiny{$\pm$0.36}  & 75.65\tiny{$\pm$0.19} \\
 Input Weak Aug.             & 14.67\tiny{$\pm$0.14} & 18.59\tiny{$\pm$0.14} & \phantom{0}5.39\tiny{$\pm$0.17} & 27.86\tiny{$\pm$0.37} & 26.51\tiny{$\pm$0.10} & 52.57\tiny{$\pm$0.13} & 39.73\tiny{$\pm$0.27}  & 67.73\tiny{$\pm$0.13} \\
 Input Gaussian           & 16.45\tiny{$\pm$0.23} & 20.23\tiny{$\pm$0.22} & 12.02\tiny{$\pm$0.05}           & 31.97\tiny{$\pm$0.31} & 44.34\tiny{$\pm$1.70} & 61.14\tiny{$\pm$1.60} & 53.28\tiny{$\pm$0.24}  & 75.12\tiny{$\pm$0.06} \\
 Input ODS                   & 16.35\tiny{$\pm$0.20} & 20.13\tiny{$\pm$0.15} & 12.01\tiny{$\pm$0.16}           & 30.42\tiny{$\pm$0.45} & 44.28\tiny{$\pm$0.64} & 61.34\tiny{$\pm$0.56} & 66.47\tiny{$\pm$15.18} & 82.46\tiny{$\pm$8.72} \\
 Input AugMix                & 12.28\tiny{$\pm$0.07} & 15.67\tiny{$\pm$0.07} & \phantom{0}7.48\tiny{$\pm$0.06} & 18.75\tiny{$\pm$0.25} & 30.09\tiny{$\pm$0.25} & 46.05\tiny{$\pm$0.13} & 42.49\tiny{$\pm$0.24}  & 61.58\tiny{$\pm$0.16} \\
 Input-Target MixUp          & 13.95\tiny{$\pm$0.01} & 17.71\tiny{$\pm$0.11} & 10.97\tiny{$\pm$0.13}           & 28.38\tiny{$\pm$0.70} & 46.15\tiny{$\pm$0.05} & 63.00\tiny{$\pm$0.02} & 54.02\tiny{$\pm$0.53}  & 75.33\tiny{$\pm$0.44} \\
\bottomrule
\end{tabular}
\end{adjustbox}
\end{sc}
\end{small}
\end{center}
\caption{CV classification: Error $(\downarrow,\%)$ comparison on in-distribution (ID) and out-of-distribution (OOD) test sets and with tuned hyperparameters.}
\label{tab:cv_classification:error}
\end{table}
\begin{table}
\begin{center}
    \begin{small}
    \begin{sc}
    \begin{adjustbox}{max width=\linewidth}
\begin{tabular}{lcccccccc}
\toprule
\multirow{2}{*}{\textbf{Noise Type}}& \multicolumn{2}{c}{\textbf{SVHN}} & \multicolumn{2}{c}{\textbf{CIFAR-10}}  & \multicolumn{2}{c}{\textbf{CIFAR-100}} & \multicolumn{2}{c}{\textbf{TinyImageNet}}\\
\cmidrule(lr){2-3} \cmidrule(lr){4-5} \cmidrule(lr){6-7}\cmidrule(lr){8-9}
 & \textbf{ID}  & \textbf{OOD} & \textbf{ID}  & \textbf{OOD} & \textbf{ID}  & \textbf{OOD} & \textbf{ID}  & \textbf{OOD}\\
\midrule
 No Noise                    & 13.10\tiny{$\pm$0.05}           & 15.59\tiny{$\pm$0.06}           & \phantom{0}4.02\tiny{$\pm$0.15} & 16.40\tiny{$\pm$0.36}           & \phantom{0}5.76\tiny{$\pm$0.25} & \phantom{0}7.43\tiny{$\pm$0.09} & 16.43\tiny{$\pm$0.27}           & 10.11\tiny{$\pm$0.11}           \\
 Input Weak Aug.             & \phantom{0}6.35\tiny{$\pm$0.19} & \phantom{0}7.85\tiny{$\pm$0.24} & \phantom{0}1.92\tiny{$\pm$0.23} & 15.91\tiny{$\pm$0.67}           & \phantom{0}4.83\tiny{$\pm$0.19} & 13.28\tiny{$\pm$0.14}           & \phantom{0}6.49\tiny{$\pm$0.28} & 11.46\tiny{$\pm$0.25}           \\
 Input Gaussian           & 13.24\tiny{$\pm$0.17}           & 15.67\tiny{$\pm$0.15}           & \phantom{0}4.22\tiny{$\pm$0.38} & 16.65\tiny{$\pm$0.59}           & \phantom{0}5.42\tiny{$\pm$0.46} & \phantom{0}7.70\tiny{$\pm$0.48} & 15.96\tiny{$\pm$1.37}           & 10.15\tiny{$\pm$0.45}           \\
 Input ODS                   & 13.03\tiny{$\pm$0.13}           & 15.43\tiny{$\pm$0.12}           & \phantom{0}4.26\tiny{$\pm$0.13} & 15.38\tiny{$\pm$0.27}           & \phantom{0}5.86\tiny{$\pm$0.30} & \phantom{0}7.48\tiny{$\pm$0.17} & 26.97\tiny{$\pm$19.11}          & 28.20\tiny{$\pm$26.53}          \\
 Input AugMix                & \phantom{0}4.81\tiny{$\pm$0.07} & \phantom{0}6.03\tiny{$\pm$0.09} & \phantom{0}1.18\tiny{$\pm$0.06} & \phantom{0}6.27\tiny{$\pm$0.39} & \phantom{0}4.41\tiny{$\pm$0.39} & 11.38\tiny{$\pm$0.24}           & \phantom{0}4.87\tiny{$\pm$0.31} & 15.78\tiny{$\pm$0.31}           \\
\bottomrule
\end{tabular}
\end{adjustbox}
\end{sc}
\end{small}
\end{center}
\caption{CV classification: ECE $(\downarrow,\%)$ comparison on in-distribution (ID) and out-of-distribution (OOD) test sets and with tuned hyperparameters.}
\label{tab:cv_classification:ece}
\end{table}
\begin{table}
\begin{center}
    \begin{small}
    \begin{sc}
\begin{adjustbox}{max width=\linewidth}
\begin{tabular}{lcccccccc}
\toprule
\multirow{2}{*}{\textbf{Noise Type}}& \multicolumn{2}{c}{\textbf{SVHN}} & \multicolumn{2}{c}{\textbf{CIFAR-10}}  & \multicolumn{2}{c}{\textbf{CIFAR-100}} & \multicolumn{2}{c}{\textbf{TinyImageNet}}\\
\cmidrule(lr){2-3} \cmidrule(lr){4-5} \cmidrule(lr){6-7}\cmidrule(lr){8-9}
 & \textbf{ID}  & \textbf{OOD} & \textbf{ID}  & \textbf{OOD} & \textbf{ID}  & \textbf{OOD} & \textbf{ID}  & \textbf{OOD}\\
\midrule
 No Noise                    & \phantom{0}1.43\tiny{$\pm$0.00} & \phantom{0}1.64\tiny{$\pm$0.01} & \phantom{0}0.42\tiny{$\pm$0.01} & \phantom{0}1.21\tiny{$\pm$0.02} & \phantom{0}1.85\tiny{$\pm$0.03} & \phantom{0}2.69\tiny{$\pm$0.03} & \phantom{0}2.80\tiny{$\pm$0.02} & \phantom{0}3.90\tiny{$\pm$0.01} \\
 Input Weak Aug.             & \phantom{0}0.62\tiny{$\pm$0.01} & \phantom{0}0.75\tiny{$\pm$0.01} & \phantom{0}0.20\tiny{$\pm$0.00} & \phantom{0}1.10\tiny{$\pm$0.03} & \phantom{0}1.07\tiny{$\pm$0.01} & \phantom{0}2.44\tiny{$\pm$0.00} & \phantom{0}1.84\tiny{$\pm$0.02} & \phantom{0}3.46\tiny{$\pm$0.02} \\
 Input Gaussian           & \phantom{0}1.43\tiny{$\pm$0.01} & \phantom{0}1.65\tiny{$\pm$0.01} & \phantom{0}0.42\tiny{$\pm$0.01} & \phantom{0}1.21\tiny{$\pm$0.03} & \phantom{0}1.84\tiny{$\pm$0.08} & \phantom{0}2.67\tiny{$\pm$0.08} & \phantom{0}2.75\tiny{$\pm$0.06} & \phantom{0}3.86\tiny{$\pm$0.02} \\
 Input ODS                   & \phantom{0}1.40\tiny{$\pm$0.02} & \phantom{0}1.60\tiny{$\pm$0.02} & \phantom{0}0.42\tiny{$\pm$0.01} & \phantom{0}1.14\tiny{$\pm$0.02} & \phantom{0}1.84\tiny{$\pm$0.03} & \phantom{0}2.69\tiny{$\pm$0.03} & \phantom{0}4.79\tiny{$\pm$2.79} & \phantom{0}6.23\tiny{$\pm$3.27} \\
 Input AugMix                & \phantom{0}0.49\tiny{$\pm$0.01} & \phantom{0}0.61\tiny{$\pm$0.00} & \phantom{0}0.25\tiny{$\pm$0.00} & \phantom{0}0.63\tiny{$\pm$0.01} & \phantom{0}1.17\tiny{$\pm$0.00} & \phantom{0}2.04\tiny{$\pm$0.01} & \phantom{0}1.87\tiny{$\pm$0.01} & \phantom{0}3.14\tiny{$\pm$0.01} \\
\bottomrule
\end{tabular}
\end{adjustbox}
\end{sc}
\end{small}
\end{center}
\caption{CV classification: NLL $(\downarrow)$ comparison on in-distribution (ID) and out-of-distribution (OOD) test sets and with tuned hyperparameters.}
\label{tab:cv_classification:nll}
\end{table}

\begin{table}
\begin{center}
\begin{small}
\begin{sc}
\begin{adjustbox}{max width=\linewidth}
\begin{tabular}{lcccccc}
\toprule
\multirow{2}{*}{\textbf{Noise Type}} & \multicolumn{3}{c}{\textbf{Scores}} & \multicolumn{3}{c}{\textbf{Ranks}} \\
\cmidrule(lr){2-4} \cmidrule(lr){5-7} 
& \textbf{ID} & \textbf{OOD} & \textbf{Sketch} & \textbf{ID} & \textbf{OOD} & \textbf{Sketch} \\
\midrule
No Noise                    & 54.15\tiny{$\pm$0.36}  & 75.65\tiny{$\pm$0.19} & 89.61\tiny{$\pm$0.35} & 13.0 & 13.0 & 16.0 \\
Input Weak Aug.             & 39.73\tiny{$\pm$0.27}  & 67.73\tiny{$\pm$0.13} & 85.90\tiny{$\pm$0.56} & \phantom{0}5.0 &  \phantom{0}6.0 &  \phantom{0}6.0 \\
Input Gaussian              & 53.28\tiny{$\pm$0.24}  & 75.12\tiny{$\pm$0.06} & 89.29\tiny{$\pm$0.07} & 10.0 & 10.0 & 13.0 \\
Input ODS                   & 66.47\tiny{$\pm$15.18} & 82.46\tiny{$\pm$8.72} & 89.54\tiny{$\pm$0.61} & 17.0 & 17.0 & 15.0 \\
Input AugMix                & 42.49\tiny{$\pm$0.24}  & 61.58\tiny{$\pm$0.16} & 83.12\tiny{$\pm$0.23} & \phantom{0}6.0 &  \phantom{0}5.0 &  \phantom{0}4.0 \\
Input-Target MixUp          & 54.02\tiny{$\pm$0.53}  & 75.33\tiny{$\pm$0.44} & 89.27\tiny{$\pm$0.09} & 12.0 & 11.0 & 12.0 \\
\bottomrule
\end{tabular}
\end{adjustbox}
\end{sc}
\end{small}
\end{center}
\caption{TinyImageNet classification: Error $(\downarrow,\%)$ comparison on in-distribution (ID) and out-of-distribution (OOD) and Sketch test sets and with tuned hyperparameters.}
\label{tab:tiny_classification:error}
\end{table}
\begin{table}
\begin{center}
\begin{small}
\begin{sc}
\begin{adjustbox}{max width=\linewidth}
\begin{tabular}{lcccccc}
\toprule
\multirow{2}{*}{\textbf{Noise Type}} & \multicolumn{3}{c}{\textbf{Scores}} & \multicolumn{3}{c}{\textbf{Ranks}} \\
\cmidrule(lr){2-4} \cmidrule(lr){5-7} 
& \textbf{ID} & \textbf{OOD} & \textbf{Sketch} & \textbf{ID} & \textbf{OOD} & \textbf{Sketch} \\
\midrule
No Noise                    & 16.43\tiny{$\pm$0.27}           & 10.11\tiny{$\pm$0.11}           & \phantom{0}8.67\tiny{$\pm$0.46} & 14.0 &  \phantom{0}4.0 &  \phantom{0}5.0 \\
Input Weak Aug.             & \phantom{0}6.49\tiny{$\pm$0.28} & 11.46\tiny{$\pm$0.25}           & 18.10\tiny{$\pm$0.52}           & \phantom{0}4.0 &  \phantom{0}8.0 & 10.0 \\
Input Gaussian              & 15.96\tiny{$\pm$1.37}           & 10.15\tiny{$\pm$0.45}           & \phantom{0}7.53\tiny{$\pm$1.13} & 12.0 &  \phantom{0}6.0 &  \phantom{0}3.0 \\
Input ODS                   & 26.97\tiny{$\pm$19.11}          & 28.20\tiny{$\pm$26.53}          & 11.31\tiny{$\pm$1.74}           & 16.0 & 16.0 &  \phantom{0}9.0 \\
Input AugMix                & \phantom{0}4.87\tiny{$\pm$0.31} & 15.78\tiny{$\pm$0.31}           & 20.08\tiny{$\pm$0.82}           & \phantom{0}3.0 & 11.0 & 11.0 \\
Input-Target MixUp          & 16.33\tiny{$\pm$1.40}           & 10.26\tiny{$\pm$0.32}           & \phantom{0}7.28\tiny{$\pm$1.06} & 13.0 &  \phantom{0}7.0 &  \phantom{0}2.0 \\
\bottomrule
\end{tabular}
\end{adjustbox}
\end{sc}
\end{small}
\end{center}
\caption{TinyImageNet classification: ECE $(\downarrow,\%)$ comparison on in-distribution (ID) and out-of-distribution (OOD) and Sketch test sets and with tuned hyperparameters.}
\label{tab:tiny_classification:ece}
\end{table}
\begin{table}
\begin{center}
\begin{small}
\begin{sc}
\begin{adjustbox}{max width=\linewidth}
\begin{tabular}{lcccccc}
\toprule
\multirow{2}{*}{\textbf{Noise Type}} & \multicolumn{3}{c}{\textbf{Scores}} & \multicolumn{3}{c}{\textbf{Ranks}} \\
\cmidrule(lr){2-4} \cmidrule(lr){5-7} 
& \textbf{ID} & \textbf{OOD} & \textbf{Sketch} & \textbf{ID} & \textbf{OOD} & \textbf{Sketch} \\
\midrule
No Noise                    & \phantom{0}2.80\tiny{$\pm$0.02} & \phantom{0}3.90\tiny{$\pm$0.01} & \phantom{0}4.68\tiny{$\pm$0.01} & 14.0 &  14.0 &  \phantom{0}8.0 \\
Input Weak Aug.             & \phantom{0}1.84\tiny{$\pm$0.02} & \phantom{0}3.46\tiny{$\pm$0.02} & \phantom{0}4.69\tiny{$\pm$0.02} & \phantom{0}5.0 &  \phantom{0}6.0 &  \phantom{0}9.0 \\
Input Gaussian              & \phantom{0}2.75\tiny{$\pm$0.06} & \phantom{0}3.86\tiny{$\pm$0.02} & \phantom{0}4.65\tiny{$\pm$0.02} & 10.0 &  \phantom{0}9.0 &  \phantom{0}5.0 \\
Input ODS                   & \phantom{0}4.79\tiny{$\pm$2.79} & \phantom{0}6.23\tiny{$\pm$3.27} & \phantom{0}4.72\tiny{$\pm$0.06} & 17.0 & 17.0 & 12.0 \\
Input AugMix                & \phantom{0}1.87\tiny{$\pm$0.01} & \phantom{0}3.14\tiny{$\pm$0.01} & \phantom{0}4.52\tiny{$\pm$0.01} & \phantom{0}6.0 &  \phantom{0}3.0 &  \phantom{0}1.0 \\
Input-Target MixUp          & \phantom{0}2.79\tiny{$\pm$0.07} & \phantom{0}3.88\tiny{$\pm$0.04} & \phantom{0}4.66\tiny{$\pm$0.01} & 13.0 & 11.0 &  \phantom{0}6.0 \\
\bottomrule
\end{tabular}
\end{adjustbox}
\end{sc}
\end{small}
\end{center}
\caption{TinyImageNet classification: NLL $(\downarrow)$ comparison on in-distribution (ID), out-of-distribution (OOD) and Sketch test sets and with tuned hyperparameters.}
\label{tab:tiny_classification:nll}
\end{table}

\begin{table}
\begin{center}
\begin{small}
\begin{sc}
\begin{adjustbox}{max width=\linewidth}
\begin{tabular}{lccc}
\toprule
\textbf{Metric} & ID vs OOD & ID vs Sketch & OOD vs Sketch \\
\midrule
Error & 0.912\tiny{$\pm$0.000} & 0.794\tiny{$\pm$0.000} & 0.824\tiny{$\pm$0.000} \\
ECE & 0.118\tiny{$\pm$0.542} & -0.397\tiny{$\pm$0.027} & 0.397\tiny{$\pm$0.027} \\
NLL & 0.878\tiny{$\pm$0.000} & 0.029\tiny{$\pm$0.903} & 0.081\tiny{$\pm$0.650} \\
\bottomrule
\end{tabular}
\end{adjustbox}
\end{sc}
\end{small}
\end{center}
\caption{TinyImageNet classification: Kendall Tau correlation between ID, OOD and Sketch rankings of different noise types.}
\label{tab:tiny_classification:kendalltau}
\end{table}

\begin{table}
\begin{center}
    \begin{small}
    \begin{sc}
\begin{adjustbox}{max width=\linewidth}
\begin{tabular}{lcccccc}
\toprule
\multirow{2}{*}{\textbf{Noise Type}}& \multicolumn{2}{c}{\textbf{CIFAR-10}} & \multicolumn{2}{c}{\textbf{CIFAR-100}}  & \multicolumn{2}{c}{\textbf{TinyImageNet}} \\
\cmidrule(lr){2-3} \cmidrule(lr){4-5} \cmidrule(lr){6-7}
 & \textbf{ID}  & \textbf{OOD} & \textbf{ID}  & \textbf{OOD} & \textbf{ID}  & \textbf{OOD} \\
\midrule
 No Noise               & 16.09\tiny{$\pm$0.18}           & 30.81\tiny{$\pm$0.76} & 40.97\tiny{$\pm$0.40}  & 61.56\tiny{$\pm$0.11}  & 54.70\tiny{$\pm$0.81} & 75.96\tiny{$\pm$0.52} \\
 Input Weak Aug.        & \phantom{0}7.99\tiny{$\pm$0.07} & 27.60\tiny{$\pm$0.42} & 24.03\tiny{$\pm$0.09}  & 52.82\tiny{$\pm$0.13}  & 39.60\tiny{$\pm$0.26} & 67.41\tiny{$\pm$0.02} \\
 Input Gaussian      & 16.72\tiny{$\pm$0.13}           & 30.28\tiny{$\pm$0.85} & 41.07\tiny{$\pm$0.25}  & 61.04\tiny{$\pm$0.14}  & 54.01\tiny{$\pm$0.60} & 75.31\tiny{$\pm$0.33} \\
 Input ODS              & 15.79\tiny{$\pm$0.13}           & 29.54\tiny{$\pm$0.71} & 41.13\tiny{$\pm$0.43}  & 60.43\tiny{$\pm$0.39}  & 52.77\tiny{$\pm$0.54} & 74.61\tiny{$\pm$0.11} \\
 Input AugMix           & 10.26\tiny{$\pm$0.04}           & 20.95\tiny{$\pm$0.06} & 30.18\tiny{$\pm$0.46}  & 45.94\tiny{$\pm$0.29}  & 40.04\tiny{$\pm$0.29} & 60.67\tiny{$\pm$0.15} \\
 Input-Target MixUp     & 16.90\tiny{$\pm$0.17}           & 32.62\tiny{$\pm$0.36} & 39.04\tiny{$\pm$0.30}  & 58.48\tiny{$\pm$0.03}  & 51.49\tiny{$\pm$0.27} & 72.87\tiny{$\pm$0.12} \\
\bottomrule
\end{tabular}
\end{adjustbox}
\end{sc}
\end{small}
\end{center}
\caption{CV classification: Error $(\downarrow,\%)$ comparison on in-distribution (ID) and out-of-distribution (OOD) test sets and with hyperparameters transferred across datasets.}
\label{tab:transfer:dataset:cv_classification:error}
\end{table}
\begin{table}
\begin{center}
    \begin{small}
    \begin{sc}
\begin{adjustbox}{max width=\linewidth}
\begin{tabular}{lcccccc}
\toprule
\multirow{2}{*}{\textbf{Noise Type}}& \multicolumn{2}{c}{\textbf{CIFAR-10}} & \multicolumn{2}{c}{\textbf{CIFAR-100}}  & \multicolumn{2}{c}{\textbf{TinyImageNet}} \\
\cmidrule(lr){2-3} \cmidrule(lr){4-5} \cmidrule(lr){6-7}
 & \textbf{ID}  & \textbf{OOD} & \textbf{ID}  & \textbf{OOD} & \textbf{ID}  & \textbf{OOD} \\
\midrule
 No Noise               & \phantom{0}9.76\tiny{$\pm$0.07} & 20.73\tiny{$\pm$0.67}           & 12.77\tiny{$\pm$0.86}           & 10.73\tiny{$\pm$0.38}           & 20.11\tiny{$\pm$2.25}           & 11.66\tiny{$\pm$0.63}           \\
 Input Weak Aug.        & \phantom{0}5.40\tiny{$\pm$0.09} & 20.98\tiny{$\pm$0.62}           & \phantom{0}3.65\tiny{$\pm$0.38} & 11.12\tiny{$\pm$0.59}           & \phantom{0}5.97\tiny{$\pm$0.40} & 11.75\tiny{$\pm$0.48}           \\
 Input Gaussian      & 10.39\tiny{$\pm$0.15}           & 20.21\tiny{$\pm$0.76}           & 12.81\tiny{$\pm$0.25}           & 10.66\tiny{$\pm$0.16}           & 16.74\tiny{$\pm$0.67}           & 10.72\tiny{$\pm$0.16}           \\
 Input ODS              & \phantom{0}9.89\tiny{$\pm$0.12} & 20.40\tiny{$\pm$0.75}           & 11.18\tiny{$\pm$2.18}           & 10.28\tiny{$\pm$1.07}           & 15.98\tiny{$\pm$0.37}           & 10.52\tiny{$\pm$0.22}           \\
 Input AugMix           & \phantom{0}5.53\tiny{$\pm$0.08} & 11.65\tiny{$\pm$0.07}           & \phantom{0}8.94\tiny{$\pm$0.07} & \phantom{0}8.86\tiny{$\pm$0.05} & \phantom{0}3.97\tiny{$\pm$0.20} & 15.43\tiny{$\pm$0.53}           \\
 Input-Target MixUp     & \phantom{0}6.27\tiny{$\pm$0.51} & \phantom{0}8.12\tiny{$\pm$0.20} & 14.76\tiny{$\pm$0.48}           & 10.99\tiny{$\pm$0.06}           & 19.33\tiny{$\pm$0.27}           & 10.08\tiny{$\pm$0.32}           \\
\bottomrule
\end{tabular}
\end{adjustbox}
\end{sc}
\end{small}
\end{center}
\caption{CV classification: ECE $(\downarrow,\%)$ comparison on in-distribution (ID) and out-of-distribution (OOD) test sets and with hyperparameters transferred across datasets.}
\label{tab:transfer:dataset:cv_classification:ece}
\end{table}
\begin{table}
\begin{center}
    \begin{small}
    \begin{sc}
\begin{adjustbox}{max width=\linewidth}
\begin{tabular}{lcccccc}
\toprule
\multirow{2}{*}{\textbf{Noise Type}}& \multicolumn{2}{c}{\textbf{CIFAR-10}} & \multicolumn{2}{c}{\textbf{CIFAR-100}}  & \multicolumn{2}{c}{\textbf{TinyImageNet}} \\
\cmidrule(lr){2-3} \cmidrule(lr){4-5} \cmidrule(lr){6-7}
 & \textbf{ID}  & \textbf{OOD} & \textbf{ID}  & \textbf{OOD} & \textbf{ID}  & \textbf{OOD} \\
\midrule
 No Noise               & \phantom{0}0.71\tiny{$\pm$0.01} & \phantom{0}1.55\tiny{$\pm$0.05} & \phantom{0}1.88\tiny{$\pm$0.02} & \phantom{0}2.87\tiny{$\pm$0.01} & \phantom{0}2.96\tiny{$\pm$0.12} & \phantom{0}3.99\tiny{$\pm$0.06} \\
 Input Weak Aug.        & \phantom{0}0.40\tiny{$\pm$0.00} & \phantom{0}1.84\tiny{$\pm$0.09} & \phantom{0}1.02\tiny{$\pm$0.01} & \phantom{0}2.45\tiny{$\pm$0.02} & \phantom{0}1.80\tiny{$\pm$0.01} & \phantom{0}3.43\tiny{$\pm$0.02} \\
 Input Gaussian      & \phantom{0}0.72\tiny{$\pm$0.01} & \phantom{0}1.50\tiny{$\pm$0.06} & \phantom{0}1.89\tiny{$\pm$0.01} & \phantom{0}2.84\tiny{$\pm$0.01} & \phantom{0}2.81\tiny{$\pm$0.06} & \phantom{0}3.91\tiny{$\pm$0.03} \\
 Input ODS              & \phantom{0}0.71\tiny{$\pm$0.00} & \phantom{0}1.55\tiny{$\pm$0.07} & \phantom{0}1.87\tiny{$\pm$0.07} & \phantom{0}2.80\tiny{$\pm$0.05} & \phantom{0}2.72\tiny{$\pm$0.02} & \phantom{0}3.85\tiny{$\pm$0.00} \\
 Input AugMix           & \phantom{0}0.39\tiny{$\pm$0.00} & \phantom{0}0.84\tiny{$\pm$0.00} & \phantom{0}1.30\tiny{$\pm$0.01} & \phantom{0}2.02\tiny{$\pm$0.01} & \phantom{0}1.73\tiny{$\pm$0.01} & \phantom{0}3.10\tiny{$\pm$0.02} \\
 Input-Target MixUp     & \phantom{0}0.56\tiny{$\pm$0.01} & \phantom{0}1.04\tiny{$\pm$0.02} & \phantom{0}1.71\tiny{$\pm$0.01} & \phantom{0}2.59\tiny{$\pm$0.00} & \phantom{0}2.62\tiny{$\pm$0.02} & \phantom{0}3.65\tiny{$\pm$0.01} \\
\bottomrule
\end{tabular}
\end{adjustbox}
\end{sc}
\end{small}
\end{center}
\caption{CV classification: NLL $(\downarrow)$ comparison on in-distribution (ID) and out-of-distribution (OOD) test sets and with hyperparameters transferred across datasets.}
\label{tab:transfer:dataset:cv_classification:nll}
\end{table}
\begin{table}
\begin{center}
    \begin{small}
    \begin{sc}
\begin{adjustbox}{max width=\linewidth}
\begin{tabular}{lcccccc}
\toprule
 \multirow{2}{*}{\textbf{Noise Type}}& \multicolumn{2}{c}{\textbf{SVHN}} & \multicolumn{2}{c}{\textbf{CIFAR-100}}  & \multicolumn{2}{c}{\textbf{TinyImageNet}} \\
\cmidrule(lr){2-3} \cmidrule(lr){4-5} \cmidrule(lr){6-7}
 & \textbf{ID}  & \textbf{OOD} & \textbf{ID}  & \textbf{OOD} & \textbf{ID}  & \textbf{OOD} \\
\midrule
 No Noise               & \phantom{0}5.12\tiny{$\pm$0.13} & \phantom{0}9.20\tiny{$\pm$0.10} & 43.87\tiny{$\pm$0.48} & 61.25\tiny{$\pm$0.46} & 53.63\tiny{$\pm$0.17} & 75.20\tiny{$\pm$0.32} \\
 Input Weak Aug.        & \phantom{0}4.13\tiny{$\pm$0.10} & \phantom{0}8.51\tiny{$\pm$0.23} & 27.33\tiny{$\pm$0.35} & 51.09\tiny{$\pm$0.18} & 38.21\tiny{$\pm$0.24} & 64.75\tiny{$\pm$0.29} \\
 Input Gaussian      & \phantom{0}5.01\tiny{$\pm$0.08} & \phantom{0}9.09\tiny{$\pm$0.06} & 43.05\tiny{$\pm$0.35} & 58.71\tiny{$\pm$0.32} & 54.15\tiny{$\pm$1.59} & 74.61\tiny{$\pm$1.09} \\
 Input AugMix           & \phantom{0}3.51\tiny{$\pm$0.05} & \phantom{0}8.27\tiny{$\pm$0.04} & 30.23\tiny{$\pm$0.06} & 45.51\tiny{$\pm$0.17} & 42.05\tiny{$\pm$0.29} & 59.93\tiny{$\pm$0.10} \\
 Input-Target MixUp     & \phantom{0}5.58\tiny{$\pm$0.12} & 12.75\tiny{$\pm$0.11}           & 43.93\tiny{$\pm$0.68} & 59.98\tiny{$\pm$0.23} & 52.95\tiny{$\pm$1.26} & 73.82\tiny{$\pm$0.54} \\
 Label Smoothing        & \phantom{0}5.04\tiny{$\pm$0.03} & \phantom{0}8.88\tiny{$\pm$0.01} & 42.99\tiny{$\pm$0.36} & 57.27\tiny{$\pm$0.43} & 53.77\tiny{$\pm$0.32} & 74.25\tiny{$\pm$0.11} \\
\bottomrule
\end{tabular}
\end{adjustbox}
\end{sc}
\end{small}
\end{center}
\caption{CV classification: Error $(\downarrow,\%)$ comparison on in-distribution (ID) and out-of-distribution (OOD) test sets and with hyperparameters transferred across architectures.}
\label{tab:transfer:architecture:cv_classification:error}
\end{table}
\begin{table}
\begin{center}
    \begin{small}
    \begin{sc}
\begin{adjustbox}{max width=\linewidth}
\begin{tabular}{lcccccc}
\toprule
 \multirow{2}{*}{\textbf{Noise Type}}& \multicolumn{2}{c}{\textbf{SVHN}} & \multicolumn{2}{c}{\textbf{CIFAR-100}}  & \multicolumn{2}{c}{\textbf{TinyImageNet}} \\
\cmidrule(lr){2-3} \cmidrule(lr){4-5} \cmidrule(lr){6-7}
 & \textbf{ID}  & \textbf{OOD} & \textbf{ID}  & \textbf{OOD} & \textbf{ID}  & \textbf{OOD} \\
\midrule
 No Noise               & \phantom{0}2.73\tiny{$\pm$0.10} & \phantom{0}5.13\tiny{$\pm$0.06} & \phantom{0}5.76\tiny{$\pm$0.34} & \phantom{0}7.50\tiny{$\pm$0.11} & 14.68\tiny{$\pm$0.47}           & \phantom{0}9.93\tiny{$\pm$0.08} \\
 Input Weak Aug.        & \phantom{0}2.66\tiny{$\pm$0.05} & \phantom{0}5.75\tiny{$\pm$0.09} & \phantom{0}6.33\tiny{$\pm$0.32} & 15.26\tiny{$\pm$0.80}           & \phantom{0}8.20\tiny{$\pm$1.23} & \phantom{0}9.88\tiny{$\pm$0.38} \\
 Input Gaussian      & \phantom{0}2.68\tiny{$\pm$0.05} & \phantom{0}5.09\tiny{$\pm$0.03} & \phantom{0}5.07\tiny{$\pm$0.61} & \phantom{0}9.59\tiny{$\pm$1.63} & 13.79\tiny{$\pm$0.49}           & \phantom{0}8.75\tiny{$\pm$0.46} \\
 Input AugMix           & \phantom{0}1.46\tiny{$\pm$0.04} & \phantom{0}3.04\tiny{$\pm$0.07} & \phantom{0}6.56\tiny{$\pm$0.71} & 14.17\tiny{$\pm$0.59}           & \phantom{0}6.23\tiny{$\pm$0.63} & 16.60\tiny{$\pm$0.23}           \\
 Input-Target MixUp     & \phantom{0}5.61\tiny{$\pm$1.19} & \phantom{0}5.27\tiny{$\pm$0.90} & \phantom{0}2.68\tiny{$\pm$0.26} & \phantom{0}6.77\tiny{$\pm$0.43} & 12.24\tiny{$\pm$0.82}           & \phantom{0}8.32\tiny{$\pm$0.10} \\
 Label Smoothing        & \phantom{0}0.95\tiny{$\pm$0.07} & \phantom{0}1.16\tiny{$\pm$0.03} & 10.18\tiny{$\pm$2.84}           & \phantom{0}8.54\tiny{$\pm$1.72} & 17.47\tiny{$\pm$0.22}           & \phantom{0}9.76\tiny{$\pm$0.12} \\
\bottomrule
\end{tabular}
\end{adjustbox}
\end{sc}
\end{small}
\end{center}
\caption{CV classification: ECE $(\downarrow,\%)$ comparison on in-distribution (ID) and out-of-distribution (OOD) test sets and with hyperparameters transferred across architectures.}
\label{tab:transfer:architecture:cv_classification:ece}
\end{table}
\begin{table}
\begin{center}
    \begin{small}
    \begin{sc}
\begin{adjustbox}{max width=\linewidth}
\begin{tabular}{lcccccc}
\toprule
 \multirow{2}{*}{\textbf{Noise Type}}& \multicolumn{2}{c}{\textbf{SVHN}} & \multicolumn{2}{c}{\textbf{CIFAR-100}}  & \multicolumn{2}{c}{\textbf{TinyImageNet}} \\
\cmidrule(lr){2-3} \cmidrule(lr){4-5} \cmidrule(lr){6-7}
 & \textbf{ID}  & \textbf{OOD} & \textbf{ID}  & \textbf{OOD} & \textbf{ID}  & \textbf{OOD} \\
\midrule
 No Noise               & \phantom{0}0.23\tiny{$\pm$0.01} & \phantom{0}0.41\tiny{$\pm$0.00} & \phantom{0}1.83\tiny{$\pm$0.03} & \phantom{0}2.69\tiny{$\pm$0.02} & \phantom{0}2.72\tiny{$\pm$0.03} & \phantom{0}3.86\tiny{$\pm$0.02} \\
 Input Weak Aug.        & \phantom{0}0.23\tiny{$\pm$0.00} & \phantom{0}0.48\tiny{$\pm$0.00} & \phantom{0}1.09\tiny{$\pm$0.01} & \phantom{0}2.36\tiny{$\pm$0.02} & \phantom{0}1.84\tiny{$\pm$0.05} & \phantom{0}3.28\tiny{$\pm$0.04} \\
 Input Gaussian      & \phantom{0}0.23\tiny{$\pm$0.00} & \phantom{0}0.41\tiny{$\pm$0.00} & \phantom{0}1.75\tiny{$\pm$0.03} & \phantom{0}2.56\tiny{$\pm$0.02} & \phantom{0}2.67\tiny{$\pm$0.08} & \phantom{0}3.76\tiny{$\pm$0.06} \\
 Input AugMix           & \phantom{0}0.16\tiny{$\pm$0.00} & \phantom{0}0.31\tiny{$\pm$0.00} & \phantom{0}1.19\tiny{$\pm$0.01} & \phantom{0}2.05\tiny{$\pm$0.00} & \phantom{0}1.90\tiny{$\pm$0.01} & \phantom{0}3.05\tiny{$\pm$0.01} \\
 Input-Target MixUp     & \phantom{0}0.24\tiny{$\pm$0.01} & \phantom{0}0.46\tiny{$\pm$0.01} & \phantom{0}1.76\tiny{$\pm$0.03} & \phantom{0}2.58\tiny{$\pm$0.01} & \phantom{0}2.56\tiny{$\pm$0.03} & \phantom{0}3.68\tiny{$\pm$0.03} \\
 Label Smoothing        & \phantom{0}0.19\tiny{$\pm$0.00} & \phantom{0}0.31\tiny{$\pm$0.00} & \phantom{0}1.99\tiny{$\pm$0.06} & \phantom{0}2.65\tiny{$\pm$0.06} & \phantom{0}2.82\tiny{$\pm$0.02} & \phantom{0}3.86\tiny{$\pm$0.01} \\
\bottomrule
\end{tabular}
\end{adjustbox}
\end{sc}
\end{small}
\end{center}
\caption{CV classification: NLL $(\downarrow)$ comparison on in-distribution (ID) and out-of-distribution (OOD) test set and with hyperparameters transferred across architectures.}
\label{tab:transfer:architecture:cv_classification:nll}
\end{table}
\begin{table}
\begin{center}
    \begin{small}
    \begin{sc}
    \begin{adjustbox}{max width=\linewidth}
\begin{tabular}{lcccccccccc}
\toprule
\multirow{2}{*}{\textbf{Noise Type}}& \multicolumn{2}{c}{\textbf{Wine}} & \multicolumn{2}{c}{\textbf{Toxicity}}  & \multicolumn{2}{c}{\textbf{Abalone}} & \multicolumn{2}{c}{\textbf{Students}} & \multicolumn{2}{c}{\textbf{Adult}}\\
\cmidrule(lr){2-3} \cmidrule(lr){4-5} \cmidrule(lr){6-7}\cmidrule(lr){8-9}\cmidrule(lr){10-11}
 & \textbf{ID}  & \textbf{OOD} & \textbf{ID}  & \textbf{OOD} & \textbf{ID}  & \textbf{OOD} & \textbf{ID}  & \textbf{OOD} & \textbf{ID}  & \textbf{OOD}\\
 
\midrule
 No Noise            & 35.73\tiny{$\pm$2.20}     & 36.37\tiny{$\pm$2.07}      & 47.06\tiny{$\pm$8.32}         & 44.98\tiny{$\pm$7.10}          & 43.30\tiny{$\pm$0.29}        & 43.34\tiny{$\pm$0.20}         & 65.40\tiny{$\pm$4.88}         & 65.89\tiny{$\pm$3.72}          & 15.54\tiny{$\pm$0.23}      & 15.73\tiny{$\pm$0.19}       \\
 Input Gaussian      & 35.94\tiny{$\pm$2.43}     & 36.28\tiny{$\pm$2.15}      & 47.06\tiny{$\pm$8.32}         & 44.78\tiny{$\pm$7.26}          & 43.30\tiny{$\pm$0.43}        & 43.53\tiny{$\pm$0.26}         & 65.40\tiny{$\pm$4.88}         & 65.81\tiny{$\pm$3.62}          & 15.58\tiny{$\pm$0.42}      & 15.74\tiny{$\pm$0.37}       \\
 Input ODS           & 35.83\tiny{$\pm$1.64}     & 36.17\tiny{$\pm$1.53}      & 48.04\tiny{$\pm$5.55}         & 45.65\tiny{$\pm$5.41}          & 43.38\tiny{$\pm$0.73}        & 43.61\tiny{$\pm$0.60}         & 64.56\tiny{$\pm$3.73}         & 65.74\tiny{$\pm$3.30}          & 15.47\tiny{$\pm$0.39}      & 15.67\tiny{$\pm$0.34}       \\
 Input-Target MixUp  & 38.02\tiny{$\pm$0.82}     & 38.12\tiny{$\pm$0.80}      & 50.98\tiny{$\pm$5.00}         & 46.47\tiny{$\pm$4.29}          & 43.46\tiny{$\pm$0.45}        & 43.55\tiny{$\pm$0.40}         & 65.40\tiny{$\pm$2.60}         & 66.13\tiny{$\pm$2.87}          & 15.28\tiny{$\pm$0.33}      & 15.44\tiny{$\pm$0.33}       \\
\bottomrule
\end{tabular}
\end{adjustbox}
\end{sc}
\end{small}
\end{center}
\caption{Tabular data classification: error $(\downarrow, \%)$ comparison on in-distribution (ID) and out-of-distribution (OOD) test sets and with tuned hyperparameters.}
\label{tab:tabular_classification:error}
\end{table}
\begin{table}
\begin{center}
    \begin{small}
    \begin{sc}
    \begin{adjustbox}{max width=\linewidth}
\begin{tabular}{lcccccccccc}
\toprule
\multirow{2}{*}{\textbf{Noise Type}}& \multicolumn{2}{c}{\textbf{Wine}} & \multicolumn{2}{c}{\textbf{Toxicity}}  & \multicolumn{2}{c}{\textbf{Abalone}} & \multicolumn{2}{c}{\textbf{Students}} & \multicolumn{2}{c}{\textbf{Adult}}\\
\cmidrule(lr){2-3} \cmidrule(lr){4-5} \cmidrule(lr){6-7}\cmidrule(lr){8-9}\cmidrule(lr){10-11}
 & \textbf{ID}  & \textbf{OOD} & \textbf{ID}  & \textbf{OOD} & \textbf{ID}  & \textbf{OOD} & \textbf{ID}  & \textbf{OOD} & \textbf{ID}  & \textbf{OOD}\\
 
\midrule
 No Noise            & \phantom{0}8.75\tiny{$\pm$1.34} & \phantom{0}9.07\tiny{$\pm$0.96} & 45.78\tiny{$\pm$8.12}         & 44.18\tiny{$\pm$6.80}          & \phantom{0}3.94\tiny{$\pm$0.59} & \phantom{0}4.19\tiny{$\pm$0.61} & 12.46\tiny{$\pm$2.92}           & 13.81\tiny{$\pm$2.17}           & \phantom{0}3.47\tiny{$\pm$0.29} & \phantom{0}3.85\tiny{$\pm$0.34} \\
 Input Gaussian      & \phantom{0}8.56\tiny{$\pm$0.60} & \phantom{0}8.82\tiny{$\pm$1.13} & 45.84\tiny{$\pm$8.40}         & 43.93\tiny{$\pm$7.07}          & \phantom{0}3.60\tiny{$\pm$0.41} & \phantom{0}4.05\tiny{$\pm$0.67} & 13.36\tiny{$\pm$0.44}           & 14.37\tiny{$\pm$1.68}           & \phantom{0}3.50\tiny{$\pm$0.44} & \phantom{0}3.84\tiny{$\pm$0.42} \\
 Input ODS           & \phantom{0}5.75\tiny{$\pm$0.71} & \phantom{0}6.02\tiny{$\pm$0.52} & 42.82\tiny{$\pm$4.17}         & 39.93\tiny{$\pm$3.67}          & \phantom{0}3.04\tiny{$\pm$0.75} & \phantom{0}3.55\tiny{$\pm$1.01} & 14.04\tiny{$\pm$2.38}           & 14.07\tiny{$\pm$1.77}           & \phantom{0}3.64\tiny{$\pm$0.46} & \phantom{0}3.89\tiny{$\pm$0.41} \\
 Input-Target MixUp  & \phantom{0}4.43\tiny{$\pm$0.77} & \phantom{0}4.76\tiny{$\pm$0.77} & 43.18\tiny{$\pm$5.18}         & 41.22\tiny{$\pm$3.89}          & \phantom{0}3.41\tiny{$\pm$1.00} & \phantom{0}3.86\tiny{$\pm$0.92} & 11.07\tiny{$\pm$3.74}           & 11.74\tiny{$\pm$3.19}           & \phantom{0}2.79\tiny{$\pm$0.36} & \phantom{0}3.06\tiny{$\pm$0.43} \\
\bottomrule
\end{tabular}
\end{adjustbox}
\end{sc}
\end{small}
\end{center}
\caption{Tabular data classification: ECE $(\downarrow, \%)$ comparison on in-distribution (ID) and out-of-distribution (OOD) test sets and with tuned hyperparameters.}
\label{tab:tabular_classification:ece}
\end{table}
\begin{table}
\begin{center}
    \begin{small}
    \begin{sc}
    \begin{adjustbox}{max width=\linewidth}
\begin{tabular}{lcccccccccc}
\toprule
\multirow{2}{*}{\textbf{Noise Type}}& \multicolumn{2}{c}{\textbf{Wine}} & \multicolumn{2}{c}{\textbf{Toxicity}}  & \multicolumn{2}{c}{\textbf{Abalone}} & \multicolumn{2}{c}{\textbf{Students}} & \multicolumn{2}{c}{\textbf{Adult}}\\
\cmidrule(lr){2-3} \cmidrule(lr){4-5} \cmidrule(lr){6-7}\cmidrule(lr){8-9}\cmidrule(lr){10-11}
 & \textbf{ID}  & \textbf{OOD} & \textbf{ID}  & \textbf{OOD} & \textbf{ID}  & \textbf{OOD} & \textbf{ID}  & \textbf{OOD} & \textbf{ID}  & \textbf{OOD}\\
 
\midrule
 No Noise            & \phantom{0}0.94\tiny{$\pm$0.05} & \phantom{0}0.95\tiny{$\pm$0.04} & \phantom{0}4.85\tiny{$\pm$0.23} & \phantom{0}4.87\tiny{$\pm$0.27} & \phantom{0}0.84\tiny{$\pm$0.02} & \phantom{0}0.86\tiny{$\pm$0.02} & \phantom{0}1.87\tiny{$\pm$0.08} & \phantom{0}1.98\tiny{$\pm$0.07} & \phantom{0}0.35\tiny{$\pm$0.01} & \phantom{0}0.36\tiny{$\pm$0.01} \\
 Input Gaussian      & \phantom{0}0.94\tiny{$\pm$0.05} & \phantom{0}0.95\tiny{$\pm$0.05} & \phantom{0}4.85\tiny{$\pm$0.27} & \phantom{0}4.87\tiny{$\pm$0.28} & \phantom{0}0.84\tiny{$\pm$0.02} & \phantom{0}0.86\tiny{$\pm$0.02} & \phantom{0}1.87\tiny{$\pm$0.08} & \phantom{0}1.97\tiny{$\pm$0.07} & \phantom{0}0.34\tiny{$\pm$0.01} & \phantom{0}0.36\tiny{$\pm$0.01} \\
 Input ODS           & \phantom{0}0.91\tiny{$\pm$0.05} & \phantom{0}0.92\tiny{$\pm$0.05} & \phantom{0}2.20\tiny{$\pm$0.33} & \phantom{0}2.35\tiny{$\pm$0.39} & \phantom{0}0.84\tiny{$\pm$0.02} & \phantom{0}0.85\tiny{$\pm$0.02} & \phantom{0}1.87\tiny{$\pm$0.08} & \phantom{0}1.98\tiny{$\pm$0.07} & \phantom{0}0.35\tiny{$\pm$0.01} & \phantom{0}0.36\tiny{$\pm$0.01} \\
 Input-Target MixUp  & \phantom{0}0.91\tiny{$\pm$0.04} & \phantom{0}0.91\tiny{$\pm$0.04} & \phantom{0}2.39\tiny{$\pm$0.27} & \phantom{0}2.44\tiny{$\pm$0.19} & \phantom{0}0.84\tiny{$\pm$0.02} & \phantom{0}0.86\tiny{$\pm$0.02} & \phantom{0}1.88\tiny{$\pm$0.04} & \phantom{0}1.96\tiny{$\pm$0.04} & \phantom{0}0.34\tiny{$\pm$0.01} & \phantom{0}0.35\tiny{$\pm$0.01} \\
\bottomrule
\end{tabular}
\end{adjustbox}
\end{sc}
\end{small}
\end{center}
\caption{Tabular data classification: NLL $(\downarrow)$ comparison on in-distribution (ID) and out-of-distribution (OOD) test sets and with tuned hyperparameters.}
\label{tab:tabular_classification:nll}
\end{table}
\begin{table}
    \begin{center}
        \begin{small}
        \begin{sc}
    \begin{tabular}{lcccc}
    \toprule
\multirow{2}{*}{\textbf{Noise Type}}& \multicolumn{2}{c}{\textbf{NewsGroup}} & \multicolumn{2}{c}{\textbf{SST}} \\
\cmidrule(lr){2-3} \cmidrule(lr){4-5} 
 & \textbf{GP-CNN}  & \textbf{Transformer}  & \textbf{GP-CNN}  & \textbf{Transformer} \\
\midrule
 No Noise            & 35.67\tiny{$\pm$1.13} & 36.56\tiny{$\pm$0.83} & 19.15\tiny{$\pm$0.75} & 21.60\tiny{$\pm$0.24}       \\
 Input Gaussian      & 35.44\tiny{$\pm$0.87} & 36.59\tiny{$\pm$0.84} & 21.14\tiny{$\pm$0.33} & 21.52\tiny{$\pm$0.24}        \\
 Input ODS           & 33.56\tiny{$\pm$0.33} & 34.44\tiny{$\pm$0.74} & 18.46\tiny{$\pm$1.03} & 21.52\tiny{$\pm$1.22}        \\
 Input-Target MixUp  & 35.44\tiny{$\pm$0.78} & 36.70\tiny{$\pm$0.92} & 18.85\tiny{$\pm$0.11} & 21.33\tiny{$\pm$0.41}        \\
 Label Smoothing     & 35.59\tiny{$\pm$0.69} & 36.56\tiny{$\pm$1.10} & 21.75\tiny{$\pm$0.53} & 21.29\tiny{$\pm$0.38} \\
 Activation Gaussian & 35.56\tiny{$\pm$1.16} & 36.44\tiny{$\pm$0.87} & 19.61\tiny{$\pm$1.06} & 21.64\tiny{$\pm$0.24} \\
\bottomrule
    \end{tabular}
    \end{sc}
    \end{small}
    \end{center}
    \caption{NewsGroup NLP classification: Error $(\downarrow,\%)$ comparison on in-distribution (ID) test set and with tuned hyperparameters.}
    \label{tab:nlp_classification:error}
    \end{table}
    
\begin{table}
    \begin{center}
        \begin{small}
        \begin{sc}
    \begin{tabular}{lcccc}
    \toprule
\multirow{2}{*}{\textbf{Noise Type}}& \multicolumn{2}{c}{\textbf{NewsGroup}} & \multicolumn{2}{c}{\textbf{SST}} \\
\cmidrule(lr){2-3} \cmidrule(lr){4-5} 
 & \textbf{GP-CNN}  & \textbf{Transformer}  & \textbf{GP-CNN}  & \textbf{Transformer}\\
\midrule
 No Noise            & \phantom{0}5.12\tiny{$\pm$0.53} & \phantom{0}3.47\tiny{$\pm$0.98}  & 13.99\tiny{$\pm$0.47}           & 11.61\tiny{$\pm$3.97}                 \\
 Input Gaussian      & \phantom{0}4.78\tiny{$\pm$1.30} & \phantom{0}3.59\tiny{$\pm$0.83}  & 15.14\tiny{$\pm$0.17}           & 11.32\tiny{$\pm$3.72}                 \\
 Input ODS           & \phantom{0}2.57\tiny{$\pm$0.81} & \phantom{0}7.99\tiny{$\pm$0.59}  & 11.80\tiny{$\pm$0.93}           & 14.70\tiny{$\pm$2.64}                 \\
 Input-Target MixUp  & \phantom{0}5.07\tiny{$\pm$0.59} & \phantom{0}2.54\tiny{$\pm$1.27}  & \phantom{0}7.76\tiny{$\pm$0.32} & 11.75\tiny{$\pm$3.55}                 \\
 Label Smoothing     & \phantom{0}3.78\tiny{$\pm$0.50} & \phantom{0}4.13\tiny{$\pm$0.94}  & \phantom{0}9.10\tiny{$\pm$0.87} & 10.26\tiny{$\pm$3.30}                 \\
 Activation Gaussian & \phantom{0}5.75\tiny{$\pm$0.50} & \phantom{0}3.42\tiny{$\pm$1.05}  & 13.99\tiny{$\pm$1.22}           & 11.39\tiny{$\pm$3.71}                 \\
\bottomrule
    \end{tabular}
    \end{sc}
    \end{small}
    \end{center}
    \caption{NewsGroup NLP classification: ECE $(\downarrow,\%)$ comparison on in-distribution (ID) test set and with tuned hyperparameters.}
    \label{tab:nlp_classification:ece}
    \end{table}
\begin{table}
    \begin{center}
        \begin{small}
        \begin{sc}
    \begin{tabular}{lcccc}
    \toprule
\multirow{2}{*}{\textbf{Noise Type}}& \multicolumn{2}{c}{\textbf{NewsGroup}} & \multicolumn{2}{c}{\textbf{SST}} \\
\cmidrule(lr){2-3} \cmidrule(lr){4-5} 
 & \textbf{GP-CNN}  & \textbf{Transformer}  & \textbf{GP-CNN}  & \textbf{Transformer}\\
\midrule
 No Noise            & \phantom{0}1.14\tiny{$\pm$0.01} & \phantom{0}1.13\tiny{$\pm$0.02} & \phantom{0}0.81\tiny{$\pm$0.04} & \phantom{0}0.61\tiny{$\pm$0.10}   \\
 Input Gaussian      & \phantom{0}1.12\tiny{$\pm$0.01} & \phantom{0}1.13\tiny{$\pm$0.02} & \phantom{0}0.85\tiny{$\pm$0.05} & \phantom{0}0.59\tiny{$\pm$0.09}   \\
 Input ODS           & \phantom{0}1.03\tiny{$\pm$0.00} & \phantom{0}1.10\tiny{$\pm$0.01} & \phantom{0}0.62\tiny{$\pm$0.02} & \phantom{0}0.73\tiny{$\pm$0.12}   \\
 Input-Target MixUp  & \phantom{0}1.13\tiny{$\pm$0.01} & \phantom{0}1.13\tiny{$\pm$0.02} & \phantom{0}0.48\tiny{$\pm$0.00} & \phantom{0}0.59\tiny{$\pm$0.08}   \\
 Label Smoothing     & \phantom{0}1.12\tiny{$\pm$0.01} & \phantom{0}1.13\tiny{$\pm$0.02} & \phantom{0}0.52\tiny{$\pm$0.01} & \phantom{0}0.54\tiny{$\pm$0.05}   \\
 Activation Gaussian & \phantom{0}1.14\tiny{$\pm$0.01} & \phantom{0}1.13\tiny{$\pm$0.02} & \phantom{0}0.83\tiny{$\pm$0.03} & \phantom{0}0.59\tiny{$\pm$0.09}   \\
\bottomrule
    \end{tabular}
    \end{sc}
    \end{small}
    \end{center}
    \caption{NewsGroup NLP classification: NLL $(\downarrow)$ comparison on in-distribution (ID) test set and with tuned hyperparameters.}
    \label{tab:nlp_classification:nll}
    \end{table}
\begin{table}
\begin{center}
    \begin{small}
    \begin{sc}
\begin{adjustbox}{max width=\linewidth}
\begin{tabular}{lcccc}
\toprule
\multirow{2}{*}{\textbf{Noise Type}}& \multicolumn{2}{c}{\textbf{Rotated CIFAR-100}} & \multicolumn{2}{c}{\textbf{WikiFace}}\\
\cmidrule(lr){2-3} \cmidrule(lr){4-5} 
 & \textbf{ID}  & \textbf{OOD} & \textbf{ID}  & \textbf{OOD} \\
\midrule
 No Noise                    & \phantom{0}0.03\tiny{$\pm$0.00} & \phantom{0}0.13\tiny{$\pm$0.01} & \phantom{0}0.03\tiny{$\pm$0.00} & \phantom{0}0.04\tiny{$\pm$0.00} \\
 Input Weak Aug.             & \phantom{0}0.03\tiny{$\pm$0.00} & \phantom{0}0.13\tiny{$\pm$0.01} & \phantom{0}0.03\tiny{$\pm$0.00} & \phantom{0}0.04\tiny{$\pm$0.00} \\
 Input Gaussian           & \phantom{0}0.03\tiny{$\pm$0.00} & \phantom{0}0.14\tiny{$\pm$0.01} & \phantom{0}0.04\tiny{$\pm$0.00} & \phantom{0}0.05\tiny{$\pm$0.00} \\
 Input AugMix                & \phantom{0}0.03\tiny{$\pm$0.00} & \phantom{0}0.07\tiny{$\pm$0.00} & \phantom{0}0.03\tiny{$\pm$0.00} & \phantom{0}0.04\tiny{$\pm$0.00} \\
 Input-Target CMixUp         & \phantom{0}0.03\tiny{$\pm$0.00} & \phantom{0}0.09\tiny{$\pm$0.00} & \phantom{0}0.04\tiny{$\pm$0.00} & \phantom{0}0.04\tiny{$\pm$0.00} \\
 Activation Gaussian      & \phantom{0}0.03\tiny{$\pm$0.00} & \phantom{0}0.14\tiny{$\pm$0.01} & \phantom{0}0.04\tiny{$\pm$0.00} & \phantom{0}0.04\tiny{$\pm$0.00} \\
\bottomrule
\end{tabular}
\end{adjustbox}
\end{sc}
\end{small}
\end{center}
\caption{Rotated CV regression: MSE $(\downarrow)$ comparison on in-distribution (ID) and out-of-distribution (OOD) test sets and with tuned hyperparameters.}
\label{tab:cv_regression:mse}
\end{table}
\begin{table}
\begin{center}
    \begin{small}
    \begin{sc}
\begin{adjustbox}{max width=\linewidth}
\begin{tabular}{lcccc}
\toprule
\multirow{2}{*}{\textbf{Noise Type}}& \multicolumn{2}{c}{\textbf{Rotated CIFAR-100}} & \multicolumn{2}{c}{\textbf{WikiFace}}\\
\cmidrule(lr){2-3} \cmidrule(lr){4-5} 
 & \textbf{ID}  & \textbf{OOD} & \textbf{ID}  & \textbf{OOD} \\
\midrule
 No Noise                    & -4.81\tiny{$\pm$0.00}           & \phantom{0}6.90\tiny{$\pm$1.52} & 27.03\tiny{$\pm$2.94}           & 31.43\tiny{$\pm$3.80}           \\
 Input Weak Aug.             & -4.60\tiny{$\pm$0.06}           & \phantom{0}3.41\tiny{$\pm$0.53} & -0.82\tiny{$\pm$0.24}           & \phantom{0}0.31\tiny{$\pm$0.66} \\
 Input Gaussian           & -4.67\tiny{$\pm$0.11}           & \phantom{0}7.57\tiny{$\pm$2.14} & 27.96\tiny{$\pm$2.66}           & 31.50\tiny{$\pm$2.15}           \\
 Input AugMix                & -4.82\tiny{$\pm$0.01}           & -1.70\tiny{$\pm$0.04}           & \phantom{0}0.78\tiny{$\pm$0.27} & -0.12\tiny{$\pm$0.11}           \\
 Input-Target CMixUp         & -4.62\tiny{$\pm$0.04}           & \phantom{0}1.86\tiny{$\pm$0.35} & 21.83\tiny{$\pm$2.35}           & 25.62\tiny{$\pm$0.93}           \\
 Activation Gaussian      & -4.27\tiny{$\pm$0.07}           & \phantom{0}2.94\tiny{$\pm$0.18} & 14.93\tiny{$\pm$1.09}           & 17.61\tiny{$\pm$1.01}           \\
\bottomrule
\end{tabular}
\end{adjustbox}
\end{sc}
\end{small}
\end{center}
\caption{Rotated CV regression: NLL $(\downarrow)$ comparison on in-distribution (ID) and out-of-distribution (OOD) test sets and with tuned hyperparameters.}
\label{tab:cv_regression:nll}
\end{table}
\begin{table}
\begin{center}
    \begin{small}
    \begin{sc}
    \begin{adjustbox}{max width=\linewidth}
\begin{tabular}{lcccccccccc}
\toprule
\multirow{2}{*}{\textbf{Noise Type}}& \multicolumn{2}{c}{\textbf{Energy}} & \multicolumn{2}{c}{\textbf{Boston}}  & \multicolumn{2}{c}{\textbf{Wine}} & \multicolumn{2}{c}{\textbf{Yacht}} & \multicolumn{2}{c}{\textbf{Concrete}}\\
\cmidrule(lr){2-3} \cmidrule(lr){4-5} \cmidrule(lr){6-7}\cmidrule(lr){8-9}\cmidrule(lr){10-11}
 & \textbf{ID}  & \textbf{OOD} & \textbf{ID}  & \textbf{OOD} & \textbf{ID}  & \textbf{OOD} & \textbf{ID}  & \textbf{OOD} & \textbf{ID}  & \textbf{OOD}\\
\midrule
 No Noise            & \phantom{0}0.04\tiny{$\pm$0.00} & \phantom{0}0.05\tiny{$\pm$0.00} & \phantom{0}0.13\tiny{$\pm$0.02} & \phantom{0}0.15\tiny{$\pm$0.02} & \phantom{0}0.25\tiny{$\pm$0.04} & 29.42\tiny{$\pm$25.71} & \phantom{0}0.07\tiny{$\pm$0.07} & \phantom{0}0.16\tiny{$\pm$0.05} & \phantom{0}0.10\tiny{$\pm$0.01} & \phantom{0}0.10\tiny{$\pm$0.01} \\
 Input Gaussian      & \phantom{0}0.04\tiny{$\pm$0.00} & \phantom{0}0.05\tiny{$\pm$0.00} & \phantom{0}0.11\tiny{$\pm$0.02} & \phantom{0}0.13\tiny{$\pm$0.02} & \phantom{0}0.25\tiny{$\pm$0.04} & 29.39\tiny{$\pm$25.66} & \phantom{0}0.02\tiny{$\pm$0.02} & \phantom{0}0.09\tiny{$\pm$0.07} & \phantom{0}0.10\tiny{$\pm$0.01} & \phantom{0}0.10\tiny{$\pm$0.01} \\
 Input-Target CMixUp & \phantom{0}0.04\tiny{$\pm$0.00} & \phantom{0}0.05\tiny{$\pm$0.00} & \phantom{0}0.13\tiny{$\pm$0.01} & \phantom{0}0.15\tiny{$\pm$0.01} & \phantom{0}0.25\tiny{$\pm$0.04} & 28.95\tiny{$\pm$25.23} & \phantom{0}0.15\tiny{$\pm$0.17} & \phantom{0}0.19\tiny{$\pm$0.16} & \phantom{0}0.10\tiny{$\pm$0.00} & \phantom{0}0.11\tiny{$\pm$0.00} \\
 Activation Gaussian & \phantom{0}0.04\tiny{$\pm$0.00} & \phantom{0}0.05\tiny{$\pm$0.00} & \phantom{0}0.13\tiny{$\pm$0.03} & \phantom{0}0.15\tiny{$\pm$0.02} & \phantom{0}0.25\tiny{$\pm$0.04} & 29.57\tiny{$\pm$25.82} & \phantom{0}0.03\tiny{$\pm$0.02} & \phantom{0}0.09\tiny{$\pm$0.07} & \phantom{0}0.10\tiny{$\pm$0.01} & \phantom{0}0.10\tiny{$\pm$0.01} \\
\bottomrule
\end{tabular}
\end{adjustbox}
\end{sc}
\end{small}
\end{center}
\caption{Tabular regression: MSE $(\downarrow)$ comparison on in-distribution (ID) and out-of-distribution (OOD) test sets and with tuned hyperparameters.}
\label{tab:tabular_regression:mse}
\end{table}
\begin{table}
\begin{center}
    \begin{small}
    \begin{sc}
    \begin{adjustbox}{max width=\linewidth}
\begin{tabular}{lcccccccccc}
\toprule
\multirow{2}{*}{\textbf{Noise Type}}& \multicolumn{2}{c}{\textbf{Energy}} & \multicolumn{2}{c}{\textbf{Boston}}  & \multicolumn{2}{c}{\textbf{Wine}} & \multicolumn{2}{c}{\textbf{Yacht}} & \multicolumn{2}{c}{\textbf{Concrete}}\\
\cmidrule(lr){2-3} \cmidrule(lr){4-5} \cmidrule(lr){6-7}\cmidrule(lr){8-9}\cmidrule(lr){10-11}
 & \textbf{ID}  & \textbf{OOD} & \textbf{ID}  & \textbf{OOD} & \textbf{ID}  & \textbf{OOD} & \textbf{ID}  & \textbf{OOD} & \textbf{ID}  & \textbf{OOD}\\
\midrule
 No Noise            & -1.54\tiny{$\pm$0.05}   & \phantom{0}6.64\tiny{$\pm$3.38} & -0.19\tiny{$\pm$0.12}           & \phantom{0}0.04\tiny{$\pm$0.12} & -0.22\tiny{$\pm$0.08}           & \phantom{0}0.03\tiny{$\pm$0.12} & -1.18\tiny{$\pm$0.22}  & 10.45\tiny{$\pm$16.46}          & -0.56\tiny{$\pm$0.10}     & -0.01\tiny{$\pm$0.24}           \\
 Input Gaussian      & -1.55\tiny{$\pm$0.03}   & \phantom{0}6.55\tiny{$\pm$1.53} & -0.54\tiny{$\pm$0.14}           & -0.40\tiny{$\pm$0.18}           & -0.22\tiny{$\pm$0.08}           & \phantom{0}0.03\tiny{$\pm$0.12} & -1.46\tiny{$\pm$0.19}  & -1.04\tiny{$\pm$0.30}           & -0.53\tiny{$\pm$0.11}     & \phantom{0}0.21\tiny{$\pm$0.41} \\
 Input-Target CMixUp & -1.53\tiny{$\pm$0.03}   & \phantom{0}6.03\tiny{$\pm$2.17} & -0.42\tiny{$\pm$0.12}           & -0.29\tiny{$\pm$0.14}           & -0.19\tiny{$\pm$0.09}           & 81.98\tiny{$\pm$58.22}          & -1.05\tiny{$\pm$0.20}  & -0.87\tiny{$\pm$0.31}           & -0.63\tiny{$\pm$0.01}     & -0.26\tiny{$\pm$0.26}           \\
 Activation Gaussian & -1.56\tiny{$\pm$0.03}   & \phantom{0}6.64\tiny{$\pm$0.32} & -0.20\tiny{$\pm$0.17}           & \phantom{0}0.04\tiny{$\pm$0.11} & -0.22\tiny{$\pm$0.08}           & \phantom{0}0.04\tiny{$\pm$0.10} & -1.31\tiny{$\pm$0.15}  & -1.21\tiny{$\pm$0.14}           & -0.56\tiny{$\pm$0.09}     & \phantom{0}0.08\tiny{$\pm$0.44} \\
\bottomrule
\end{tabular}
\end{adjustbox}
\end{sc}
\end{small}
\end{center}
\caption{Tabular regression: NLL $(\downarrow)$ comparison on in-distribution (ID) and out-of-distribution (OOD) test sets and with tuned hyperparameters.}
\label{tab:tabular_regression:nll}
\end{table}
\begin{table}
\begin{center}
    \begin{small}
    \begin{sc}
\begin{tabular}{lcccccc}
\toprule
\multirow{2}{*}{\textbf{Noise Type}}& \multicolumn{2}{c}{\textbf{Energy}} & \multicolumn{2}{c}{\textbf{Wine}}  & \multicolumn{2}{c}{\textbf{Concrete}} \\
 \cmidrule(lr){2-3} \cmidrule(lr){4-5} \cmidrule(lr){6-7}
& \textbf{ID}  & \textbf{OOD} & \textbf{ID}  & \textbf{OOD} & \textbf{ID}  & \textbf{OOD} \\
 \midrule
 No Noise            & \phantom{0}0.03\tiny{$\pm$0.01} & \phantom{0}0.03\tiny{$\pm$0.01} & \phantom{0}0.21\tiny{$\pm$0.01} & 37.20\tiny{$\pm$33.29}     & \phantom{0}0.09\tiny{$\pm$0.01} & \phantom{0}0.10\tiny{$\pm$0.01} \\
 Input Gaussian      & \phantom{0}0.03\tiny{$\pm$0.01} & \phantom{0}0.03\tiny{$\pm$0.01} & \phantom{0}0.21\tiny{$\pm$0.02} & 45.36\tiny{$\pm$43.45}     & \phantom{0}0.08\tiny{$\pm$0.01} & \phantom{0}0.09\tiny{$\pm$0.01} \\
 Input-Target CMixUp & \phantom{0}0.03\tiny{$\pm$0.00} & \phantom{0}0.04\tiny{$\pm$0.00} & \phantom{0}0.18\tiny{$\pm$0.01} & 49.23\tiny{$\pm$37.16}     & \phantom{0}0.10\tiny{$\pm$0.01} & \phantom{0}0.11\tiny{$\pm$0.00} \\
 Activation Gaussian & \phantom{0}0.03\tiny{$\pm$0.01} & \phantom{0}0.03\tiny{$\pm$0.01} & \phantom{0}0.22\tiny{$\pm$0.01} & 43.40\tiny{$\pm$45.61}     & \phantom{0}0.09\tiny{$\pm$0.02} & \phantom{0}0.10\tiny{$\pm$0.01} \\
 Activation Dropout  & \phantom{0}0.03\tiny{$\pm$0.01} & \phantom{0}0.03\tiny{$\pm$0.00} & \phantom{0}0.24\tiny{$\pm$0.02} & 42.16\tiny{$\pm$33.28}     & \phantom{0}0.09\tiny{$\pm$0.02} & \phantom{0}0.10\tiny{$\pm$0.01} \\
 Gradient Gaussian   & \phantom{0}0.03\tiny{$\pm$0.01} & \phantom{0}0.04\tiny{$\pm$0.01} & 311.38\tiny{$\pm$405.17}        & 1858.44\tiny{$\pm$2585.50} & \phantom{0}0.09\tiny{$\pm$0.02} & \phantom{0}0.10\tiny{$\pm$0.02} \\
\bottomrule
\end{tabular}
\end{sc}
\end{small}
\end{center}
\caption{Tabular regression: MSE $(\downarrow)$ comparison on in-distribution (ID) and out-of-distribution (OOD) test sets and with hyperparameters transferred across datasets.}
\label{tab:transfer:dataset:tabular_regression:mse}
\end{table}
\begin{table}
\begin{center}
    \begin{small}
    \begin{sc}
\begin{tabular}{lcccccc}
\toprule
\multirow{2}{*}{\textbf{Noise Type}}& \multicolumn{2}{c}{\textbf{Energy}} & \multicolumn{2}{c}{\textbf{Wine}}  & \multicolumn{2}{c}{\textbf{Concrete}} \\
 \cmidrule(lr){2-3} \cmidrule(lr){4-5} \cmidrule(lr){6-7}
& \textbf{ID}  & \textbf{OOD} & \textbf{ID}  & \textbf{OOD} & \textbf{ID}  & \textbf{OOD} \\
 \midrule
 No Noise            & -1.70\tiny{$\pm$0.11}   & \phantom{0}1.24\tiny{$\pm$0.61} & \phantom{0}4.78\tiny{$\pm$1.31} & \phantom{0}6.66\tiny{$\pm$1.82} & -0.54\tiny{$\pm$0.15}     & \phantom{0}0.32\tiny{$\pm$0.69} \\
 Input Gaussian      & -1.74\tiny{$\pm$0.07}   & \phantom{0}1.06\tiny{$\pm$0.88} & \phantom{0}5.24\tiny{$\pm$1.58} & 19.29\tiny{$\pm$19.08}          & -0.40\tiny{$\pm$0.15}     & \phantom{0}0.23\tiny{$\pm$0.47} \\
 Input-Target CMixUp & -1.66\tiny{$\pm$0.10}   & \phantom{0}0.42\tiny{$\pm$1.30} & -0.06\tiny{$\pm$0.15}           & 95014.61\tiny{$\pm$134370.19}   & -0.65\tiny{$\pm$0.05}     & \phantom{0}0.53\tiny{$\pm$1.27} \\
 Activation Gaussian & -1.71\tiny{$\pm$0.09}   & \phantom{0}0.51\tiny{$\pm$1.20} & \phantom{0}1.84\tiny{$\pm$1.25} & 177076.49\tiny{$\pm$250421.18}  & -0.50\tiny{$\pm$0.14}     & \phantom{0}0.46\tiny{$\pm$0.74} \\
 Activation Dropout  & -1.66\tiny{$\pm$0.12}   & \phantom{0}0.02\tiny{$\pm$1.35} & -0.27\tiny{$\pm$0.05}           & -0.19\tiny{$\pm$0.05}           & -0.56\tiny{$\pm$0.11}     & \phantom{0}1.19\tiny{$\pm$2.07} \\
 Gradient Gaussian   & -1.70\tiny{$\pm$0.11}   & \phantom{0}5.71\tiny{$\pm$6.29} & \phantom{0}1.97\tiny{$\pm$1.94} & \phantom{0}2.06\tiny{$\pm$2.03} & -0.11\tiny{$\pm$0.10}     & \phantom{0}0.63\tiny{$\pm$0.34} \\
\bottomrule
\end{tabular}
\end{sc}
\end{small}
\end{center}
\caption{Tabular regression: NLL $(\downarrow)$ comparison on in-distribution (ID) and out-of-distribution (OOD) test sets and with hyperparameters transferred across datasets.}
\label{tab:transfer:dataset:tabular_regression:nll}
\end{table}
\begin{table}
\begin{center}
    \begin{small}
    \begin{sc}
\begin{tabular}{lcccccc}
\toprule
\multirow{2}{*}{\textbf{Noise Type}}& \multicolumn{2}{c}{\textbf{Boston}} & \multicolumn{2}{c}{\textbf{Yacht}}  & \multicolumn{2}{c}{\textbf{Concrete}} \\
 \cmidrule(lr){2-3} \cmidrule(lr){4-5} \cmidrule(lr){6-7}
& \textbf{ID}  & \textbf{OOD} & \textbf{ID}  & \textbf{OOD} & \textbf{ID}  & \textbf{OOD} \\
 \midrule
 No Noise            & \phantom{0}0.13\tiny{$\pm$0.04} & \phantom{0}0.16\tiny{$\pm$0.04} & \phantom{0}0.56\tiny{$\pm$0.74} & \phantom{0}0.59\tiny{$\pm$0.72} & \phantom{0}0.11\tiny{$\pm$0.01} & \phantom{0}0.11\tiny{$\pm$0.00} \\
 Input Gaussian      & \phantom{0}0.13\tiny{$\pm$0.03} & \phantom{0}0.15\tiny{$\pm$0.03} & \phantom{0}0.55\tiny{$\pm$0.75} & \phantom{0}0.59\tiny{$\pm$0.72} & \phantom{0}0.10\tiny{$\pm$0.01} & \phantom{0}0.11\tiny{$\pm$0.01} \\
 Input-Target CMixUp & \phantom{0}0.50\tiny{$\pm$0.47} & \phantom{0}0.52\tiny{$\pm$0.46} & \phantom{0}0.05\tiny{$\pm$0.02} & \phantom{0}0.11\tiny{$\pm$0.06} & \phantom{0}0.11\tiny{$\pm$0.01} & \phantom{0}0.12\tiny{$\pm$0.01} \\
 Activation Gaussian & \phantom{0}0.11\tiny{$\pm$0.03} & \phantom{0}0.13\tiny{$\pm$0.03} & \phantom{0}0.56\tiny{$\pm$0.75} & \phantom{0}0.62\tiny{$\pm$0.71} & \phantom{0}0.11\tiny{$\pm$0.01} & \phantom{0}0.11\tiny{$\pm$0.01} \\
 Activation Dropout  & \phantom{0}0.17\tiny{$\pm$0.07} & \phantom{0}0.18\tiny{$\pm$0.06} & \phantom{0}0.05\tiny{$\pm$0.02} & \phantom{0}0.08\tiny{$\pm$0.01} & \phantom{0}0.11\tiny{$\pm$0.01} & \phantom{0}0.11\tiny{$\pm$0.01} \\
 Gradient Gaussian   & \phantom{0}0.16\tiny{$\pm$0.04} & \phantom{0}0.18\tiny{$\pm$0.04} & \phantom{0}0.02\tiny{$\pm$0.01} & \phantom{0}0.06\tiny{$\pm$0.04} & \phantom{0}0.11\tiny{$\pm$0.01} & \phantom{0}0.11\tiny{$\pm$0.00} \\
\bottomrule
\end{tabular}
\end{sc}
\end{small}
\end{center}
\caption{Tabular regression: MSE $(\downarrow)$ comparison on in-distribution (ID) and out-of-distribution (OOD) test sets and with hyperparameters transferred across architectures.}
\label{tab:transfer:architecture:tabular_regression:mse}
\end{table}
\begin{table}
\begin{center}
    \begin{small}
    \begin{sc}
\begin{tabular}{lcccccc}
\toprule
\multirow{2}{*}{\textbf{Noise Type}}& \multicolumn{2}{c}{\textbf{Boston}} & \multicolumn{2}{c}{\textbf{Yacht}}  & \multicolumn{2}{c}{\textbf{Concrete}} \\
 \cmidrule(lr){2-3} \cmidrule(lr){4-5} \cmidrule(lr){6-7}
& \textbf{ID}  & \textbf{OOD} & \textbf{ID}  & \textbf{OOD} & \textbf{ID}  & \textbf{OOD} \\
 \midrule
 No Noise            & -0.25\tiny{$\pm$0.21}           & -0.01\tiny{$\pm$0.17}           & -0.42\tiny{$\pm$0.88}  & -0.02\tiny{$\pm$0.80}   & -0.57\tiny{$\pm$0.06}     & -0.02\tiny{$\pm$0.54}           \\
 Input Gaussian      & -0.35\tiny{$\pm$0.10}           & -0.17\tiny{$\pm$0.09}           & -0.94\tiny{$\pm$1.25}  & -0.72\tiny{$\pm$1.09}   & -0.58\tiny{$\pm$0.08}     & \phantom{0}0.06\tiny{$\pm$0.63} \\
 Input-Target CMixUp & -0.09\tiny{$\pm$0.29}           & -0.02\tiny{$\pm$0.23}           & -1.20\tiny{$\pm$0.07}  & -1.06\tiny{$\pm$0.11}   & -0.60\tiny{$\pm$0.04}     & -0.25\tiny{$\pm$0.32}           \\
 Activation Gaussian & -0.18\tiny{$\pm$0.66}           & \phantom{0}0.02\tiny{$\pm$0.72} & -0.44\tiny{$\pm$0.93}  & -0.09\tiny{$\pm$0.65}   & -0.56\tiny{$\pm$0.07}     & -0.09\tiny{$\pm$0.36}           \\
 Activation Dropout  & -0.56\tiny{$\pm$0.04}           & -0.54\tiny{$\pm$0.04}           & -1.11\tiny{$\pm$0.31}  & -0.81\tiny{$\pm$0.11}   & -0.58\tiny{$\pm$0.07}     & \phantom{0}0.08\tiny{$\pm$0.66} \\
 Gradient Gaussian   & \phantom{0}0.97\tiny{$\pm$1.13} & \phantom{0}1.22\tiny{$\pm$1.16} & -1.52\tiny{$\pm$0.43}  & -1.25\tiny{$\pm$0.17}   & -0.57\tiny{$\pm$0.06}     & -0.02\tiny{$\pm$0.54}           \\
\bottomrule
\end{tabular}
\end{sc}
\end{small}
\end{center}
\caption{Tabular regression: NLL $(\downarrow)$ comparison on in-distribution (ID) and out-of-distribution (OOD) test sets and with hyperparameters transferred across architectures.}
\label{tab:transfer:architecture:tabular_regression:nll}
\end{table}
\clearpage
\begin{figure}
\centering
\begin{subfigure}{0.21\textwidth}
	\centering
	\includegraphics[width=\textwidth]{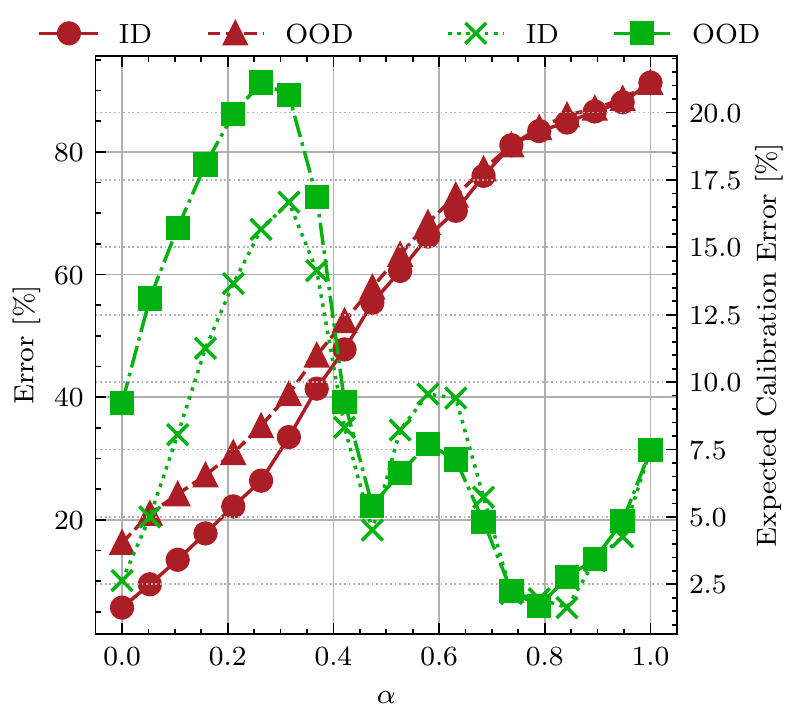}
	\caption{Error and ECE.}
	\label{fig:loss_landscape:cifar10-resnet-input_random_crop_horizontal_flip-lin_error_ece}
\end{subfigure}
\begin{subfigure}{0.21\textwidth}
	\centering
	\includegraphics[width=\textwidth]{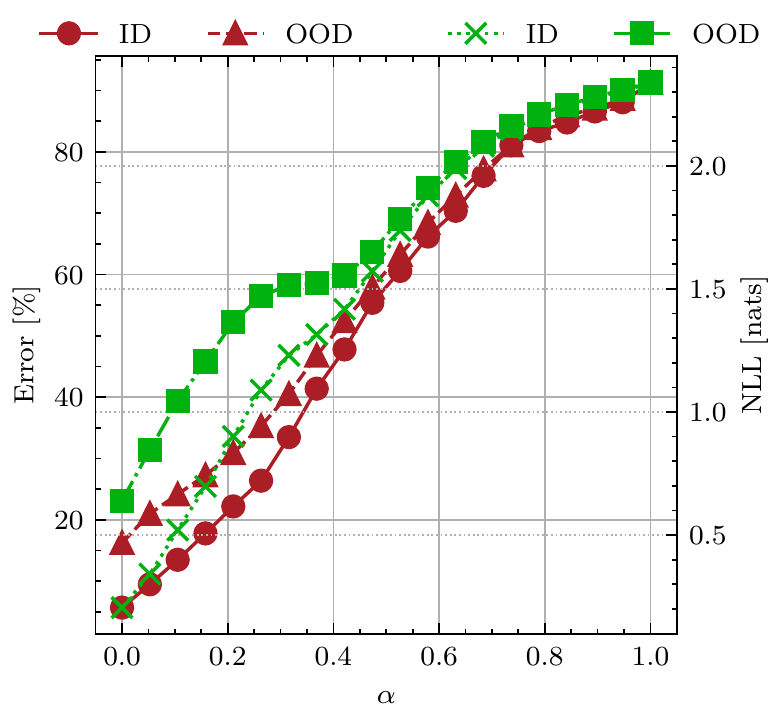}
	\caption{Error and NLL.}
	\label{fig:loss_landscape:cifar10-resnet-input_random_crop_horizontal_flip-lin_error_nll}
\end{subfigure}
\begin{subfigure}{0.25\textwidth}
	\centering
	\includegraphics[width=\textwidth]{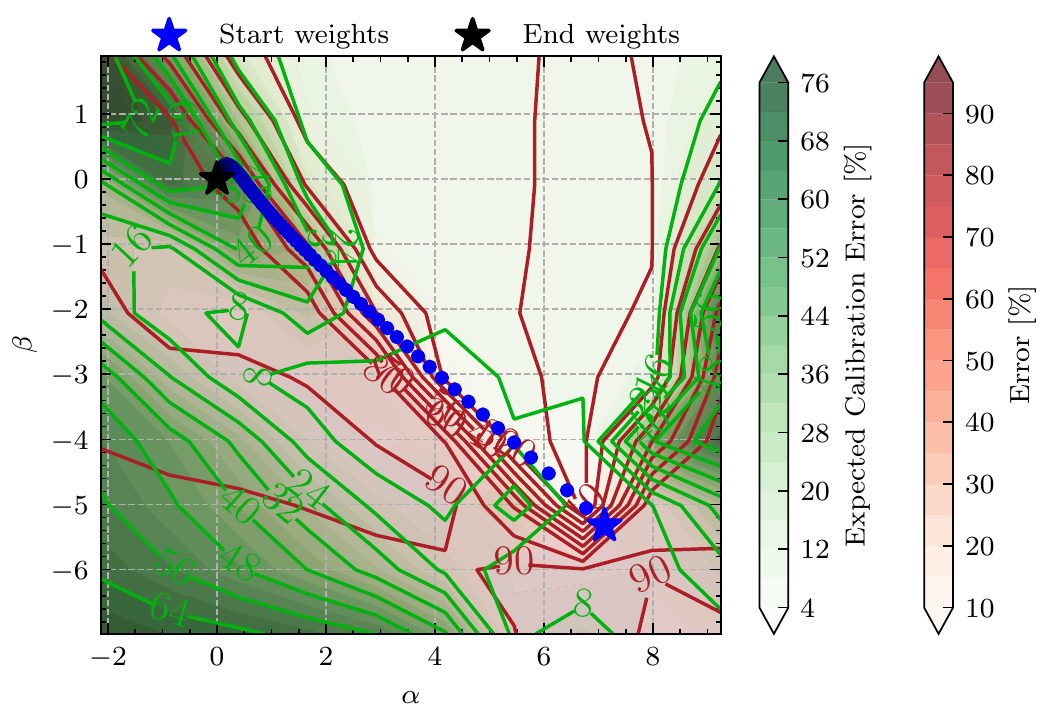}
	\caption{Error and ECE on ID.}
	\label{fig:loss_landscape:cifar10-resnet-input_random_crop_horizontal_flip-test_2d_error_ece}
\end{subfigure}
\begin{subfigure}{0.25\textwidth}
	\centering
	\includegraphics[width=\textwidth]{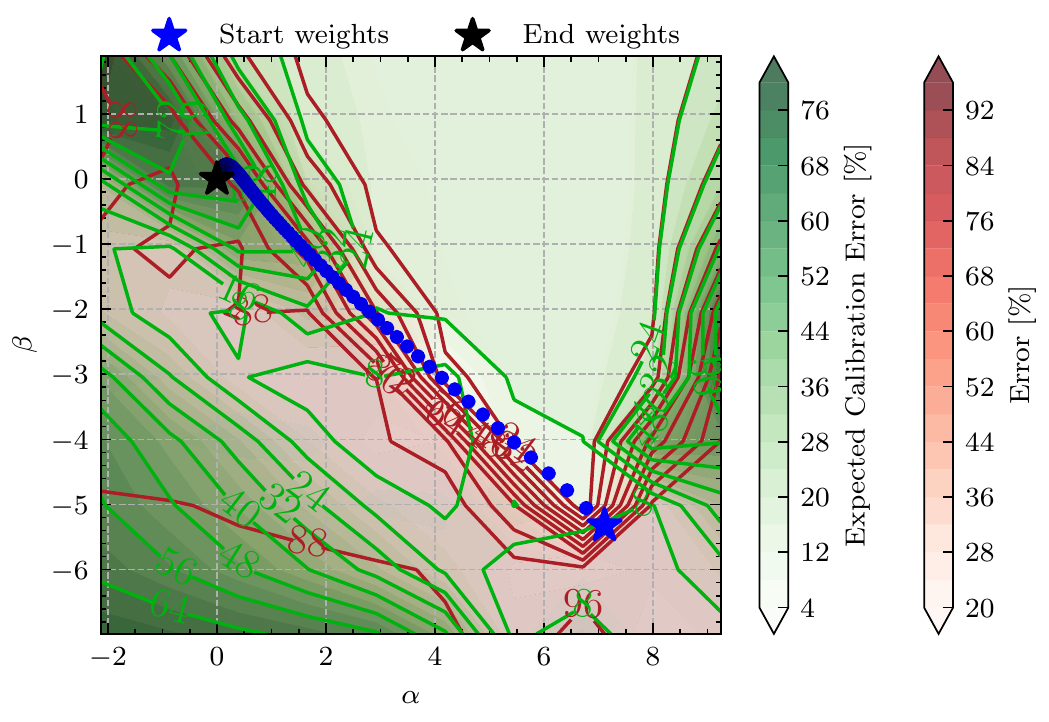}
	\caption{Error and ECE on OOD.}
	\label{fig:loss_landscape:cifar10-resnet-input_random_crop_horizontal_flip-test_2d_aug_error_ece}
\end{subfigure}
\caption{Input Random Crop, Horizontal Flip on CIFAR-10.
\textit{Observations}: Did not change the smoothness of the 1D curves or the 2D metric landscape trajectory compared to no noise.
}
\label{fig:loss_landscape:cifar10-resnet-input_random_crop_horizontal_flip}
\end{figure}
\begin{figure}
\centering
\begin{subfigure}{0.21\textwidth}
	\centering
	\includegraphics[width=\textwidth]{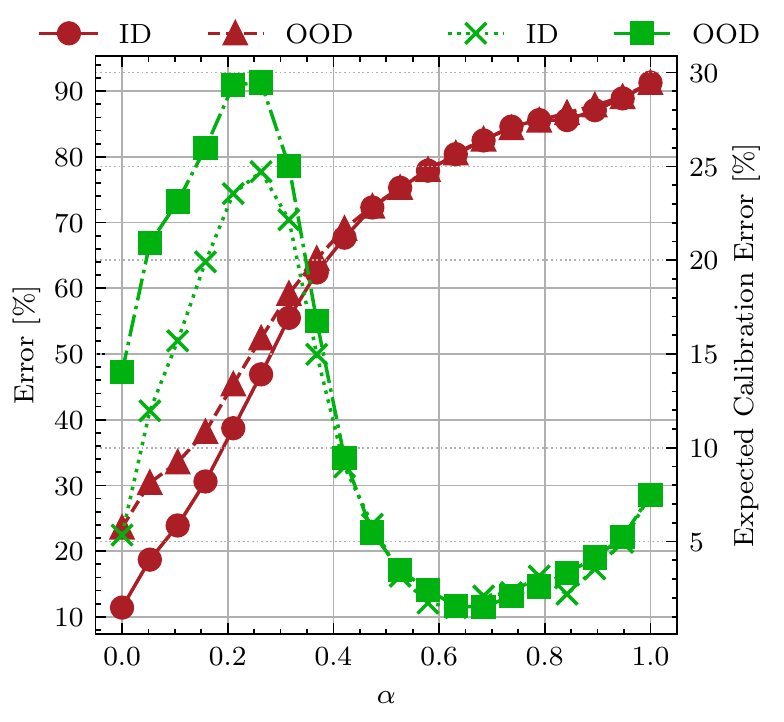}
	\caption{Error and ECE.}
	\label{fig:loss_landscape:cifar10-resnet-input_additive_gaussian-lin_error_ece}
\end{subfigure}
\begin{subfigure}{0.21\textwidth}
	\centering
	\includegraphics[width=\textwidth]{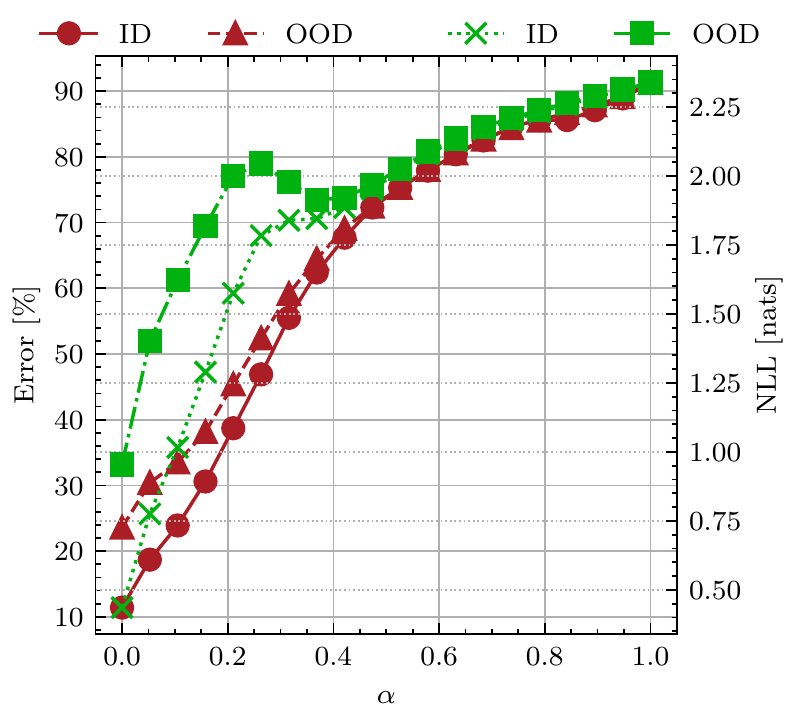}
	\caption{Error and NLL.}
	\label{fig:loss_landscape:cifar10-resnet-input_additive_gaussian-lin_error_nll}
\end{subfigure}
\begin{subfigure}{0.25\textwidth}
	\centering
	\includegraphics[width=\textwidth]{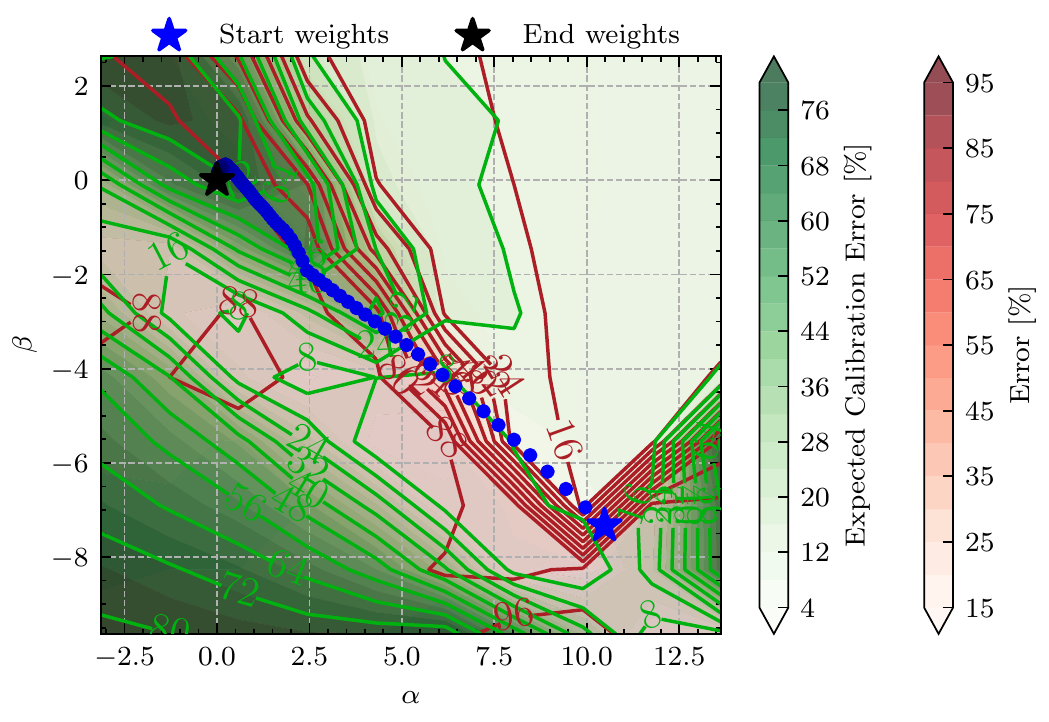}
	\caption{Error and ECE on ID.}
	\label{fig:loss_landscape:cifar10-resnet-input_additive_gaussian-test_2d_error_ece}
\end{subfigure}
\begin{subfigure}{0.25\textwidth}
	\centering
	\includegraphics[width=\textwidth]{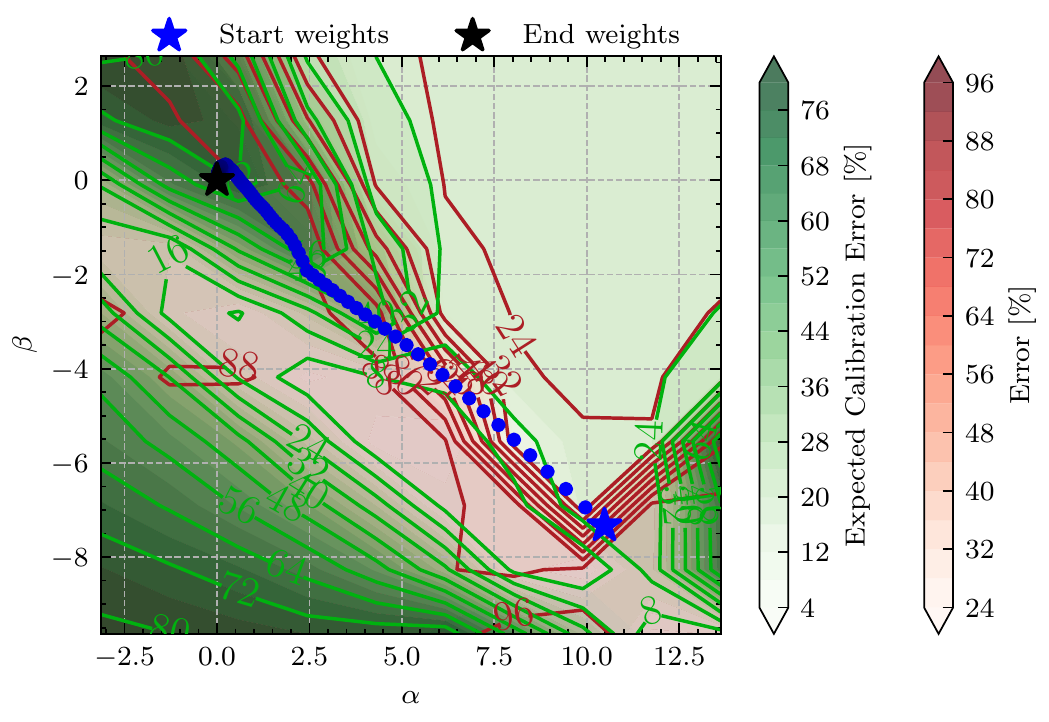}
	\caption{Error and ECE on OOD.}
	\label{fig:loss_landscape:cifar10-resnet-input_additive_gaussian-test_2d_aug_error_ece}
\end{subfigure}
\caption{Input Additive Gaussian on CIFAR-10.
\textit{Observations}: Changed the smoothness of the 1D curves where NLL became less smooth and removed the bumps in ECE for $\alpha$ approaching the initial model. 
The 2D metric landscape trajectory did not change in comparison to no noise.}
\label{fig:loss_landscape:cifar10-resnet-input_additive_gaussian}
\end{figure}
\begin{figure}
\centering
\begin{subfigure}{0.21\textwidth}
	\centering
	\includegraphics[width=\textwidth]{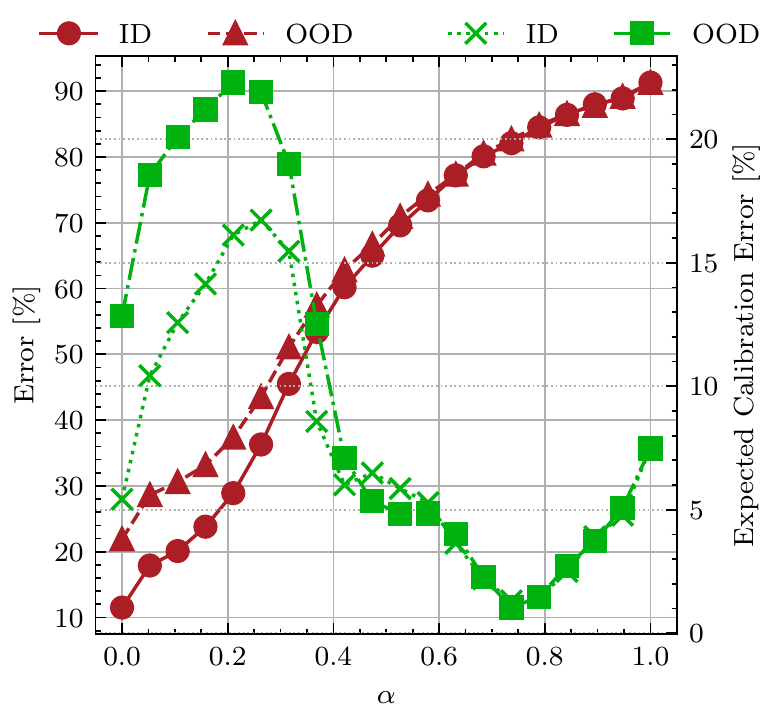}
	\caption{Error and ECE.}
	\label{fig:loss_landscape:cifar10-resnet-input_ods-lin_error_ece}
\end{subfigure}
\begin{subfigure}{0.21\textwidth}
	\centering
	\includegraphics[width=\textwidth]{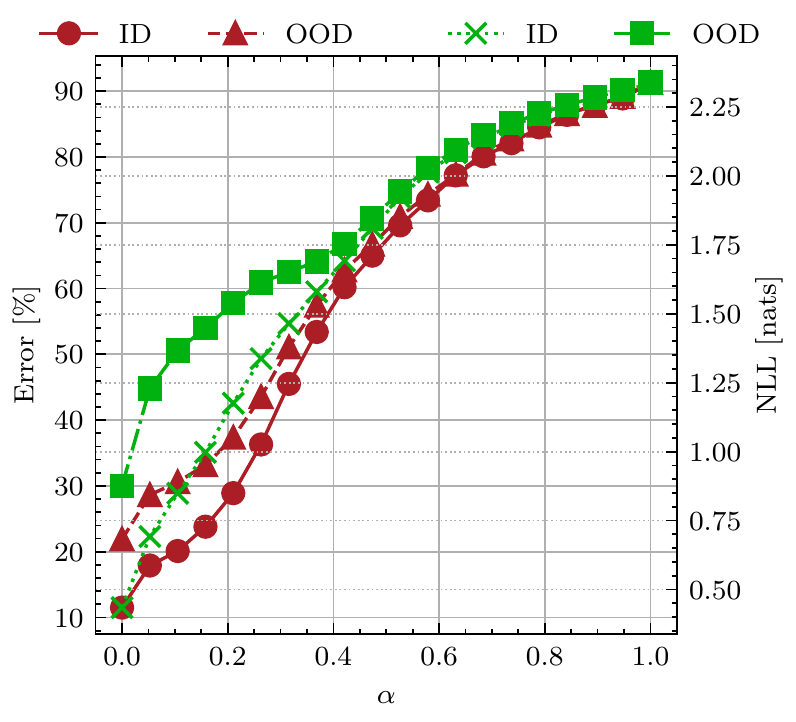}
	\caption{Error and NLL.}
	\label{fig:loss_landscape:cifar10-resnet-input_ods-lin_error_nll}
\end{subfigure}
\begin{subfigure}{0.25\textwidth}
	\centering
	\includegraphics[width=\textwidth]{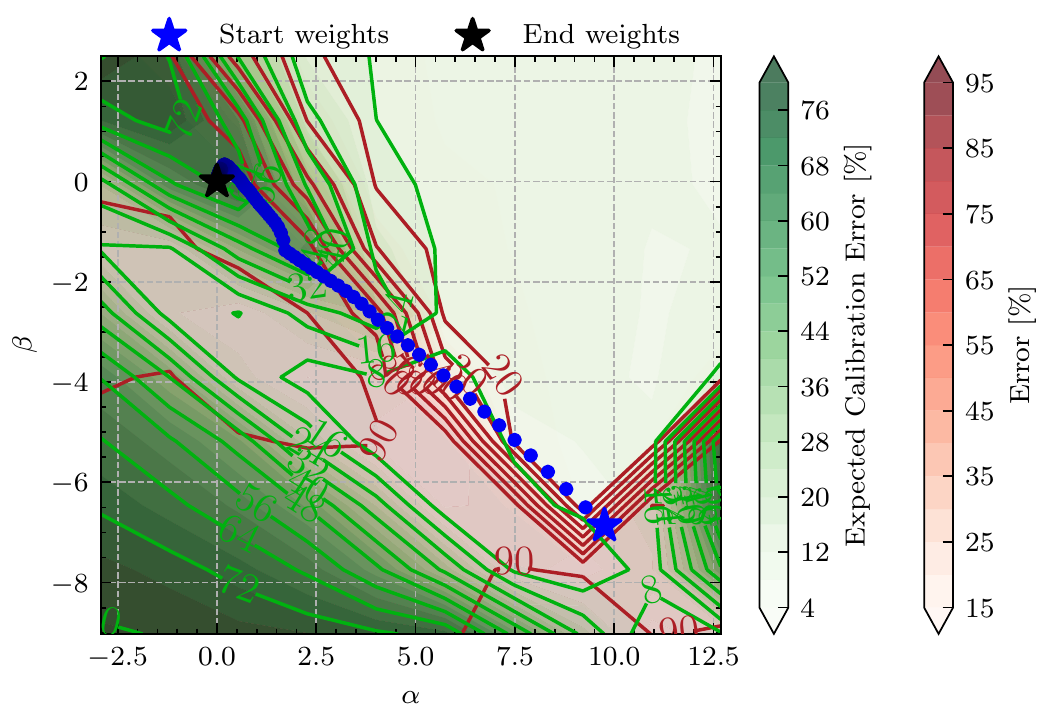}
	\caption{Error and ECE on ID.}
	\label{fig:loss_landscape:cifar10-resnet-input_ods-test_2d_error_ece}
\end{subfigure}
\begin{subfigure}{0.25\textwidth}
	\centering
	\includegraphics[width=\textwidth]{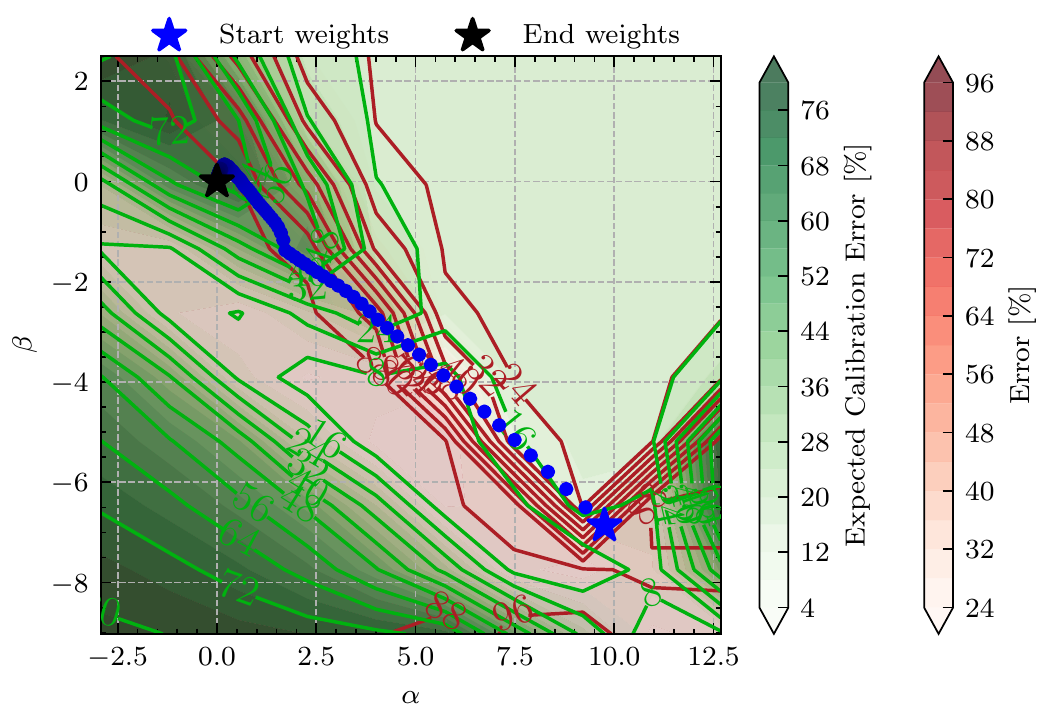}
	\caption{Error and ECE on OOD.}
	\label{fig:loss_landscape:cifar10-resnet-input_ods-test_2d_aug_error_ece}
\end{subfigure}
\caption{Input ODS on CIFAR-10.
\textit{Observations}: Marginally changed the smoothness of the 1D curves. 
The 2D metric landscape trajectory did not change in comparison to no noise.}
\label{fig:loss_landscape:cifar10-resnet-input_ods}
\end{figure}
\begin{figure}
\centering
\begin{subfigure}{0.21\textwidth}
	\centering
	\includegraphics[width=\textwidth]{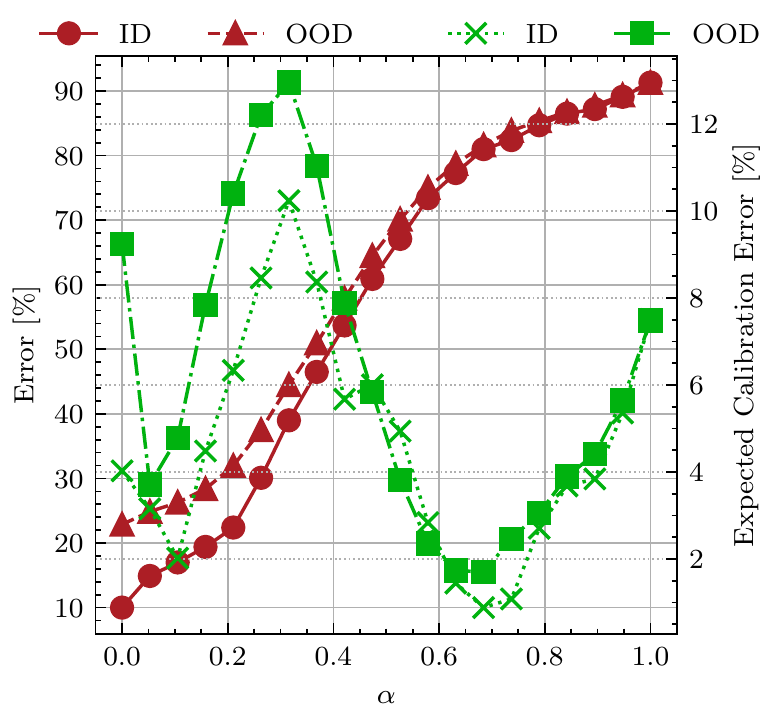}
	\caption{Error and ECE.}
	\label{fig:loss_landscape:cifar10-resnet-input_target_mixup-lin_error_ece}
\end{subfigure}
\begin{subfigure}{0.21\textwidth}
	\centering
	\includegraphics[width=\textwidth]{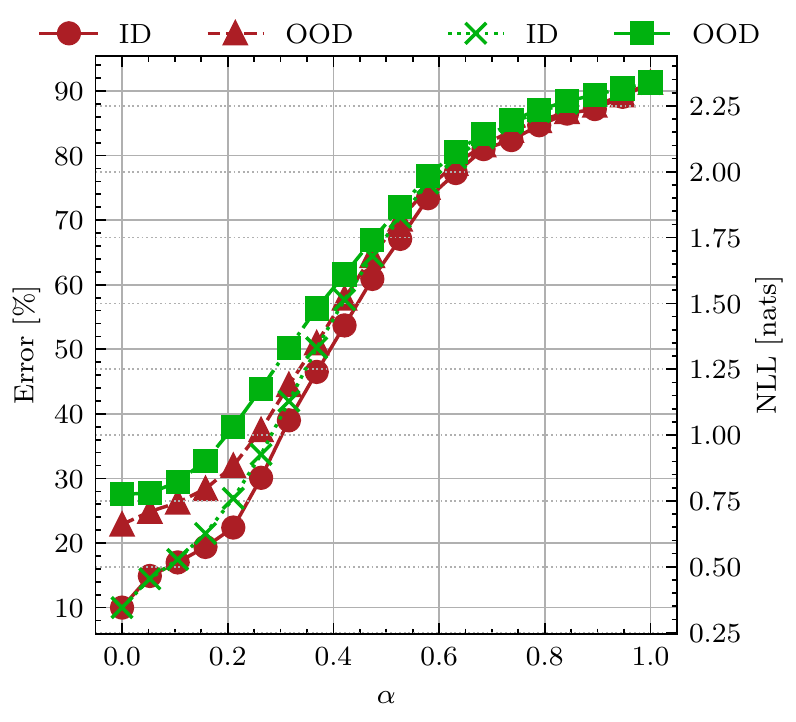}
	\caption{Error and NLL.}
	\label{fig:loss_landscape:cifar10-resnet-input_target_mixup-lin_error_nll}
\end{subfigure}
\begin{subfigure}{0.25\textwidth}
	\centering
	\includegraphics[width=\textwidth]{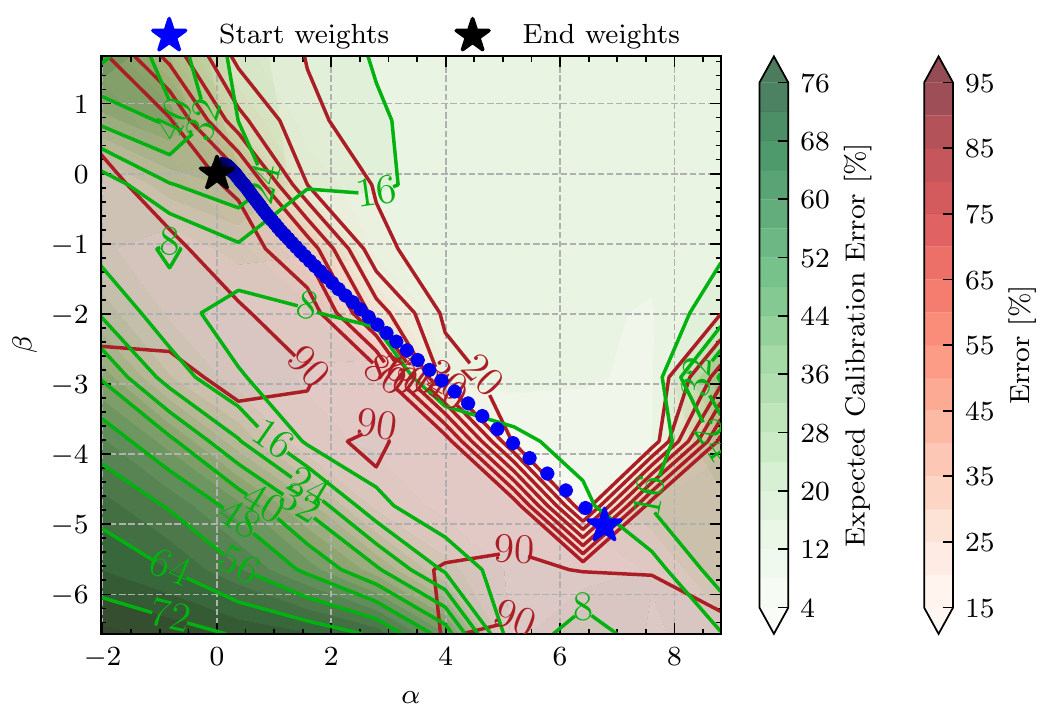}
	\caption{Error and ECE on ID.}
	\label{fig:loss_landscape:cifar10-resnet-input_target_mixup-test_2d_error_ece}
\end{subfigure}
\begin{subfigure}{0.25\textwidth}
	\centering
	\includegraphics[width=\textwidth]{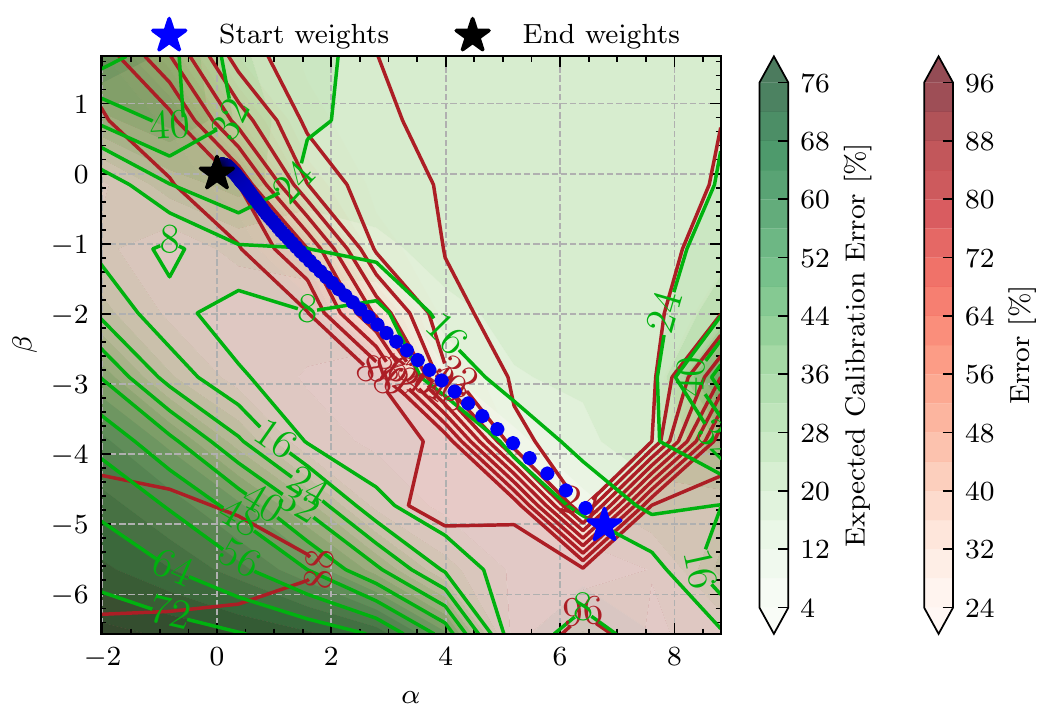}
	\caption{Error and ECE on OOD.}
	\label{fig:loss_landscape:cifar10-resnet-input_target_mixup-test_2d_aug_error_ece}
\end{subfigure}
\caption{Input-Target MixUp on CIFAR-10.
\textit{Observations}: Both the NLL and ECE 1D curves changed in comparison to no noise, and the 2D plots seem to explore wider valleys compared to no noise.
}
\label{fig:loss_landscape:cifar10-resnet-input_target_mixup}
\end{figure}
\begin{figure}
\centering
\begin{subfigure}{0.21\textwidth}
	\centering
	\includegraphics[width=\textwidth]{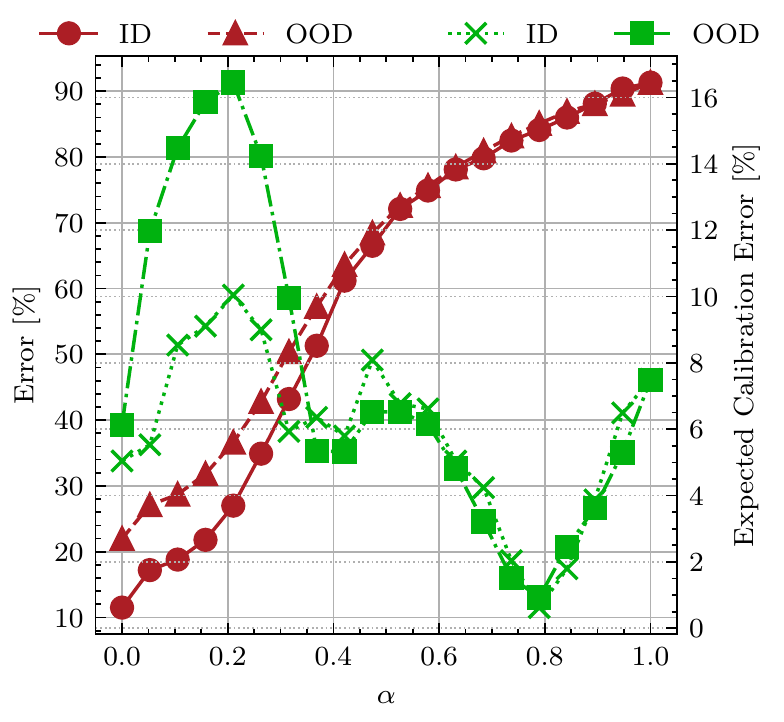}
	\caption{Error and ECE.}
	\label{fig:loss_landscape:cifar10-resnet-target_smoothing-lin_error_ece}
\end{subfigure}
\begin{subfigure}{0.21\textwidth}
	\centering
	\includegraphics[width=\textwidth]{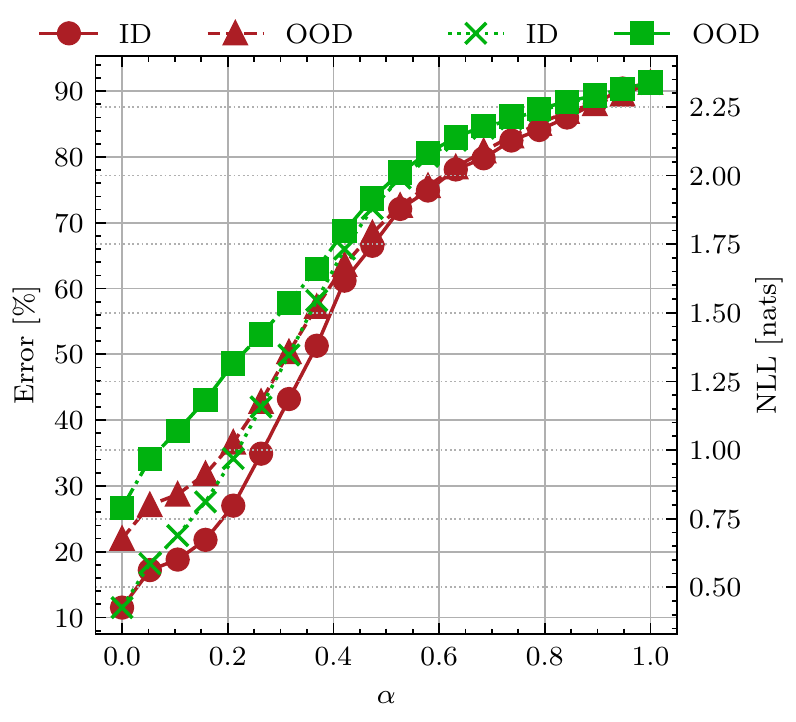}
	\caption{Error and NLL.}
	\label{fig:loss_landscape:cifar10-resnet-target_smoothing-lin_error_nll}
\end{subfigure}
\begin{subfigure}{0.25\textwidth}
	\centering
	\includegraphics[width=\textwidth]{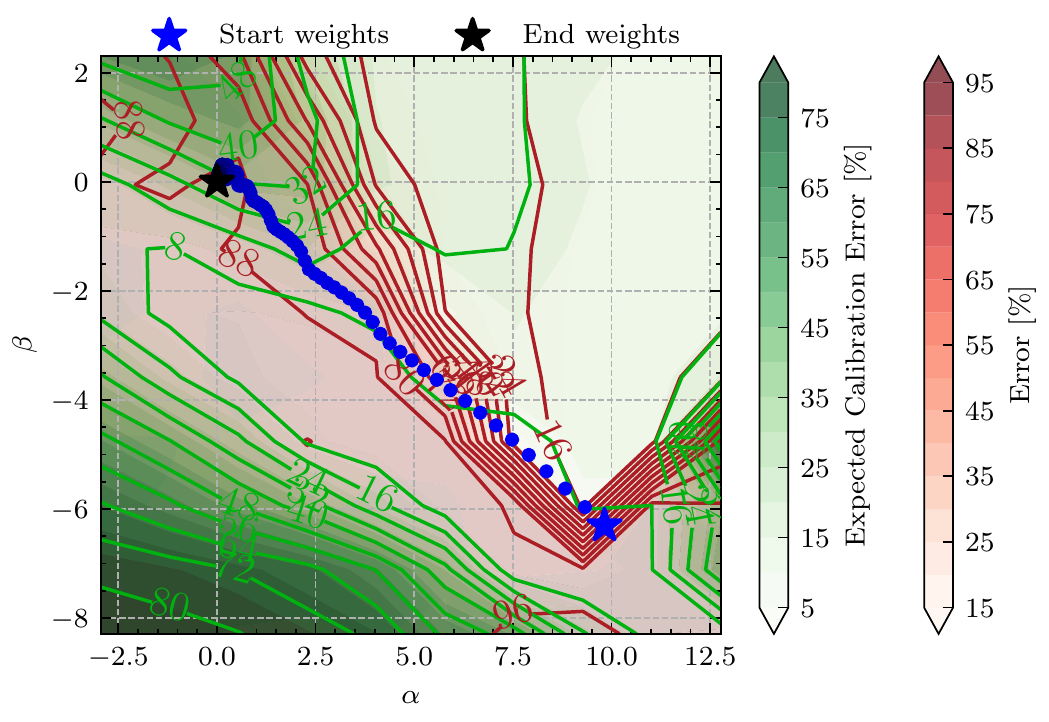}
	\caption{Error and ECE on ID.}
	\label{fig:loss_landscape:cifar10-resnet-target_smoothing-test_2d_error_ece}
\end{subfigure}
\begin{subfigure}{0.25\textwidth}
	\centering
	\includegraphics[width=\textwidth]{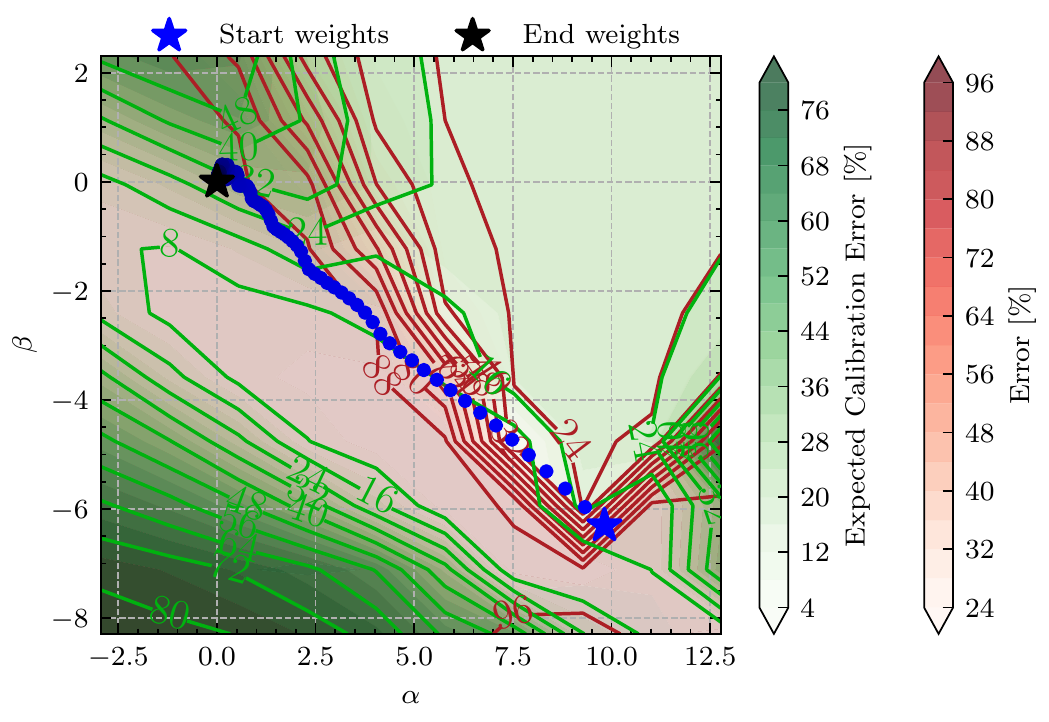}
	\caption{Error and ECE on OOD.}
	\label{fig:loss_landscape:cifar10-resnet-target_smoothing-test_2d_aug_error_ece}
\end{subfigure}
\caption{Target Smoothing on CIFAR-10.
\textit{Observations}: The NLL became more aligned with the error, not the ECE. 
The 2D plots show slightly more variation in the trajectory than no noise.}
\label{fig:loss_landscape:cifar10-resnet-target_smoothing}
\end{figure}
\begin{figure}
\centering
\begin{subfigure}{0.21\textwidth}
	\centering
	\includegraphics[width=\textwidth]{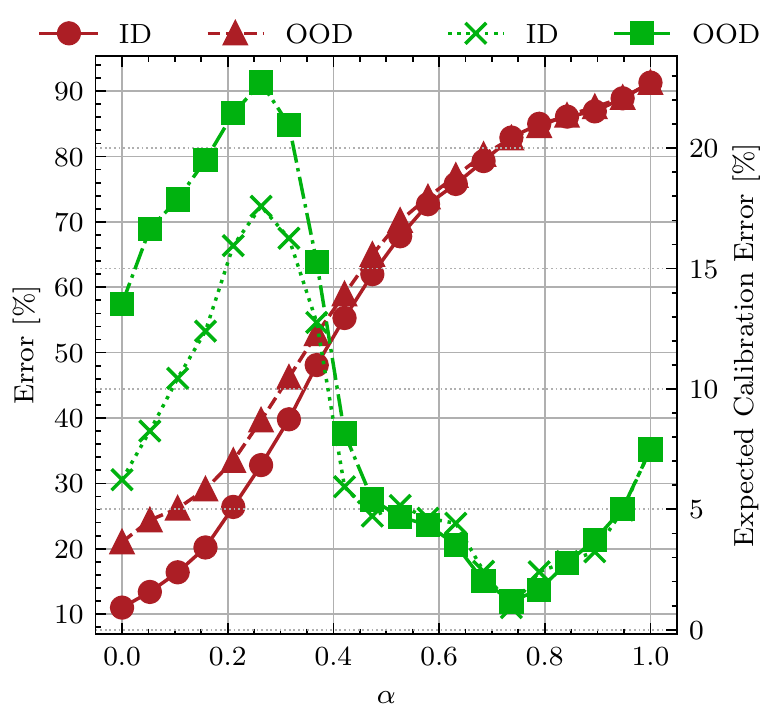}
	\caption{Error and ECE.}
	\label{fig:loss_landscape:cifar10-resnet-activation_additive_gaussian-lin_error_ece}
\end{subfigure}
\begin{subfigure}{0.21\textwidth}
	\centering
	\includegraphics[width=\textwidth]{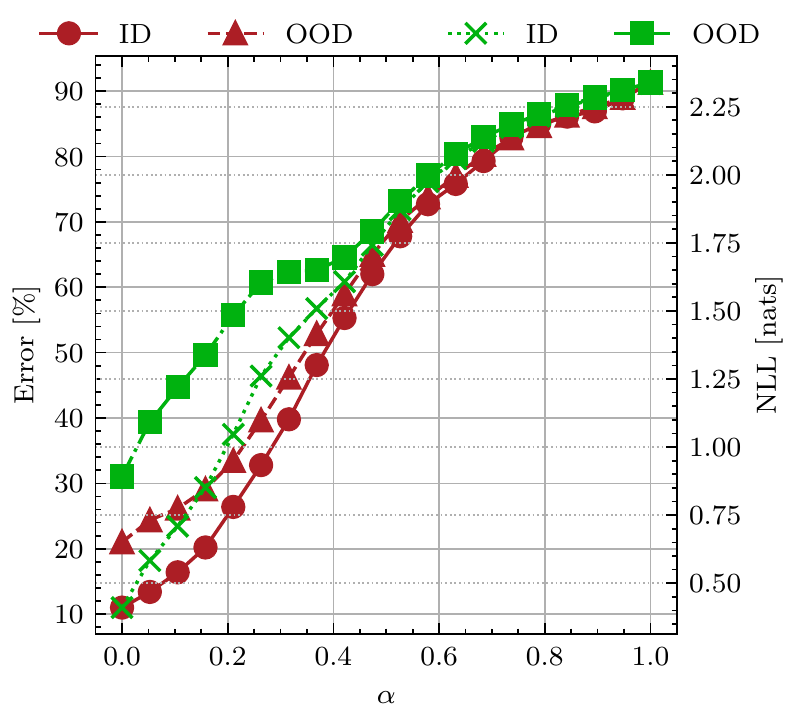}
	\caption{Error and NLL.}
	\label{fig:loss_landscape:cifar10-resnet-activation_additive_gaussian-lin_error_nll}
\end{subfigure}
\begin{subfigure}{0.25\textwidth}
	\centering
	\includegraphics[width=\textwidth]{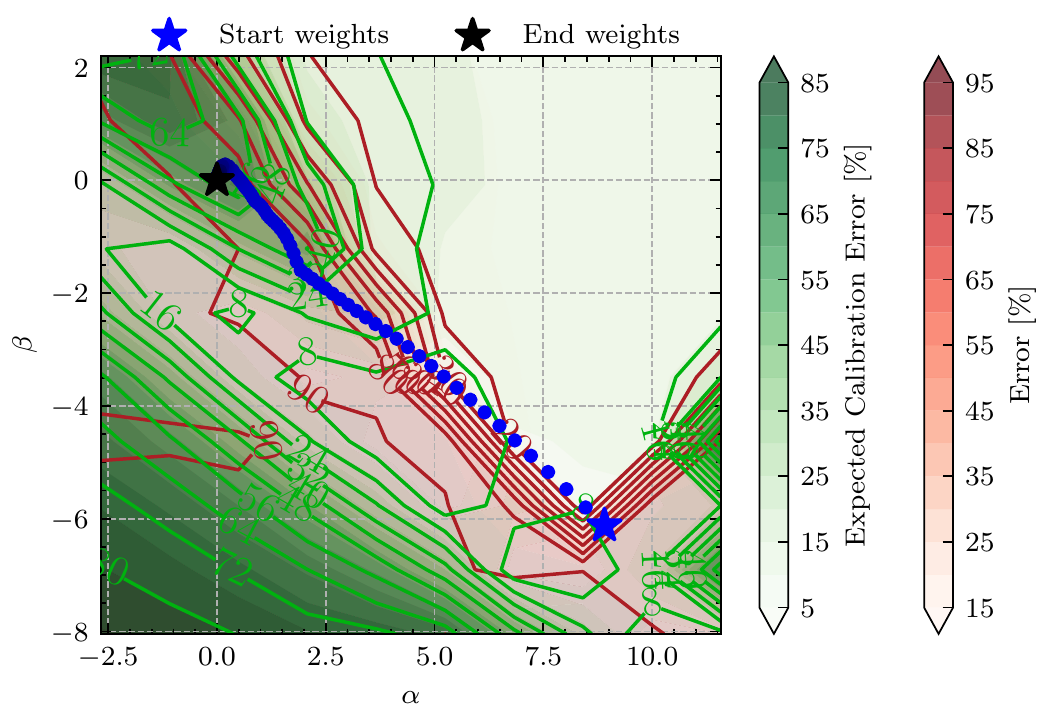}
	\caption{Error and ECE on ID.}
	\label{fig:loss_landscape:cifar10-resnet-activation_additive_gaussian-test_2d_error_ece}
\end{subfigure}
\begin{subfigure}{0.25\textwidth}
	\centering
	\includegraphics[width=\textwidth]{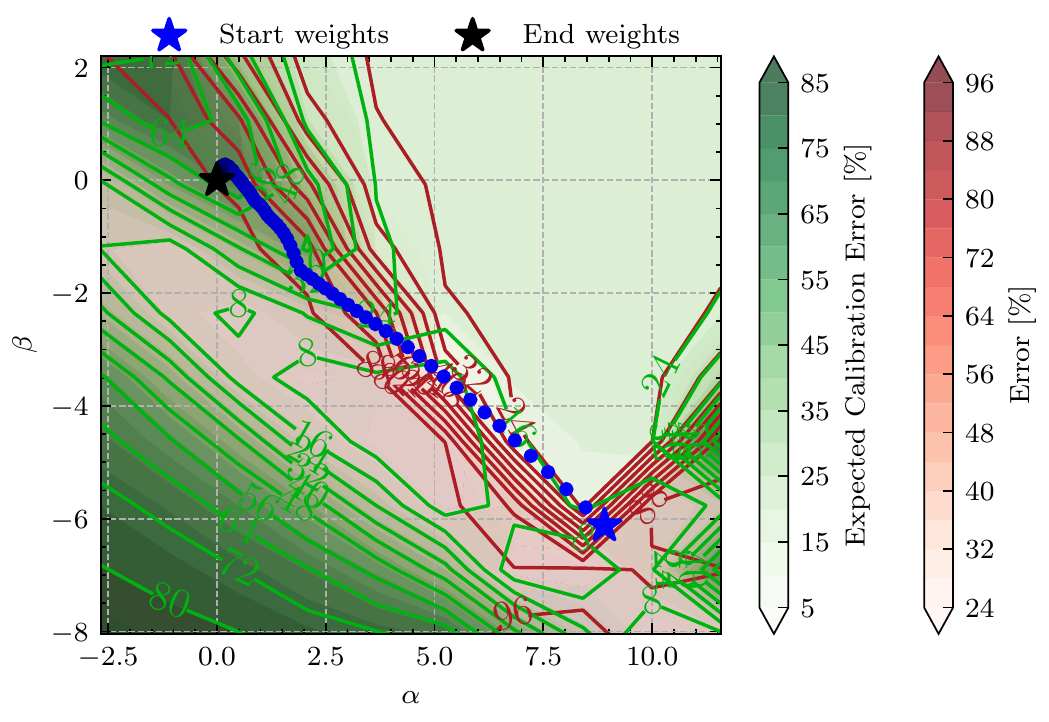}
	\caption{Error and ECE on OOD.}
	\label{fig:loss_landscape:cifar10-resnet-activation_additive_gaussian-test_2d_aug_error_ece}
\end{subfigure}
\caption{Activation Additive Gaussian on CIFAR-10.
\textit{Observations}: Did not change the smoothness of the 1D curves or the 2D metric landscape trajectory compared to no noise.}
\label{fig:loss_landscape:cifar10-resnet-activation_additive_gaussian}
\end{figure}
\begin{figure}
\centering
\begin{subfigure}{0.21\textwidth}
	\centering
	\includegraphics[width=\textwidth]{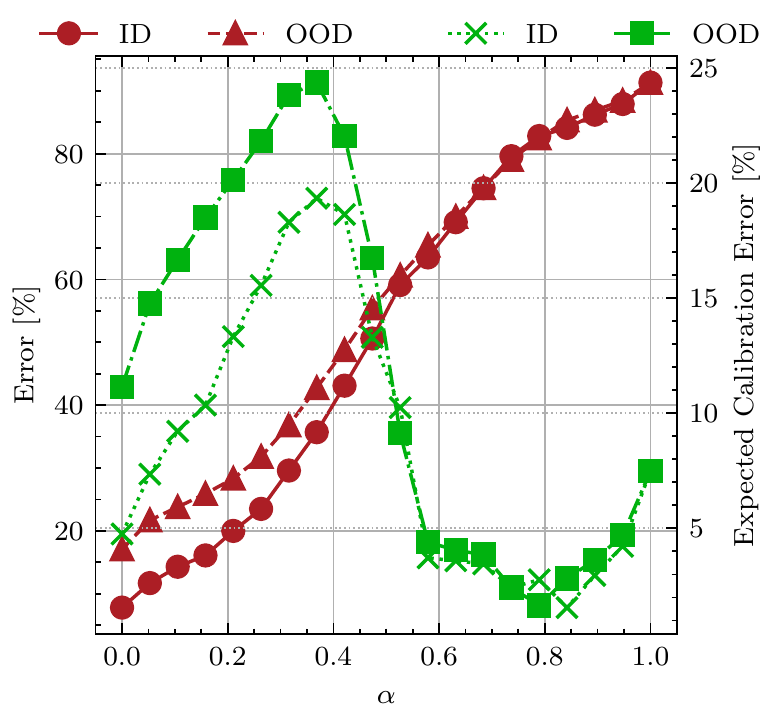}
	\caption{Error and ECE.}
	\label{fig:loss_landscape:cifar10-resnet-activation_dropout-lin_error_ece}
\end{subfigure}
\begin{subfigure}{0.21\textwidth}
	\centering
	\includegraphics[width=\textwidth]{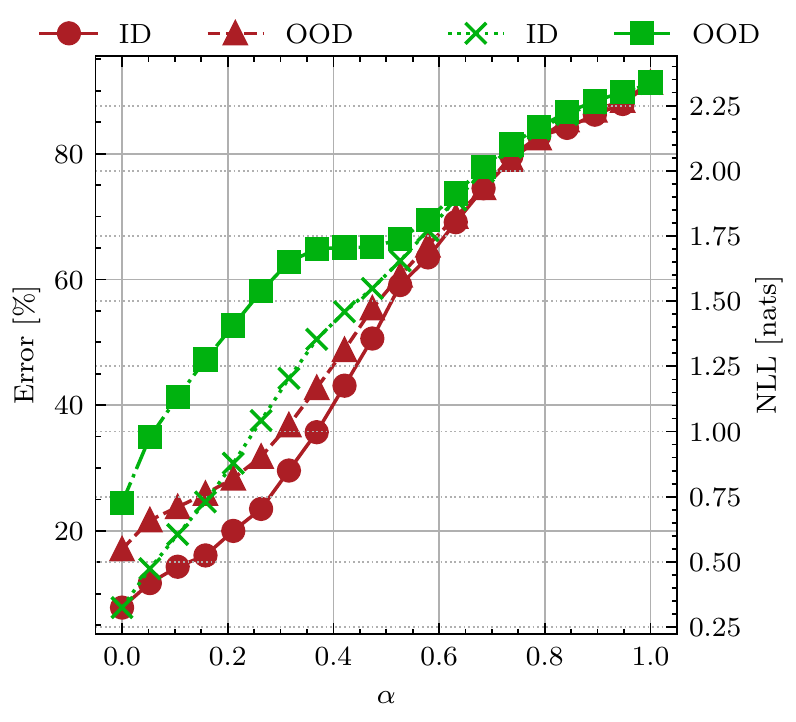}
	\caption{Error and NLL.}
	\label{fig:loss_landscape:cifar10-resnet-activation_dropout-lin_error_nll}
\end{subfigure}
\begin{subfigure}{0.25\textwidth}
	\centering
	\includegraphics[width=\textwidth]{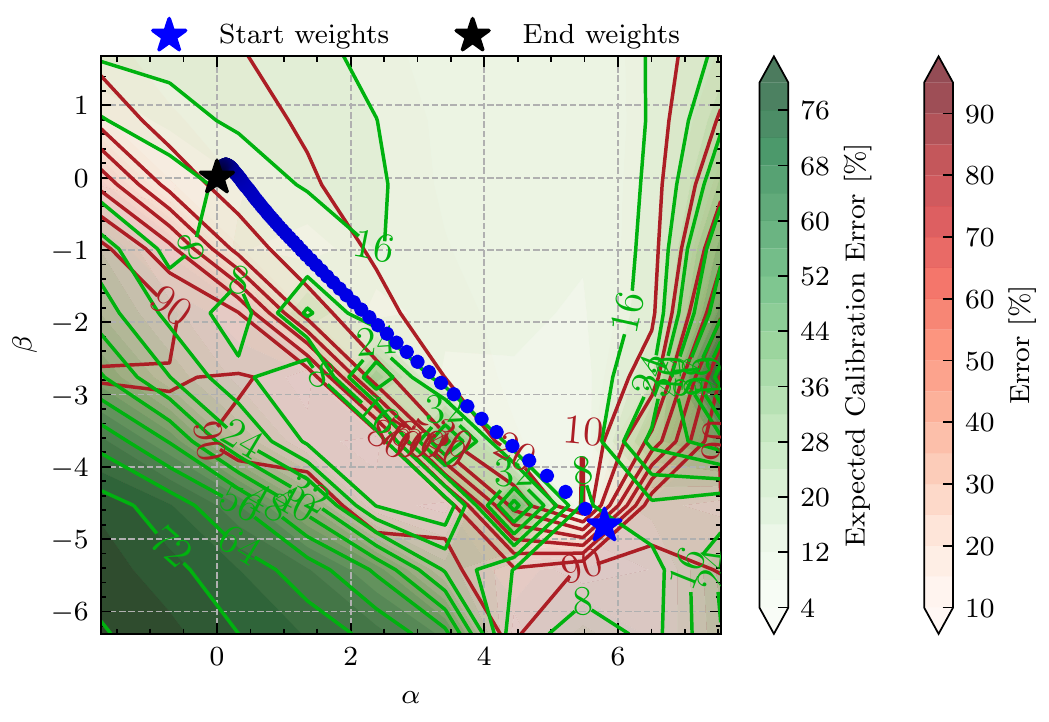}
	\caption{Error and ECE on ID.}
	\label{fig:loss_landscape:cifar10-resnet-activation_dropout-test_2d_error_ece}
\end{subfigure}
\begin{subfigure}{0.25\textwidth}
	\centering
	\includegraphics[width=\textwidth]{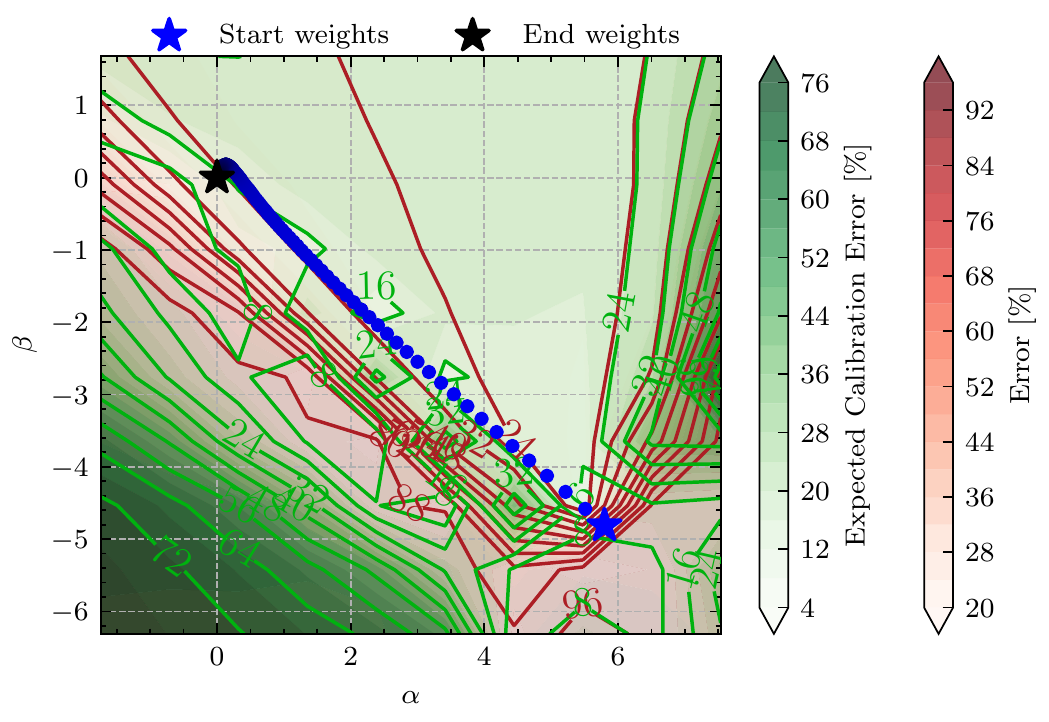}
	\caption{Error and ECE on OOD.}
	\label{fig:loss_landscape:cifar10-resnet-activation_dropout-test_2d_aug_error_ece}
\end{subfigure}
\caption{Activation Dropout on CIFAR-10.
\textit{Observations}: Dropout narrowed the gap between ID and OOD results; nevertheless, the shape of the 1D curves is similar to no noise.
The trajectories in 2D plots did not seem to converge into a narrow local minimum.}
\label{fig:loss_landscape:cifar10-resnet-activation_dropout}
\end{figure}
\begin{figure}
\centering
\begin{subfigure}{0.21\textwidth}
	\centering
	\includegraphics[width=\textwidth]{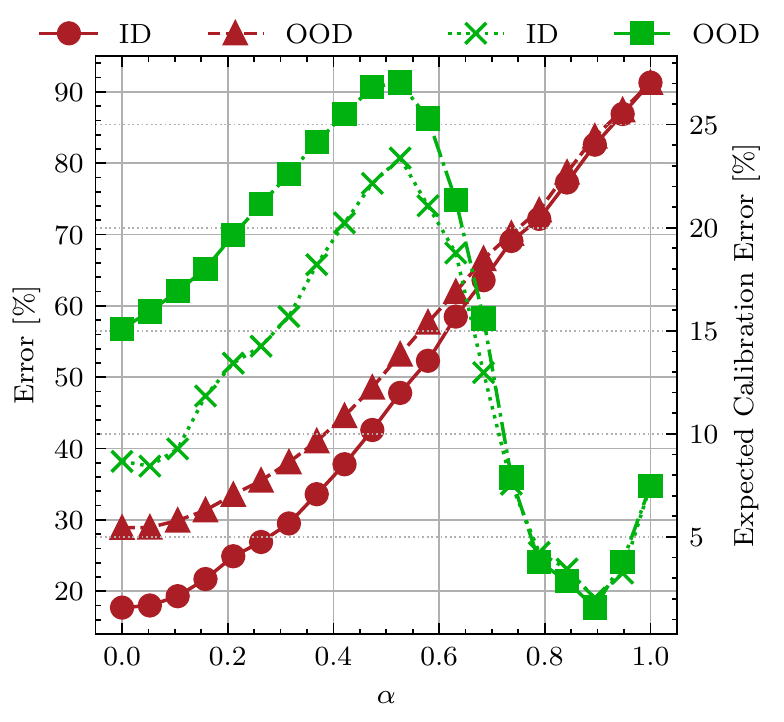}
	\caption{Error and ECE.}
	\label{fig:loss_landscape:cifar10-resnet-gradient_gaussian-lin_error_ece}
\end{subfigure}
\begin{subfigure}{0.21\textwidth}
	\centering
	\includegraphics[width=\textwidth]{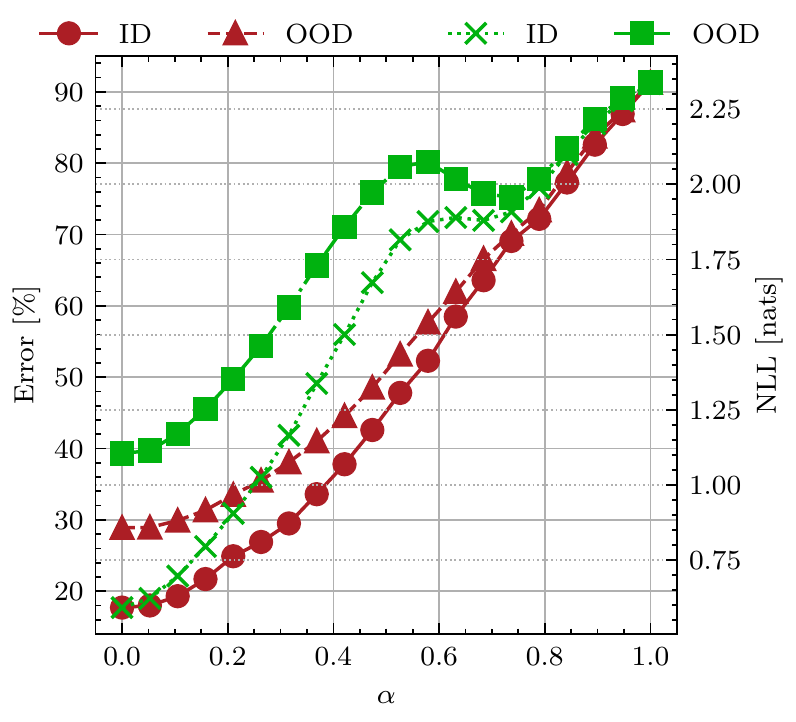}
	\caption{Error and NLL.}
	\label{fig:loss_landscape:cifar10-resnet-gradient_gaussian-lin_error_nll}
\end{subfigure}
\begin{subfigure}{0.25\textwidth}
	\centering
	\includegraphics[width=\textwidth]{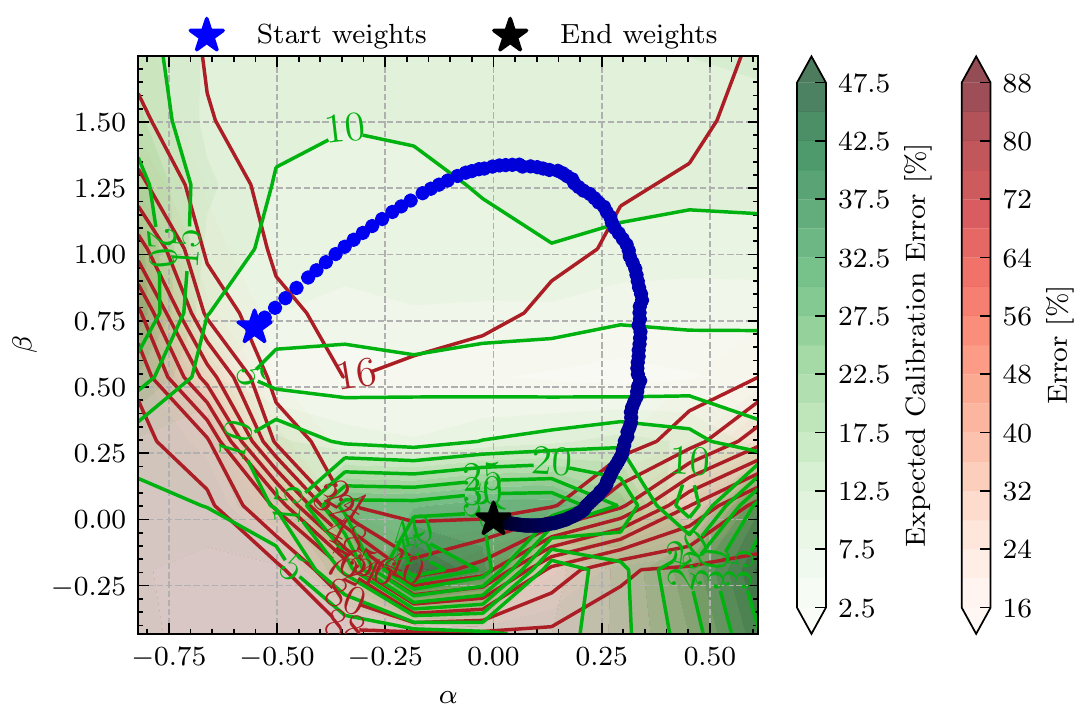}
	\caption{Error and ECE on ID.}
	\label{fig:loss_landscape:cifar10-resnet-gradient_gaussian-test_2d_error_ece}
\end{subfigure}
\begin{subfigure}{0.25\textwidth}
	\centering
	\includegraphics[width=\textwidth]{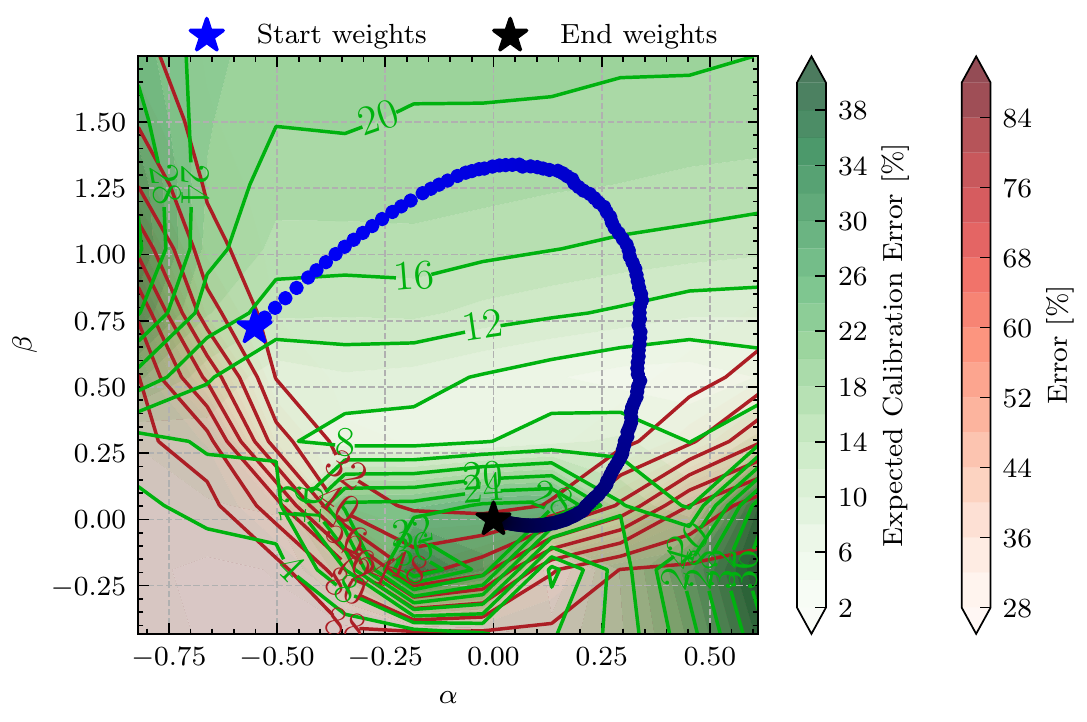}
	\caption{Error and ECE on OOD.}
	\label{fig:loss_landscape:cifar10-resnet-gradient_gaussian-test_2d_aug_error_ece}
\end{subfigure}
\caption{Gradient Gaussian on CIFAR-10.
\textit{Observations}: The 1D and 2D figures changed curvature and shape drastically, and NLL and ECE follow a non-linear pattern.
The 2D plots show a circular curvature, perhaps suggesting difficulty in convergence.
}
\label{fig:loss_landscape:cifar10-resnet-gradient_gaussian}
\end{figure}
\begin{figure}
\centering
\begin{subfigure}{0.21\textwidth}
	\centering
	\includegraphics[width=\textwidth]{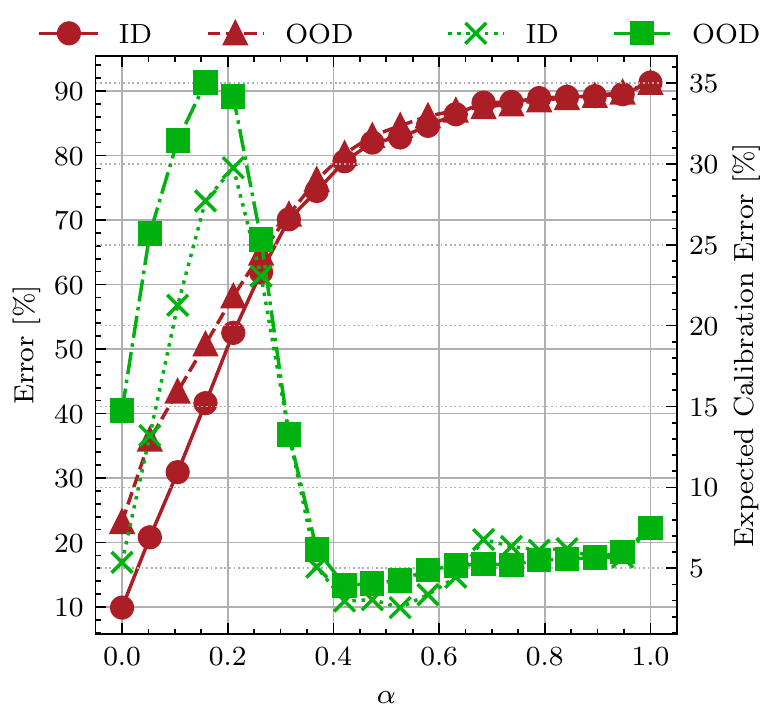}
	\caption{Error and ECE.}
	\label{fig:loss_landscape:cifar10-resnet-model_sp-lin_error_ece}
\end{subfigure}
\begin{subfigure}{0.21\textwidth}
	\centering
	\includegraphics[width=\textwidth]{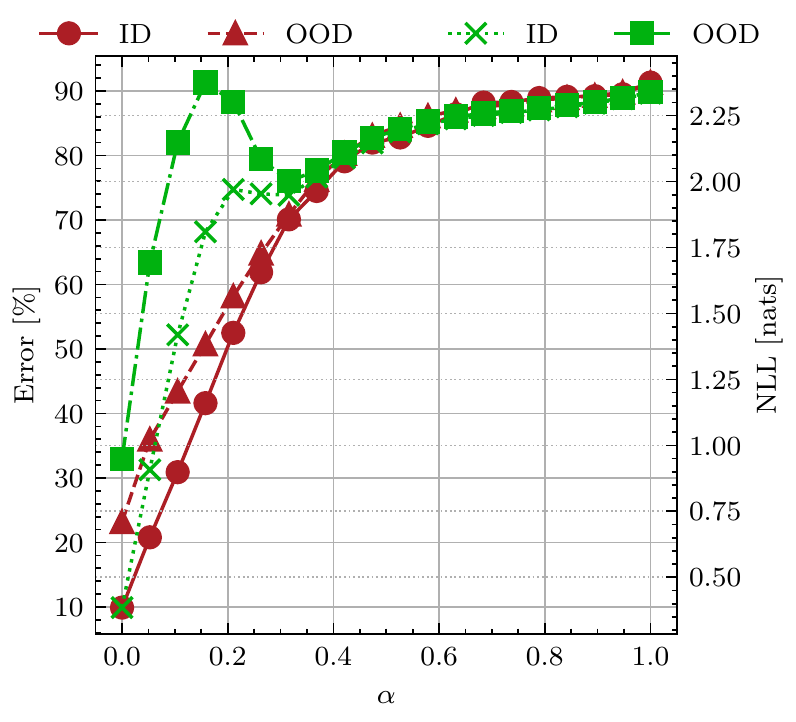}
	\caption{Error and NLL.}
	\label{fig:loss_landscape:cifar10-resnet-model_sp-lin_error_nll}
\end{subfigure}
\begin{subfigure}{0.25\textwidth}
	\centering
	\includegraphics[width=\textwidth]{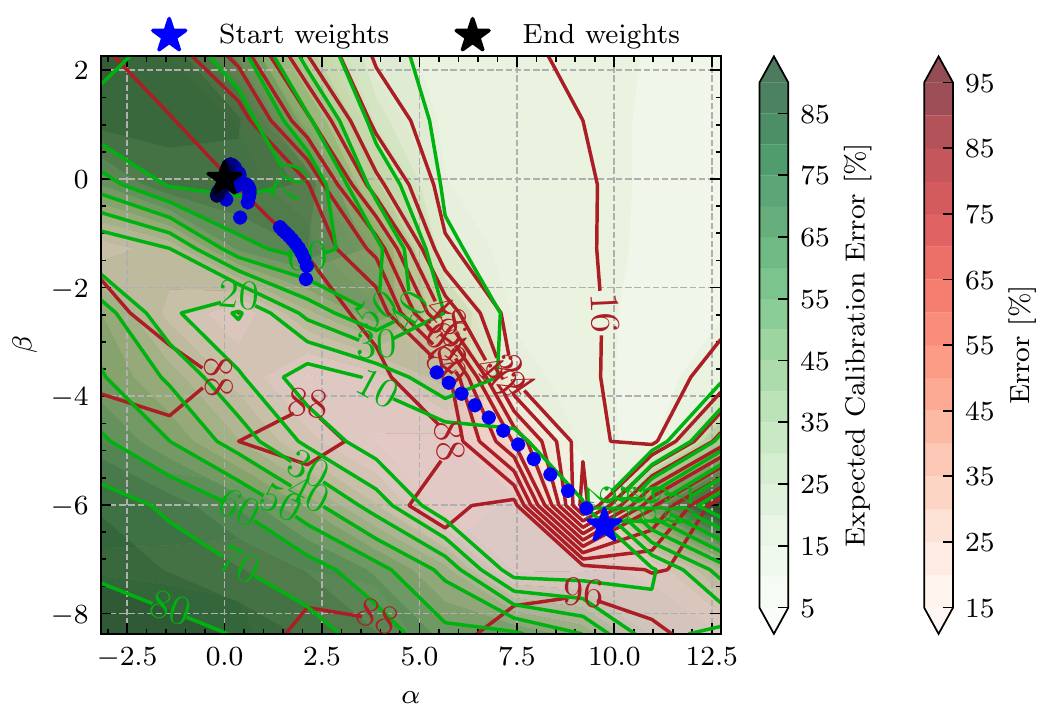}
	\caption{Error and ECE on ID.}
	\label{fig:loss_landscape:cifar10-resnet-model_sp-test_2d_error_ece}
\end{subfigure}
\begin{subfigure}{0.25\textwidth}
	\centering
	\includegraphics[width=\textwidth]{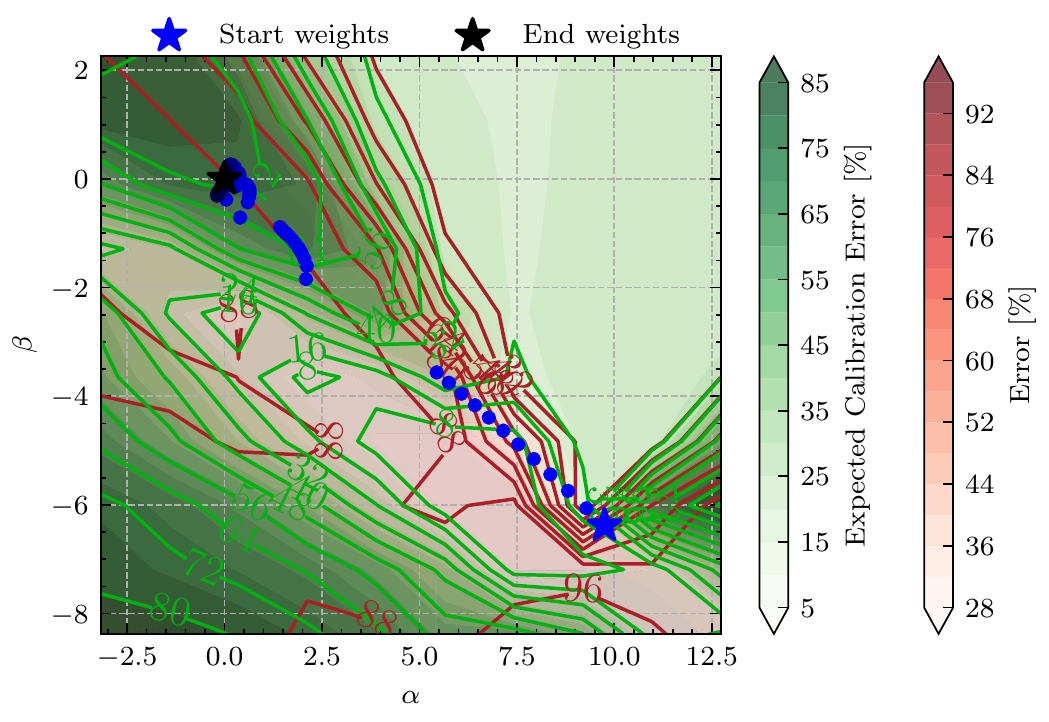}
	\caption{Error and ECE on OOD.}
	\label{fig:loss_landscape:cifar10-resnet-model_sp-test_2d_aug_error_ece}
\end{subfigure}
\caption{Model Shrink and Perturb on CIFAR-10.
\textit{Observations}: The 1D and 2D figures changed curvature and shape drastically, and all metrics show a non-linear optimisation path as hypothesised.
The point cluster around centres created by shrinking and perturbing the weights.
}
\label{fig:loss_landscape:cifar10-resnet-model_sp}
\end{figure}
\begin{figure}
\centering
\begin{subfigure}{0.21\textwidth}
	\centering
	\includegraphics[width=\textwidth]{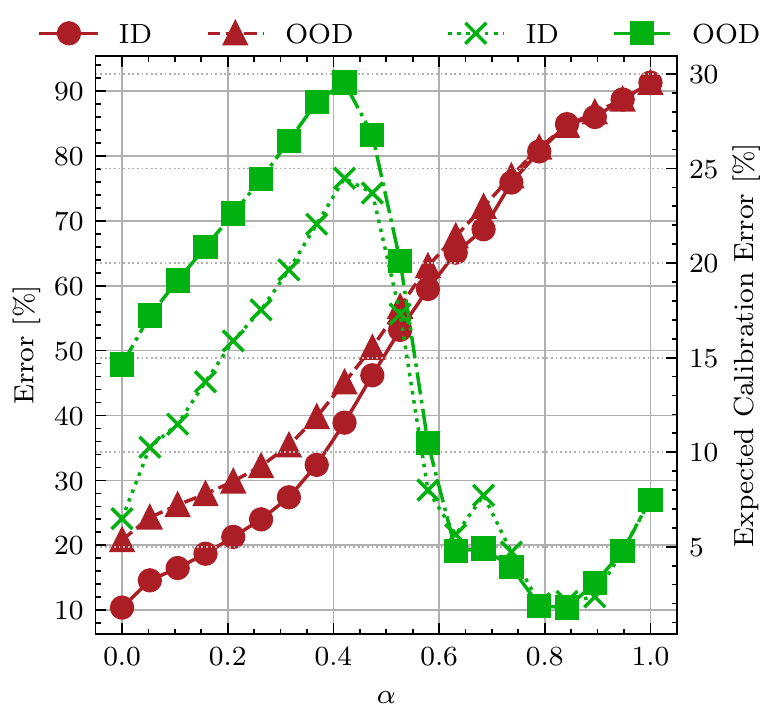}
	\caption{Error and ECE.}
	\label{fig:loss_landscape:cifar10-resnet-weight_additive_gaussian-lin_error_ece}
\end{subfigure}
\begin{subfigure}{0.21\textwidth}
	\centering
	\includegraphics[width=\textwidth]{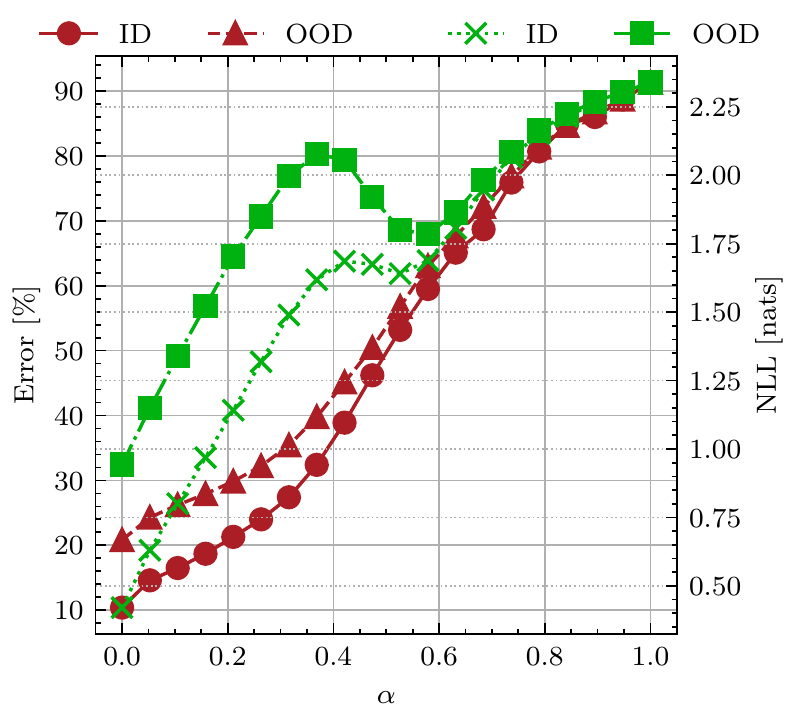}
	\caption{Error and NLL.}
	\label{fig:loss_landscape:cifar10-resnet-weight_additive_gaussian-lin_error_nll}
\end{subfigure}
\begin{subfigure}{0.25\textwidth}
	\centering
	\includegraphics[width=\textwidth]{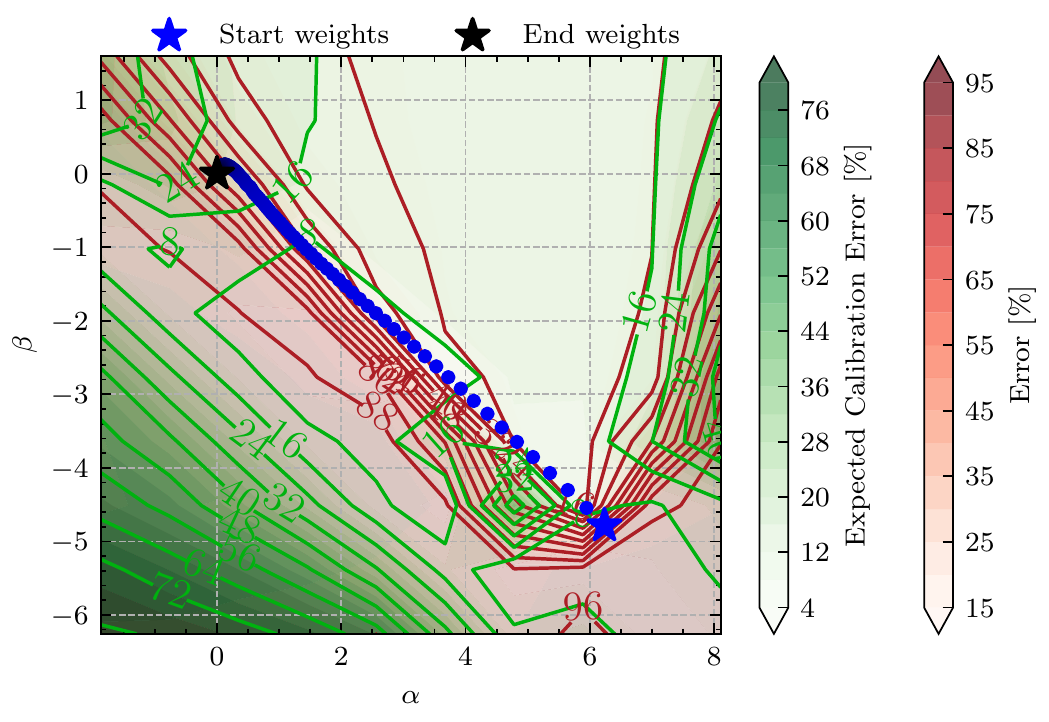}
	\caption{Error and ECE on ID.}
	\label{fig:loss_landscape:cifar10-resnet-weight_additive_gaussian-test_2d_error_ece}
\end{subfigure}
\begin{subfigure}{0.25\textwidth}
	\centering
	\includegraphics[width=\textwidth]{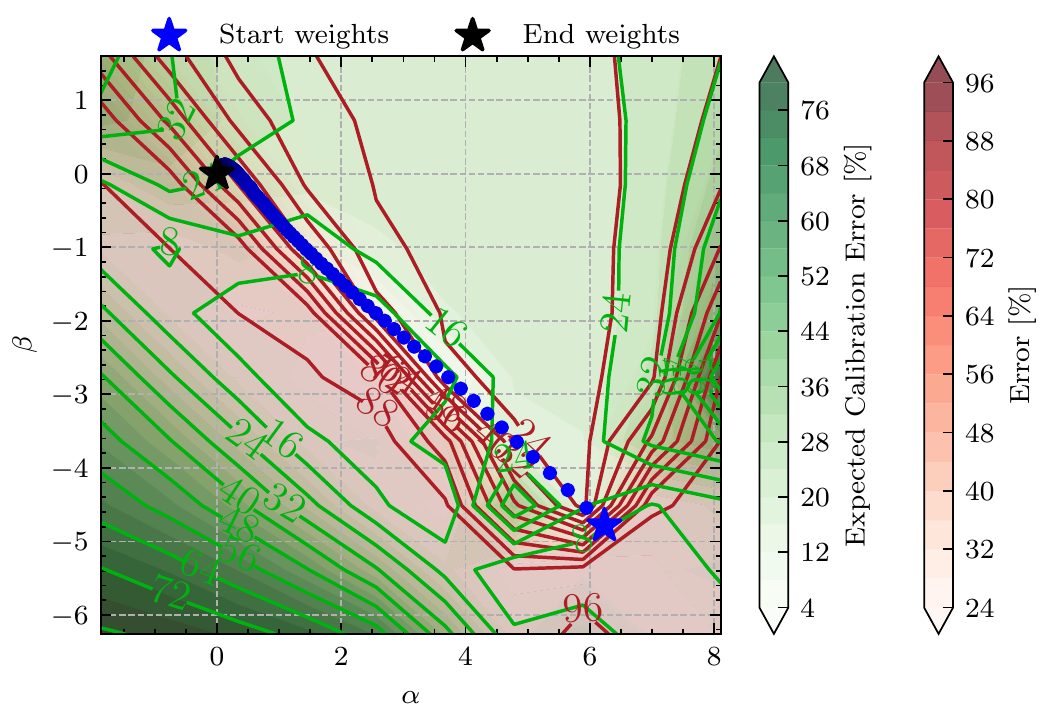}
	\caption{Error and ECE on OOD.}
	\label{fig:loss_landscape:cifar10-resnet-weight_additive_gaussian-test_2d_aug_error_ece}
\end{subfigure}
\caption{Weight Additive Gaussian on CIFAR-10.
\textit{Observations}: The 1D curves marginally changed their shape. 
However, the difference between ID and OOD metrics became more profound. 
The 2D plots suggest that the optimisation was not able to converge. 
}
\label{fig:loss_landscape:cifar10-resnet-weight_additive_gaussian}
\end{figure}
\begin{figure}
\centering
\begin{subfigure}{0.21\textwidth}
	\centering
	\includegraphics[width=\textwidth]{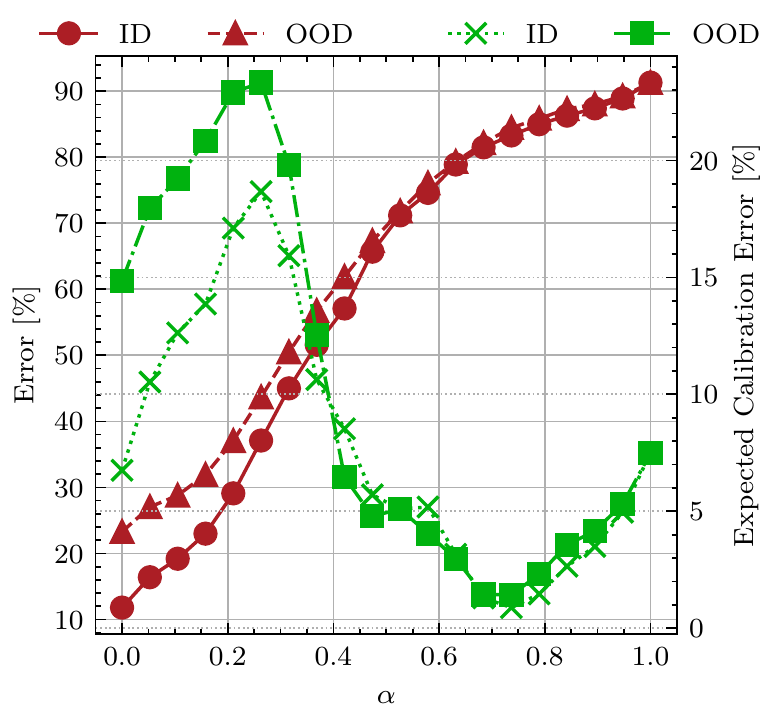}
	\caption{Error and ECE.}
	\label{fig:loss_landscape:cifar10-resnet-weight_dropconnect-lin_error_ece}
\end{subfigure}
\begin{subfigure}{0.21\textwidth}
	\centering
	\includegraphics[width=\textwidth]{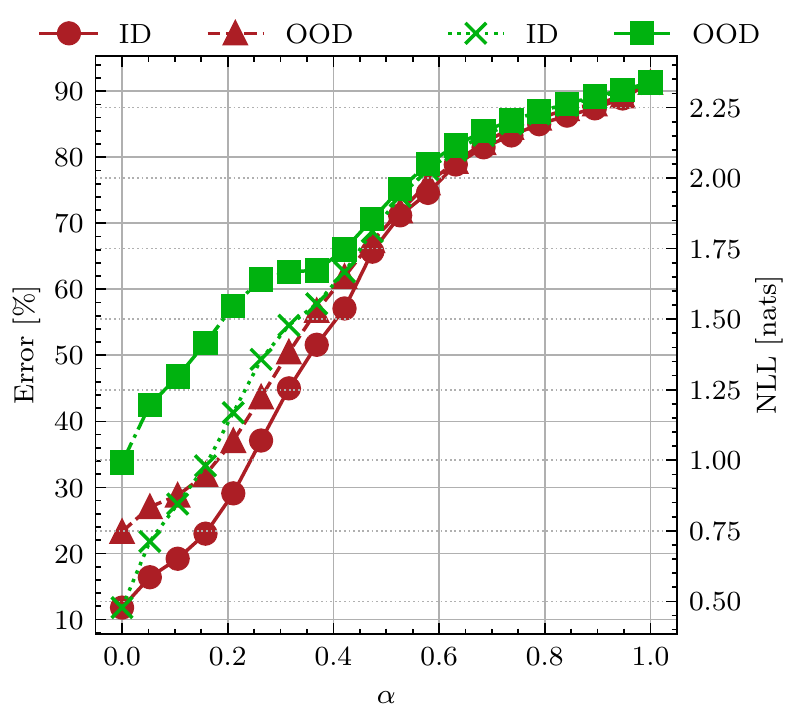}
	\caption{Error and NLL.}
	\label{fig:loss_landscape:cifar10-resnet-weight_dropconnect-lin_error_nll}
\end{subfigure}
\begin{subfigure}{0.25\textwidth}
	\centering
	\includegraphics[width=\textwidth]{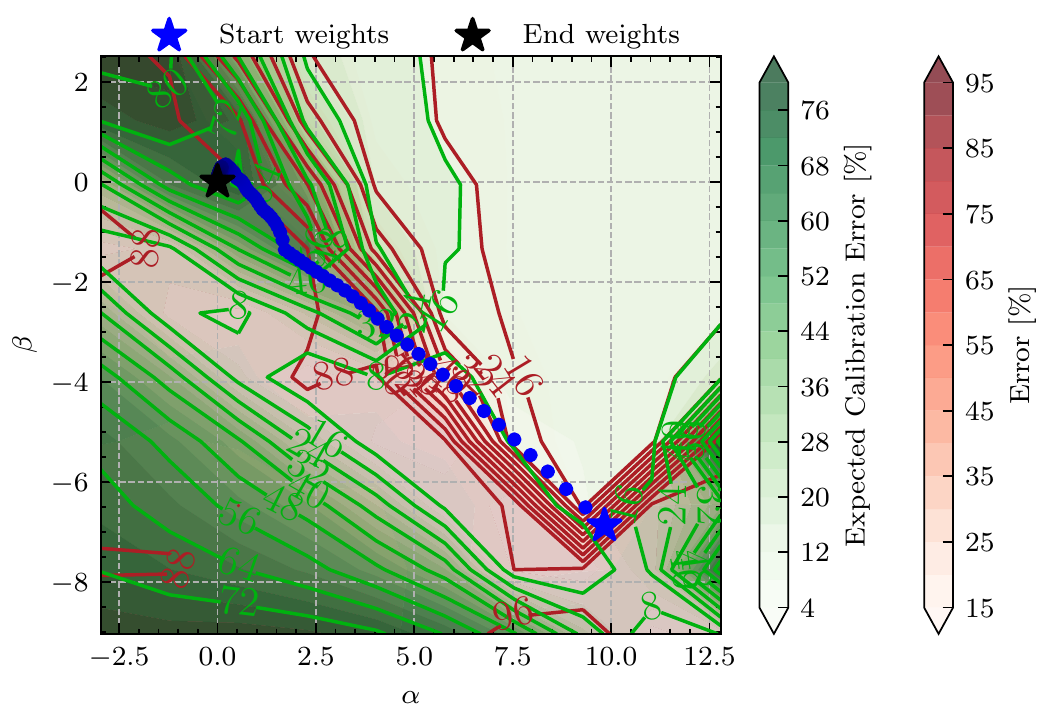}
	\caption{Error and ECE on ID.}
	\label{fig:loss_landscape:cifar10-resnet-weight_dropconnect-test_2d_error_ece}
\end{subfigure}
\begin{subfigure}{0.25\textwidth}
	\centering
	\includegraphics[width=\textwidth]{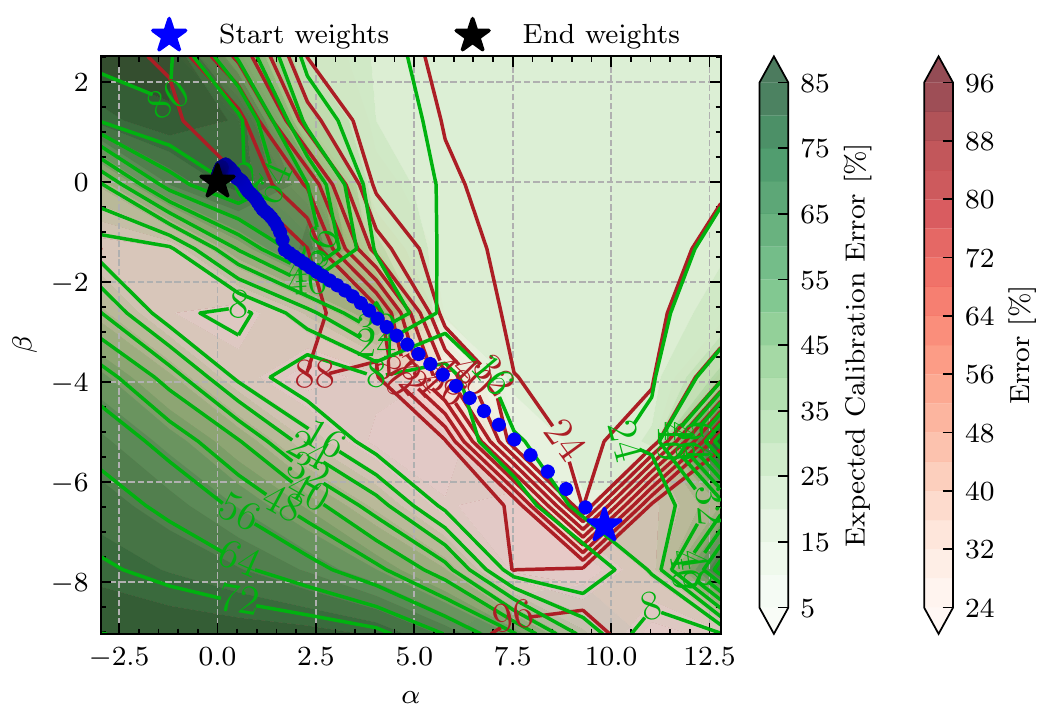}
	\caption{Error and ECE on OOD.}
	\label{fig:loss_landscape:cifar10-resnet-weight_dropconnect-test_2d_aug_error_ece}
\end{subfigure}
\caption{Weight DropConnect on CIFAR-10.
\textit{Observations}: Did not change the smoothness of the 1D curves or the 2D metric landscape trajectory compared to no noise.}
\label{fig:loss_landscape:cifar10-resnet-weight_dropconnect}
\end{figure}

\begin{figure}
\centering
\begin{subfigure}{0.21\textwidth}
	\centering
	\includegraphics[width=\textwidth]{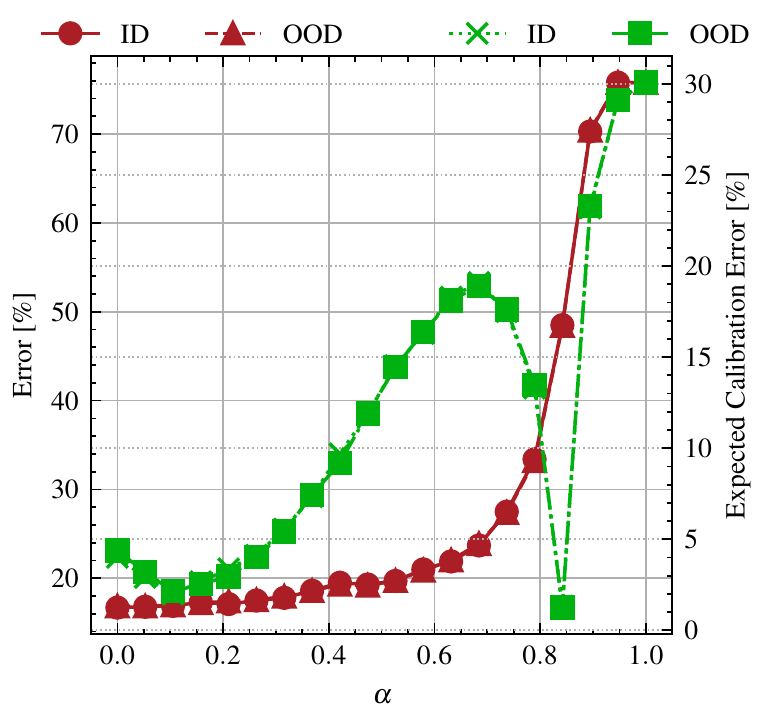}
	\caption{Error and ECE.}
	\label{fig:loss_landscape:classification_adult-fc-vanilla-lin_error_ece}
\end{subfigure}
\begin{subfigure}{0.21\textwidth}
	\centering
	\includegraphics[width=\textwidth]{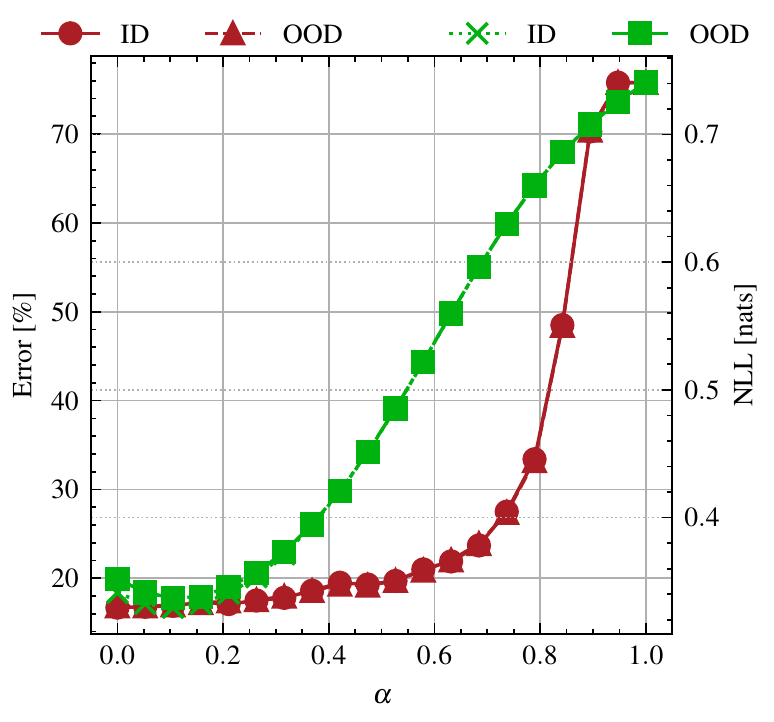}
	\caption{Error and NLL.}
	\label{fig:loss_landscape:classification_adult-fc-vanilla-lin_error_nll}
\end{subfigure}
\begin{subfigure}{0.25\textwidth}
	\centering
	\includegraphics[width=\textwidth]{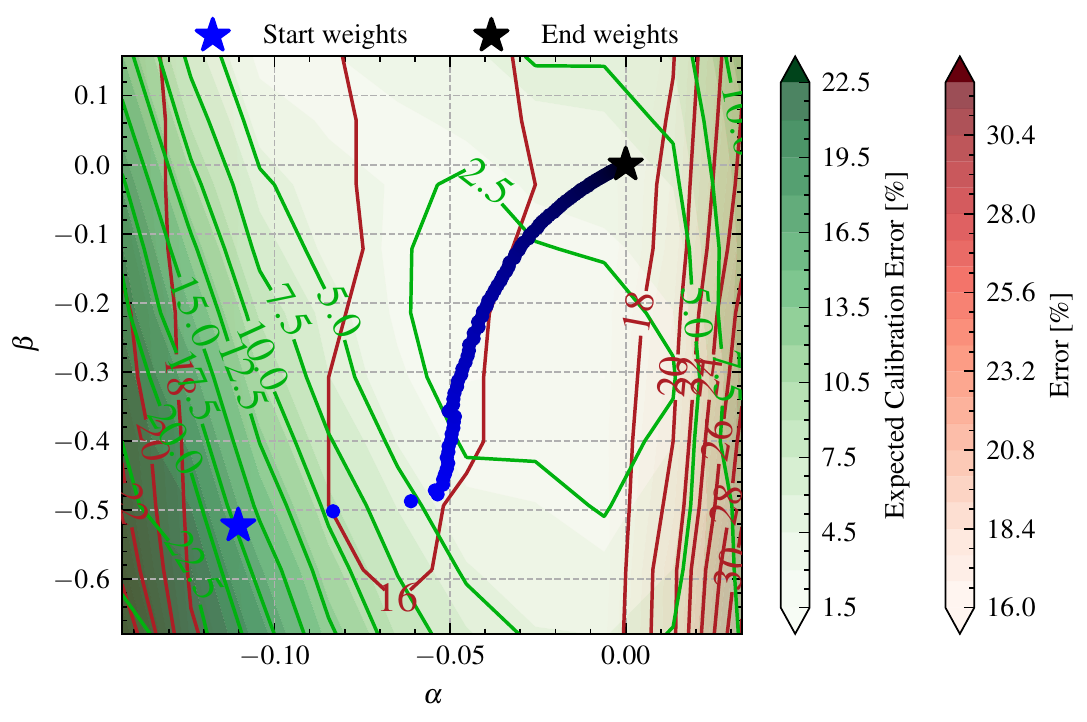}
	\caption{Error and ECE on ID.}
	\label{fig:loss_landscape:classification_adult-fc-vanilla-test_2d_error_ece}
\end{subfigure}
\begin{subfigure}{0.25\textwidth}
	\centering
	\includegraphics[width=\textwidth]{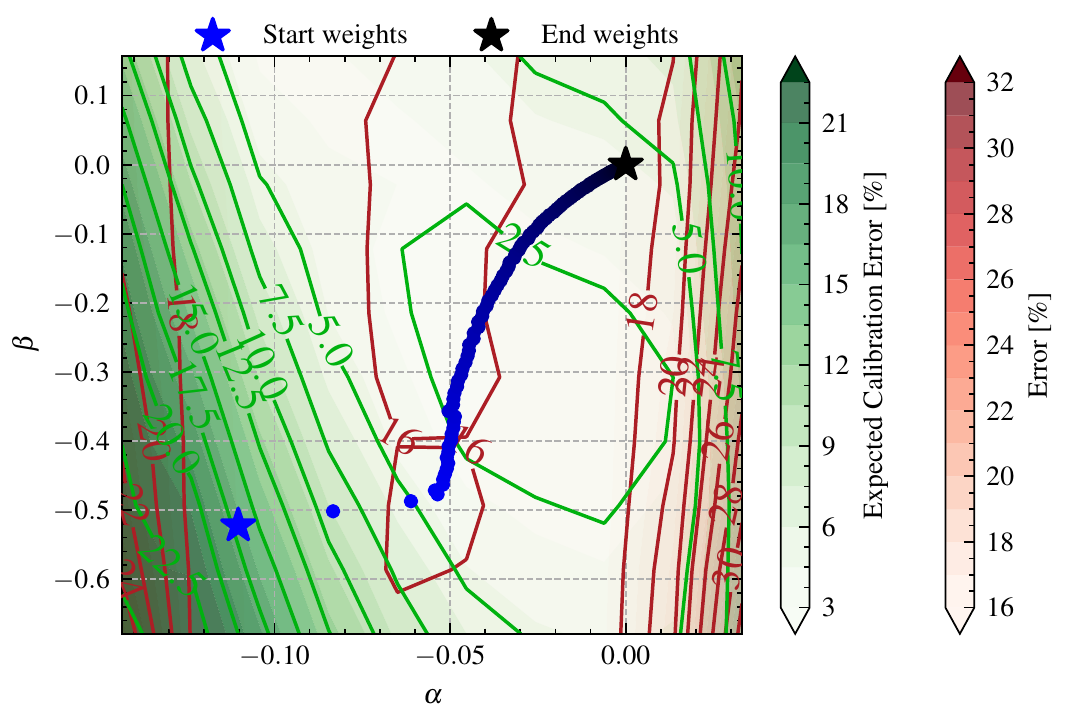}
	\caption{Error and ECE on OOD.}
	\label{fig:loss_landscape:classification_adult-fc-vanilla-test_2d_aug_error_ece}
\end{subfigure}
\caption{No noise on Adult.}
\label{fig:loss_landscape:classification_adult-fc-vanilla}
\end{figure}
\begin{figure}
\centering
\begin{subfigure}{0.21\textwidth}
	\centering
	\includegraphics[width=\textwidth]{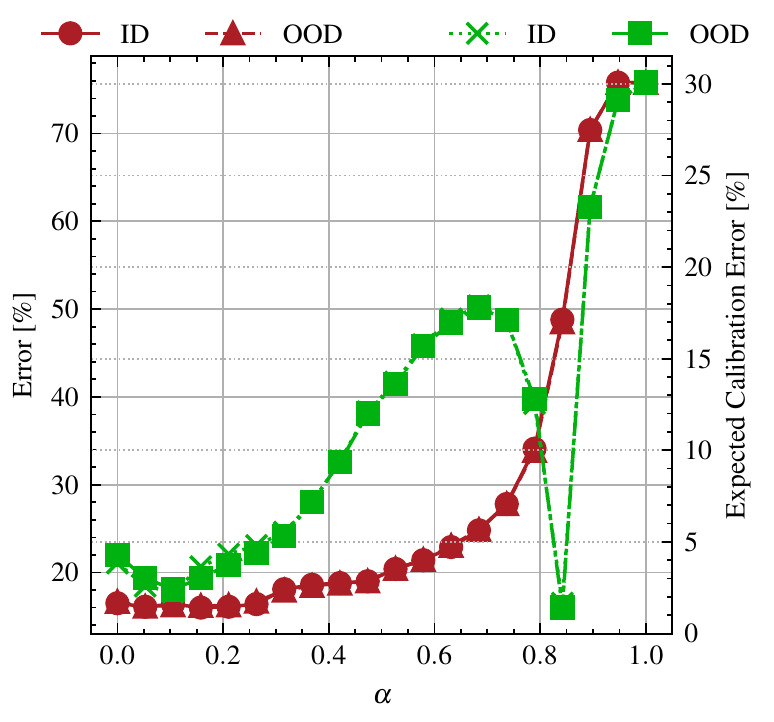}
	\caption{Error and ECE.}
	\label{fig:loss_landscape:classification_adult-fc-input_additive_gaussian-lin_error_ece}
\end{subfigure}
\begin{subfigure}{0.21\textwidth}
	\centering
	\includegraphics[width=\textwidth]{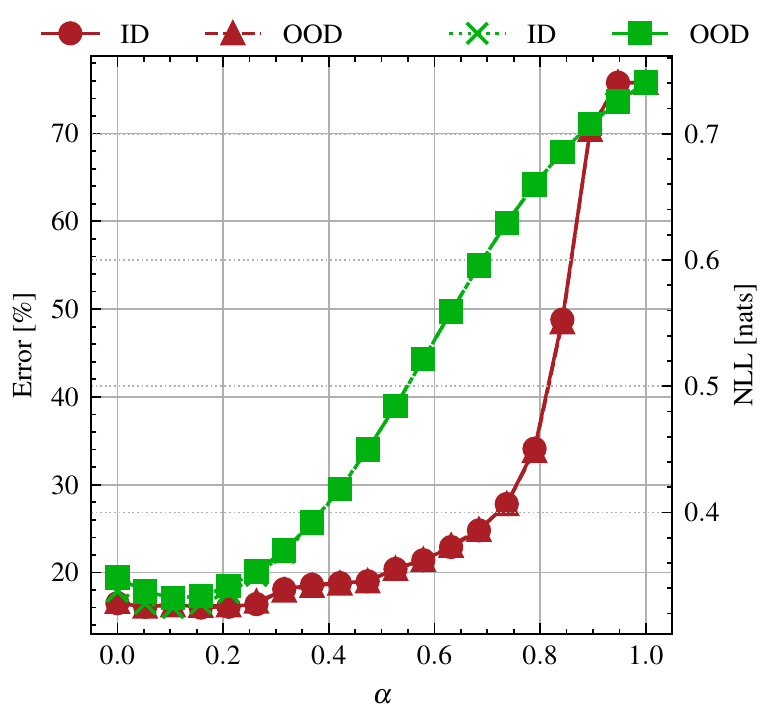}
	\caption{Error and NLL.}
	\label{fig:loss_landscape:classification_adult-fc-input_additive_gaussian-lin_error_nll}
\end{subfigure}
\begin{subfigure}{0.25\textwidth}
	\centering
	\includegraphics[width=\textwidth]{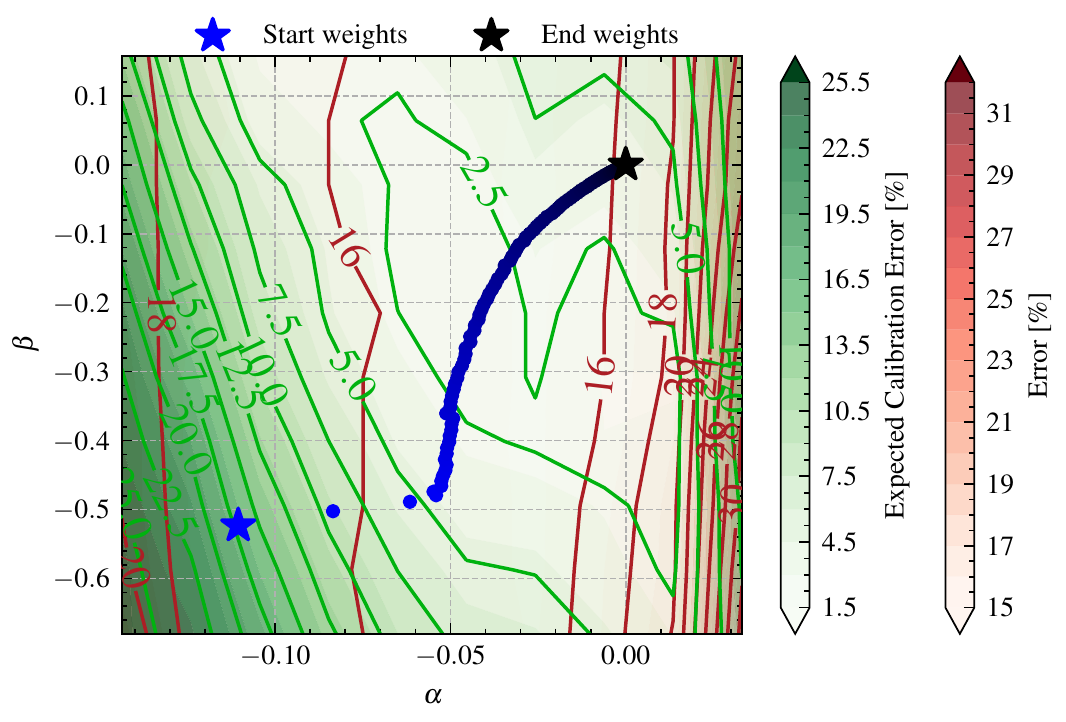}
	\caption{Error and ECE on ID.}
	\label{fig:loss_landscape:classification_adult-fc-input_additive_gaussian-test_2d_error_ece}
\end{subfigure}
\begin{subfigure}{0.25\textwidth}
	\centering
	\includegraphics[width=\textwidth]{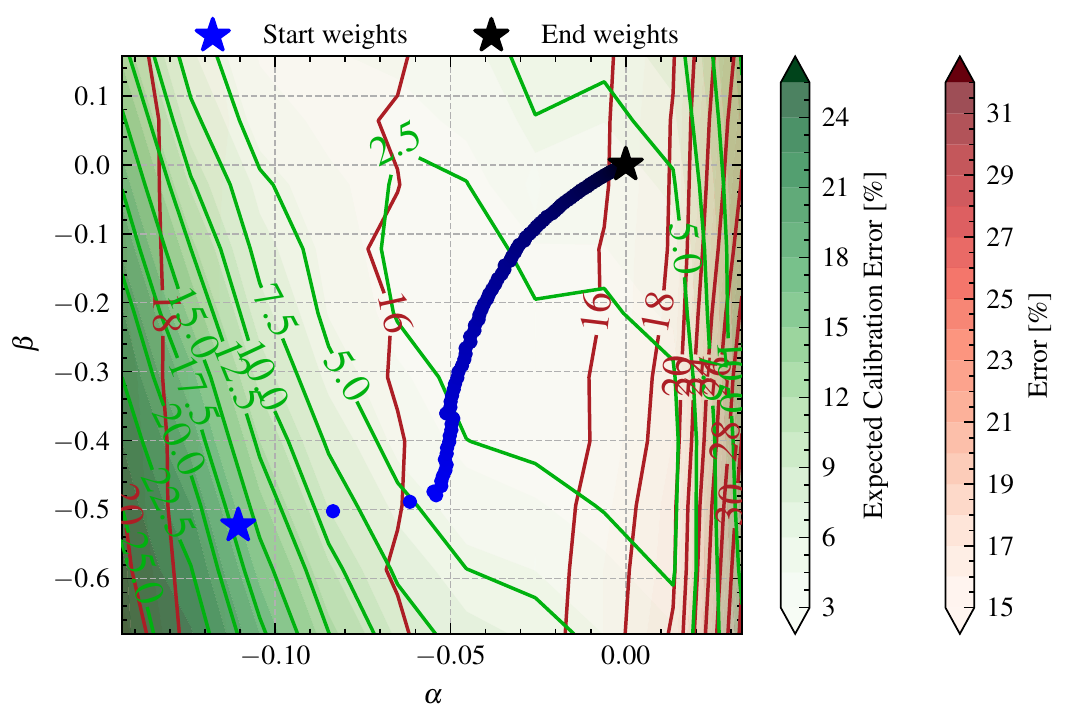}
	\caption{Error and ECE on OOD.}
	\label{fig:loss_landscape:classification_adult-fc-input_additive_gaussian-test_2d_aug_error_ece}
\end{subfigure}
\caption{Input Additive Gaussian on Adult.
\textit{Observations}: Did not change the smoothness of the 1D curves or the 2D metric landscape trajectory compared to no noise.}
\label{fig:loss_landscape:classification_adult-fc-input_additive_gaussian}
\end{figure}
\begin{figure}
\centering
\begin{subfigure}{0.21\textwidth}
	\centering
	\includegraphics[width=\textwidth]{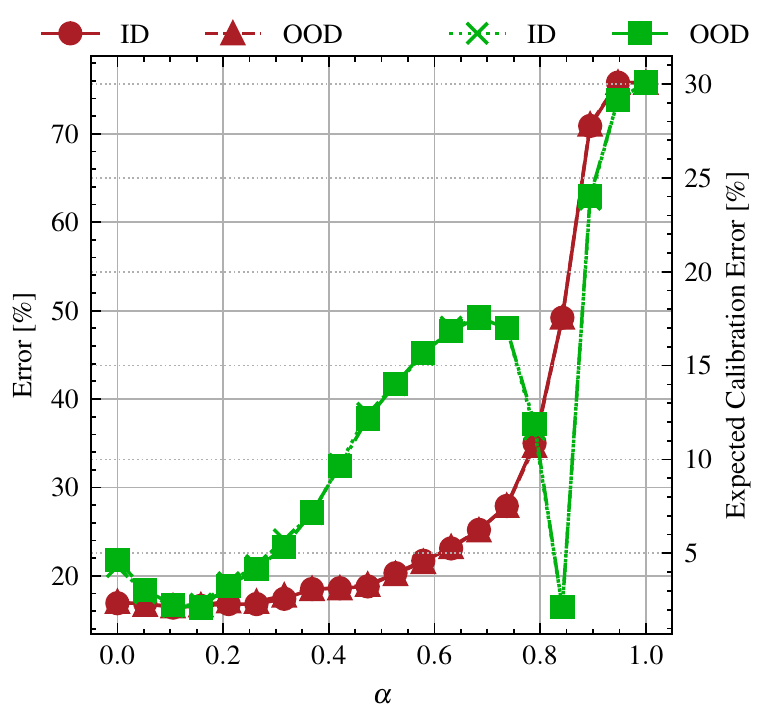}
	\caption{Error and ECE.}
	\label{fig:loss_landscape:classification_adult-fc-input_ods-lin_error_ece}
\end{subfigure}
\begin{subfigure}{0.21\textwidth}
	\centering
	\includegraphics[width=\textwidth]{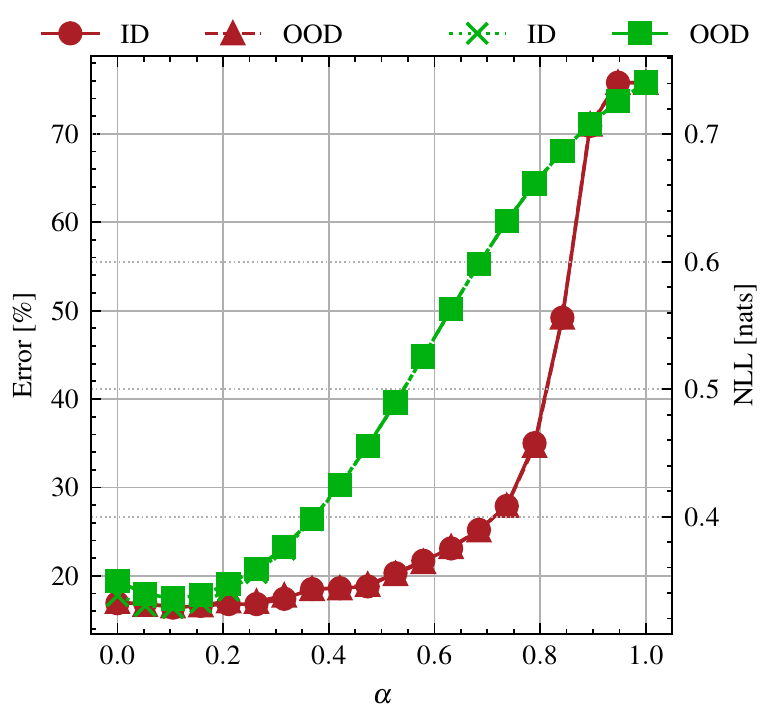}
	\caption{Error and NLL.}
	\label{fig:loss_landscape:classification_adult-fc-input_ods-lin_error_nll}
\end{subfigure}
\begin{subfigure}{0.25\textwidth}
	\centering
	\includegraphics[width=\textwidth]{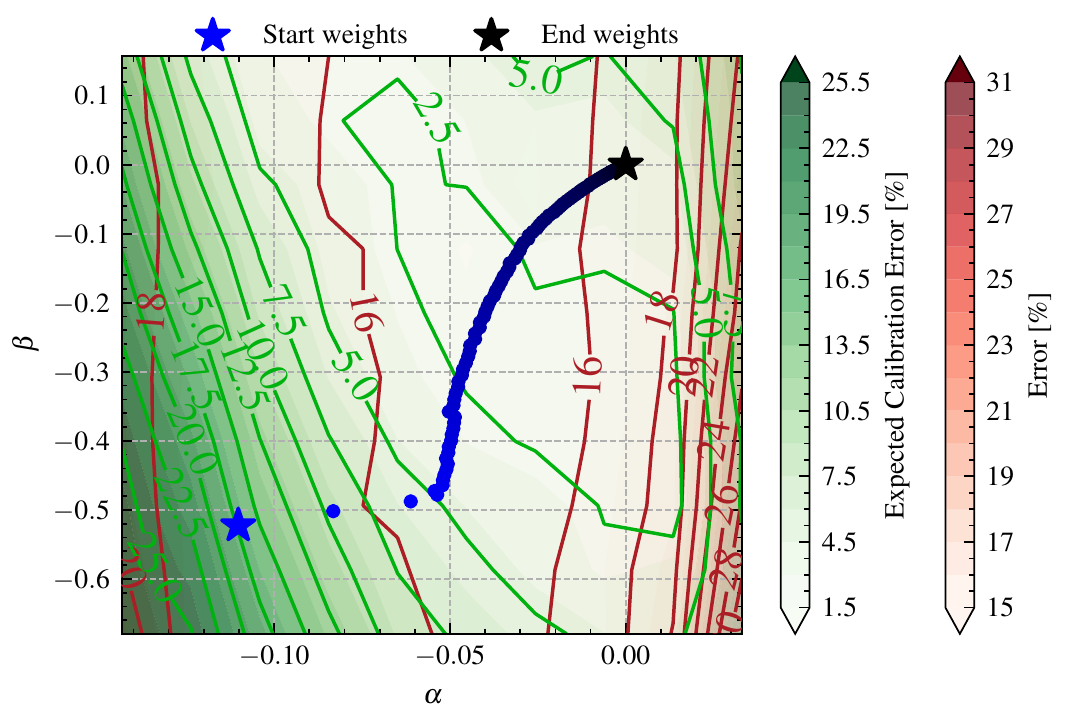}
	\caption{Error and ECE on ID.}
	\label{fig:loss_landscape:classification_adult-fc-input_ods-test_2d_error_ece}
\end{subfigure}
\begin{subfigure}{0.25\textwidth}
	\centering
	\includegraphics[width=\textwidth]{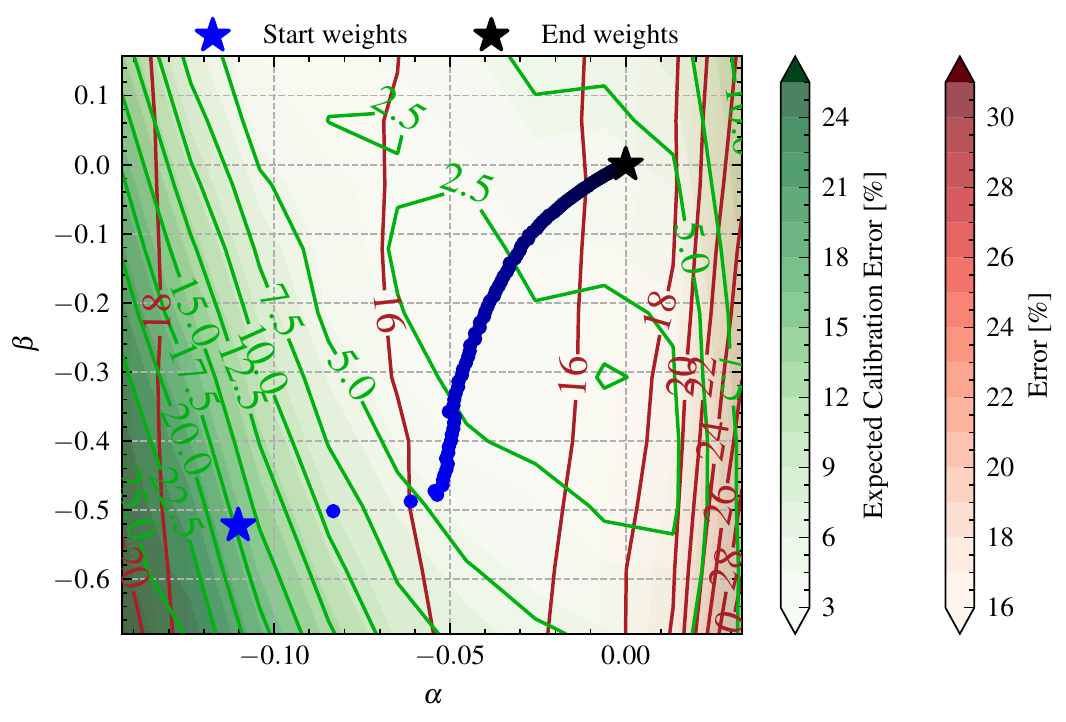}
	\caption{Error and ECE on OOD.}
	\label{fig:loss_landscape:classification_adult-fc-input_ods-test_2d_aug_error_ece}
\end{subfigure}
\caption{Input ODS on Adult.
\textit{Observations}: Did not change the smoothness of the 1D curves or the 2D metric landscape trajectory compared to no noise.}
\label{fig:loss_landscape:classification_adult-fc-input_ods}
\end{figure}
\begin{figure}
\centering
\begin{subfigure}{0.21\textwidth}
	\centering
	\includegraphics[width=\textwidth]{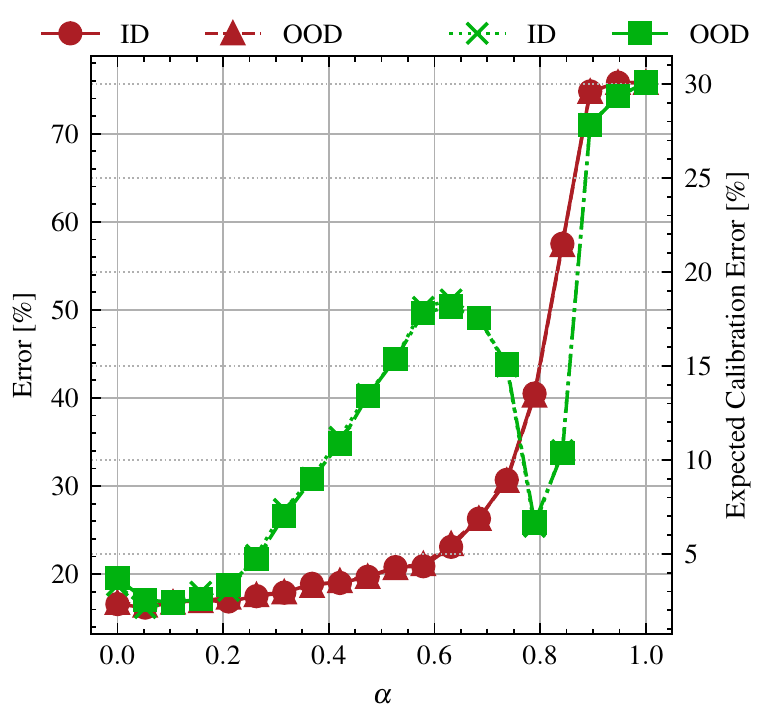}
	\caption{Error and ECE.}
	\label{fig:loss_landscape:classification_adult-fc-input_target_mixup-lin_error_ece}
\end{subfigure}
\begin{subfigure}{0.21\textwidth}
	\centering
	\includegraphics[width=\textwidth]{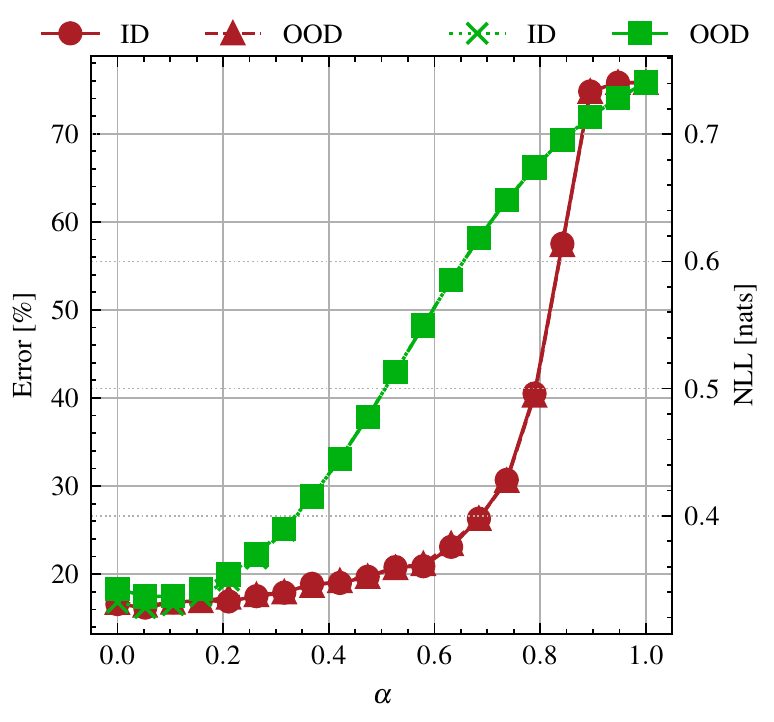}
	\caption{Error and NLL.}
	\label{fig:loss_landscape:classification_adult-fc-input_target_mixup-lin_error_nll}
\end{subfigure}
\begin{subfigure}{0.25\textwidth}
	\centering
	\includegraphics[width=\textwidth]{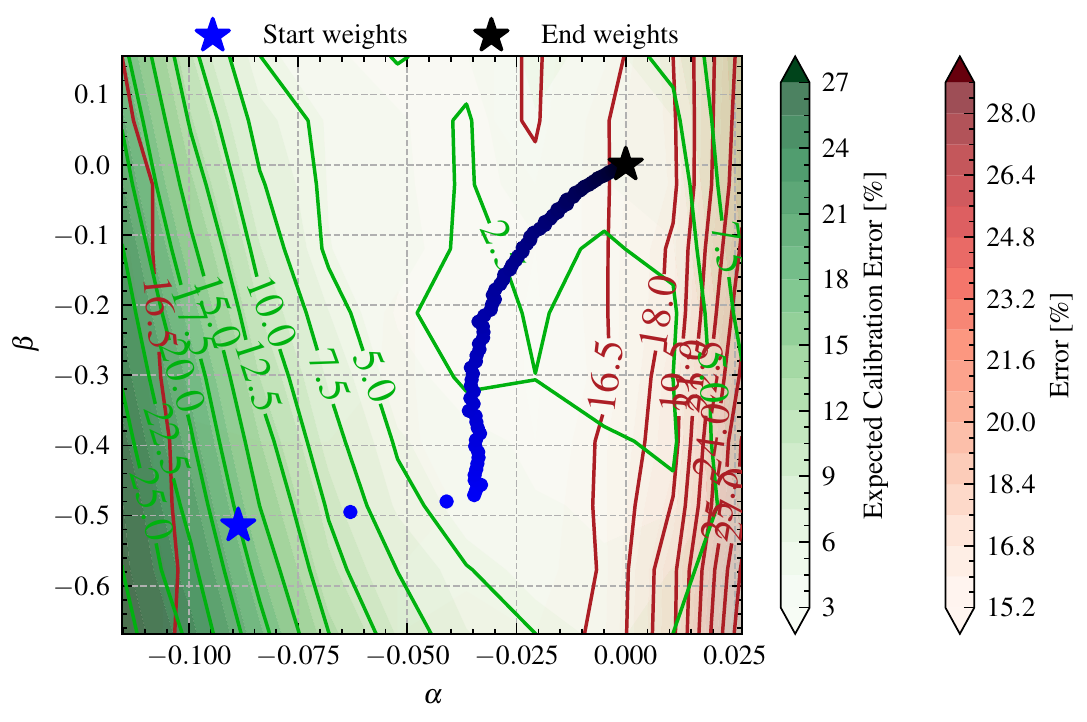}
	\caption{Error and ECE on ID.}
	\label{fig:loss_landscape:classification_adult-fc-input_target_mixup-test_2d_error_ece}
\end{subfigure}
\begin{subfigure}{0.25\textwidth}
	\centering
	\includegraphics[width=\textwidth]{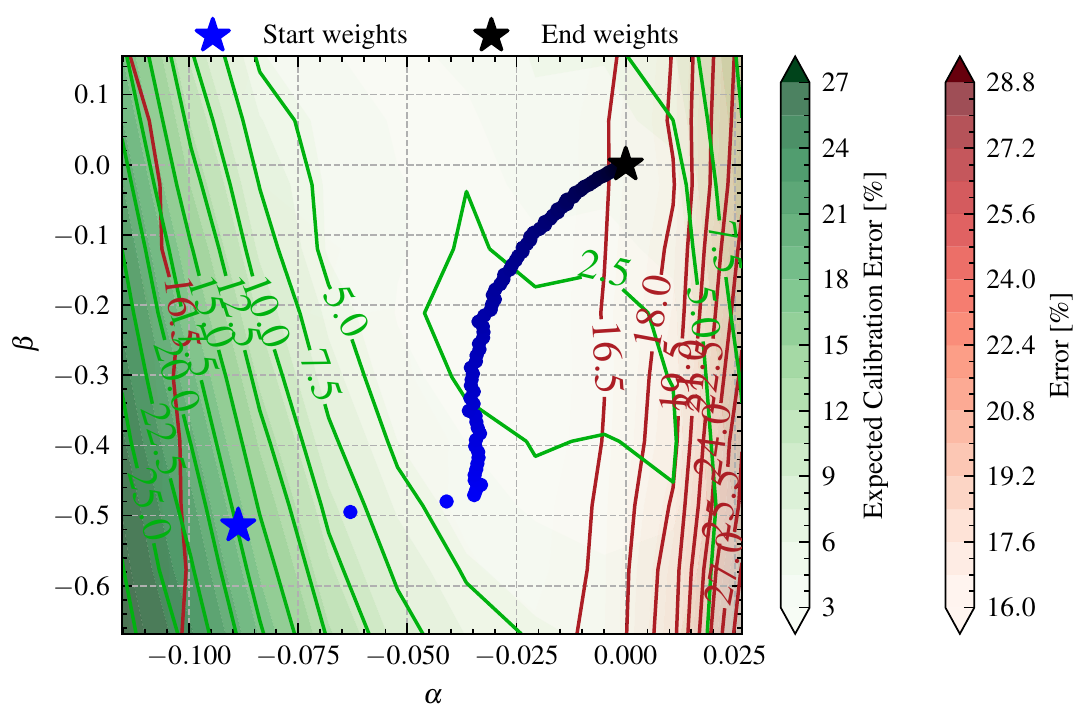}
	\caption{Error and ECE on OOD.}
	\label{fig:loss_landscape:classification_adult-fc-input_target_mixup-test_2d_aug_error_ece}
\end{subfigure}
\caption{Input-Target MixUp on Adult.
\textit{Observations}: Did not change the smoothness of the 1D curves or the 2D metric landscape trajectory compared to no noise.}
\label{fig:loss_landscape:classification_adult-fc-input_target_mixup}
\end{figure}
\begin{figure}
\centering
\begin{subfigure}{0.21\textwidth}
	\centering
	\includegraphics[width=\textwidth]{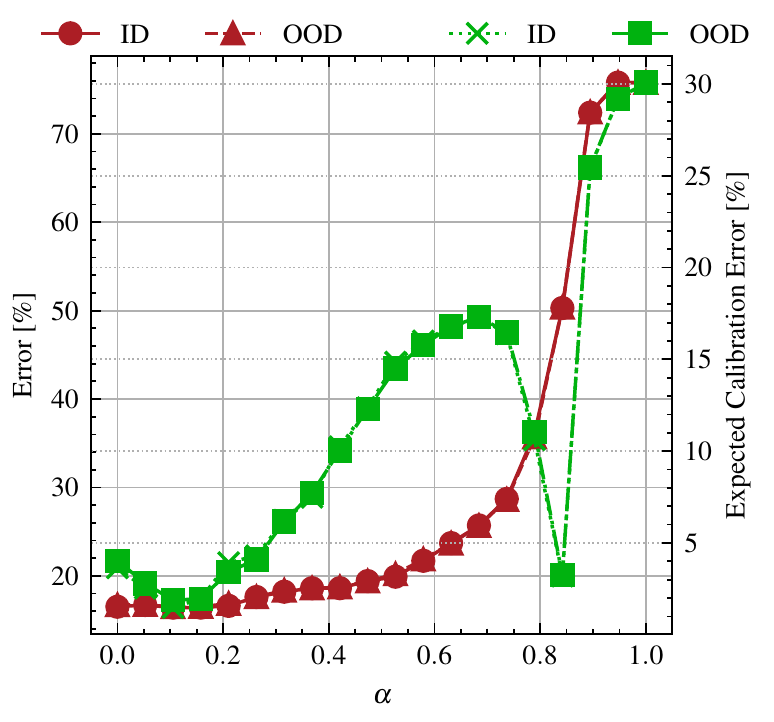}
	\caption{Error and ECE.}
	\label{fig:loss_landscape:classification_adult-fc-target_smoothing-lin_error_ece}
\end{subfigure}
\begin{subfigure}{0.21\textwidth}
	\centering
	\includegraphics[width=\textwidth]{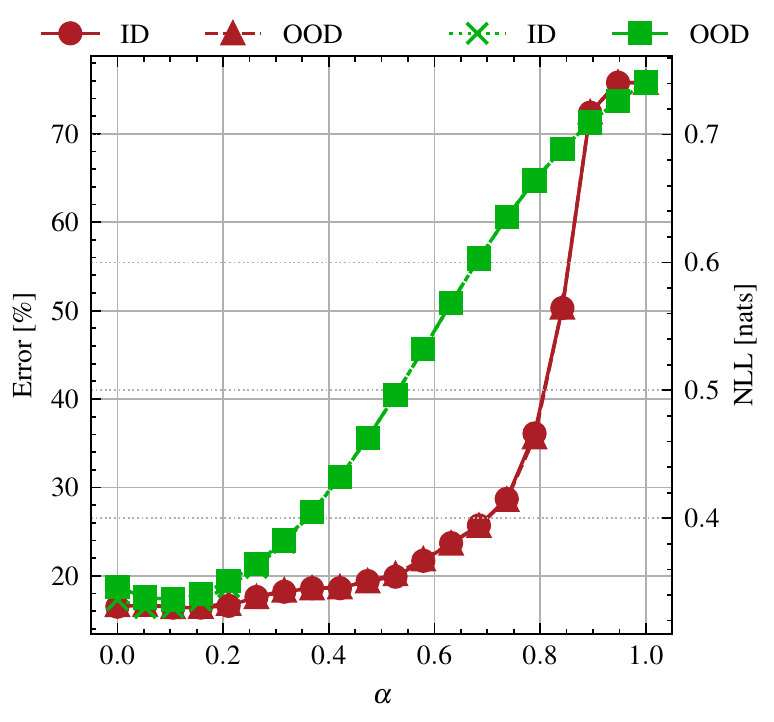}
	\caption{Error and NLL.}
	\label{fig:loss_landscape:classification_adult-fc-target_smoothing-lin_error_nll}
\end{subfigure}
\begin{subfigure}{0.25\textwidth}
	\centering
	\includegraphics[width=\textwidth]{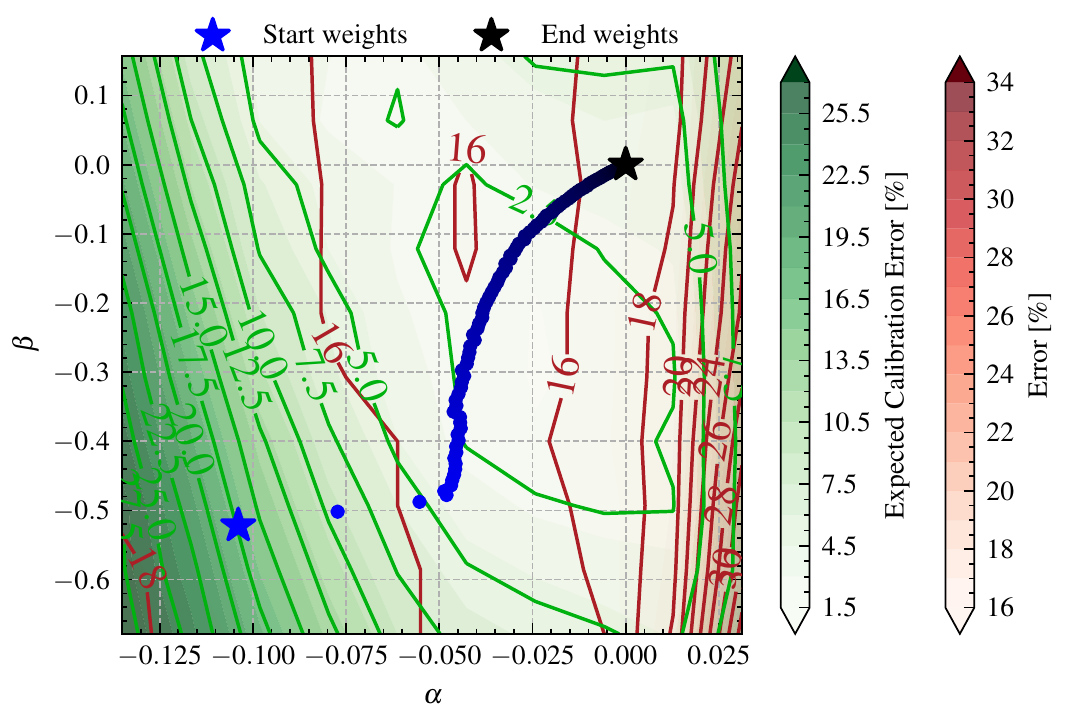}
	\caption{Error and ECE on ID.}
	\label{fig:loss_landscape:classification_adult-fc-target_smoothing-test_2d_error_ece}
\end{subfigure}
\begin{subfigure}{0.25\textwidth}
	\centering
	\includegraphics[width=\textwidth]{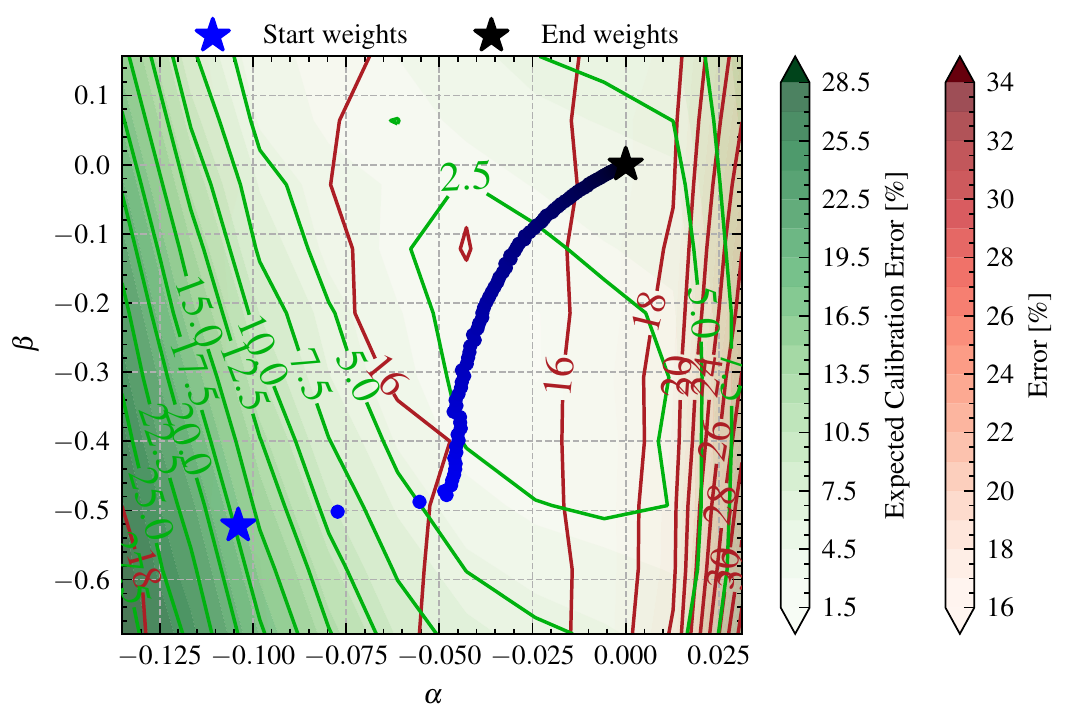}
	\caption{Error and ECE on OOD.}
	\label{fig:loss_landscape:classification_adult-fc-target_smoothing-test_2d_aug_error_ece}
\end{subfigure}
\caption{Target Smoothing on Adult.
\textit{Observations}: Did not change the smoothness of the 1D curves or the 2D metric landscape trajectory compared to no noise.}
\label{fig:loss_landscape:classification_adult-fc-target_smoothing}
\end{figure}
\begin{figure}
\centering
\begin{subfigure}{0.21\textwidth}
	\centering
	\includegraphics[width=\textwidth]{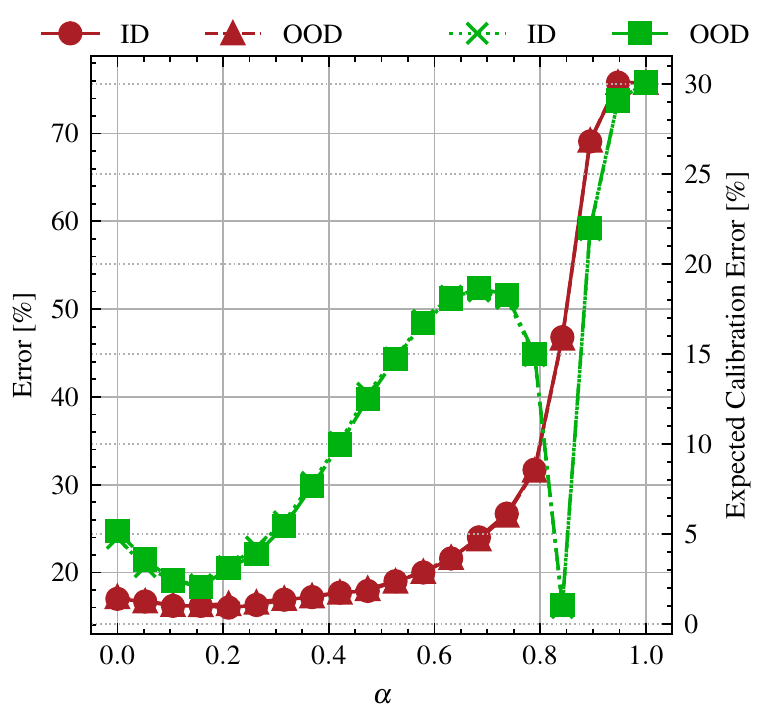}
	\caption{Error and ECE.}
	\label{fig:loss_landscape:classification_adult-fc-activation_additive_gaussian-lin_error_ece}
\end{subfigure}
\begin{subfigure}{0.21\textwidth}
	\centering
	\includegraphics[width=\textwidth]{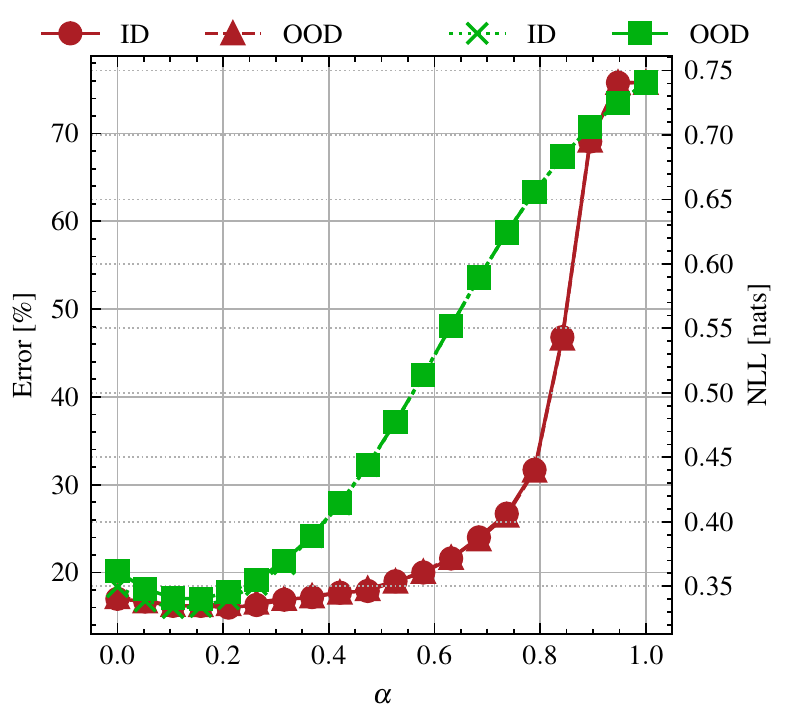}
	\caption{Error and NLL.}
	\label{fig:loss_landscape:classification_adult-fc-activation_additive_gaussian-lin_error_nll}
\end{subfigure}
\begin{subfigure}{0.25\textwidth}
	\centering
	\includegraphics[width=\textwidth]{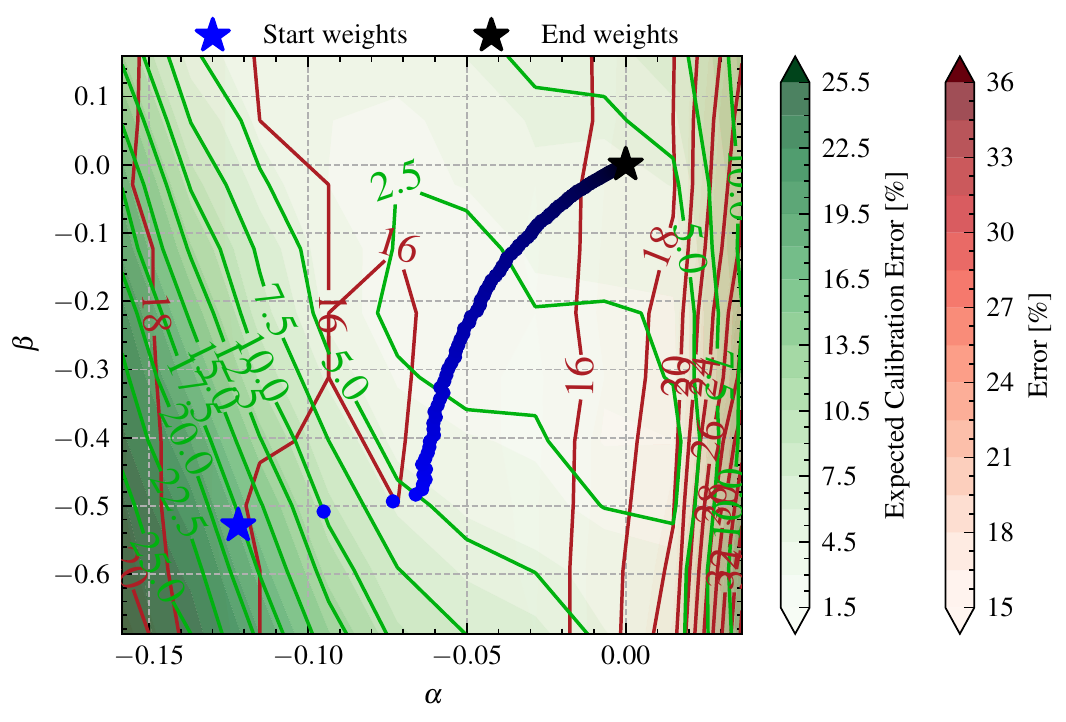}
	\caption{Error and ECE on ID.}
	\label{fig:loss_landscape:classification_adult-fc-activation_additive_gaussian-test_2d_error_ece}
\end{subfigure}
\begin{subfigure}{0.25\textwidth}
	\centering
	\includegraphics[width=\textwidth]{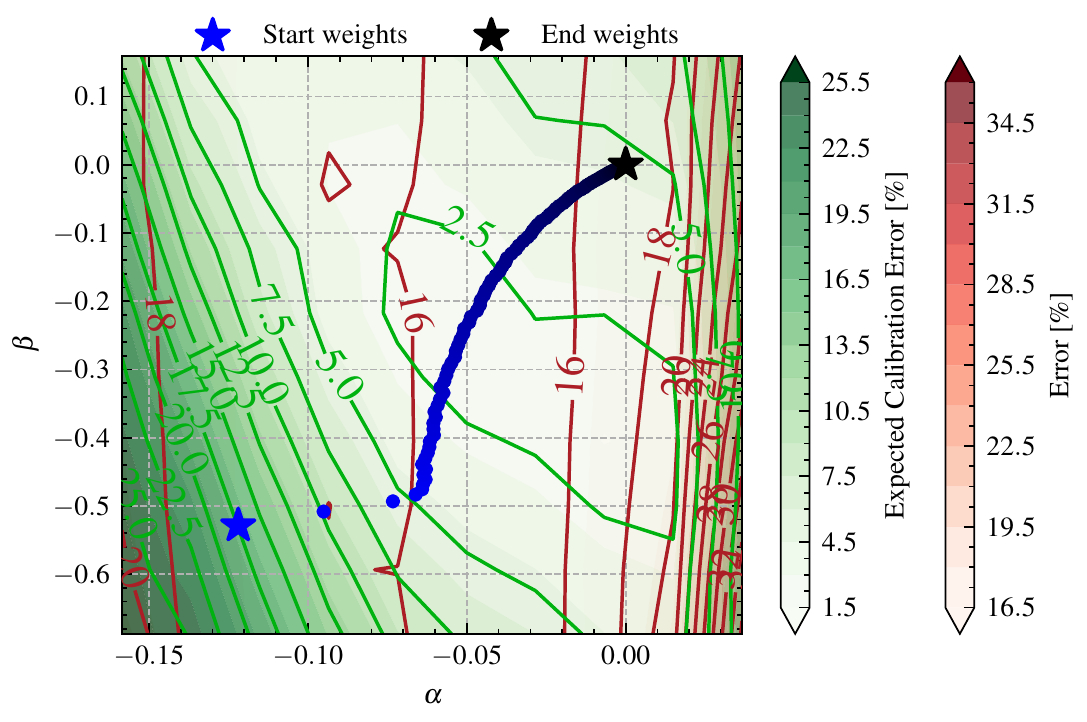}
	\caption{Error and ECE on OOD.}
	\label{fig:loss_landscape:classification_adult-fc-activation_additive_gaussian-test_2d_aug_error_ece}
\end{subfigure}
\caption{Activation Additive Gaussian on Adult.
\textit{Observations}: Did not change the smoothness of the 1D curves or the 2D metric landscape trajectory compared to no noise.}
\label{fig:loss_landscape:classification_adult-fc-activation_additive_gaussian}
\end{figure}
\begin{figure}
\centering
\begin{subfigure}{0.21\textwidth}
	\centering
	\includegraphics[width=\textwidth]{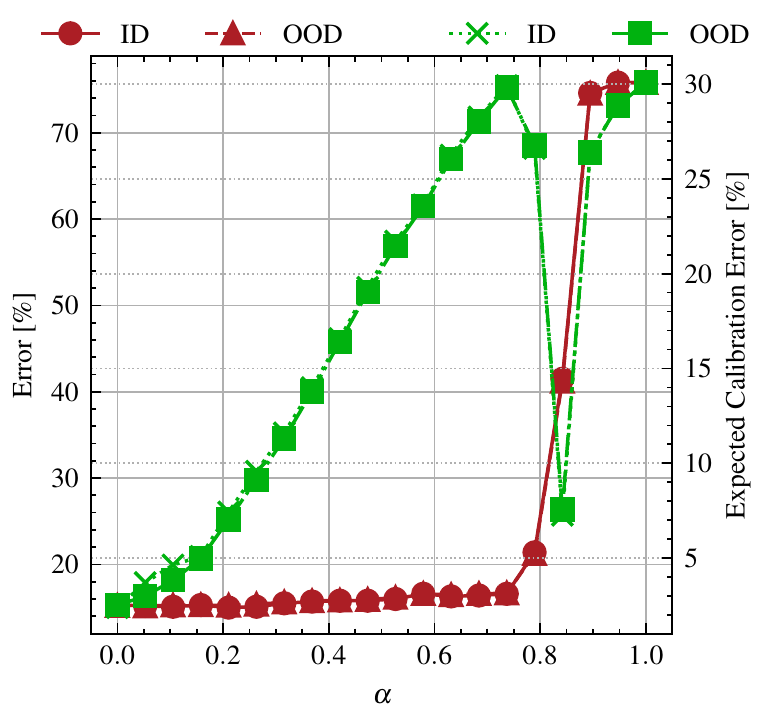}
	\caption{Error and ECE.}
	\label{fig:loss_landscape:classification_adult-fc-activation_dropout-lin_error_ece}
\end{subfigure}
\begin{subfigure}{0.21\textwidth}
	\centering
	\includegraphics[width=\textwidth]{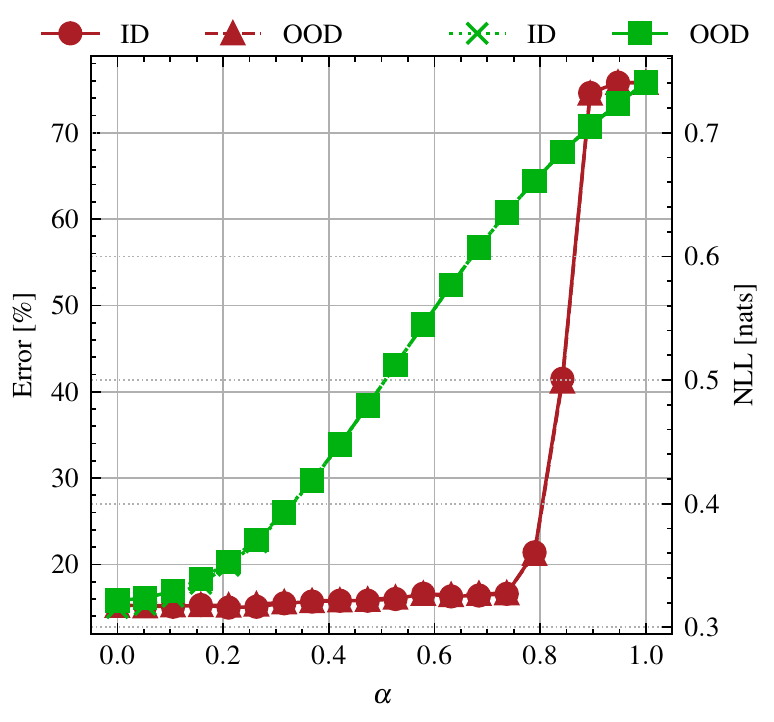}
	\caption{Error and NLL.}
	\label{fig:loss_landscape:classification_adult-fc-activation_dropout-lin_error_nll}
\end{subfigure}
\begin{subfigure}{0.25\textwidth}
	\centering
	\includegraphics[width=\textwidth]{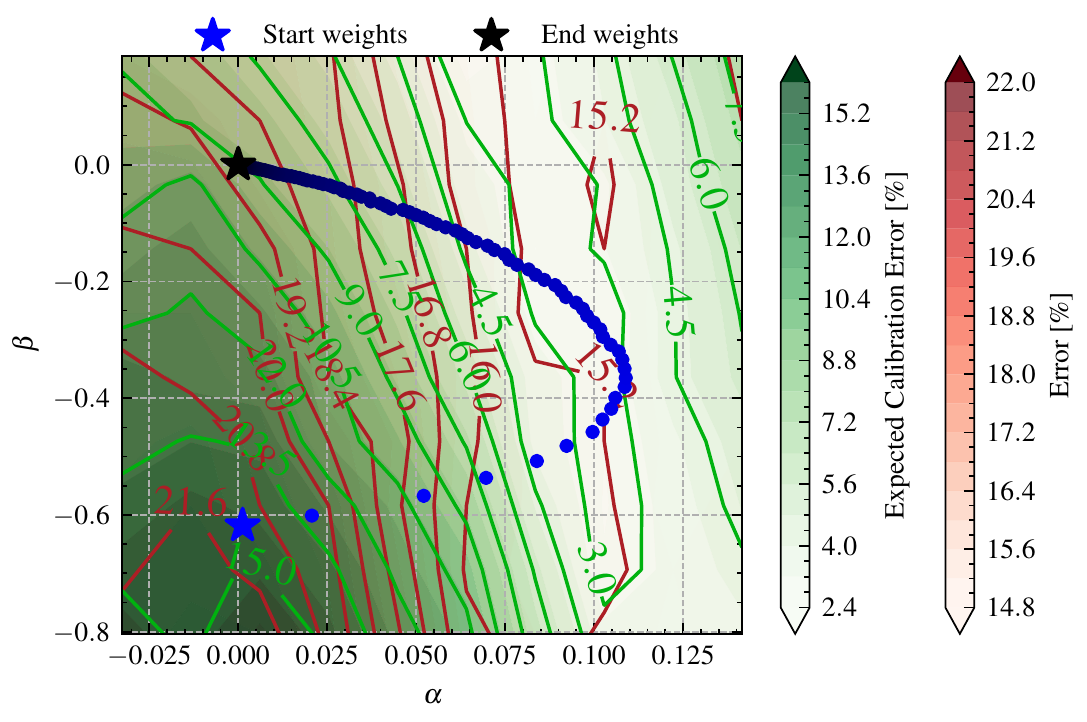}
	\caption{Error and ECE on ID.}
	\label{fig:loss_landscape:classification_adult-fc-activation_dropout-test_2d_error_ece}
\end{subfigure}
\begin{subfigure}{0.25\textwidth}
	\centering
	\includegraphics[width=\textwidth]{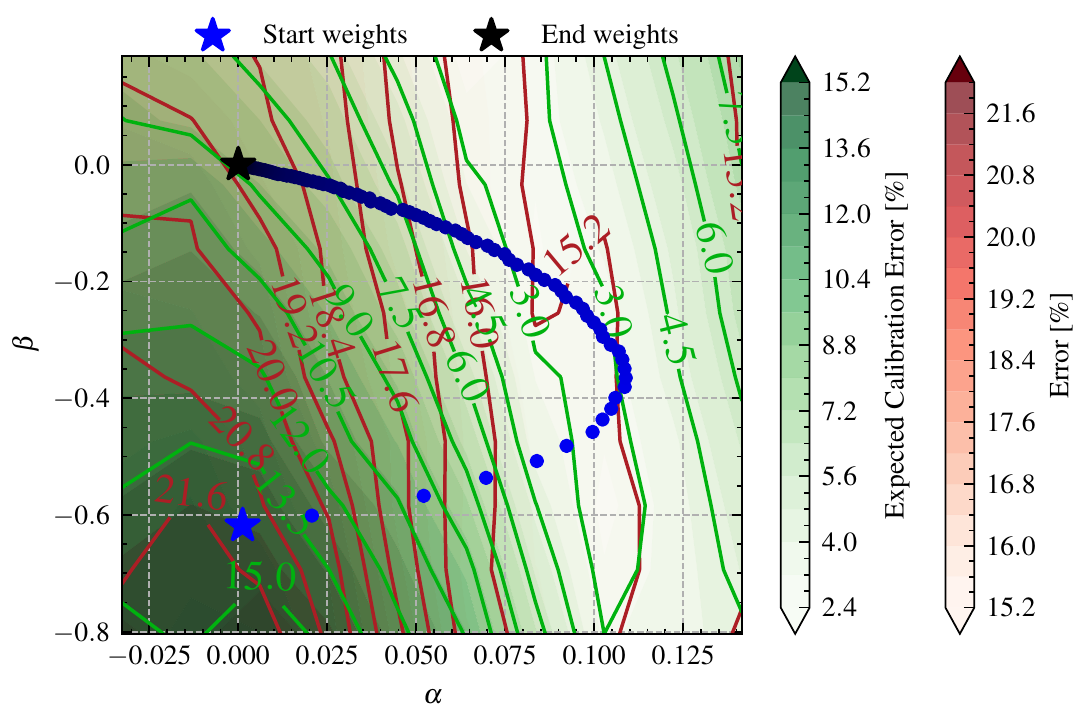}
	\caption{Error and ECE on OOD.}
	\label{fig:loss_landscape:classification_adult-fc-activation_dropout-test_2d_aug_error_ece}
\end{subfigure}
\caption{Activation Dropout on Adult.
\textit{Observations}: Changed the ECE curvature and made the NLL plots smoother in the 1D case.
In the 2D plots, the ECE and error appear aligned during optimisation.
The curvature of the 2D plots has changed and there is a higher alignment between the ECE and error.}
\label{fig:loss_landscape:classification_adult-fc-activation_dropout}
\end{figure}

\begin{figure}
\centering
\begin{subfigure}{0.21\textwidth}
	\centering
	\includegraphics[width=\textwidth]{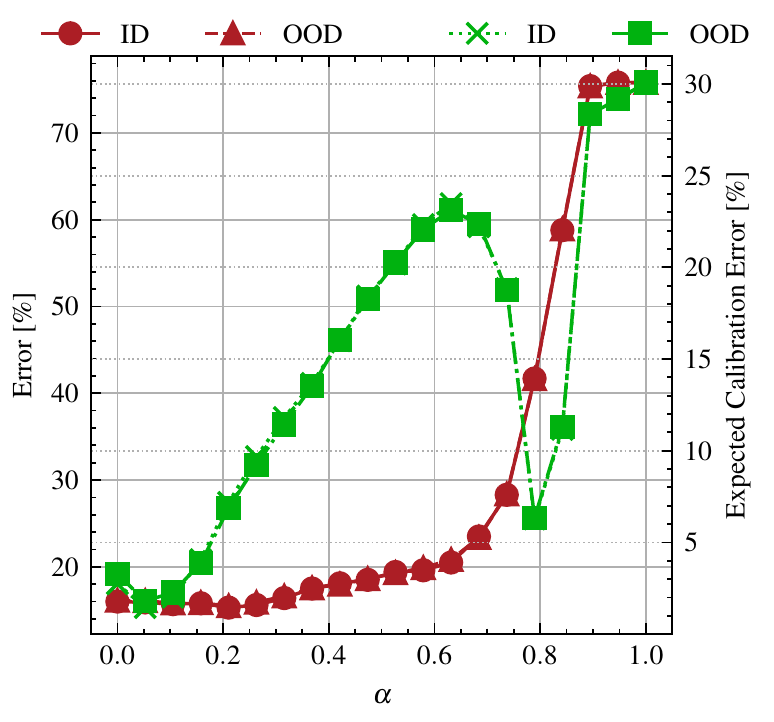}
	\caption{Error and ECE.}
	\label{fig:loss_landscape:classification_adult-fc-gradient_gaussian-lin_error_ece}
\end{subfigure}
\begin{subfigure}{0.21\textwidth}
	\centering
	\includegraphics[width=\textwidth]{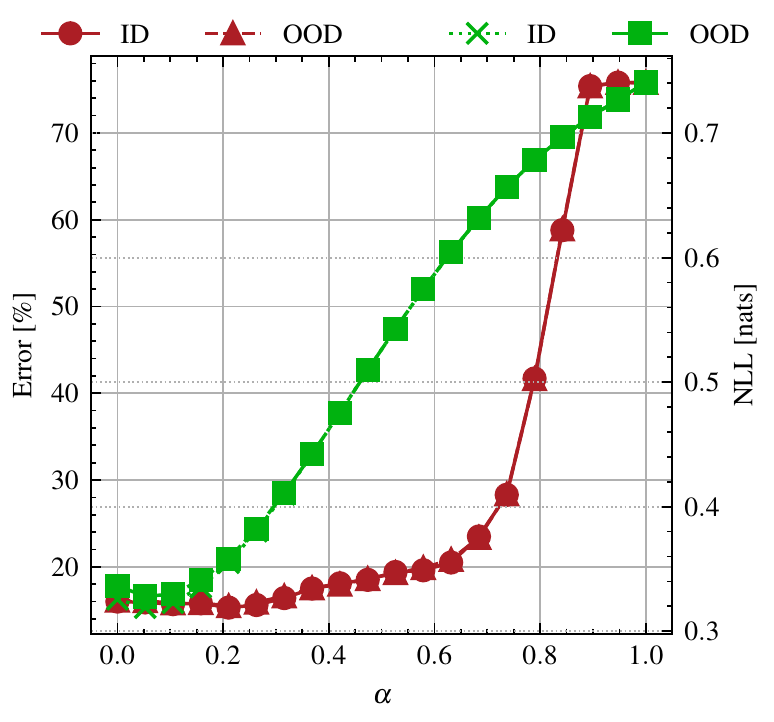}
	\caption{Error and NLL.}
	\label{fig:loss_landscape:classification_adult-fc-gradient_gaussian-lin_error_nll}
\end{subfigure}
\begin{subfigure}{0.25\textwidth}
	\centering
	\includegraphics[width=\textwidth]{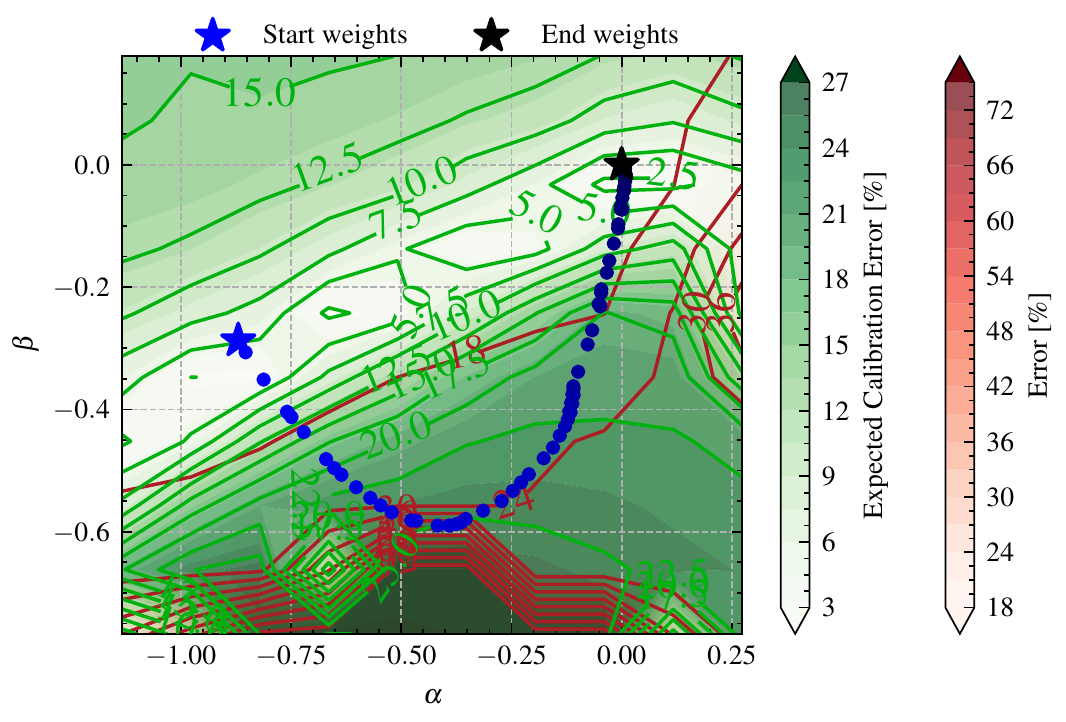}
	\caption{Error and ECE on ID.}
	\label{fig:loss_landscape:classification_adult-fc-gradient_gaussian-test_2d_error_ece}
\end{subfigure}
\begin{subfigure}{0.25\textwidth}
	\centering
	\includegraphics[width=\textwidth]{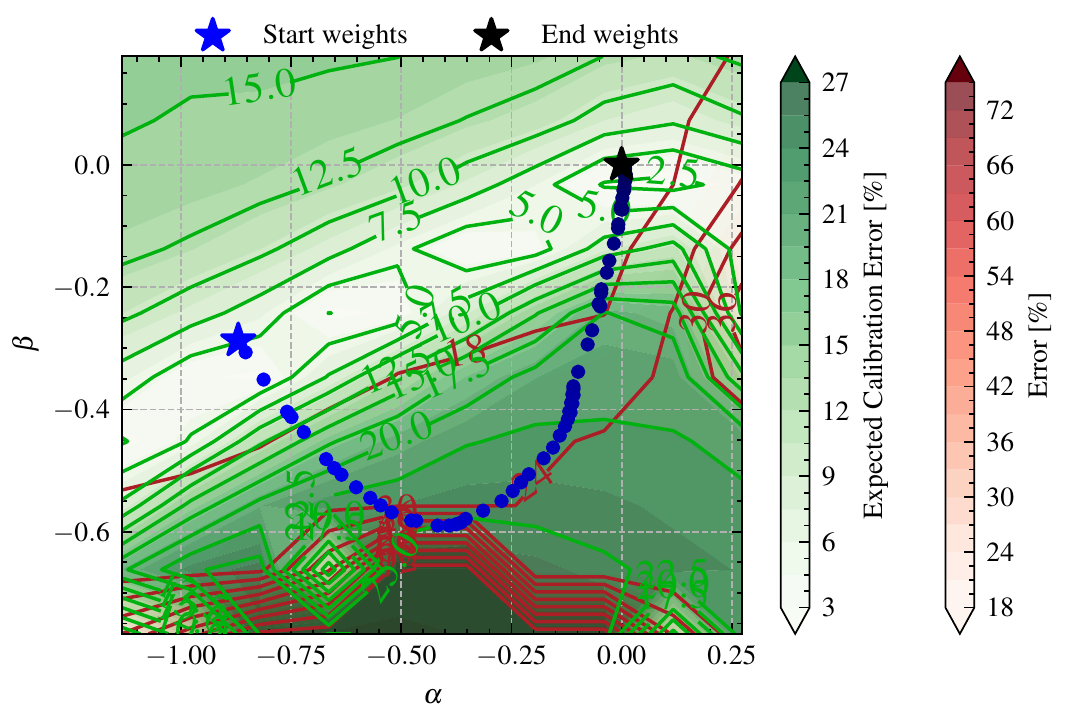}
	\caption{Error and ECE on OOD.}
	\label{fig:loss_landscape:classification_adult-fc-gradient_gaussian-test_2d_aug_error_ece}
\end{subfigure}
\caption{Gradient Gaussian on Adult.
\textit{Observations}: Did not change the smoothness of the 1D curves, but the 2D trajectory appears more exploratory compared to no noise.}
\label{fig:loss_landscape:classification_adult-fc-gradient_gaussian}
\end{figure}

\begin{figure}
\centering
\begin{subfigure}{0.21\textwidth}
	\centering
	\includegraphics[width=\textwidth]{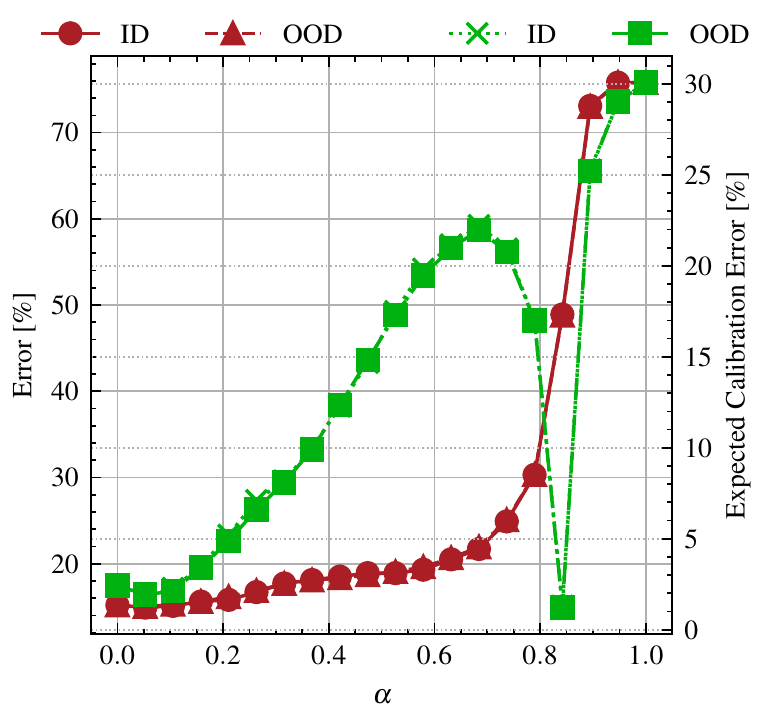}
	\caption{Error and ECE.}
	\label{fig:loss_landscape:classification_adult-fc-model_sp-lin_error_ece}
\end{subfigure}
\begin{subfigure}{0.21\textwidth}
	\centering
	\includegraphics[width=\textwidth]{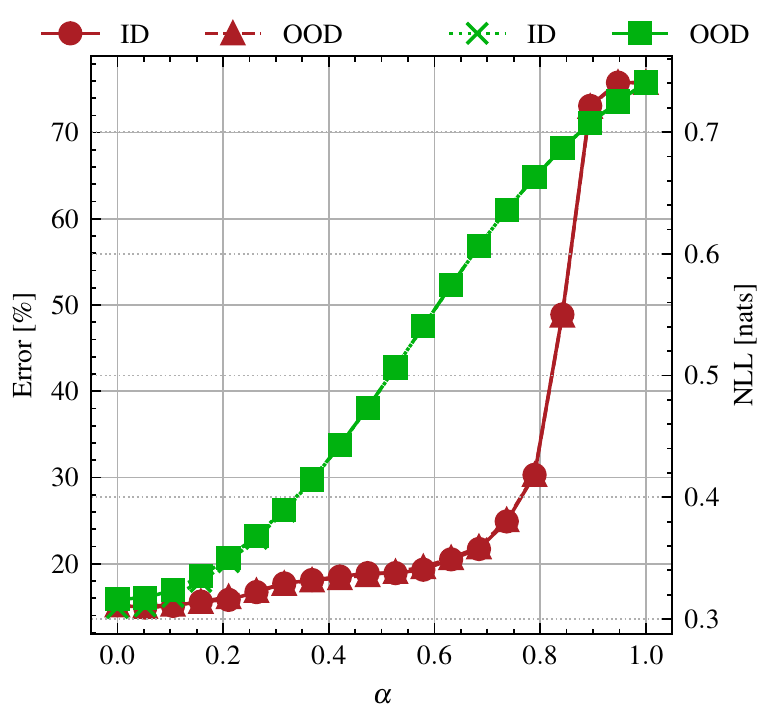}
	\caption{Error and NLL.}
	\label{fig:loss_landscape:classification_adult-fc-model_sp-lin_error_nll}
\end{subfigure}
\begin{subfigure}{0.25\textwidth}
	\centering
	\includegraphics[width=\textwidth]{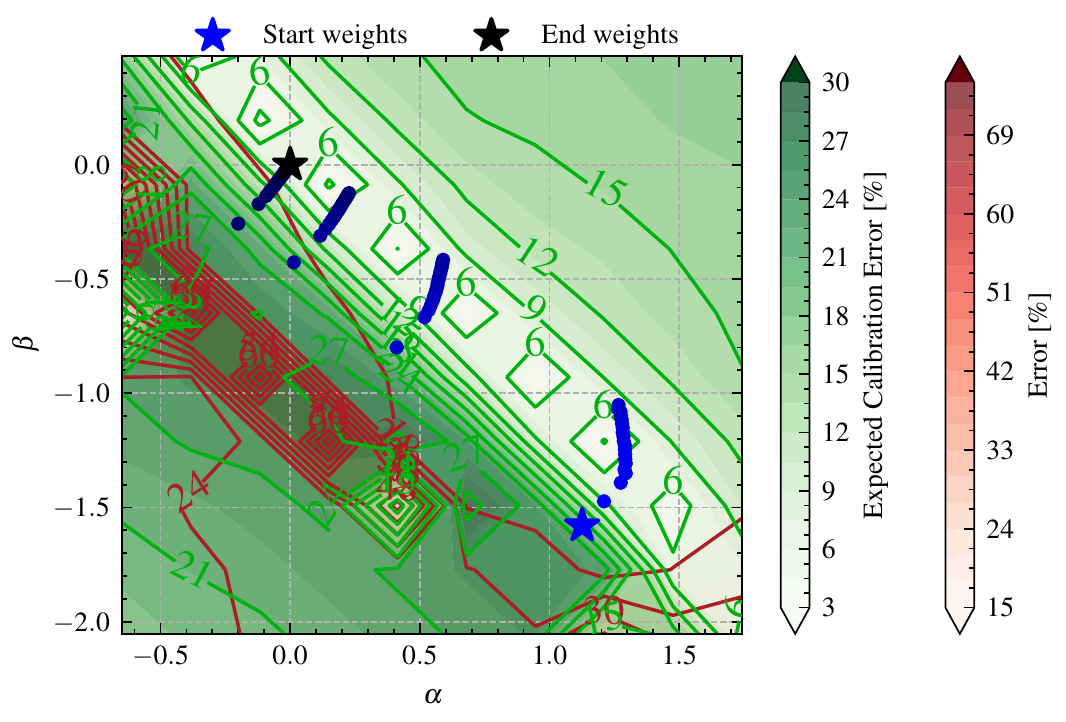}
	\caption{Error and ECE on ID.}
	\label{fig:loss_landscape:classification_adult-fc-model_sp-test_2d_error_ece}
\end{subfigure}
\begin{subfigure}{0.25\textwidth}
	\centering
	\includegraphics[width=\textwidth]{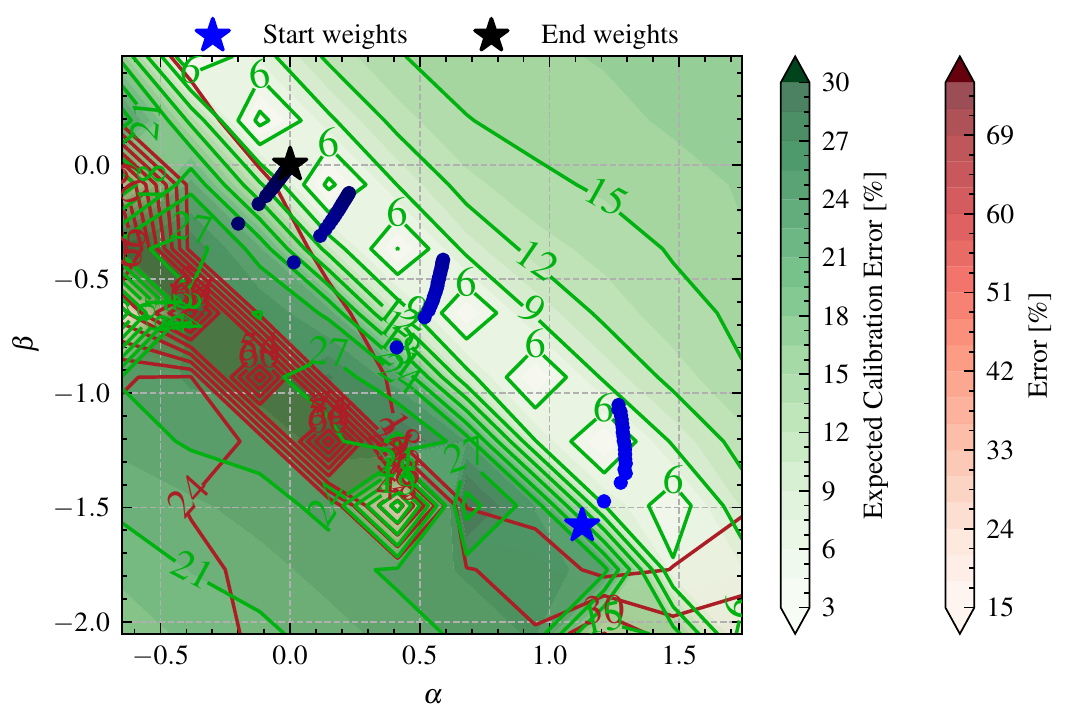}
	\caption{Error and ECE on OOD.}
	\label{fig:loss_landscape:classification_adult-fc-model_sp-test_2d_aug_error_ece}
\end{subfigure}
\caption{Model Shrink and Perturb on Adult.
\textit{Observations}: Did not change the smoothness of the 1D curves, but the 2D trajectory appears more exploratory compared to no noise.}
\label{fig:loss_landscape:classification_adult-fc-model_sp}
\end{figure}
\begin{figure}
\centering
\begin{subfigure}{0.21\textwidth}
	\centering
	\includegraphics[width=\textwidth]{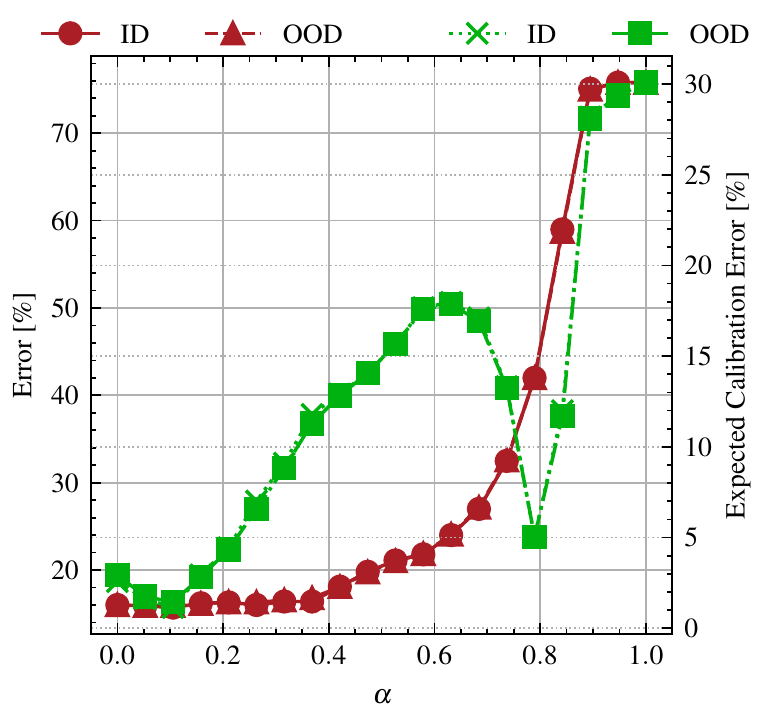}
	\caption{Error and ECE.}
	\label{fig:loss_landscape:classification_adult-fc-weight_additive_gaussian-lin_error_ece}
\end{subfigure}
\begin{subfigure}{0.21\textwidth}
	\centering
	\includegraphics[width=\textwidth]{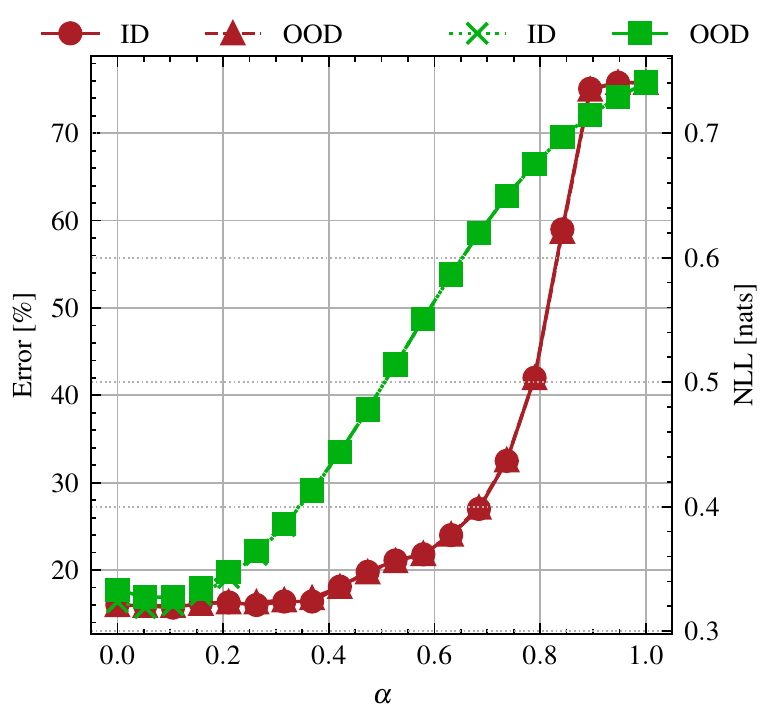}
	\caption{Error and NLL.}
	\label{fig:loss_landscape:classification_adult-fc-weight_additive_gaussian-lin_error_nll}
\end{subfigure}
\begin{subfigure}{0.25\textwidth}
	\centering
	\includegraphics[width=\textwidth]{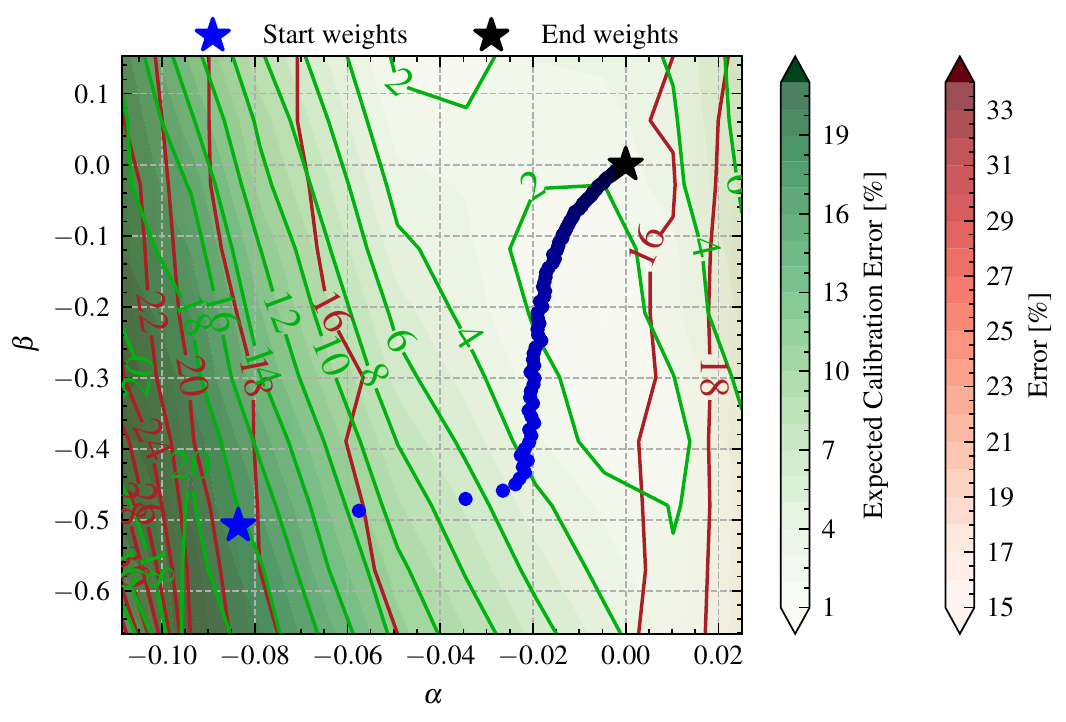}
	\caption{Error and ECE on ID.}
	\label{fig:loss_landscape:classification_adult-fc-weight_additive_gaussian-test_2d_error_ece}
\end{subfigure}
\begin{subfigure}{0.25\textwidth}
	\centering
	\includegraphics[width=\textwidth]{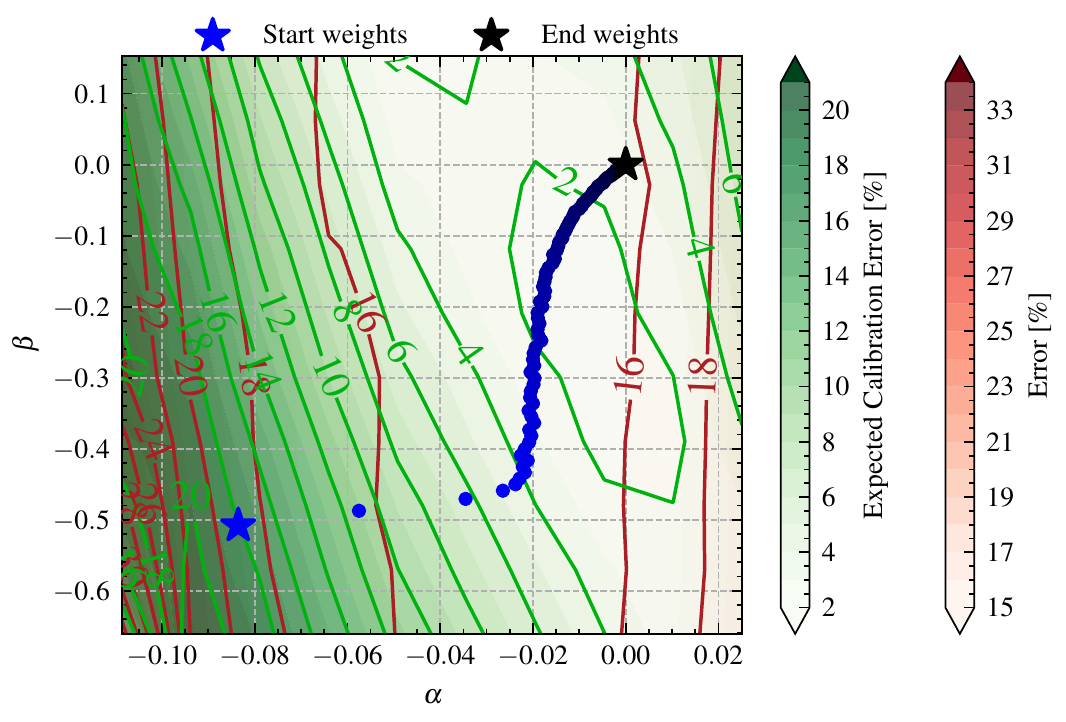}
	\caption{Error and ECE on OOD.}
	\label{fig:loss_landscape:classification_adult-fc-weight_additive_gaussian-test_2d_aug_error_ece}
\end{subfigure}
\caption{Weight Additive Gaussian on Adult.
\textit{Observations}:
Did not change the smoothness of the 1D curves or the 2D metric landscape trajectory compared to no noise.}
\label{fig:loss_landscape:classification_adult-fc-weight_additive_gaussian}
\end{figure}
\begin{figure}
\centering
\begin{subfigure}{0.21\textwidth}
	\centering
	\includegraphics[width=\textwidth]{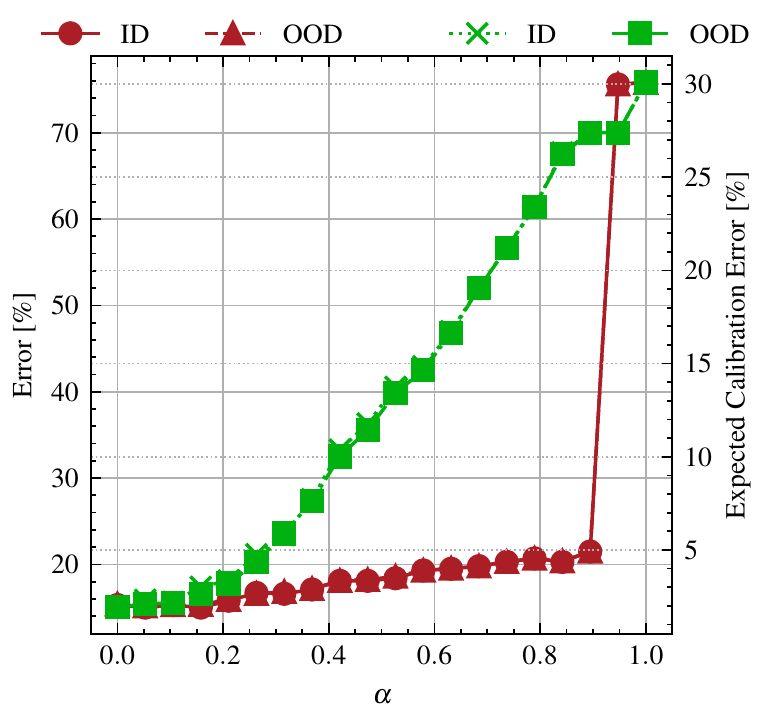}
	\caption{Error and ECE.}
	\label{fig:loss_landscape:classification_adult-fc-weight_dropconnect-lin_error_ece}
\end{subfigure}
\begin{subfigure}{0.21\textwidth}
	\centering
	\includegraphics[width=\textwidth]{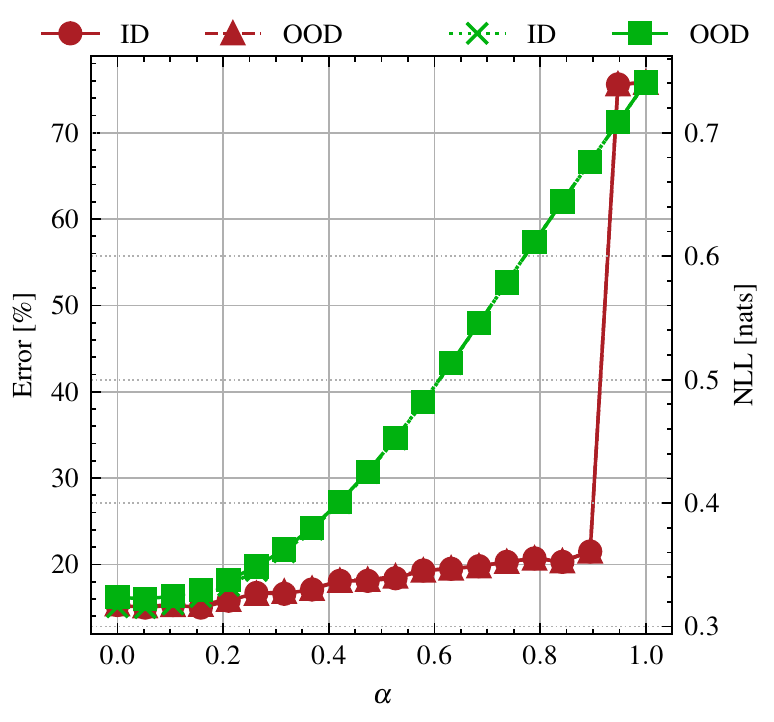}
	\caption{Error and NLL.}
	\label{fig:loss_landscape:classification_adult-fc-weight_dropconnect-lin_error_nll}
\end{subfigure}
\begin{subfigure}{0.25\textwidth}
	\centering
	\includegraphics[width=\textwidth]{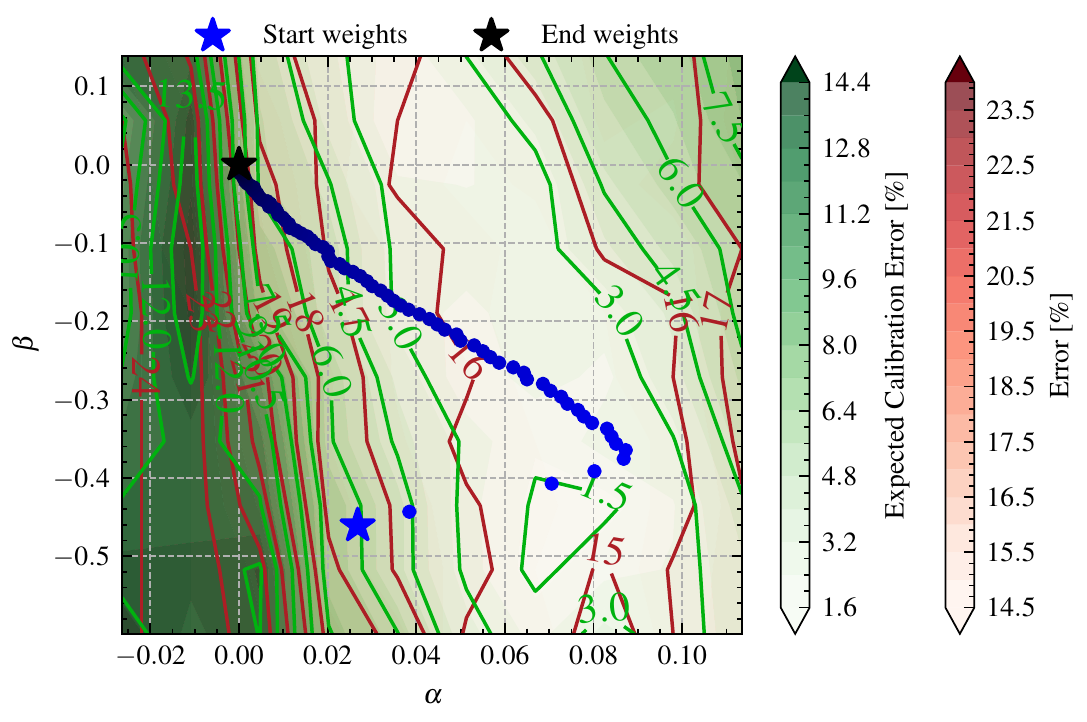}
	\caption{Error and ECE on ID.}
	\label{fig:loss_landscape:classification_adult-fc-weight_dropconnect-test_2d_error_ece}
\end{subfigure}
\begin{subfigure}{0.25\textwidth}
	\centering
	\includegraphics[width=\textwidth]{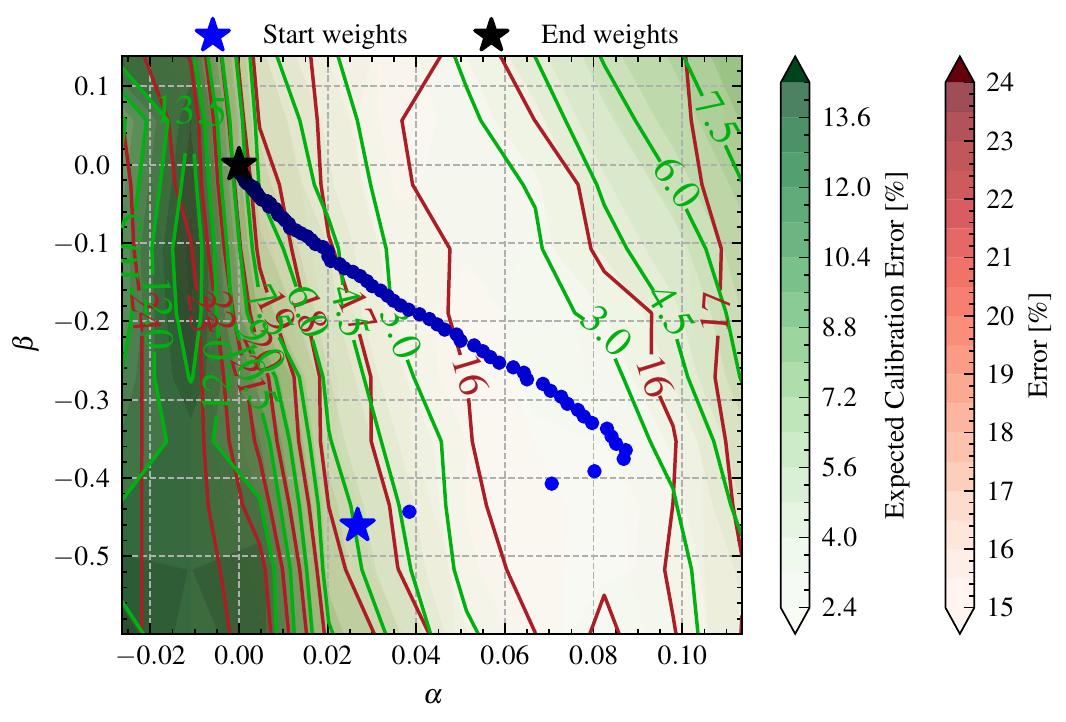}
	\caption{Error and ECE on OOD.}
	\label{fig:loss_landscape:classification_adult-fc-weight_dropconnect-test_2d_aug_error_ece}
\end{subfigure}
\caption{Weight DropConnect on Adult. 
\textit{Observations}:
Changed the ECE curvature and made the NLL and ECE plots smoother in the 1D case.
In the 2D plots, the ECE and error appear aligned during optimisation.
}
\label{fig:loss_landscape:classification_adult-fc-weight_dropconnect}
\end{figure}
\begin{figure}
\centering
\begin{subfigure}{0.25\textwidth}
	\centering
	\includegraphics[width=\textwidth]{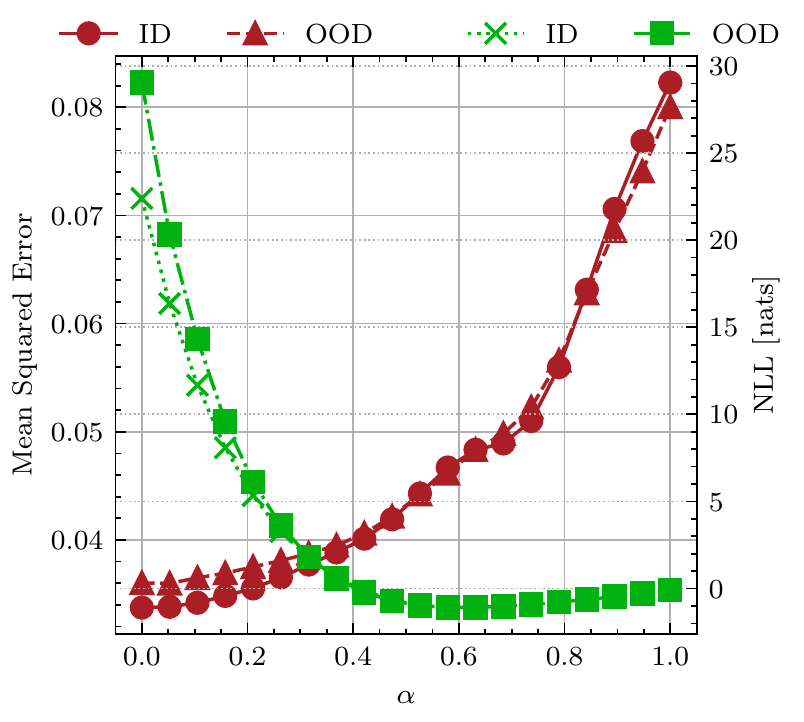}
	\caption{MSE and NLL.}
	\label{fig:loss_landscape:wiki_face-resnet-input_additive_gaussian-lin_mse_nll}
\end{subfigure}
\begin{subfigure}{0.35\textwidth}
	\centering
	\includegraphics[width=\textwidth]{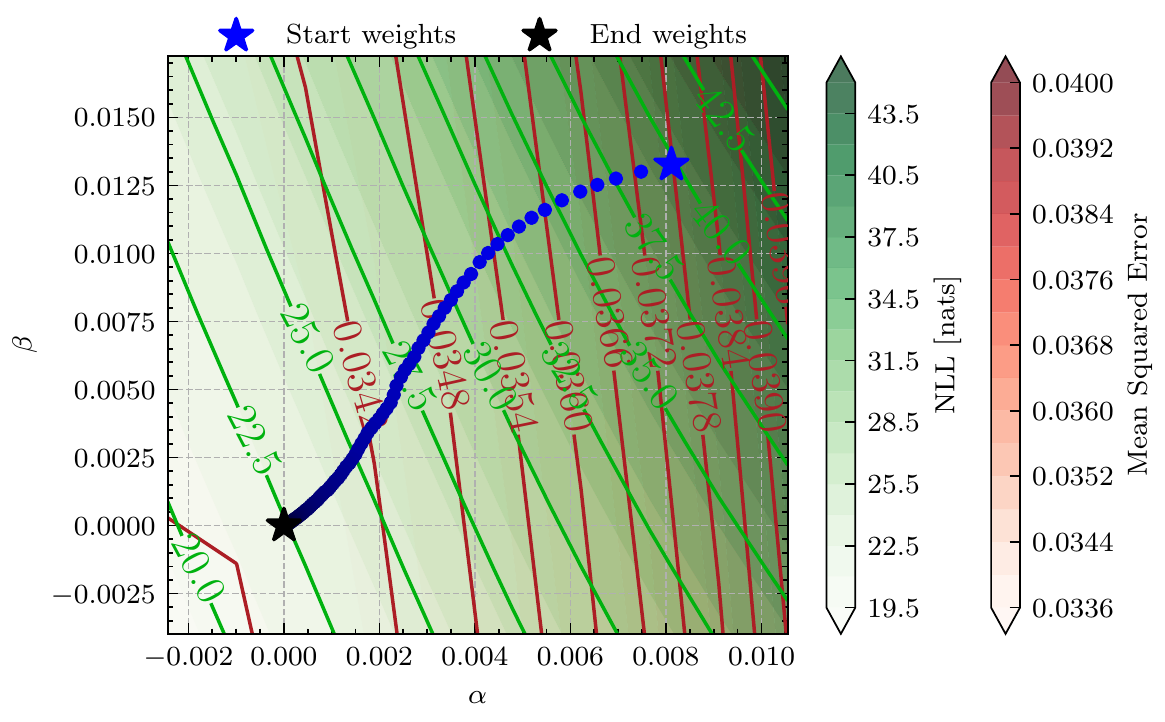}
	\caption{MSE and NLL on ID.}
	\label{fig:loss_landscape:wiki_face-resnet-input_additive_gaussian-test_2d_mse_nll}
\end{subfigure}
\begin{subfigure}{0.35\textwidth}
	\centering
	\includegraphics[width=\textwidth]{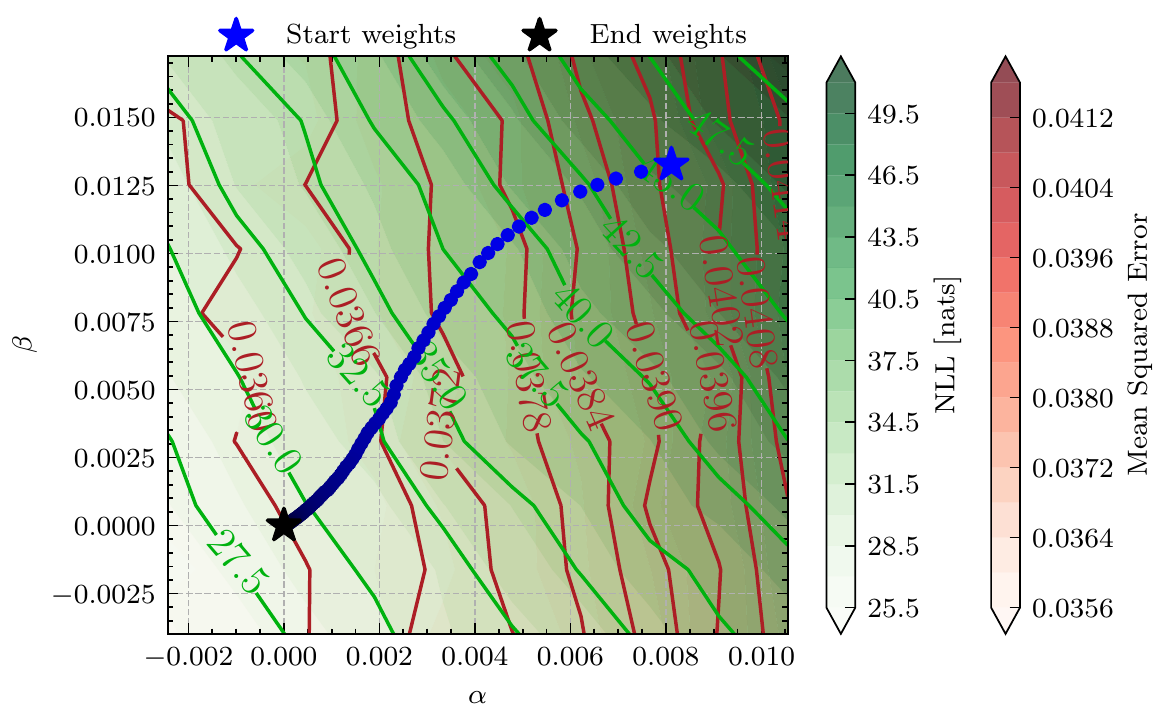}
	\caption{MSE and NLL on OOD.}
	\label{fig:loss_landscape:wiki_face-resnet-input_additive_gaussian-test_2d_aug_mse_nll}
\end{subfigure}
\caption{Input Additive Gaussian on WikiFace.
\textit{Observations}: Did not change the smoothness of the 1D curves, or the 2D trajectory.}
\label{fig:loss_landscape:wiki_face-resnet-input_additive_gaussian}
\end{figure}
\begin{figure}
\centering
\begin{subfigure}{0.25\textwidth}
	\centering
	\includegraphics[width=\textwidth]{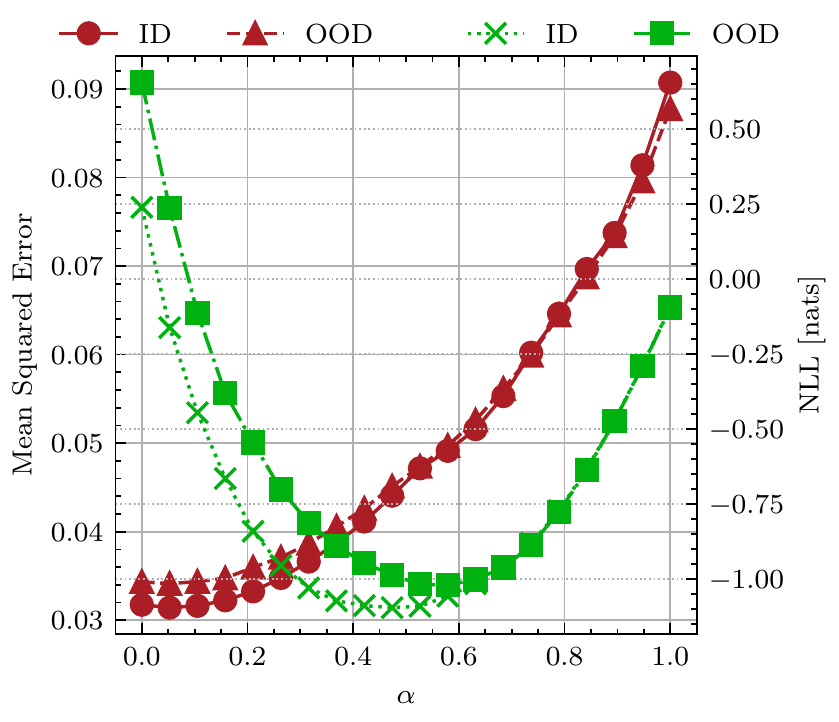}
	\caption{MSE and NLL.}
	\label{fig:loss_landscape:wiki_face-resnet-input_random_crop_horizontal_flip-lin_mse_nll}
\end{subfigure}
\begin{subfigure}{0.35\textwidth}
	\centering
	\includegraphics[width=\textwidth]{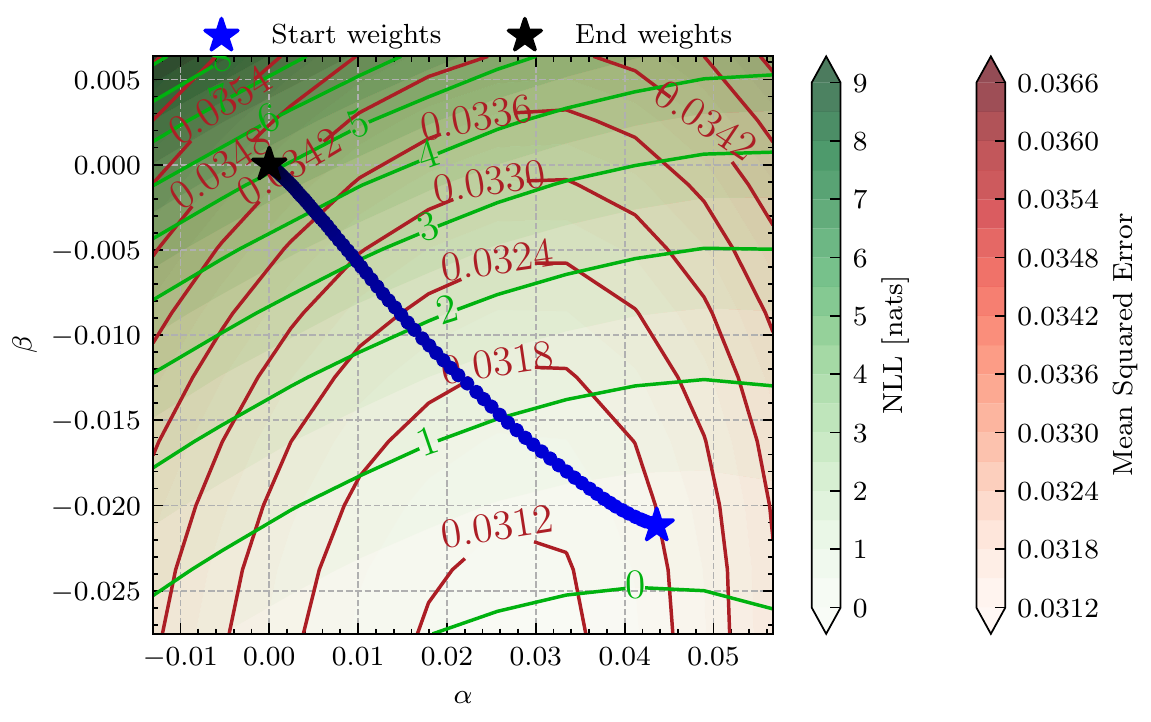}
	\caption{MSE and NLL on ID.}
	\label{fig:loss_landscape:wiki_face-resnet-input_random_crop_horizontal_flip-test_2d_mse_nll}
\end{subfigure}
\begin{subfigure}{0.35\textwidth}
	\centering
	\includegraphics[width=\textwidth]{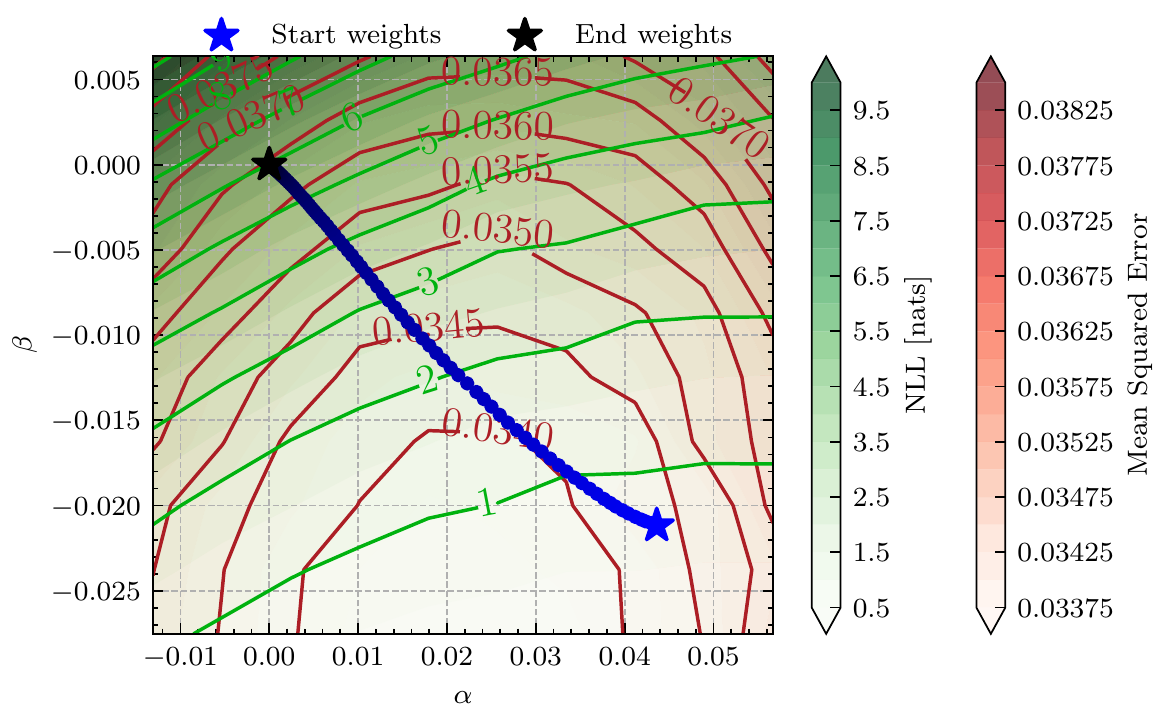}
	\caption{MSE and NLL on OOD.}
	\label{fig:loss_landscape:wiki_face-resnet-input_random_crop_horizontal_flip-test_2d_aug_mse_nll}
\end{subfigure}
\caption{Input Random Crop, Horizontal Flip on WikiFace.
\textit{Observations}: Surprisingly, the NLL starts decreasing compared to MSE as the model is interpolated between the final and the initial model in the 1D plots.
The 2D plots demonstrate that the model was able to explore a deeper optimal from the start where NLL was slower to converge than MSE.}
\label{fig:loss_landscape:wiki_face-resnet-input_random_crop_horizontal_flip}
\end{figure}
\begin{figure}
\centering
\begin{subfigure}{0.25\textwidth}
	\centering
	\includegraphics[width=\textwidth]{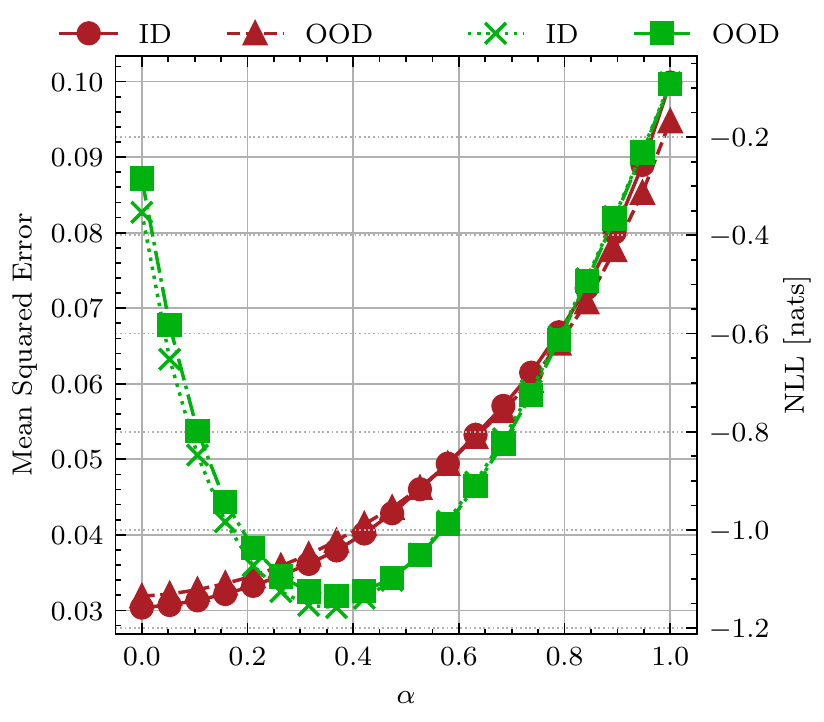}
	\caption{MSE and NLL.}
	\label{fig:loss_landscape:wiki_face-resnet-input_augmix-lin_mse_nll}
\end{subfigure}
\begin{subfigure}{0.35\textwidth}
	\centering
	\includegraphics[width=\textwidth]{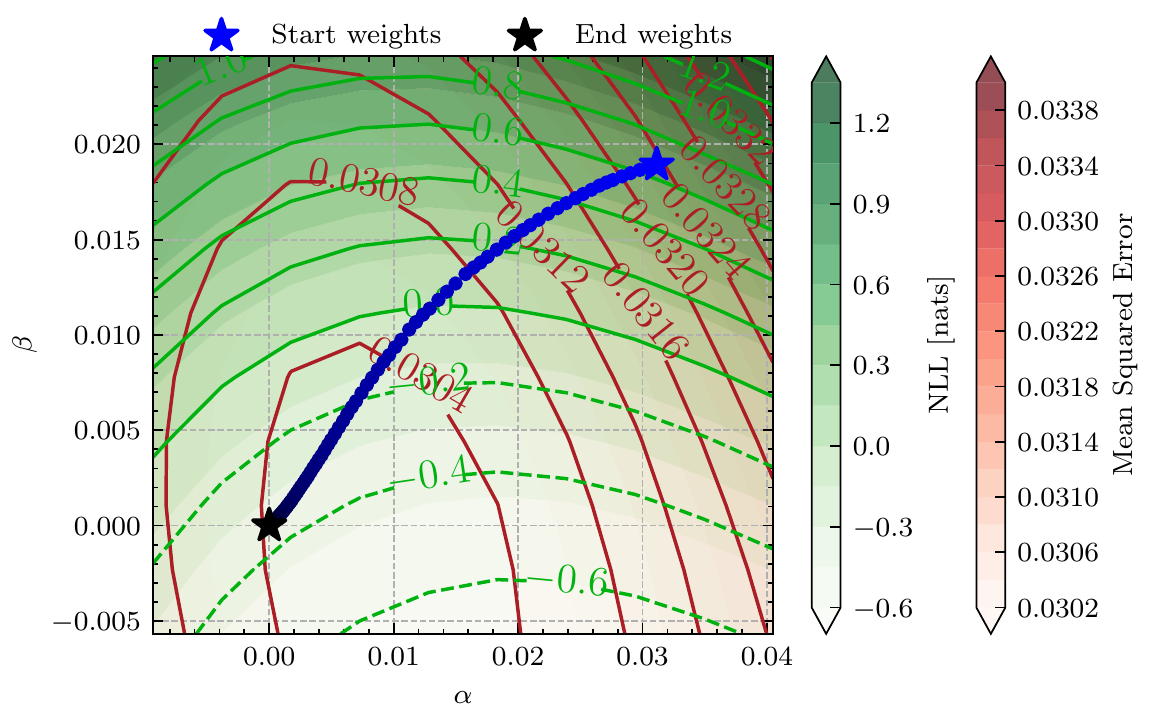}
	\caption{MSE and NLL on ID.}
	\label{fig:loss_landscape:wiki_face-resnet-input_augmix-test_2d_mse_nll}
\end{subfigure}
\begin{subfigure}{0.35\textwidth}
	\centering
	\includegraphics[width=\textwidth]{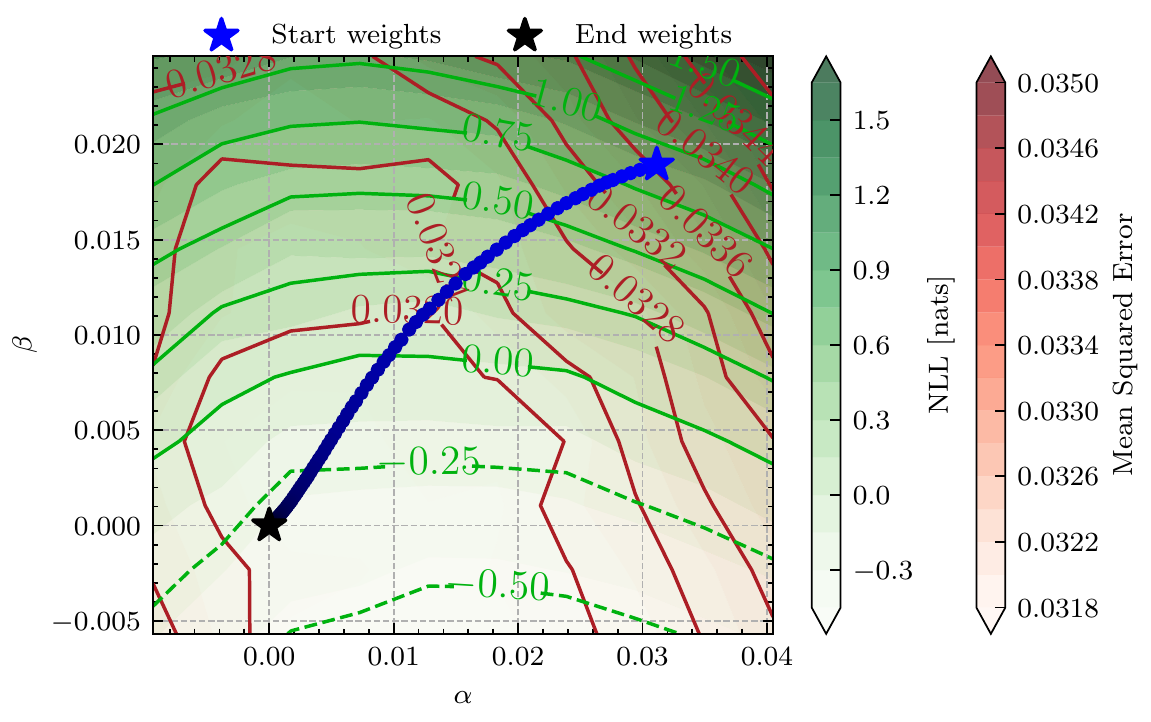}
	\caption{MSE and NLL on OOD.}
	\label{fig:loss_landscape:wiki_face-resnet-input_augmix-test_2d_aug_mse_nll}
\end{subfigure}
\caption{Input AugMix on WikiFace. 
\textit{Observations}: Surprisingly, the NLL starts decreasing compared to MSE as the model is interpolated between the final and the initial model in the 1D plots.
The 2D plots demonstrate that the model was able to explore a deeper optimal from the start where NLL was slower to converge than MSE, and it did not converge in the optima from the perspective of NLL.}
\label{fig:loss_landscape:wiki_face-resnet-input_augmix}
\end{figure}
\begin{figure}
\centering
\begin{subfigure}{0.25\textwidth}
	\centering
	\includegraphics[width=\textwidth]{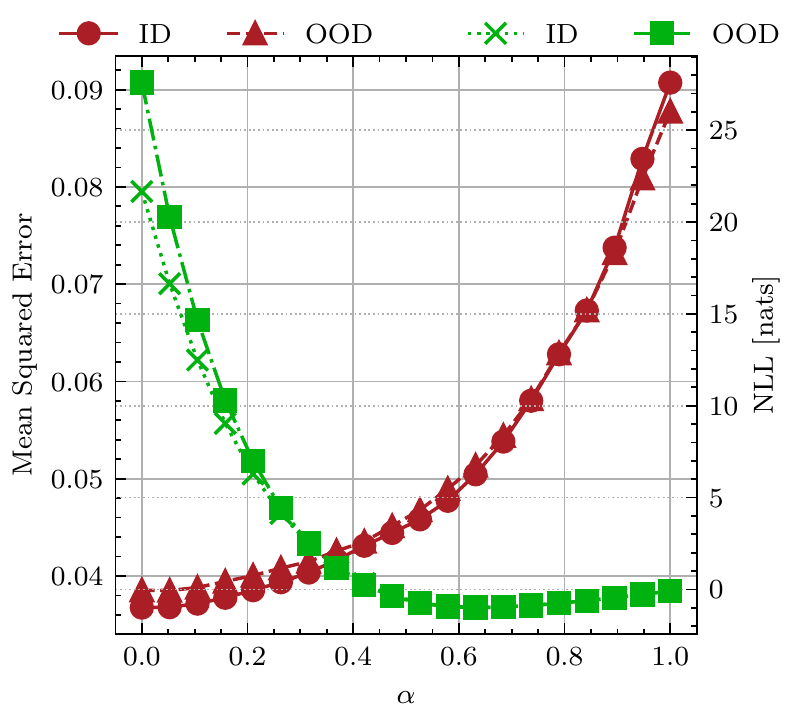}
	\caption{MSE and NLL.}
	\label{fig:loss_landscape:wiki_face-resnet-input_target_cmixup-lin_mse_nll}
\end{subfigure}
\begin{subfigure}{0.35\textwidth}
	\centering
	\includegraphics[width=\textwidth]{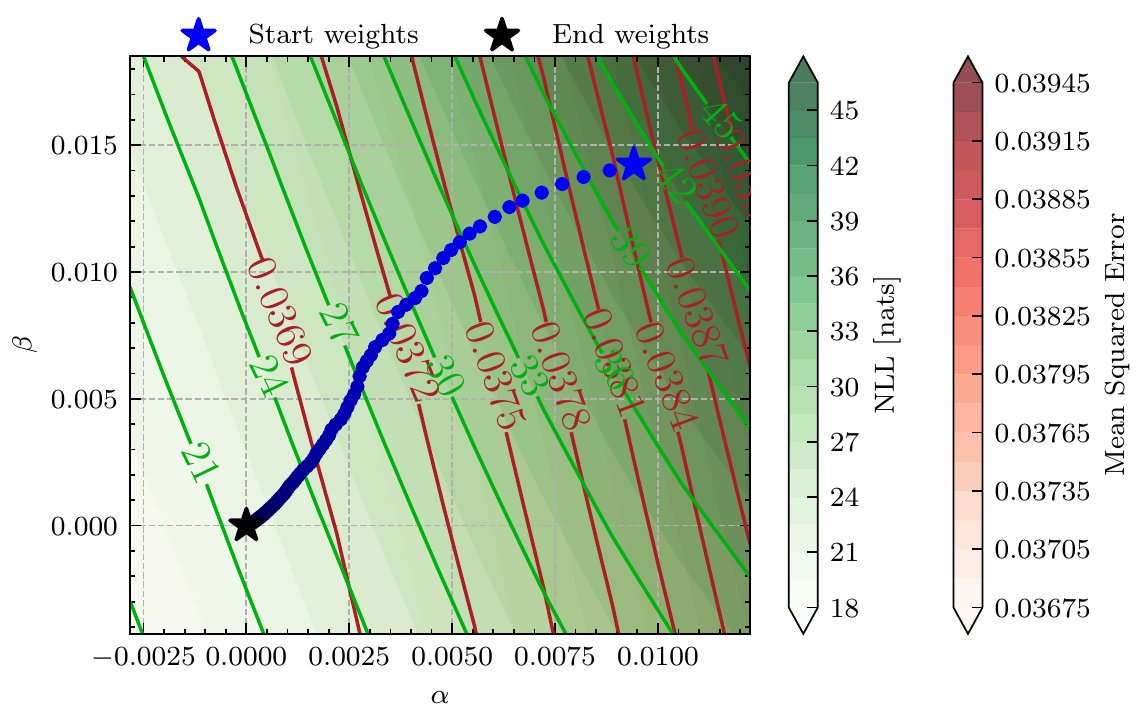}
	\caption{MSE and NLL on ID.}
	\label{fig:loss_landscape:wiki_face-resnet-input_target_cmixup-test_2d_mse_nll}
\end{subfigure}
\begin{subfigure}{0.35\textwidth}
	\centering
	\includegraphics[width=\textwidth]{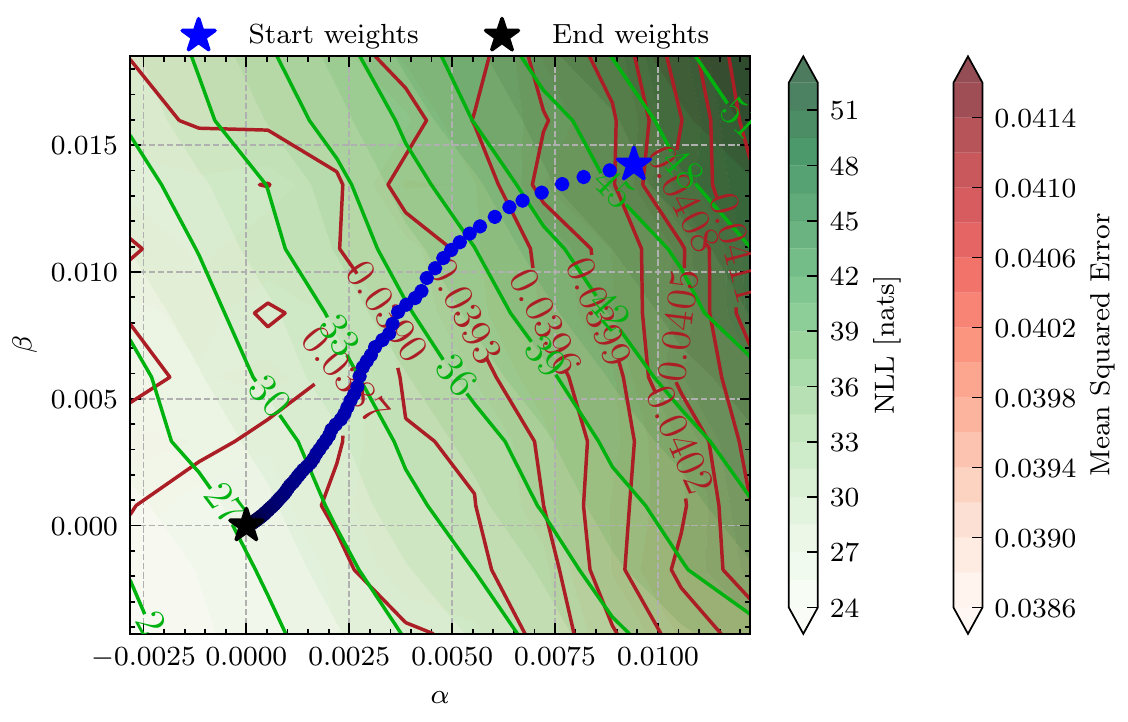}
	\caption{MSE and NLL on OOD.}
	\label{fig:loss_landscape:wiki_face-resnet-input_target_cmixup-test_2d_aug_mse_nll}
\end{subfigure}
\caption{Input-Target CMixUp on WikiFace.
\textit{Observations}: Did not change the smoothness of the 1D curves, or the 2D trajectory appears more exploratory compared to no noise.}
\label{fig:loss_landscape:wiki_face-resnet-input_target_cmixup}
\end{figure}
\begin{figure}
\centering
\begin{subfigure}{0.25\textwidth}
	\centering
	\includegraphics[width=\textwidth]{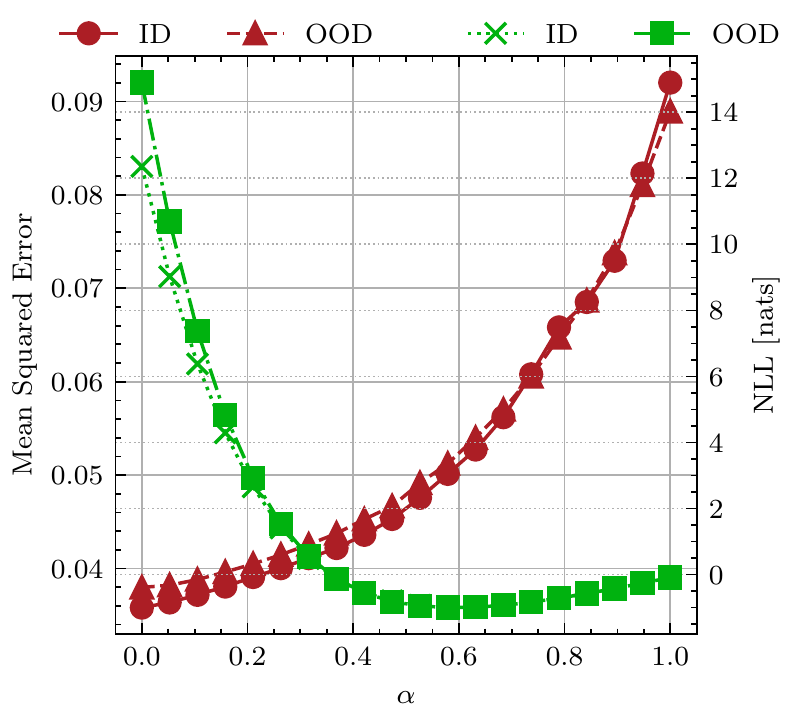}
	\caption{MSE and NLL.}
	\label{fig:loss_landscape:wiki_face-resnet-activation_additive_gaussian-lin_mse_nll}
\end{subfigure}
\begin{subfigure}{0.35\textwidth}
	\centering
	\includegraphics[width=\textwidth]{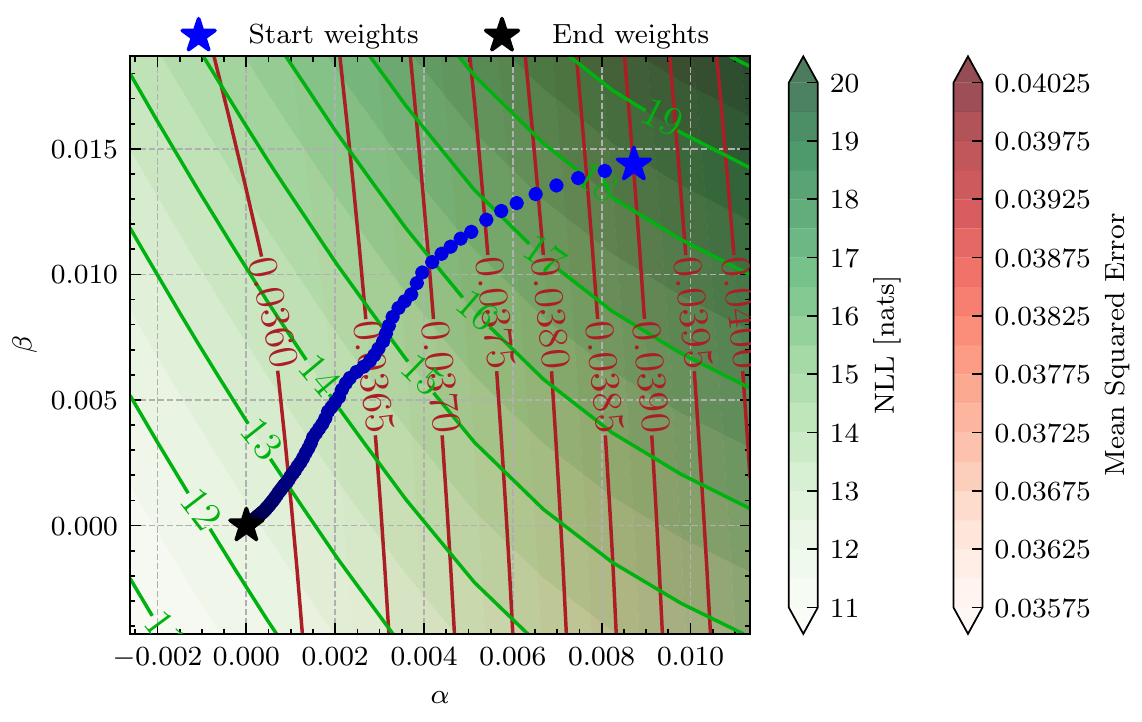}
	\caption{MSE and NLL on ID.}
	\label{fig:loss_landscape:wiki_face-resnet-activation_additive_gaussian-test_2d_mse_nll}
\end{subfigure}
\begin{subfigure}{0.35\textwidth}
	\centering
	\includegraphics[width=\textwidth]{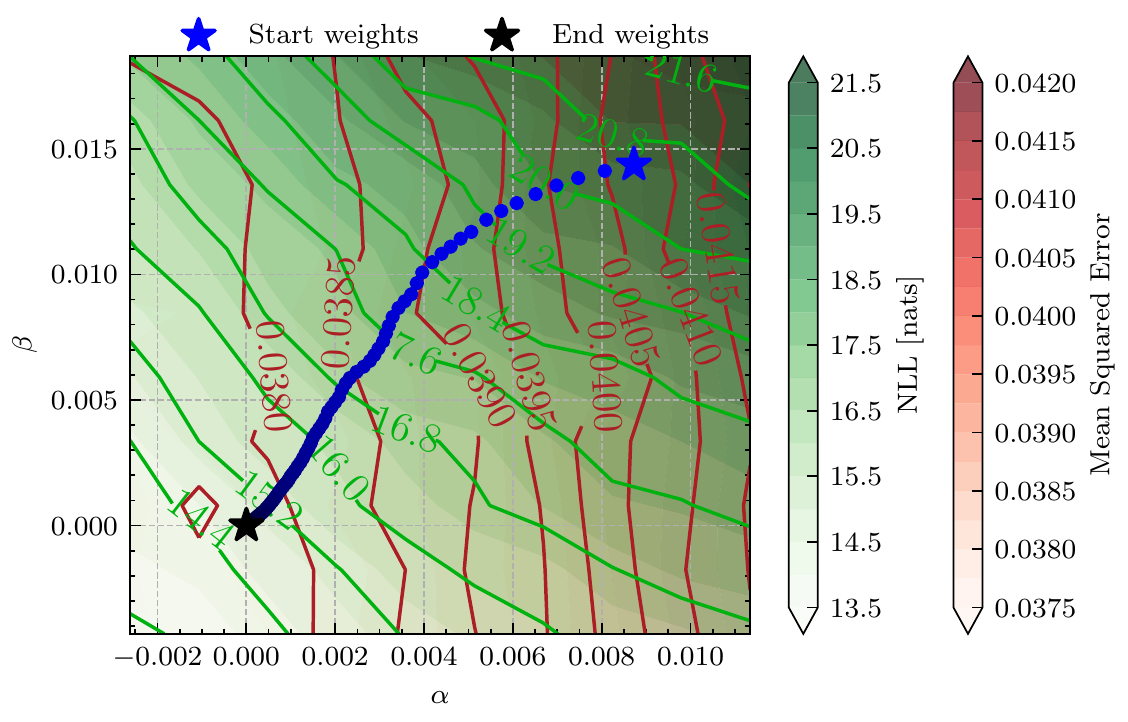}
	\caption{MSE and NLL on OOD.}
	\label{fig:loss_landscape:wiki_face-resnet-activation_additive_gaussian-test_2d_aug_mse_nll}
\end{subfigure}
\caption{Activation Additive Gaussian on WikiFace.
\textit{Observations}: Did not change the smoothness of the 1D curves, but the 2D trajectory appears more exploratory compared to no noise.}
\label{fig:loss_landscape:wiki_face-resnet-activation_additive_gaussian}
\end{figure}
\begin{figure}
\centering
\begin{subfigure}{0.25\textwidth}
	\centering
	\includegraphics[width=\textwidth]{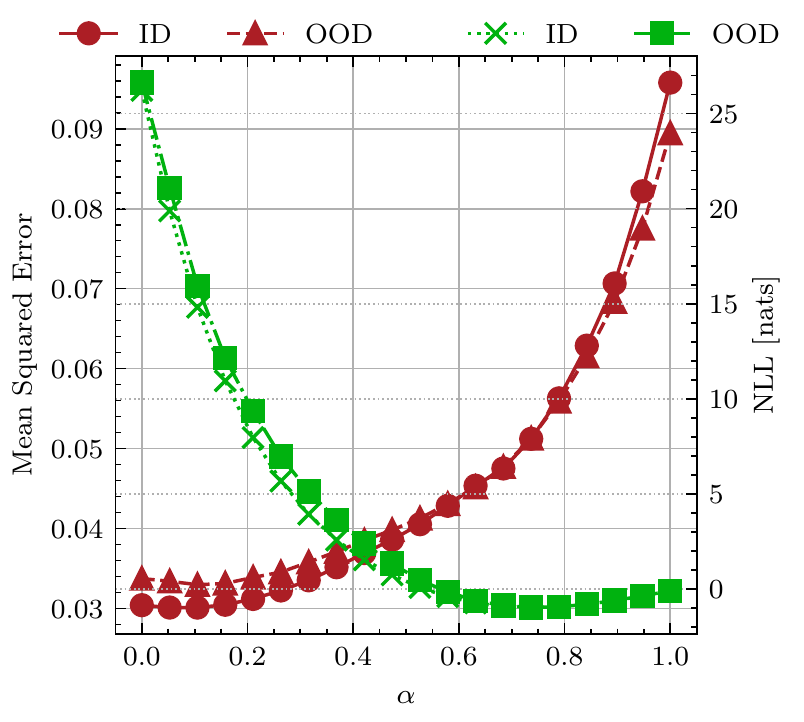}
	\caption{MSE and NLL.}
	\label{fig:loss_landscape:wiki_face-resnet-gradient_gaussian-lin_mse_nll}
\end{subfigure}
\begin{subfigure}{0.35\textwidth}
	\centering
	\includegraphics[width=\textwidth]{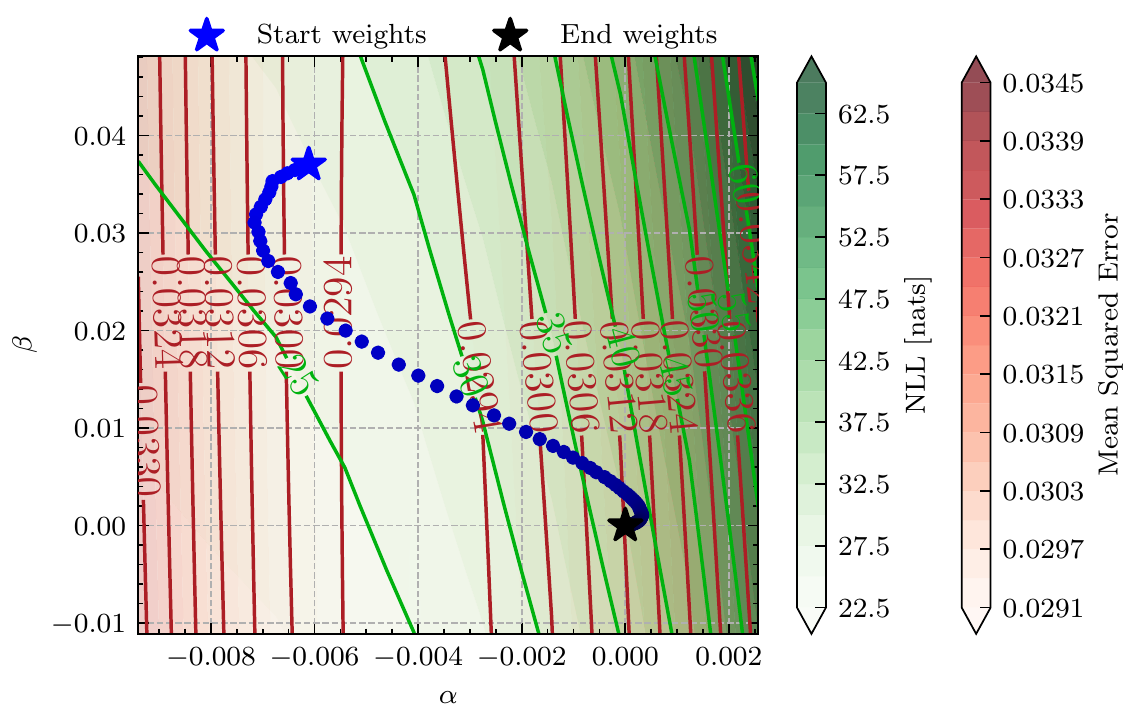}
	\caption{MSE and NLL on ID.}
	\label{fig:loss_landscape:wiki_face-resnet-gradient_gaussian-test_2d_mse_nll}
\end{subfigure}
\begin{subfigure}{0.35\textwidth}
	\centering
	\includegraphics[width=\textwidth]{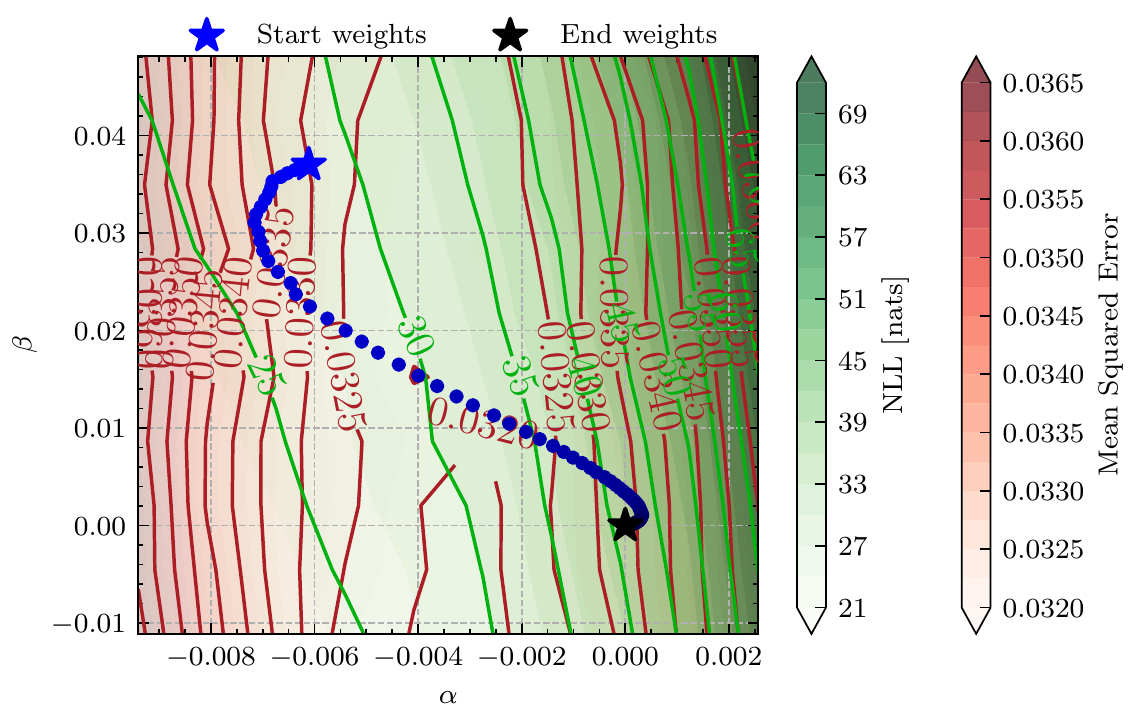}
	\caption{MSE and NLL on OOD.}
	\label{fig:loss_landscape:wiki_face-resnet-gradient_gaussian-test_2d_aug_mse_nll}
\end{subfigure}
\caption{Gradient Gaussian on WikiFace.
\textit{Observations}: Did not change the smoothness of the 1D curves, but the 2D trajectory appears to align MSE and NLL.
However, it seems that the optimisation missed a local minimum during training.}
\label{fig:loss_landscape:wiki_face-resnet-gradient_gaussian}
\end{figure}
\begin{figure}
\centering
\begin{subfigure}{0.25\textwidth}
	\centering
	\includegraphics[width=\textwidth]{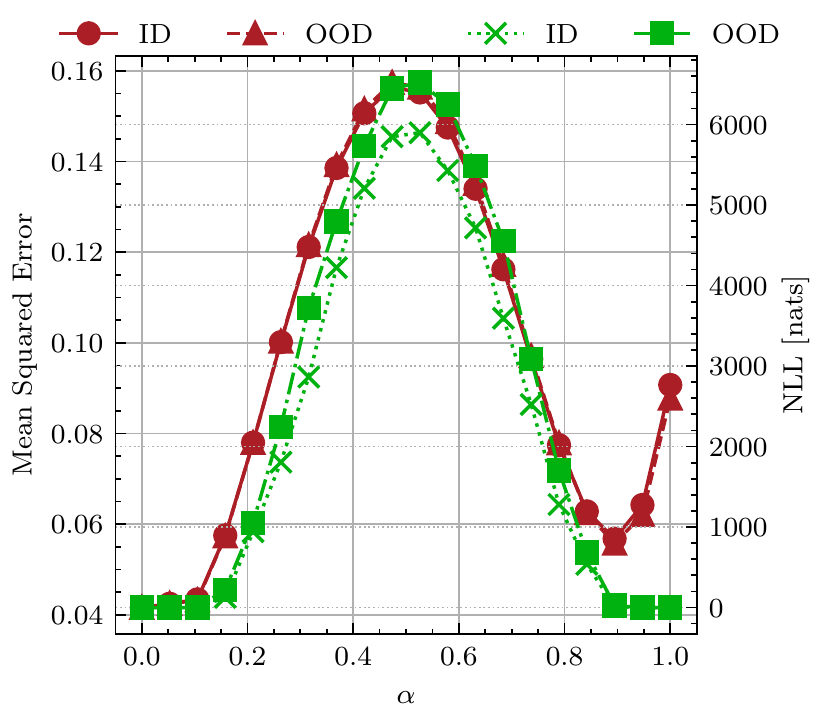}
	\caption{MSE and NLL.}
	\label{fig:loss_landscape:wiki_face-resnet-model_sp-lin_mse_nll}
\end{subfigure}
\begin{subfigure}{0.35\textwidth}
	\centering
	\includegraphics[width=\textwidth]{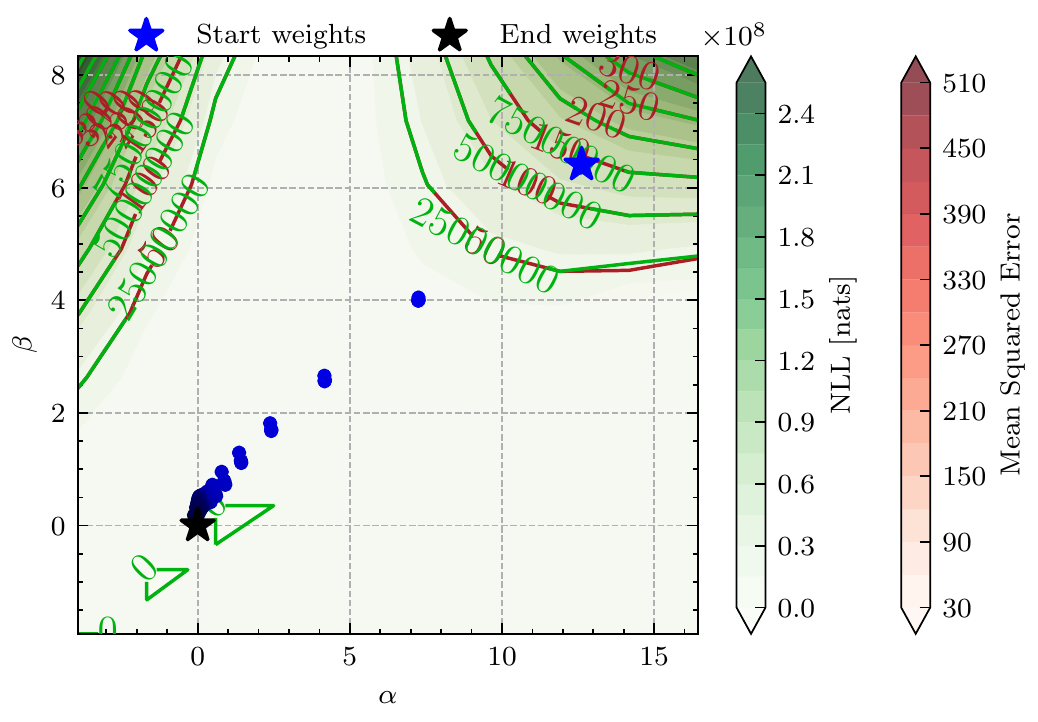}
	\caption{MSE and NLL on ID.}
	\label{fig:loss_landscape:wiki_face-resnet-model_sp-test_2d_mse_nll}
\end{subfigure}
\begin{subfigure}{0.35\textwidth}
	\centering
	\includegraphics[width=\textwidth]{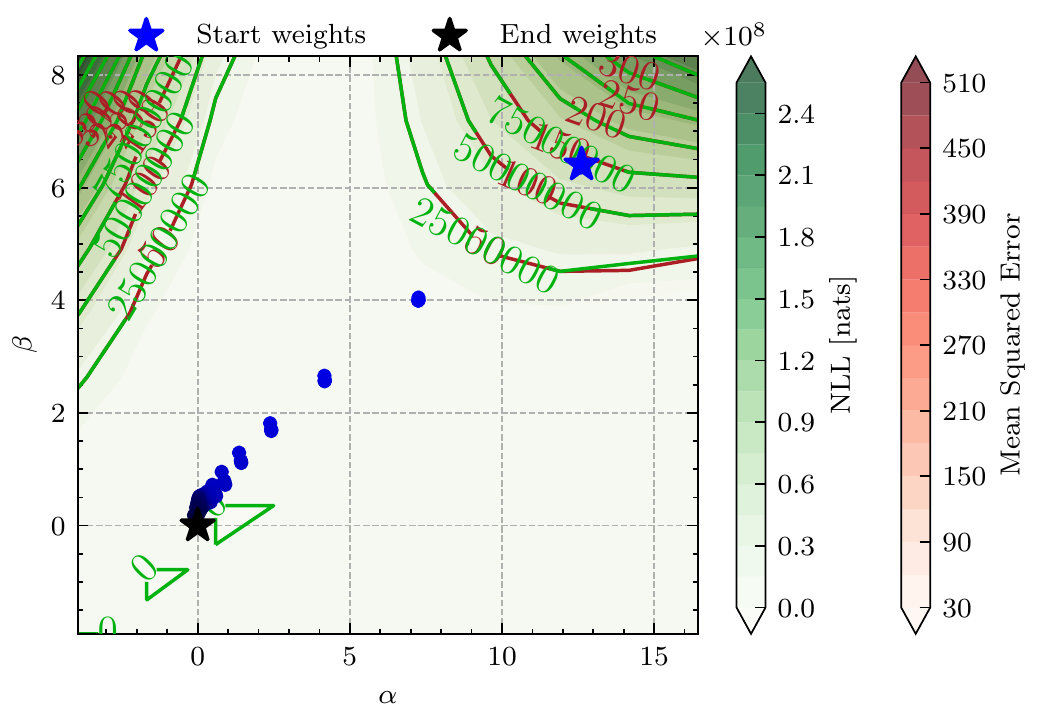}
	\caption{MSE and NLL on OOD.}
	\label{fig:loss_landscape:wiki_face-resnet-model_sp-test_2d_aug_mse_nll}
\end{subfigure}
\caption{Model Shrink and Perturb on WikiFace.
\textit{Observations}:
Due to shrinking and perturbation, the experiment appears to converge in a narrow basin and as seed in the 1D plots, the optimisation was completely non-linear and unrecoverable. }
\label{fig:loss_landscape:wiki_face-resnet-model_sp}
\end{figure}
\begin{figure}
\centering
\begin{subfigure}{0.25\textwidth}
	\centering
	\includegraphics[width=\textwidth]{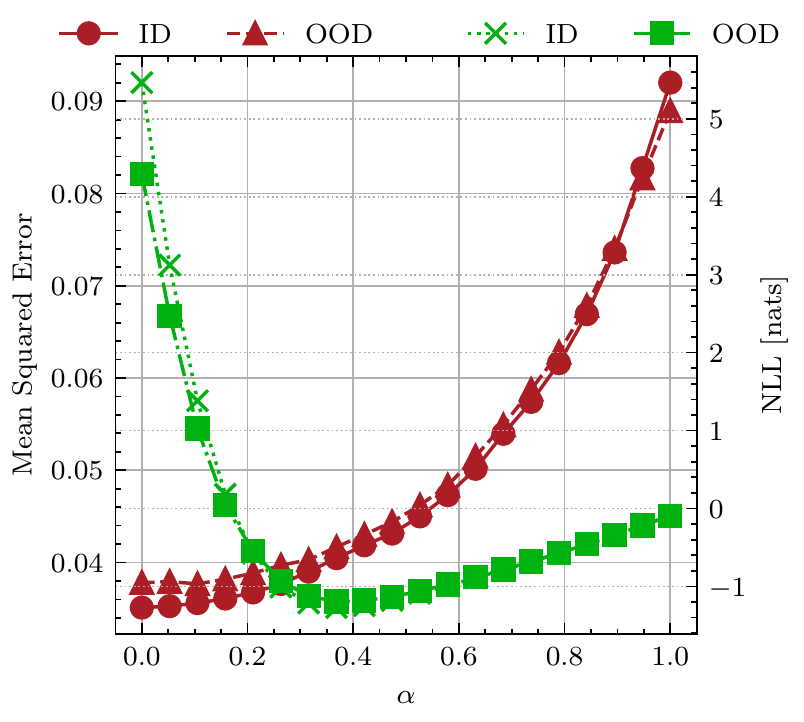}
	\caption{MSE and NLL.}
	\label{fig:loss_landscape:wiki_face-resnet-weight_additive_gaussian-lin_mse_nll}
\end{subfigure}
\begin{subfigure}{0.35\textwidth}
	\centering
	\includegraphics[width=\textwidth]{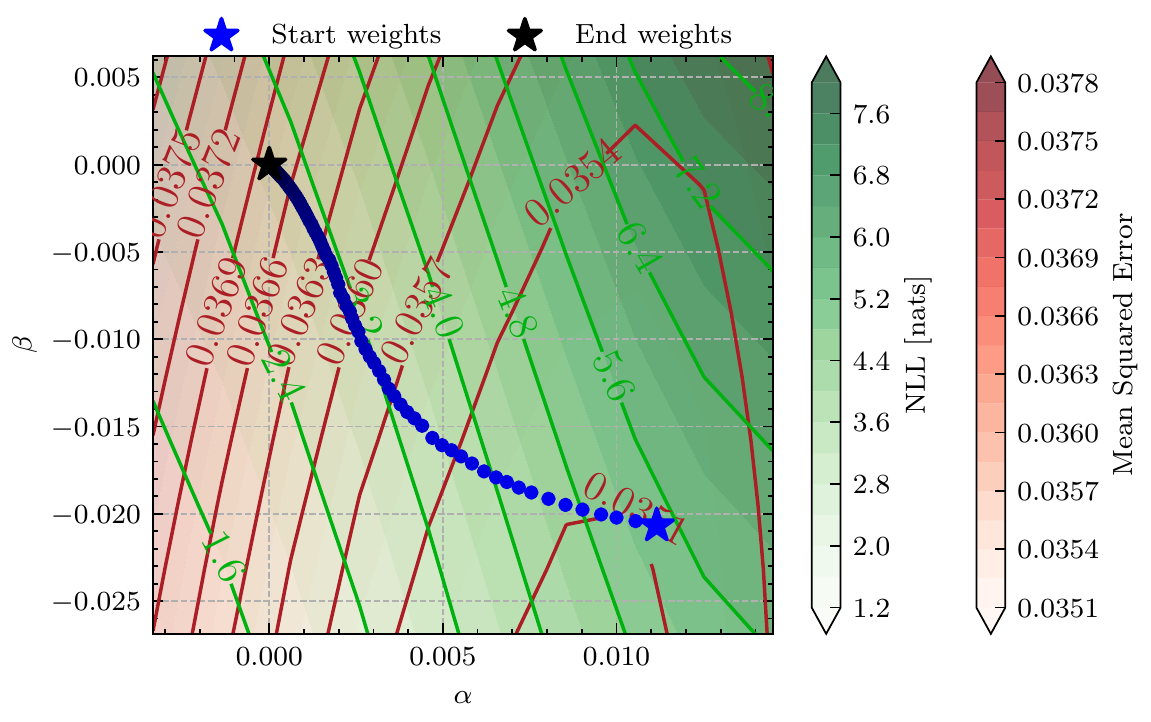}
	\caption{MSE and NLL on ID.}
	\label{fig:loss_landscape:wiki_face-resnet-weight_additive_gaussian-test_2d_mse_nll}
\end{subfigure}
\begin{subfigure}{0.35\textwidth}
	\centering
	\includegraphics[width=\textwidth]{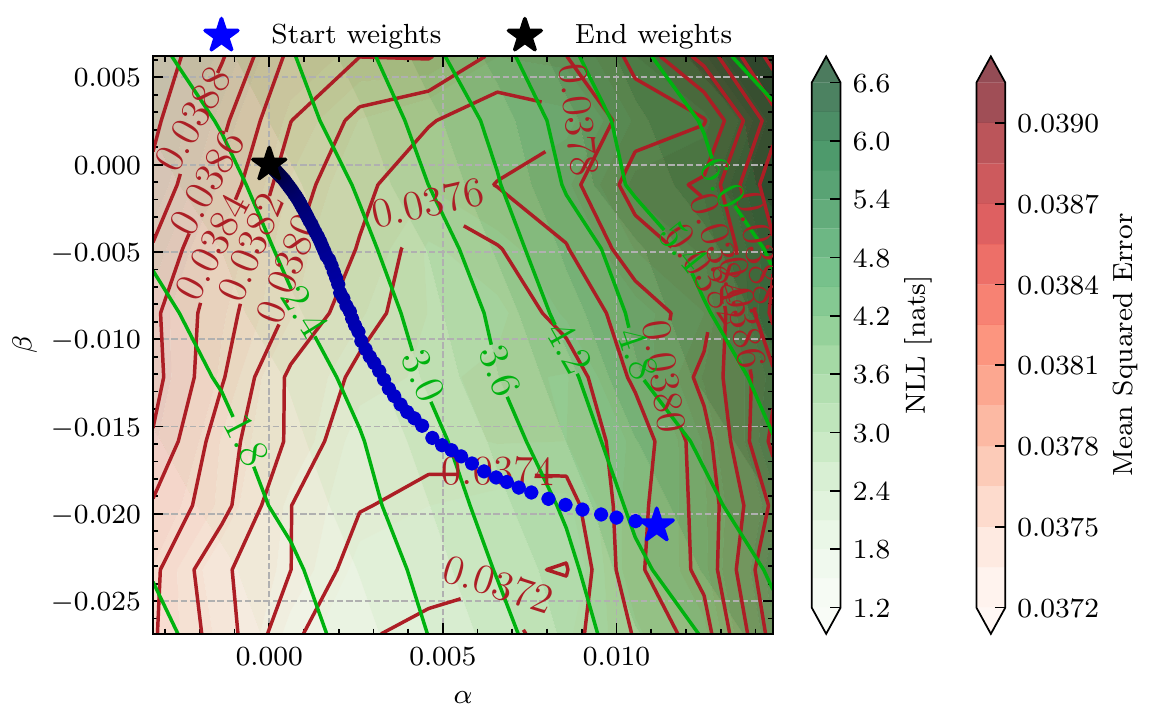}
	\caption{MSE and NLL on OOD.}
	\label{fig:loss_landscape:wiki_face-resnet-weight_additive_gaussian-test_2d_aug_mse_nll}
\end{subfigure}
\caption{Weight Additive Gaussian on WikiFace.
\textit{Observations}:
The 1D curves look similar to no noise, although with respect to a different scale for NLL.
The 2D plots explore a similar trajectory to no noise; however, the 2D landscape appears more distorted. }
\label{fig:loss_landscape:wiki_face-resnet-weight_additive_gaussian}
\end{figure}
\begin{figure}
\centering
\begin{subfigure}{0.25\textwidth}
	\centering
	\includegraphics[width=\textwidth]{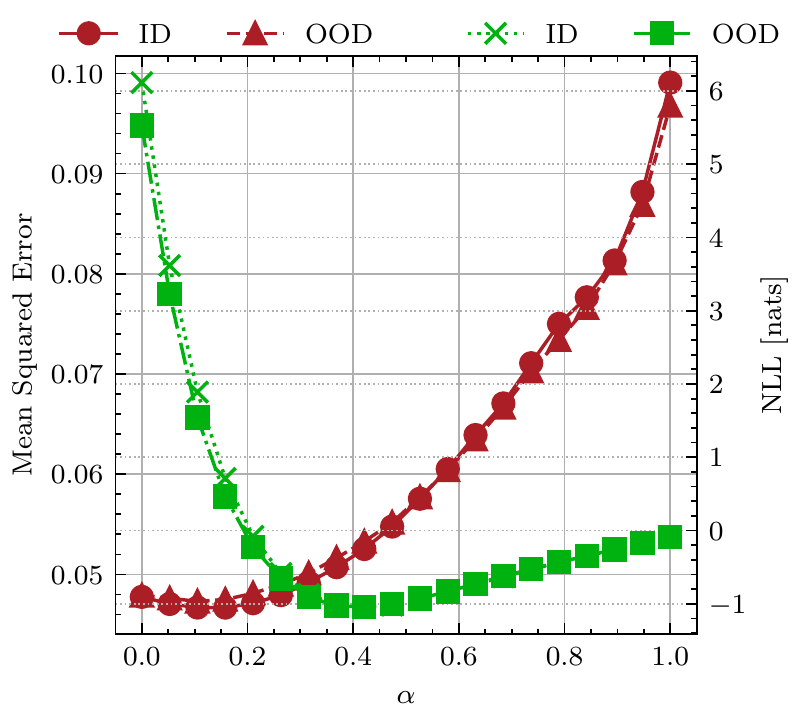}
	\caption{MSE and NLL.}
	\label{fig:loss_landscape:wiki_face-resnet-weight_dropconnect-lin_mse_nll}
\end{subfigure}
\begin{subfigure}{0.35\textwidth}
	\centering
	\includegraphics[width=\textwidth]{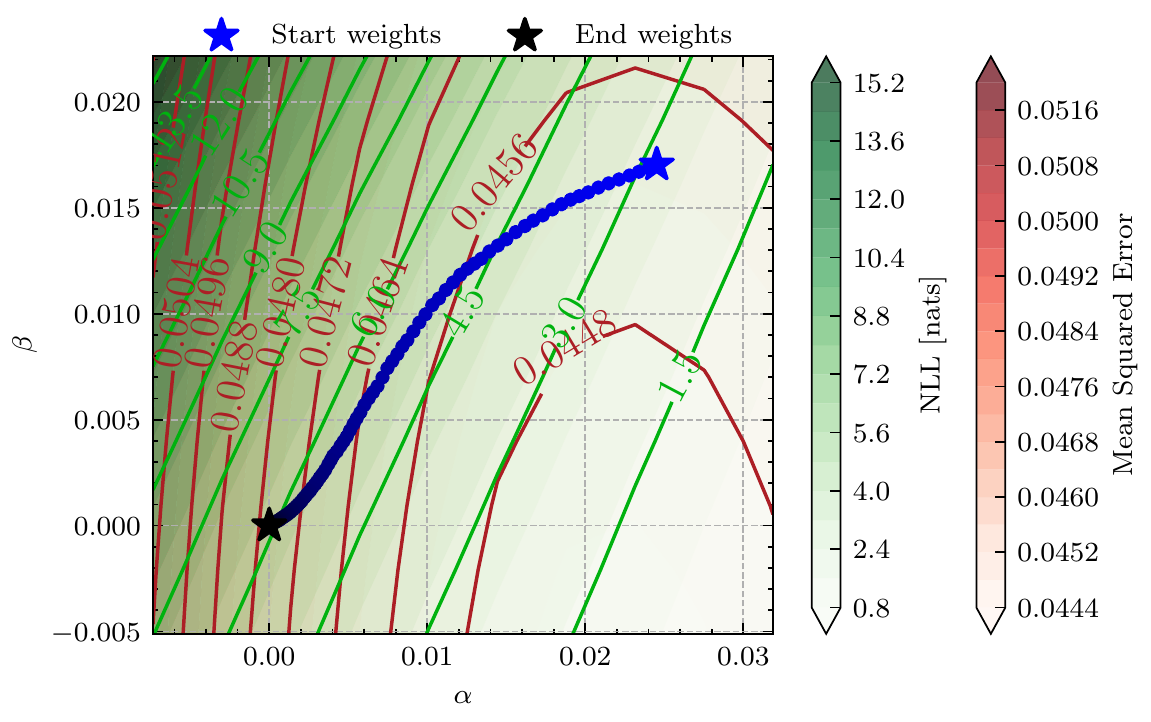}
	\caption{MSE and NLL on ID.}
	\label{fig:loss_landscape:wiki_face-resnet-weight_dropconnect-test_2d_mse_nll}
\end{subfigure}
\begin{subfigure}{0.35\textwidth}
	\centering
	\includegraphics[width=\textwidth]{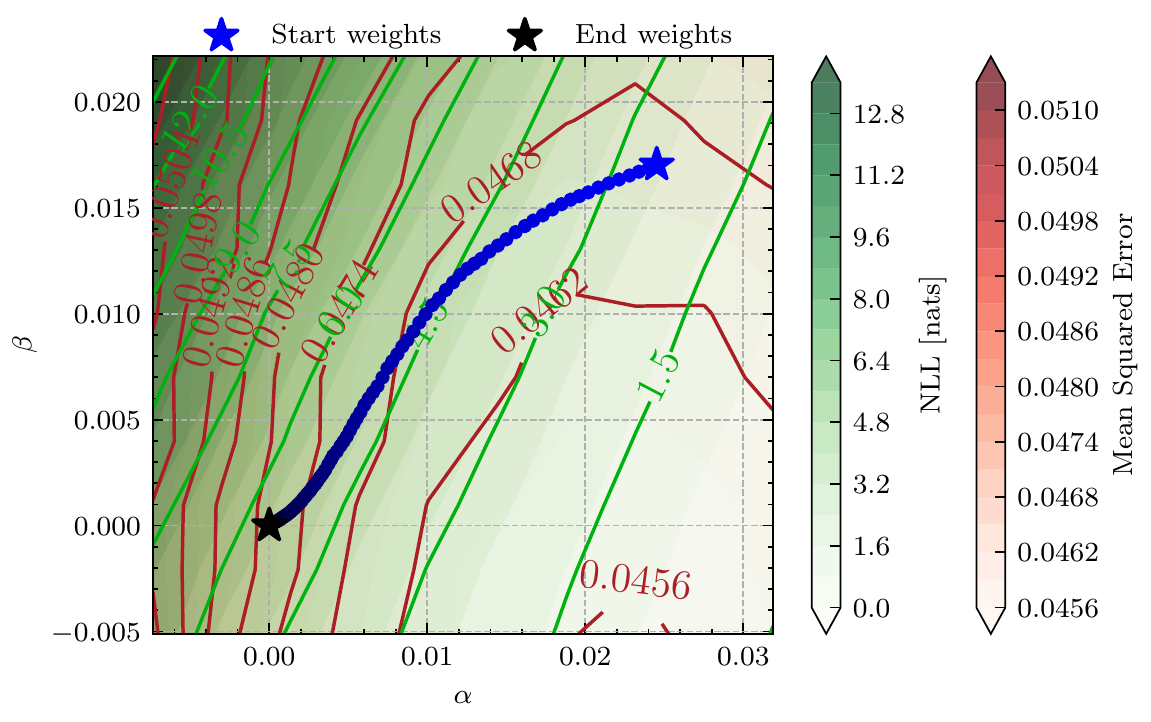}
	\caption{MSE and NLL on OOD.}
	\label{fig:loss_landscape:wiki_face-resnet-weight_dropconnect-test_2d_aug_mse_nll}
\end{subfigure}
\caption{Weight DropConnect on WikiFace.
\textit{Observations}:
The 1D curves look similar to no noise, although with respect to a different scale for NLL.
The 2D plots explore a similar trajectory to no noise.}
\label{fig:loss_landscape:wiki_face-resnet-weight_dropconnect}
\end{figure}
\begin{figure}
\centering
\begin{subfigure}{0.25\textwidth}
	\centering
	\includegraphics[width=\textwidth]{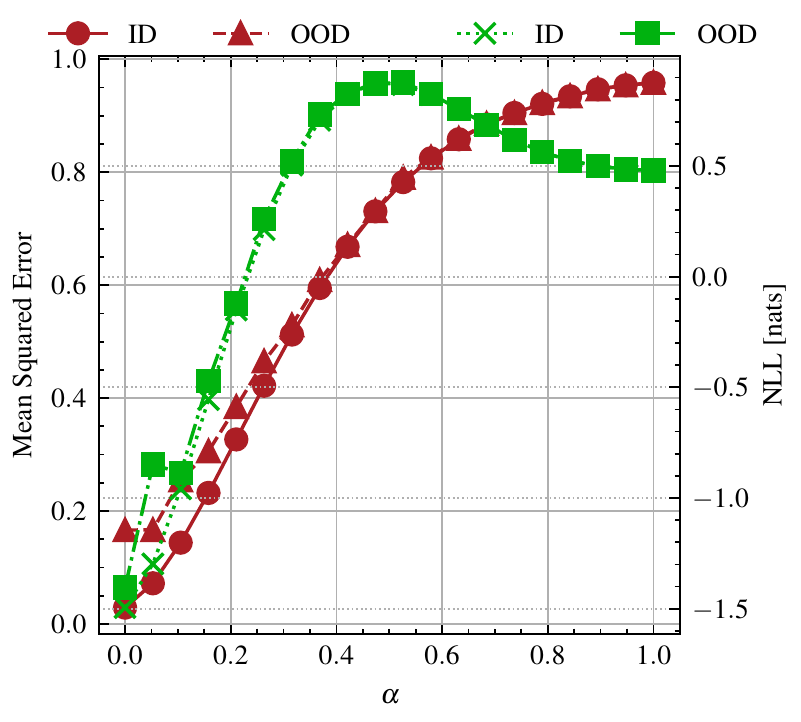}
	\caption{MSE and NLL.}
	\label{fig:loss_landscape:regression_yacht-fc-vanilla-lin_mse_nll}
\end{subfigure}
\begin{subfigure}{0.35\textwidth}
	\centering
	\includegraphics[width=\textwidth]{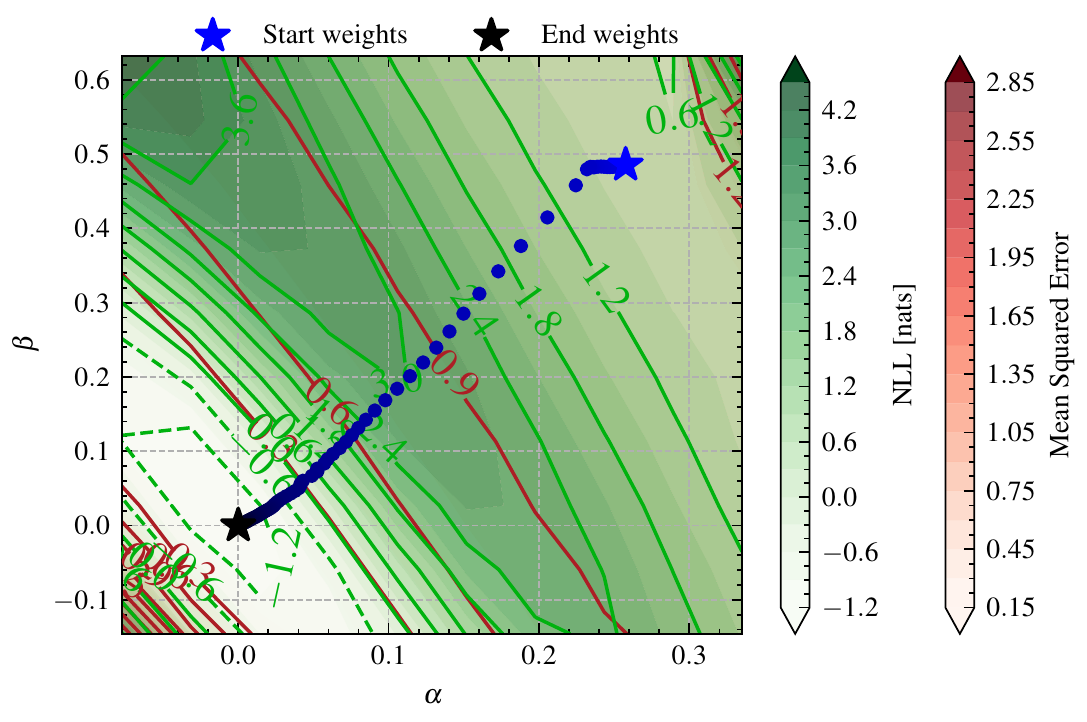}
	\caption{MSE and NLL on ID.}
	\label{fig:loss_landscape:regression_yacht-fc-vanilla-test_2d_mse_nll}
\end{subfigure}
\begin{subfigure}{0.35\textwidth}
	\centering
	\includegraphics[width=\textwidth]{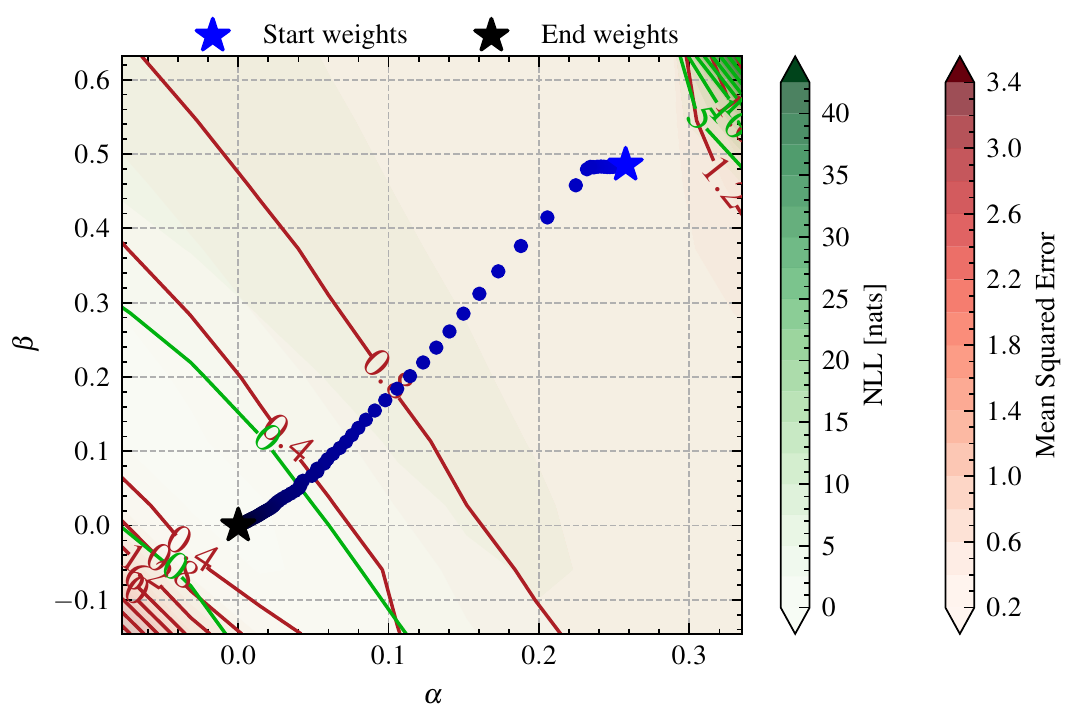}
	\caption{MSE and NLL on OOD.}
	\label{fig:loss_landscape:regression_yacht-fc-vanilla-test_2d_aug_mse_nll}
\end{subfigure}
\caption{No noise on Yacht.}
\label{fig:loss_landscape:regression_yacht-fc-vanilla}
\end{figure}
\begin{figure}
\centering
\begin{subfigure}{0.25\textwidth}
	\centering
	\includegraphics[width=\textwidth]{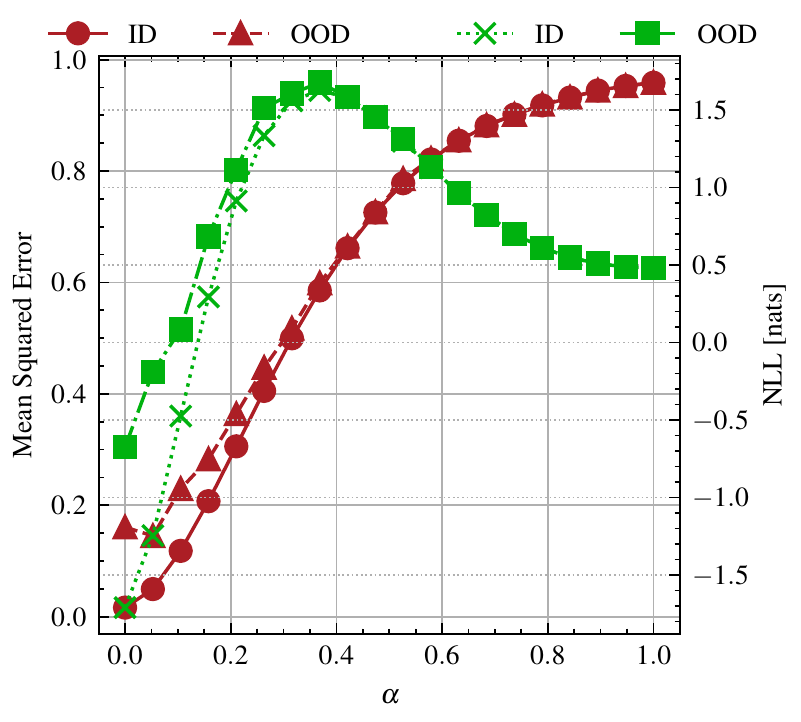}
	\caption{MSE and NLL.}
	\label{fig:loss_landscape:regression_yacht-fc-input_additive_gaussian-lin_mse_nll}
\end{subfigure}
\begin{subfigure}{0.35\textwidth}
	\centering
	\includegraphics[width=\textwidth]{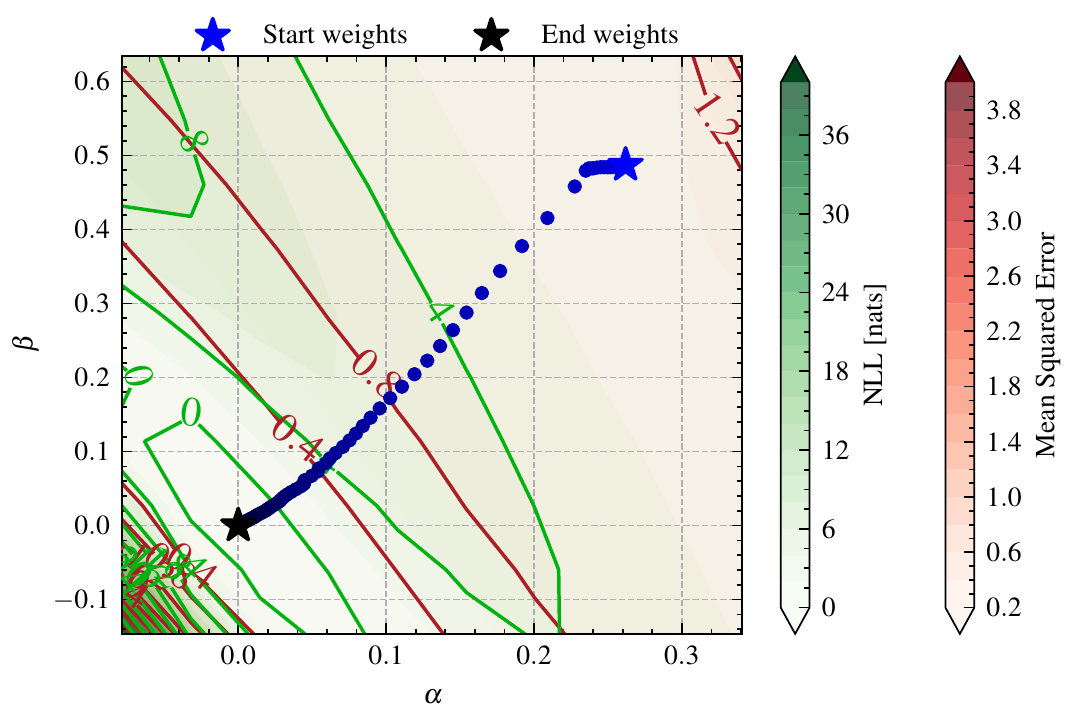}
	\caption{MSE and NLL on ID.}
	\label{fig:loss_landscape:regression_yacht-fc-input_additive_gaussian-test_2d_mse_nll}
\end{subfigure}
\begin{subfigure}{0.35\textwidth}
	\centering
	\includegraphics[width=\textwidth]{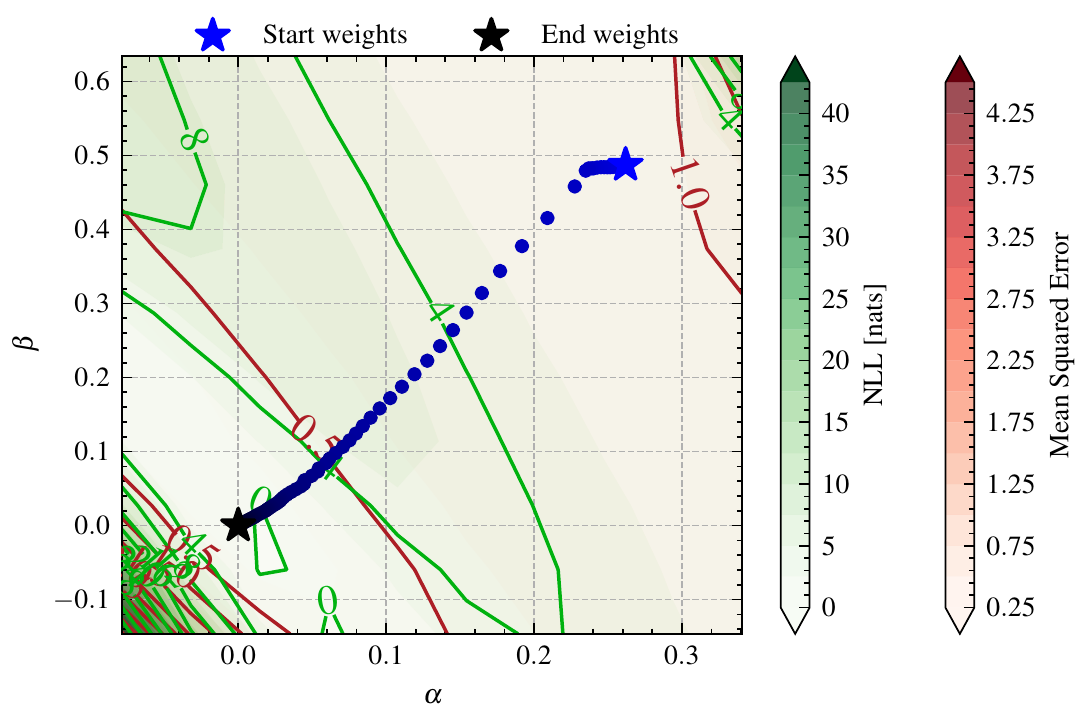}
	\caption{MSE and NLL on OOD.}
	\label{fig:loss_landscape:regression_yacht-fc-input_additive_gaussian-test_2d_aug_mse_nll}
\end{subfigure}
\caption{Input Additive Gaussian on Yacht.
\textit{Observations}:
While the shape of the 1D curves looks similar to no noise, the MSE and NLL magnitudes are different.
The OOD NLL is substantially higher than the OOD NLL for no noise.
The 2D plots demonstrate a wider landscape of feasible solutions than no noise.}
\label{fig:loss_landscape:regression_yacht-fc-input_additive_gaussian}
\end{figure}
\begin{figure}
\centering
\begin{subfigure}{0.25\textwidth}
	\centering
	\includegraphics[width=\textwidth]{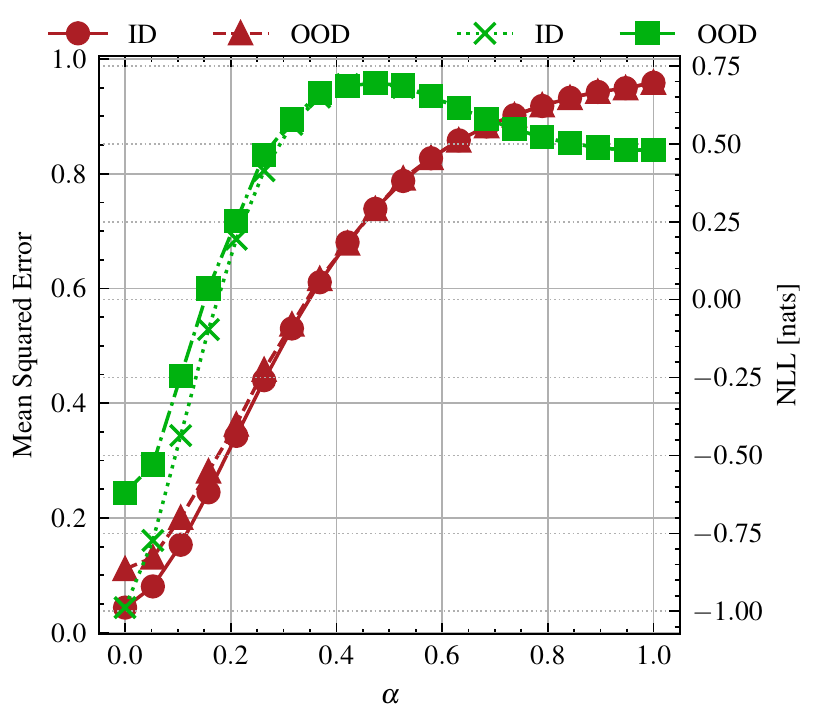}
	\caption{MSE and NLL.}
	\label{fig:loss_landscape:regression_yacht-fc-input_target_cmixup-lin_mse_nll}
\end{subfigure}
\begin{subfigure}{0.35\textwidth}
	\centering
	\includegraphics[width=\textwidth]{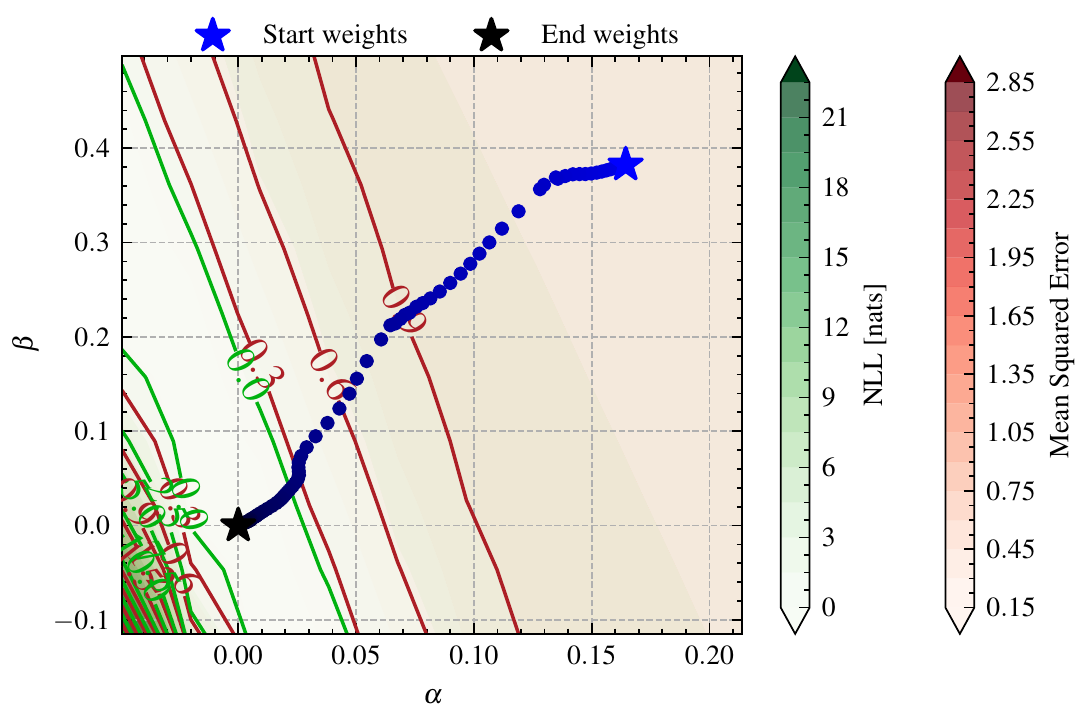}
	\caption{MSE and NLL on ID.}
	\label{fig:loss_landscape:regression_yacht-fc-input_target_cmixup-test_2d_mse_nll}
\end{subfigure}
\begin{subfigure}{0.35\textwidth}
	\centering
	\includegraphics[width=\textwidth]{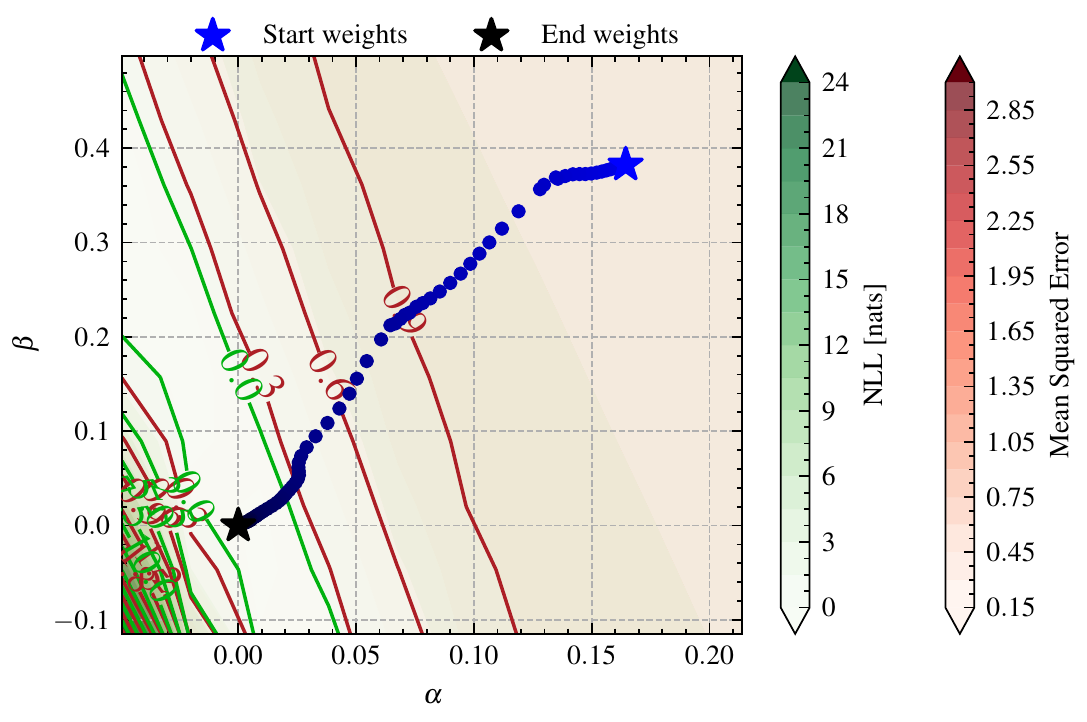}
	\caption{MSE and NLL on OOD.}
	\label{fig:loss_landscape:regression_yacht-fc-input_target_cmixup-test_2d_aug_mse_nll}
\end{subfigure}
\caption{Input-Target CMixUp on Yacht.
\textit{Observations}:
While the shape of the 1D curves looks similar to no noise, the MSE and NLL magnitudes are different.
The 2D plots demonstrate a wider landscape of feasible solutions than no noise.}
\label{fig:loss_landscape:regression_yacht-fc-input_target_cmixup}
\end{figure}
\begin{figure}
\centering
\begin{subfigure}{0.25\textwidth}
	\centering
	\includegraphics[width=\textwidth]{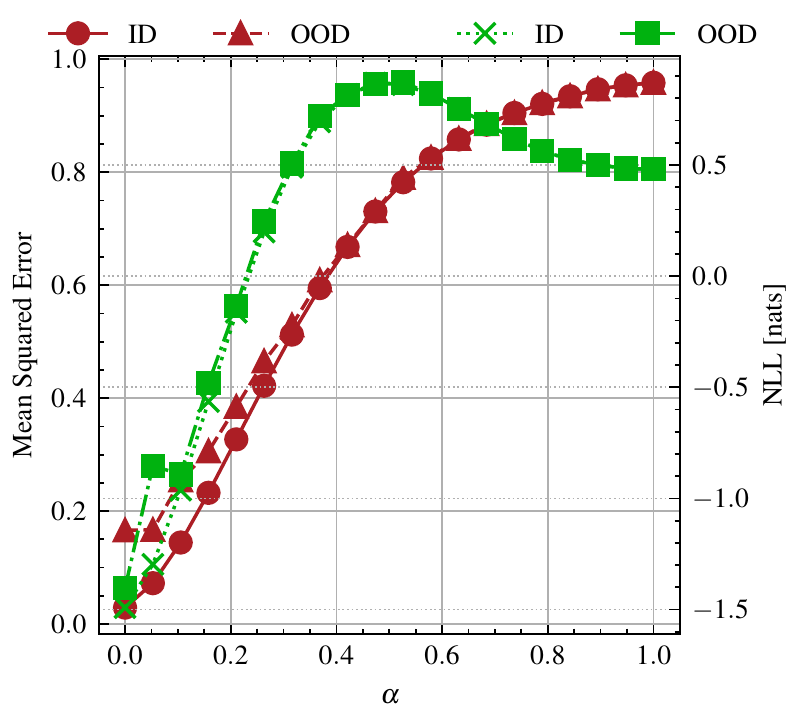}
	\caption{MSE and NLL.}
	\label{fig:loss_landscape:regression_yacht-fc-activation_additive_gaussian-lin_mse_nll}
\end{subfigure}
\begin{subfigure}{0.35\textwidth}
	\centering
	\includegraphics[width=\textwidth]{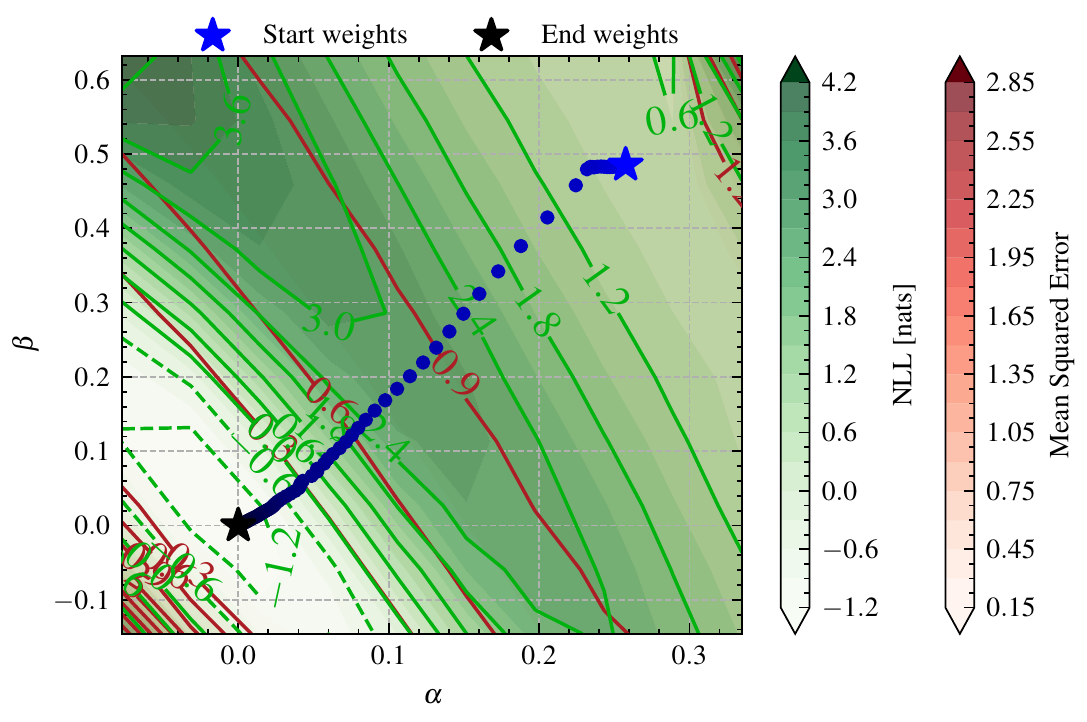}
	\caption{MSE and NLL on ID.}
	\label{fig:loss_landscape:regression_yacht-fc-activation_additive_gaussian-test_2d_mse_nll}
\end{subfigure}
\begin{subfigure}{0.35\textwidth}
	\centering
	\includegraphics[width=\textwidth]{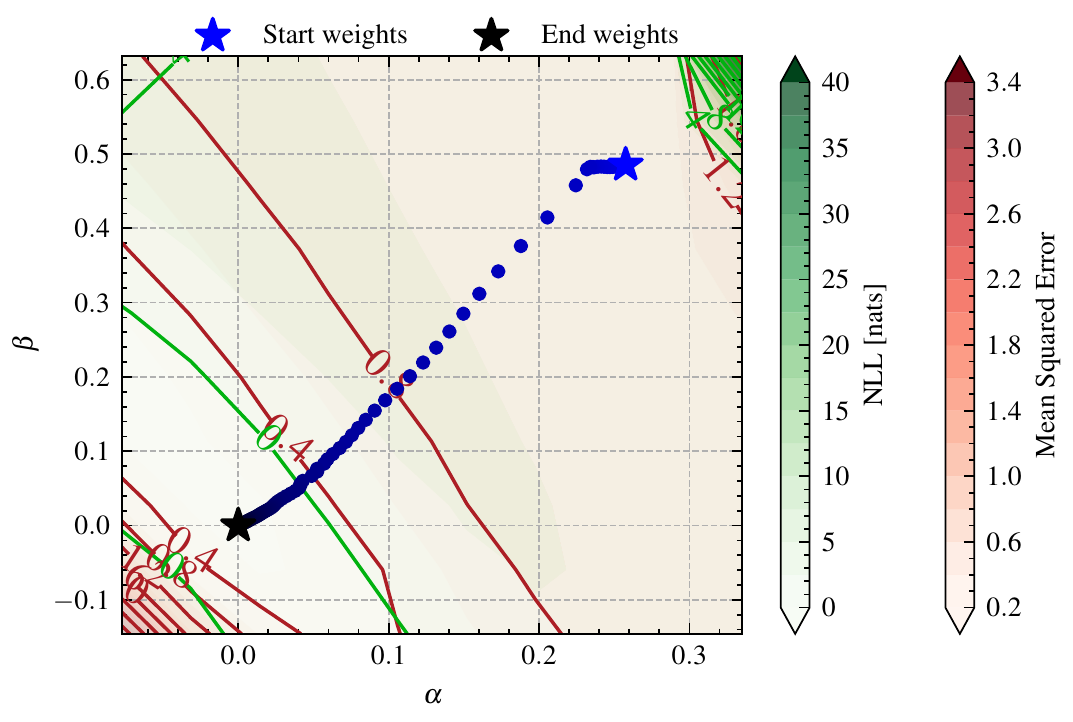}
	\caption{MSE and NLL on OOD.}
	\label{fig:loss_landscape:regression_yacht-fc-activation_additive_gaussian-test_2d_aug_mse_nll}
\end{subfigure}
\caption{Activation Additive Gaussian on Yacht.
\textit{Observations}:
While the shape of the 1D curves looks similar to no noise, the MSE and NLL magnitudes are different.
The 2D plots are close to the no-noise ones, showing marginal differences.}
\label{fig:loss_landscape:regression_yacht-fc-activation_additive_gaussian}
\end{figure}
\begin{figure}
\centering
\begin{subfigure}{0.25\textwidth}
	\centering
	\includegraphics[width=\textwidth]{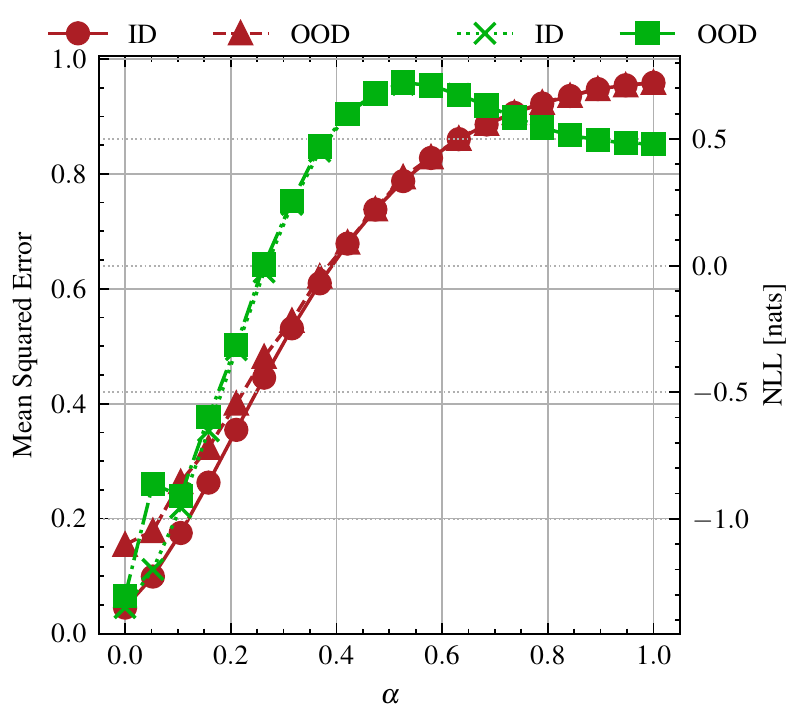}
	\caption{MSE and NLL.}
	\label{fig:loss_landscape:regression_yacht-fc-activation_dropout-lin_mse_nll}
\end{subfigure}
\begin{subfigure}{0.35\textwidth}
	\centering
	\includegraphics[width=\textwidth]{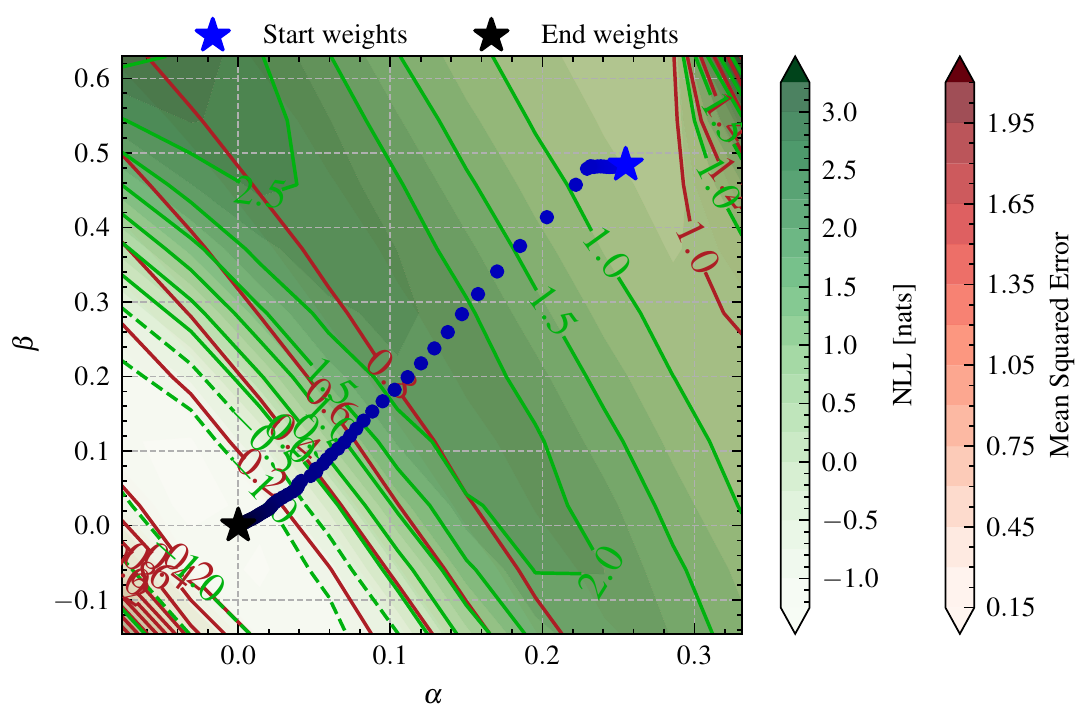}
	\caption{MSE and NLL on ID.}
	\label{fig:loss_landscape:regression_yacht-fc-activation_dropout-test_2d_mse_nll}
\end{subfigure}
\begin{subfigure}{0.35\textwidth}
	\centering
	\includegraphics[width=\textwidth]{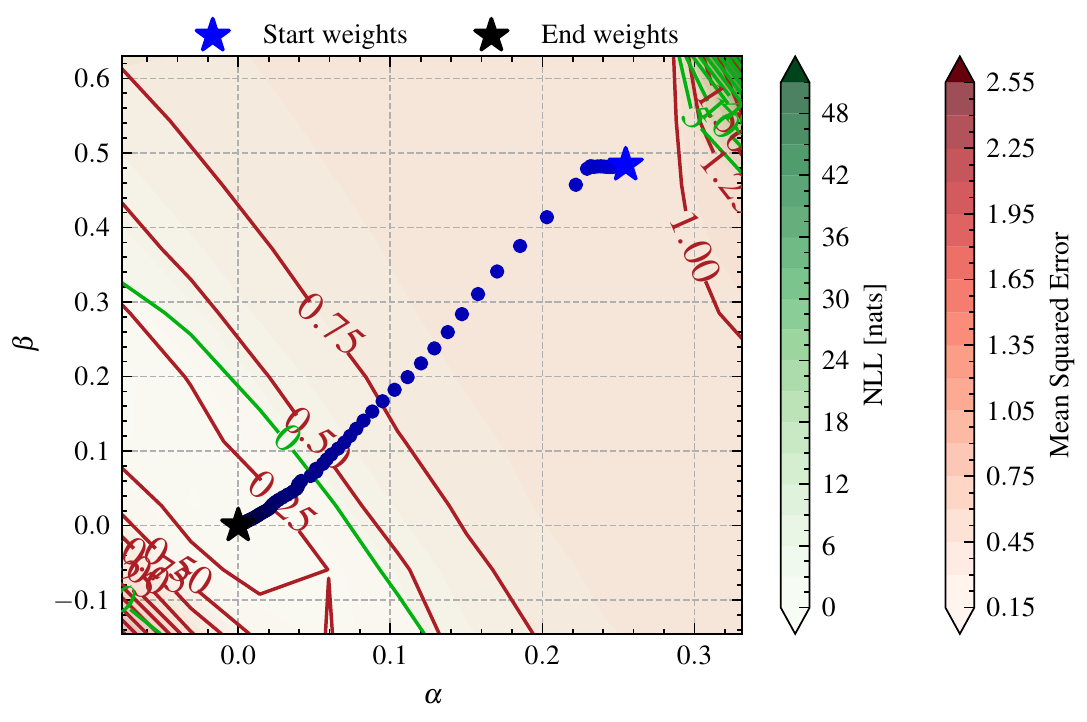}
	\caption{MSE and NLL on OOD.}
	\label{fig:loss_landscape:regression_yacht-fc-activation_dropout-test_2d_aug_mse_nll}
\end{subfigure}
\caption{Activation Dropout on Yacht.
\textit{Observations}:
{The 1D curves look similar to no noise but Dropout converged in a narrow valley, as demonstrated in the 2D plots.}}
\label{fig:loss_landscape:regression_yacht-fc-activation_dropout}
\end{figure}
\begin{figure}
\centering
\begin{subfigure}{0.25\textwidth}
	\centering
	\includegraphics[width=\textwidth]{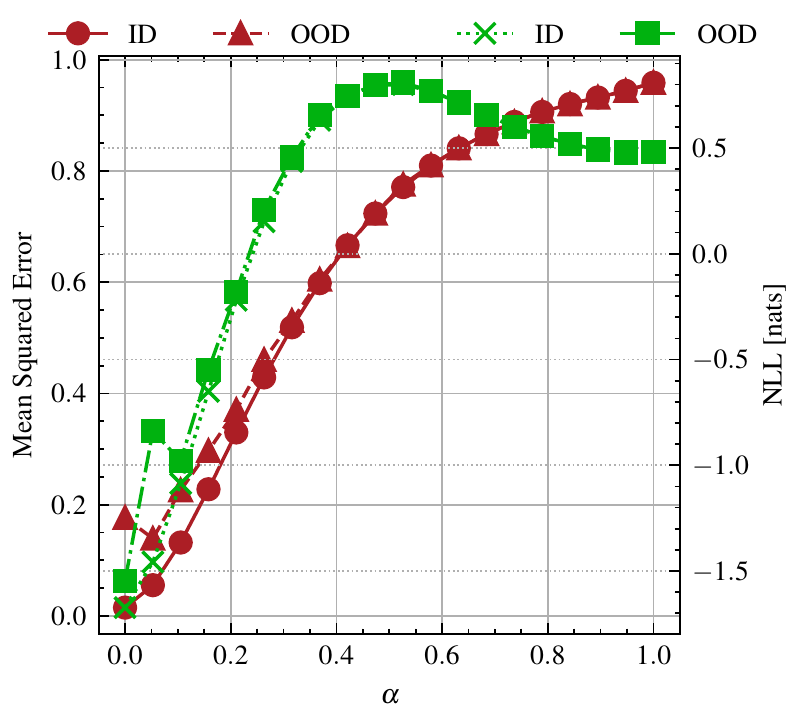}
	\caption{MSE and NLL.}
	\label{fig:loss_landscape:regression_yacht-fc-gradient_gaussian-lin_mse_nll}
\end{subfigure}
\begin{subfigure}{0.35\textwidth}
	\centering
	\includegraphics[width=\textwidth]{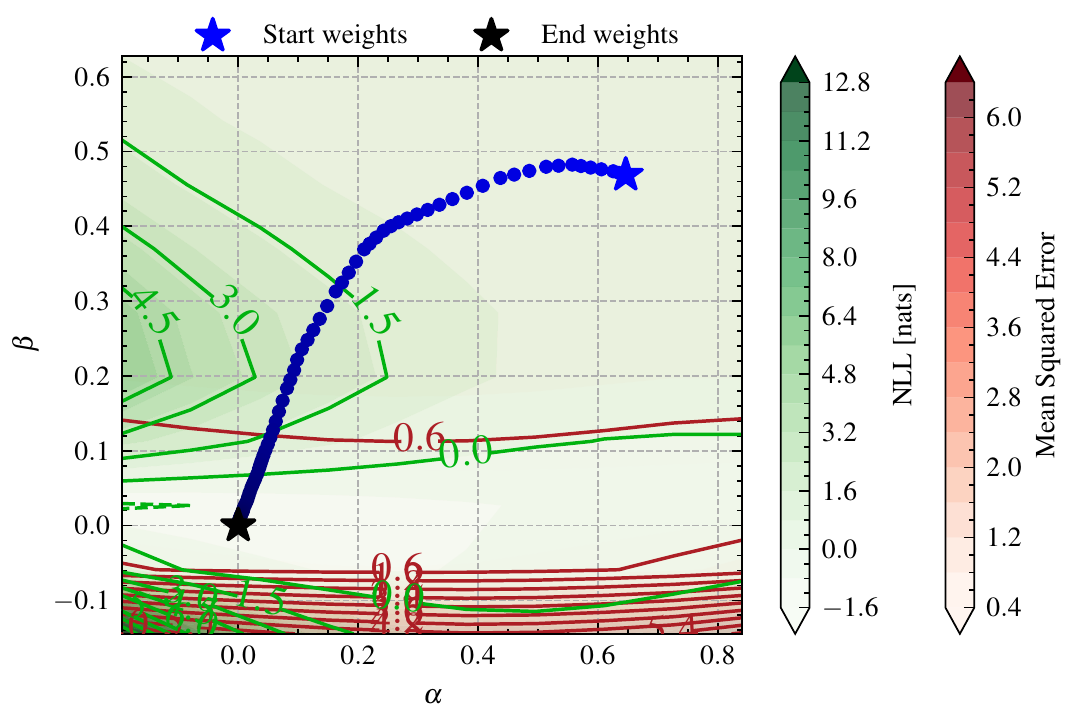}
	\caption{MSE and NLL on ID.}
	\label{fig:loss_landscape:regression_yacht-fc-gradient_gaussian-test_2d_mse_nll}
\end{subfigure}
\begin{subfigure}{0.35\textwidth}
	\centering
	\includegraphics[width=\textwidth]{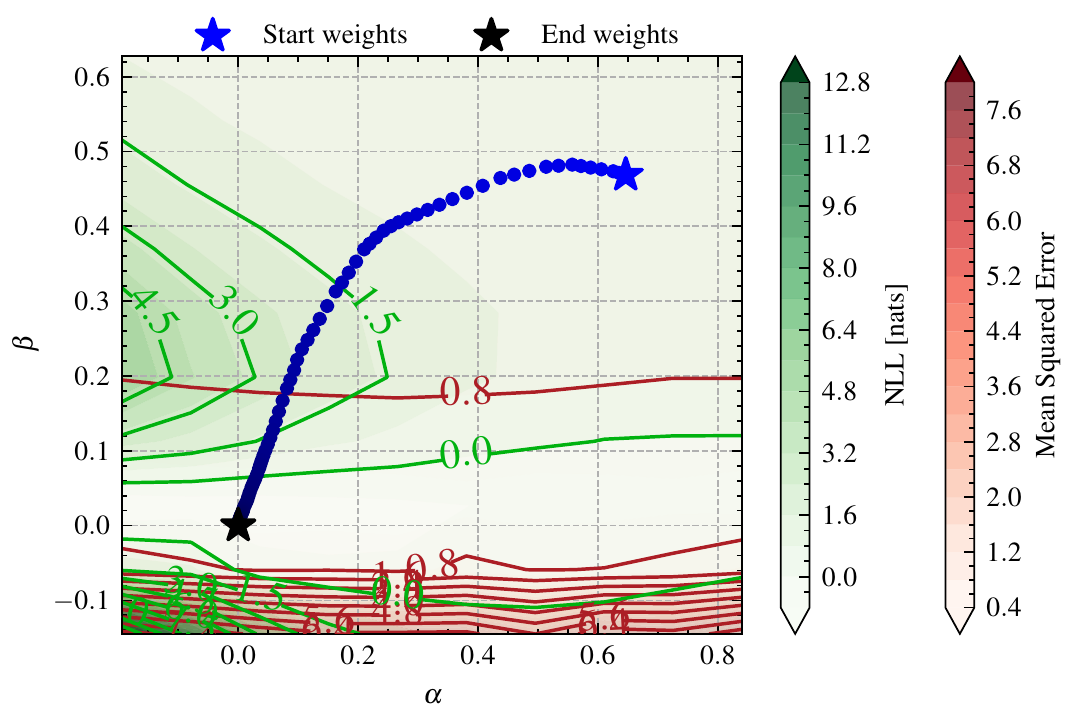}
	\caption{MSE and NLL on OOD.}
	\label{fig:loss_landscape:regression_yacht-fc-gradient_gaussian-test_2d_aug_mse_nll}
\end{subfigure}
\caption{Gradient Gaussian on Yacht.
\textit{Observations}: The 1D curves remained unchanged except for the magnitude of NLL or MSE.
Nevertheless, the 2D plots show us that the optimisation trajectory significantly differed from no noise where the landscape of potential optimal solutions was wider.}
\label{fig:loss_landscape:regression_yacht-fc-gradient_gaussian}
\end{figure}
\begin{figure}
\centering
\begin{subfigure}{0.25\textwidth}
	\centering
	\includegraphics[width=\textwidth]{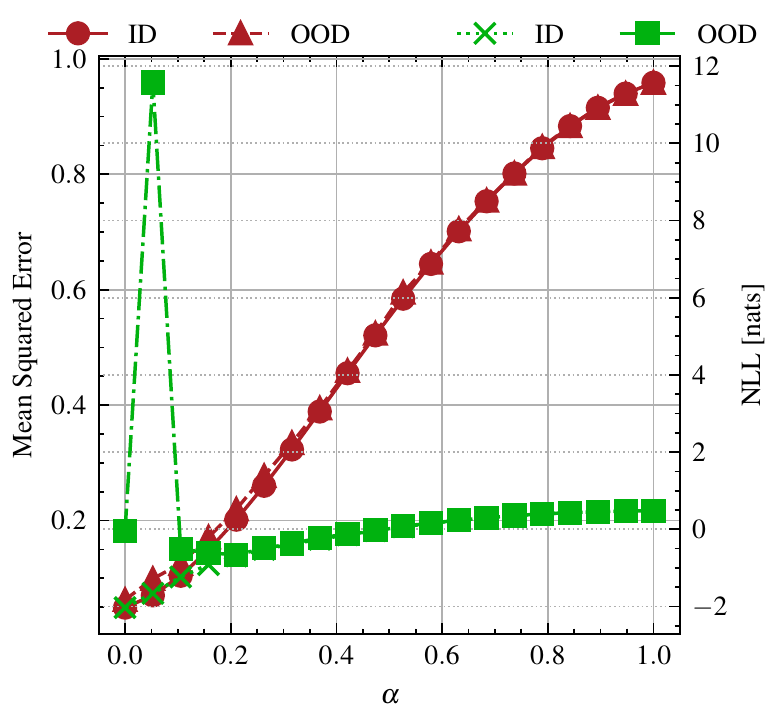}
	\caption{MSE and NLL.}
	\label{fig:loss_landscape:regression_yacht-fc-model_sp-lin_mse_nll}
\end{subfigure}
\begin{subfigure}{0.35\textwidth}
	\centering
	\includegraphics[width=\textwidth]{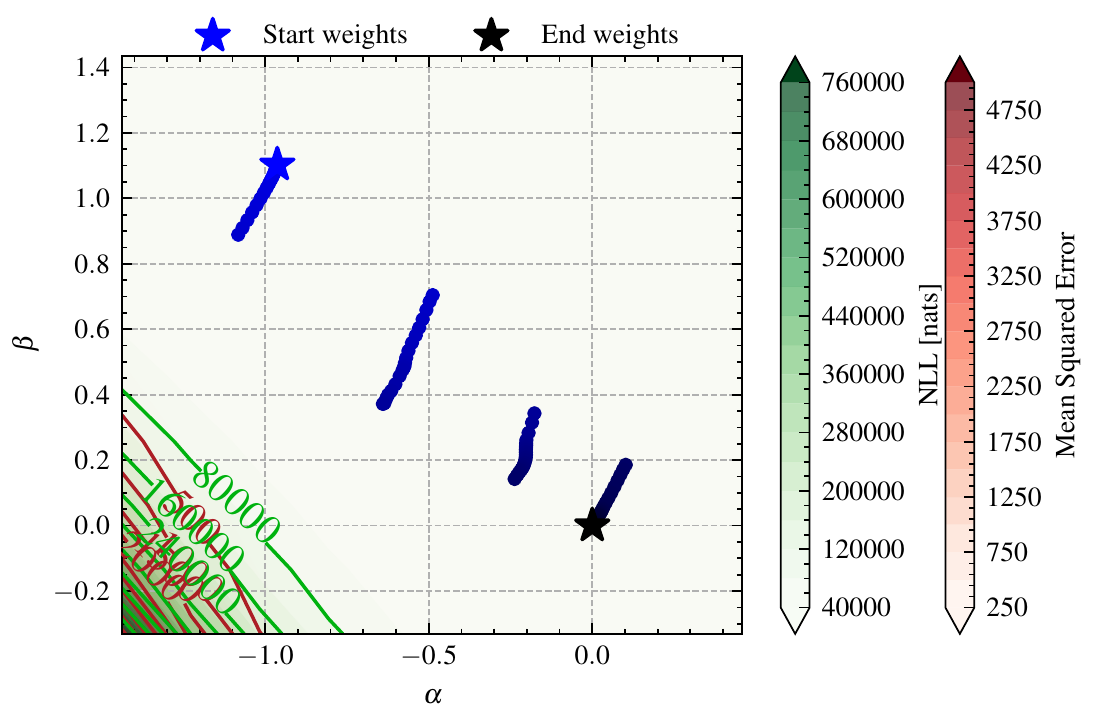}
	\caption{MSE and NLL on ID.}
	\label{fig:loss_landscape:regression_yacht-fc-model_sp-test_2d_mse_nll}
\end{subfigure}
\begin{subfigure}{0.35\textwidth}
	\centering
	\includegraphics[width=\textwidth]{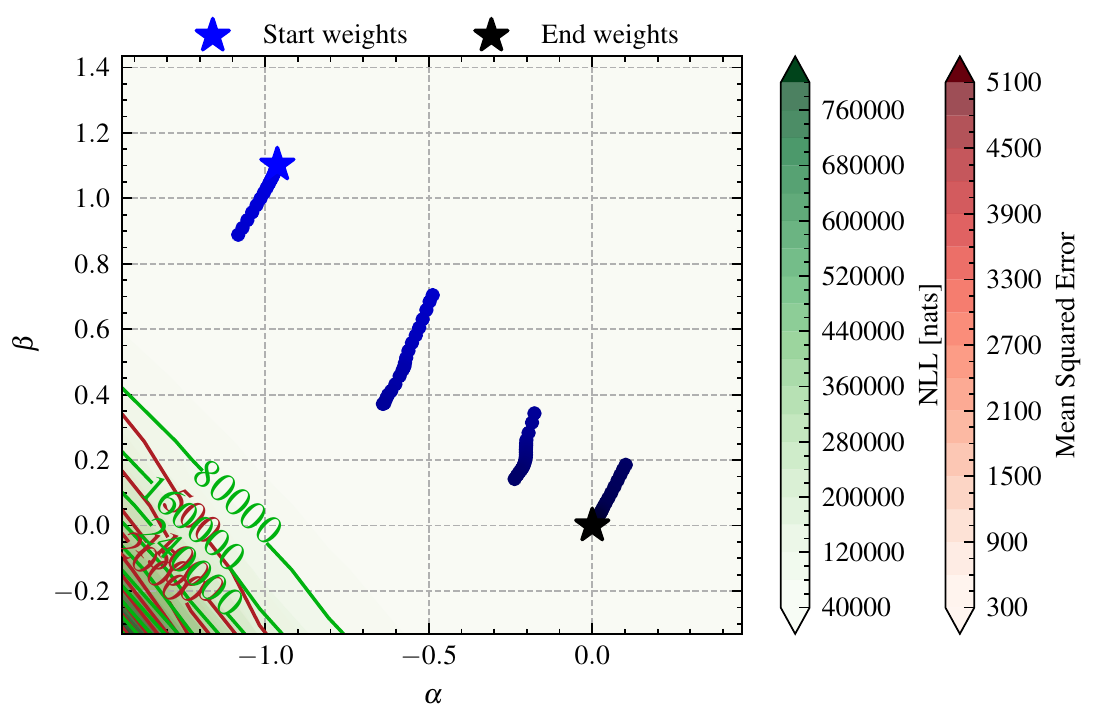}
	\caption{MSE and NLL on OOD.}
	\label{fig:loss_landscape:regression_yacht-fc-model_sp-test_2d_aug_mse_nll}
\end{subfigure}
\caption{Model Shrink and Perturb on Yacht.
\textit{Observations}: The model jumped between narrow valleys as seed in the 2D plots and the 1D plots show smoother behaviour from the OOD perspective for MSE but not NLL.}
\label{fig:loss_landscape:regression_yacht-fc-model_sp}
\end{figure}
\begin{figure}
\centering
\begin{subfigure}{0.25\textwidth}
	\centering
	\includegraphics[width=\textwidth]{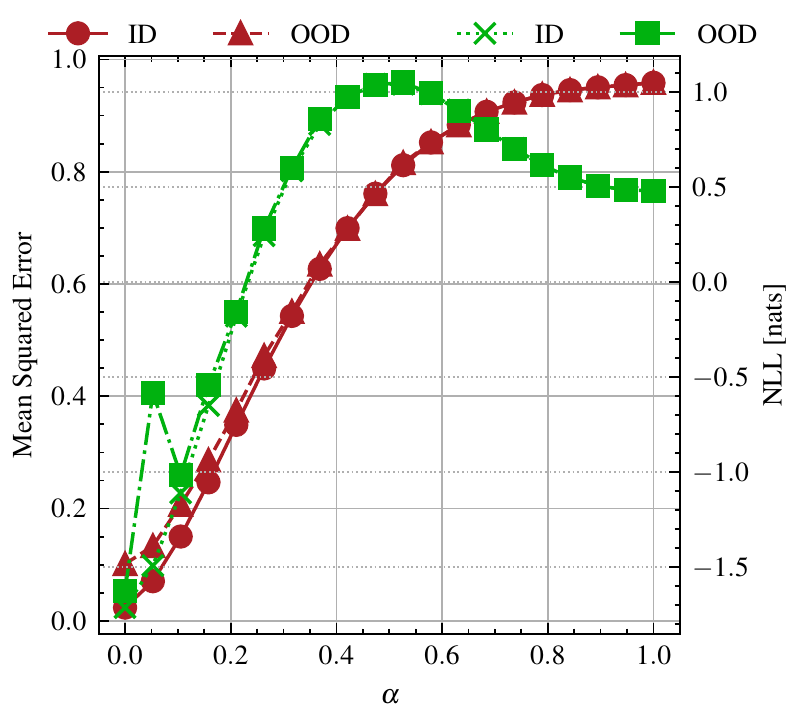}
	\caption{MSE and NLL.}
	\label{fig:loss_landscape:regression_yacht-fc-weight_additive_gaussian-lin_mse_nll}
\end{subfigure}
\begin{subfigure}{0.35\textwidth}
	\centering
	\includegraphics[width=\textwidth]{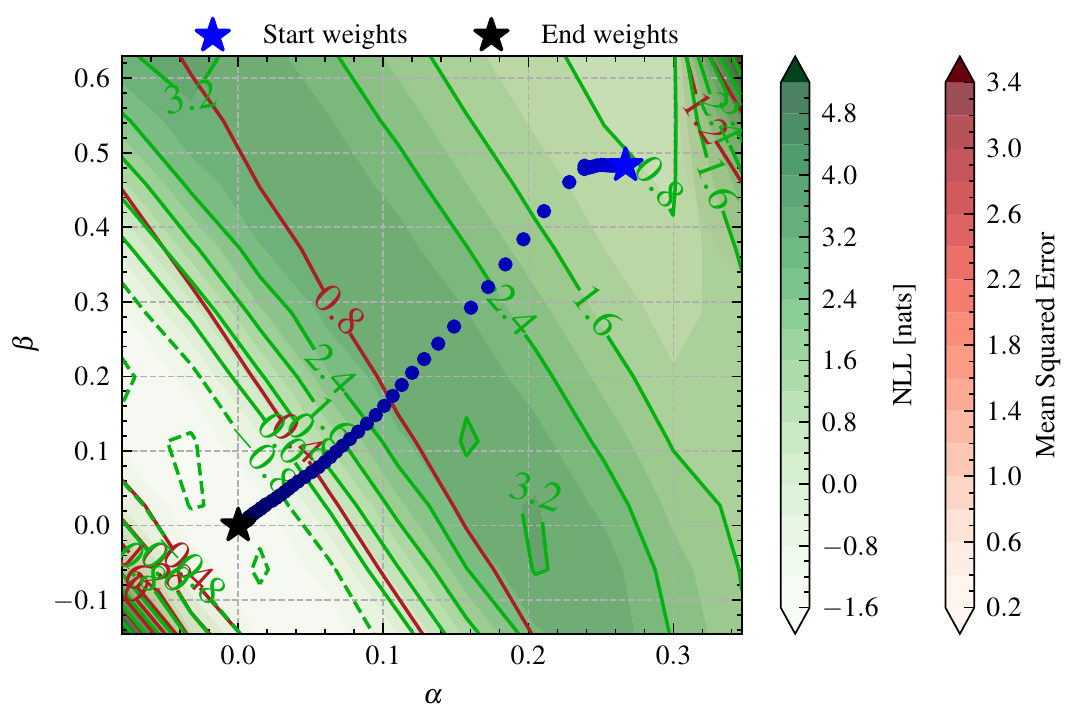}
	\caption{MSE and NLL on ID.}
	\label{fig:loss_landscape:regression_yacht-fc-weight_additive_gaussian-test_2d_mse_nll}
\end{subfigure}
\begin{subfigure}{0.35\textwidth}
	\centering
	\includegraphics[width=\textwidth]{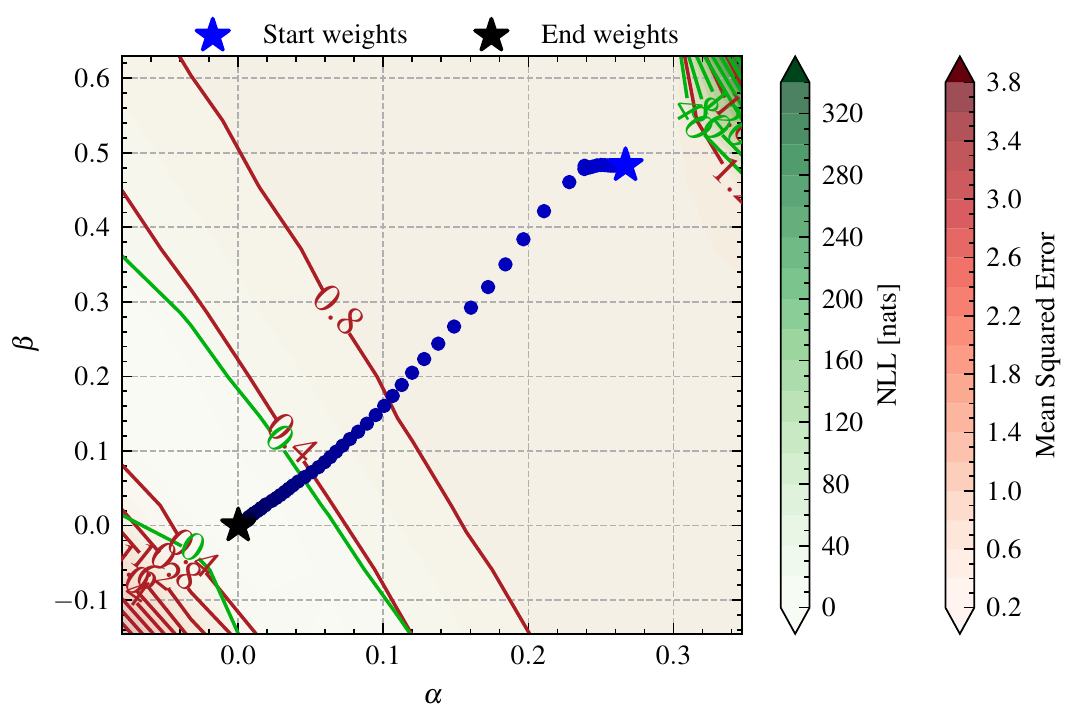}
	\caption{MSE and NLL on OOD.}
	\label{fig:loss_landscape:regression_yacht-fc-weight_additive_gaussian-test_2d_aug_mse_nll}
\end{subfigure}
\caption{Weight Additive Gaussian on Yacht.
\textit{Observations}: Did not change the smoothness of the 1D curves or the 2D trajectory.}
\label{fig:loss_landscape:regression_yacht-fc-weight_additive_gaussian}
\end{figure}
\begin{figure}
\centering
\begin{subfigure}{0.25\textwidth}
	\centering
	\includegraphics[width=\textwidth]{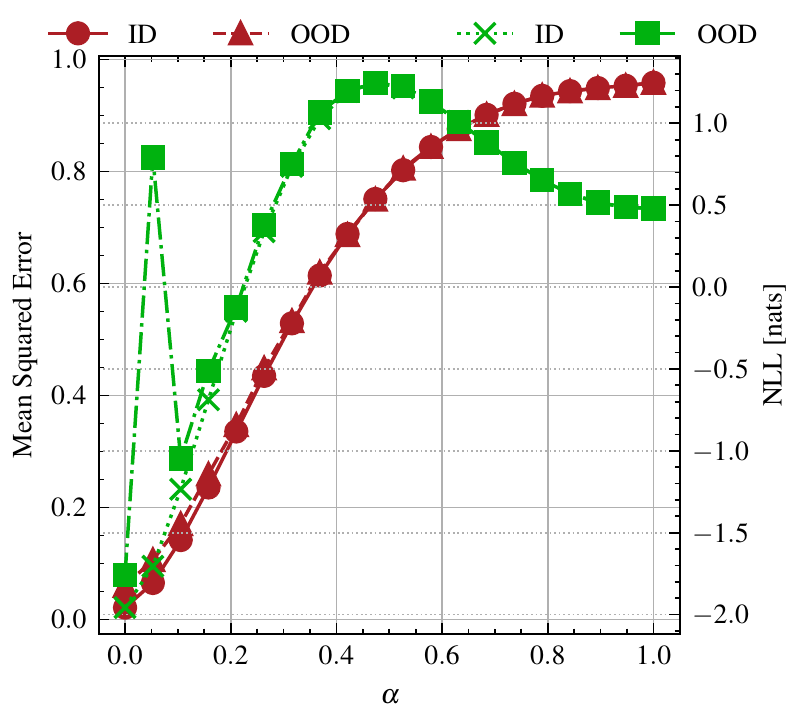}
	\caption{MSE and NLL.}
	\label{fig:loss_landscape:regression_yacht-fc-weight_dropconnect-lin_mse_nll}
\end{subfigure}
\begin{subfigure}{0.35\textwidth}
	\centering
	\includegraphics[width=\textwidth]{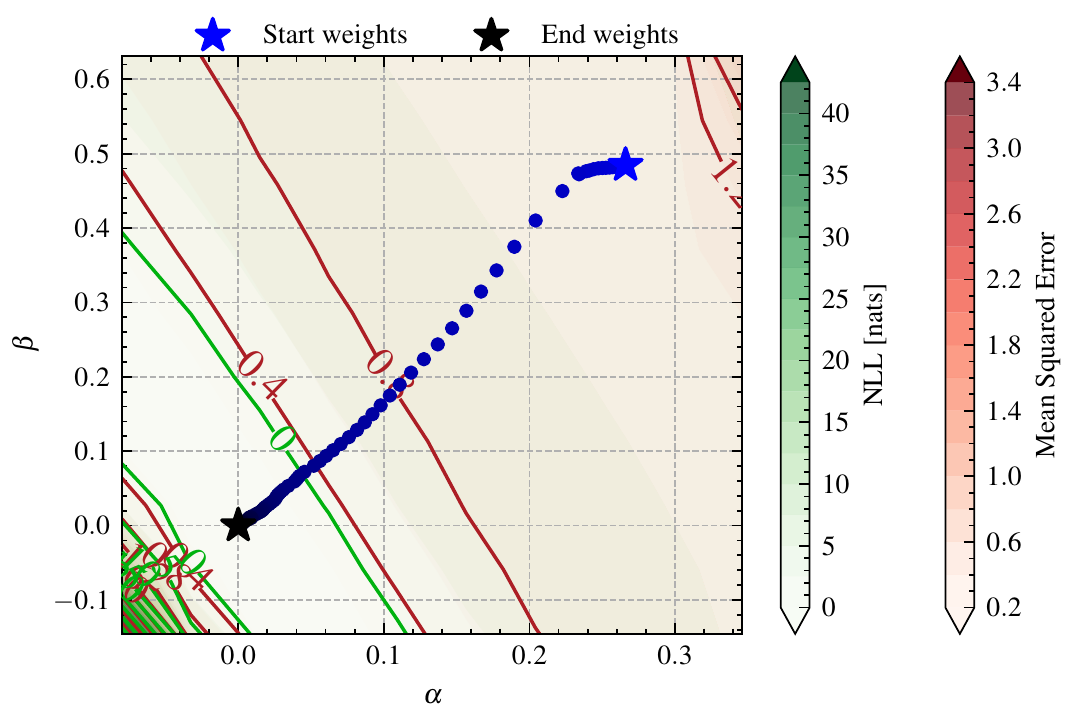}
	\caption{MSE and NLL on ID.}
	\label{fig:loss_landscape:regression_yacht-fc-weight_dropconnect-test_2d_mse_nll}
\end{subfigure}
\begin{subfigure}{0.35\textwidth}
	\centering
	\includegraphics[width=\textwidth]{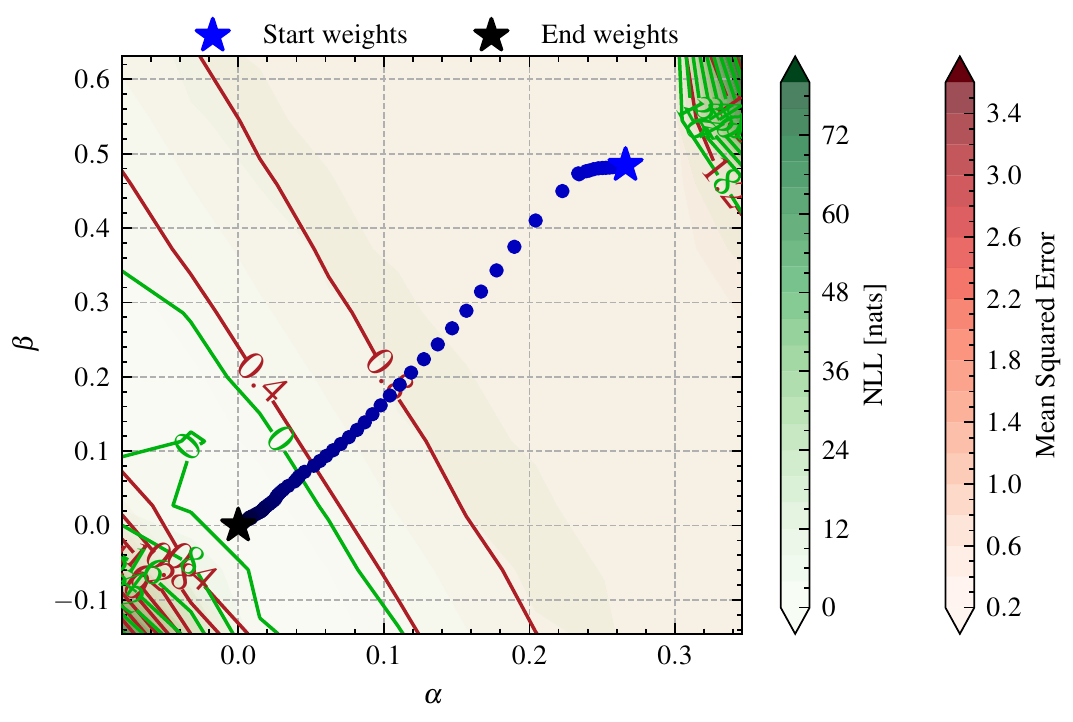}
	\caption{MSE and NLL on OOD.}
	\label{fig:loss_landscape:regression_yacht-fc-weight_dropconnect-test_2d_aug_mse_nll}
\end{subfigure}
\caption{Weight DropConnect on Yacht.
\textit{Observations}:
While the shape of the 1D curves looks similar to no noise, the MSE and NLL magnitudes are different.
The 2D plots demonstrate a wider landscape of feasible solutions than no noise.}
\label{fig:loss_landscape:regression_yacht-fc-weight_dropconnect}
\end{figure}

\end{document}